\documentclass[preprint,10pt]{elsarticle}
\usepackage{amssymb}
\usepackage[para,online,flushleft]{threeparttable}
\graphicspath{ {./figures/} }
\usepackage{hyperref}
\usepackage{float}
\usepackage{verbatim}
\restylefloat{figure}
\restylefloat{table}

\journal{Journal}

\usepackage{titlesec}
\titleclass{\subsubsubsection}{straight}[\subsection]
\newcounter{subsubsubsection}[subsubsection]
\renewcommand\thesubsubsubsection{\thesubsubsection.\arabic{subsubsubsection}}
\titleformat{\subsubsubsection}{\normalfont\normalsize\itshape}{\thesubsubsubsection.\space}{0em}{}
\titlespacing*{\subsubsubsection}{0pt}{2ex plus 1ex minus .2ex}{0.75ex plus .2ex}
\makeatletter
\def\toclevel@subsubsubsection{4}
\def\l@subsubsubsection{\@dottedtocline{4}{7em}{4em}}
\makeatother

\usepackage{amssymb}
\usepackage{pifont}
\newcommand{\cmark}{\ding{51}}%
\newcommand{\xmark}{\ding{55}}%

\usepackage[table]{xcolor}
\usepackage{graphicx}
\usepackage{amsmath,amssymb,amsfonts}
\usepackage{algorithmic}
\usepackage{graphicx}
\usepackage{textcomp}
\usepackage{booktabs}
\usepackage{xcolor}
\usepackage{array}
\usepackage{multirow}
\usepackage{url}
\usepackage{float}
\usepackage{pifont}
\usepackage{enumitem}

\floatstyle{plaintop}
\restylefloat{table}
\usepackage{caption}
\usepackage{subcaption}

\newcommand{\ph}[1]{\textcolor{black}{#1}}

\newcommand{\red}[1]{{\textcolor{red}{#1}}}

\usepackage{booktabs}
\newcommand{\tabitem}{~~\llap{\textbullet}~~}

\newif\ifblackandwhite
\blackandwhitetrue
\usepackage{pdflscape}
\usepackage{colortbl}
\newcommand{\myrowcolour}{\rowcolor[gray]{0.925}}
\usepackage{booktabs}

\usepackage{adjustbox}
\usepackage{rotating}

\def\BibTeX{{\rm B\kern-.05em{\sc i\kern-.025em b}\kern-.08em
    T\kern-.1667em\lower.7ex\hbox{E}\kern-.125emX}}

\usepackage{geometry}
\begin{document}
\begin{frontmatter}

\begin{titlepage}
\begin{center}
\vspace*{0.5cm}

\textbf{All You Need for Object Detection: From Pixels, Points, and Prompts to Next-Gen Fusion and Multimodal LLMs/VLMs in Autonomous Vehicles} 
\footnote{This material is based upon the work supported by the National Science Foundation (NSF) under Grant Numbers 2008784 and
2204721.}

\vspace{2cm}

Sayed Pedram Haeri Boroujeni$^{a}$ (shaerib@g.clemson.edu)\\
Niloufar Mehrabi$^a$ (nmehrab@g.clemson.edu)\\
Hazim Alzorgan$^{a}$(halzorg@g.clemson.edu)\\
Mahlagha Fazeli$^{a}$ (mfazeli@clemson.edu)\\
Abolfazl Razi$^{a*}$ (arazi@clemson.edu)\\

\hspace{10pt}

\begin{flushleft}
\small  
$^a$School of Computing, Clemson University, Clemson, SC 29632, USA\\[1mm]

\vspace{2.5cm}
\textbf{Corresponding Author:} \\
Abolfazl Razi\\
School of Computing, Clemson University, Clemson, SC 29632, USA \\
Email$^{1}$: arazi@clemson.edu\\
Email$^{2}$: shaerib@g.clemson.edu\\

\end{flushleft}        
\end{center}
\end{titlepage}

\title{All You Need for Object Detection: From Pixels, Points, and Prompts to Next-Gen Fusion and Multimodal LLMs/VLMs in Autonomous Vehicles}



\author{Sayed Pedram Haeri Boroujeni$^{a}$, Niloufar Mehrabi$^{a}$, Hazim Alzorgan$^{a}$, Mahlagha Fazeli$^{a}$, Abolfazl Razi$^{a*}$}

\affiliation{organization={School of Computing},
            addressline={Clemson University}, 
            city={Clemson},
            postcode={29632}, 
            state={SC},
            country={USA}}

\begin{abstract}
\small{Autonomous Vehicles (AVs) are transforming the future of transportation through advances in intelligent perception, decision-making, and control systems. However, their success is tied to one core capability, reliable object detection in complex and multimodal environments. While recent breakthroughs in Computer Vision (CV) and Artificial Intelligence (AI) have driven remarkable progress, the field still faces a critical challenge as knowledge remains fragmented across multimodal perception, contextual reasoning, and cooperative intelligence. This survey bridges that gap by delivering a forward-looking analysis of object detection in AVs, emphasizing emerging paradigms such as Vision-Language Models (VLMs), Large Language Models (LLMs), and Generative AI rather than re-examining outdated techniques. We begin by systematically reviewing the fundamental spectrum of AV sensors (camera, ultrasonic, LiDAR, and Radar) and their fusion strategies, highlighting not only their capabilities and limitations in dynamic driving environments but also their potential to integrate with recent advances in LLM/VLM-driven perception frameworks. We also review autonomous vehicle simulators as a critical layer for safe development, scalable testing, and reproducible benchmarking of perception and detection pipelines before real-world deployment. Next, we introduce a structured categorization of AV datasets that moves beyond simple collections, positioning ego-vehicle, infrastructure-based, and cooperative datasets (e.g., V2V, V2I, V2X, I2I), followed by a cross-analysis of data structures and characteristics. Ultimately, we analyze cutting-edge detection methodologies, ranging from 2D and 3D pipelines to hybrid sensor fusion, with particular attention to emerging transformer-driven approaches powered by Vision Transformers (ViTs), Large and Small Language Models (SLMs), and VLMs. By synthesizing these perspectives, our survey delivers a clear roadmap of current capabilities, open challenges, and future opportunities, highlighting underexplored avenues such as multimodal reasoning, cooperative perception, and foundation-model integration. We aim to establish this work as a definitive reference for researchers, practitioners, and developers, fostering accelerated innovation toward safer and more intelligent autonomous driving systems.}

\end{abstract}

\end{frontmatter}

\section{Introduction }
\label{sec:Introduction}
\begin{figure}[htbp]
   \centering
   \centerline{\includegraphics[width=\textwidth]{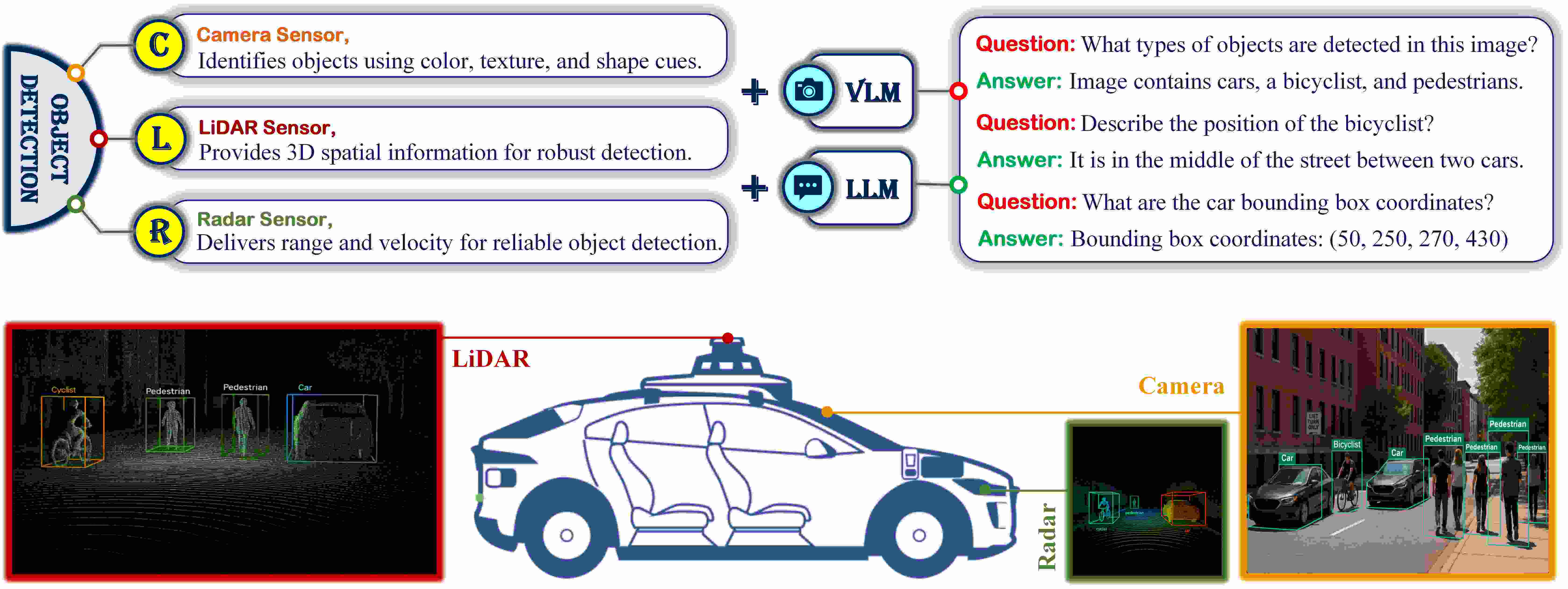}}
   \caption{Visualization of object detection across multiple sensor modalities in autonomous vehicles. The RGB image demonstrates 2D detection with colored bounding boxes for cars, pedestrians, and cyclists. LiDAR and Radar point cloud showcases 3D detection through spatially aligned boxes. The integration of multimodal AI, including LLMs and VLMs, supports contextual understanding and enhances detection accuracy by fusing visual and spatial cues from diverse sensor inputs.}
   \label{fig: detection}
\end{figure}

Object detection is a cornerstone task in autonomous driving systems and stands as a critical component for environmental understanding and safe navigation. It enables Autonomous Vehicles (AVs) to identify surrounding vehicles, pedestrians, cyclists, and obstacles using data captured from various sensors such as cameras, LiDAR, and Radar. Beyond traditional sensing methods, emerging multimodal models, including LLMs and VLMs are being integrated into AV perception systems to enhance scene interpretation and contextual reasoning. Figure \ref{fig: detection} illustrates examples of object detection across these sensor modalities, highlighting the complementary roles of multimodal AI in advancing AVs’ perception.

AVs have surfaced as a disruptive technology that can revolutionize transportation, surveillance, logistics, agriculture, environmental monitoring, and public safety. Underneath such systems lie sophisticated sensing, decision-making, and control infrastructures that enable their autonomous operation without human oversight \cite{elallid2025secure}. Their optimal execution depends on the full integration of the perception, localization, planning, and control modules, whose performance is determined by innovations in sensing technologies and intelligent algorithms. The rapid advance of technology in AI, Machine Learning (ML), and sensors has greatly influenced the rapid development and fielding of these robotic systems \cite{khosravian2024robust}. Despite the great progress, autonomous systems still encounter many technical and practical challenges, particularly in ensuring reliable performance in complex, dynamic, and unpredictable environments. These systems must robustly handle inaccurate sensors, hardware faults, and the computational challenge of not just high-dimensional but also high-frequency real-time data. In parallel with these technical developments, AVs have also drawn great attention from industry and academia with regard to their promising road safety, traffic congestion mitigation, reduction in transportation cost, and social inclusion for elderly and disabled people \cite{gerhard2025scoping}. Their deployment spans a broad range of applications, from self-driving taxis and delivery robots to autonomous machinery and emergency response systems \cite{grosse2024qualitative}. 


Effective adoption and integration of AVs depend on the careful consideration and coordination of multiple interrelated factors. Robustness, scalability, reliability, real-time efficiency, accuracy, and safety are fundamental requirements for autonomous driving systems to operate effectively and gain acceptance in real-world applications \cite{wang2025uncertainty}. First and foremost, safety is the most challenging issue, and AVs must function consistently in a dynamic and sometimes unpredictable scenario to protect passengers, pedestrians, and other road users. The ultimate goal is to significantly minimize, or ideally remove, accidents from human error. This requires AV systems to incorporate multiple layers of safety mechanisms, fail-safe redundancies, and rigorous testing to reduce operational risks. Reliability is closely linked to safety, as it ensures that the AV performs consistently across a broad spectrum of use cases, from typical situations to rare edge cases. Reliable behavior is essential for the public to gain confidence in AVs and support their integration into complex traffic systems. Moreover, robustness is equally important. The AV should be able to operate effectively even in the presence of noisy sensors, hardware faults, extreme weather conditions, uncertain road environments, and adversarial inputs. These disruptions can be managed by a well-designed system that not only endures but continues to function reliably and reinforce user trust. Timely responsiveness is also a critical requirement for AV systems. AVs must process data from their sensors, interpret their surroundings, and make instantaneous decisions in rapidly evolving environments. The ability to act within milliseconds is crucial for avoiding collisions and ensuring smooth operation in dynamic. Together, these elements lay the foundation for a reliable, resilient, and publicly trusted autonomous vehicle system. At the same time, perception, localization, and prediction must achieve high levels of accuracy, as AVs are required to detect objects, estimate their positions, and anticipate the behavior of surrounding agents to navigate safely and make informed decisions. Efficiency is also important, as the system has to effectively utilize computational and energy resources \cite{chen2025hierarchical}. Finally, the scalability of AV systems to accommodate multiple types of vehicles, levels of autonomy, and operational conditions (without sacrificing performance or increasing the complexity of the system) is required.

Another critical aspect influencing the performance of autonomous vehicles is their strong dependence on data throughout both development and deployment. Developing high-performing AV systems requires access to large-scale, high-quality, diverse, and well-labeled datasets that comprehensively represent the full spectrum of real-world driving scenarios. From crowded city streets to rare situations like road construction or extreme weather, the system must be trained on and exposed to varied conditions to develop the ability to generalize beyond controlled test environments \cite{zha2025real}. The size of the dataset is especially important, as complex models demand extensive training data to capture the variability of real-world driving. In addition, features extracted from multi-sensor data collected by cameras, LiDAR, and Radar, as well as mechanisms that provide high-quality annotations, are necessary to handle tasks such as object detection, tracking, and behavior prediction. Annotating this multimodal data, however, remains a labor-intensive and time-consuming task that typically involves manual work or human expert supervision of the process to maintain reliable and valid information across modalities \cite{huang2025robotron}. In addition, real-world datasets typically exhibit a long-tail distribution, where critical events such as pedestrian crossings in low-light conditions or near-collision incidents are rarely captured. This lack of edge cases makes it difficult to train models that are resilient to rare but high-stakes occurrences. To mitigate these problems, several techniques for synthetic data generation and data augmentation have been developed in order to augment diversity in the data and to counterbalance the class distribution. It should be noted that transferring knowledge of learned synthetic data to real-world scenarios is still a challenge, in terms of retaining fidelity, transferability, and semantic consistency. It must also be consistently sustained to maintain AV systems in tune with dynamic infrastructure, traffic laws, and driver actions, ultimately underscoring the need for scalable and adaptable data pipelines. In the absence of enough coverage or representative samples, models become biased or too specialized, which decreases their generalization capabilities in unseen conditions. Ultimately, a data-driven technique is necessary for designing intelligent AV systems that can perform reliably in real-world driving environments, not just in theoretical settings \cite{tekkesinoglu2025advancing}.

Autonomous vehicles must perceive and operate with high precision as they navigate environments enriched with diverse sensors, each designed to fulfill distinct functional roles. Cameras acquire high-resolution visual information for subsequent recognition and interpretation of the environment. LiDAR systems provide high-resolution three-dimensional depth measurements, enabling precise spatial mapping of objects and structural elements within a scene \cite{chang2023using}. In contrast, Radar is particularly effective under conditions of severe visual degradation, such as rain, fog, or low-light environments, offering consistent and robust object detection and tracking in adverse weather and limited visibility scenarios \cite{thottempudi2025resilient}. Ultrasonic sensors assist with close-range obstacle detection, which is especially useful during parking or low-speed maneuvers. These sensors are often supplemented by high-resolution maps with precise information about road trajectories, lane boundaries (lines), and traffic signs \cite{wang2025depth}. Furthermore, vehicle-to-everything (V2X) communication increases the situational awareness of vehicles, as they can share data with surrounding infrastructure, other vehicles, and pedestrians \cite{zha2025heterogeneous}. In particular, vehicle-to-vehicle (V2V) communication allows the AVs to exchange their positions, velocities, accelerations, and intended trajectories, and to perform cooperative actions to reduce the risk of collision. Vehicle-to-infrastructure (V2I) communication, frequently enabled by roadside units (RSUs), serves as a crucial source of environmental and regulatory information, such as traffic signal timings and phases, work zones, and speed limits \cite{salari2022optimal}. Moreover, Infrastructure-to-infrastructure (I2I) communication is a crucial part of this ecosystem that serves as the link between fixed infrastructure nodes, such as traffic lights, and RSUs. In addition to external sensing, AVs depend on ego-centric information: their own position, orientation, velocity, and acceleration, as retrieved from Global Positioning System (GPS) \cite{praveen2025autonomous}, Internal Measurement Units (IMUs) \cite{mohammadi2025detection}, and wheel odometry. The states of these internal variables are important for localization, path planning, and motion control of the vehicle. Novel technologies such as vehicle-to-language (V2L) are also under development, aiming to transform multimodal sensory data into natural language descriptions \cite{rs2025embedded}. This enables AVs to produce interpretable explanations of their environment, internal states, and behavioral choices. 


Effective processing of rich sensorimotor data in AVs depends on sophisticated perception algorithms to retrieve, interpret, and aggregate scene descriptions. These algorithms are targeted to address key perception tasks, such as object detection, classification, and segmentation, that are necessary to develop a comprehensive understanding of the operational environment. These perception pipelines include 2D vision-based models, 3D LiDAR-driven methods, and increasingly, hybrid 2D–3D fusion strategies that combine complementary features across modalities. The latest deep learning models, such as convolutional neural networks (CNNs) and transformer-based models, can learn automatic discriminative feature representations from multimodal sensor measurements. Object detection in real time, for instance, is achieved by detection systems such as Faster R-CNN \cite{kumar2025improving}, and Single Shot MultiBox Detector (SSD) \cite{shrivastava2025ai}, while pixel-level scene understanding is obtained through segmentation models such as DeepLab \cite{subhedar2025insights} and Mask R-CNN \cite{chen2025ranging}. In addition to detecting objects in a single frame, such models have now been adapted to understand scene dynamics over time. This allows the models to follow object motion and anticipate their near-future positions. They have also begun to integrate multiple tasks (detection and segmentation) into a single model, which improves processing efficiency and contributes to richer and more semantically meaningful representations of the surrounding environment. To further improve the performance and robustness of such systems, sensor fusion methods are used systematically to integrate heterogeneous data from diverse sensors. These fusion techniques, which can operate at raw data, feature, or decision stages, frequently utilize probabilistic filters, attention models, or deep nets to map and reconcile heterogeneous inputs. This sophisticated algorithmic pipeline is tailored to ensure real-time responsiveness, precision, and resilience, thereby enabling AVs to operate reliably in unpredictable conditions.

Large Language Models (LLMs) \cite{tian2025large, boroujeni2026don} have received widespread interest over the past few years as they can process and generate human-like language within extensive contextual settings. Their incorporation in autonomous and assistive systems changes the way machines perceive, understand, and interact in the environment. In AVs, LLMs are investigated for improving perception and reasoning through natural language interpretation of sensor data, scene descriptions as well as high-level decision-making \cite{tian2025uavs}. This results in better semantic comprehension and an increased ability to deal with complex or ambiguous situations, which traditional models can misinterpret. Additionally, LLMs support natural language querying of sensory inputs and map-based information, enabling intuitive human-vehicle interaction and enhancing explainability in decision pipelines. Concurrently, LLMs are also being adopted in assistive technologies to help people with disabilities, where they can help implement functionalities such as voice-controlled navigation, real-time description of the environment, and personalized interaction. Building on LLMs, Vision-Language Models (VLMs) integrate visual and textual context for a better understanding of an environment \cite{guo2024vlm}. In AVs, this combination enables self-driving cars to describe traffic scenes in human-like language, better understand road conditions, and explain why they perceive and react as they do. Connecting visual input with semantic understanding allows for more efficient and flexible operations of AVs in complex and dynamic environments, making interactions with human users clearer and more intuitive. The multimodal nature of this multi-source integration promotes informed and context-aware decision-making, which is crucial for reliably and safely dealing with real-world traffic scenes. For automated aids, they increase accessibility by allowing machines to verbally describe visual content during use, providing support for low-vision and cognitively impaired users. VLMs can also help close the gap between vision and language, making intelligent systems more interpretable, interactive, and user-oriented.

\subsection{Contributions of This Survey}
This survey provides a comprehensive and structured review of object detection in autonomous vehicles, emphasizing the integration of diverse sensor technologies and the evolution of detection methods. After examining over seven hundred research and survey articles, the study identifies key trends, innovations, and research gaps across multiple facets of AV perception systems. The survey is uniquely organized into three core components: AV sensor technologies, AV datasets, and object detection methods. In the first part, we analyze different sensor types, including camera, ultrasonic, LiDAR, and Radar sensors, as well as their fusion strategies. The second part introduces a novel taxonomy of AV datasets categorized by ego-vehicle, infrastructure, and communication paradigms (e.g., V2L, V2V, V2I, V2X, I2I). Finally, the third part presents a detailed review of object detection techniques, including 2D camera-based, 3D LiDAR-based, 2D–3D fusion, and LLM/VLM-based approaches. By synthesizing these diverse elements, our survey not only highlights the current state of object detection in AVs but also offers insights into underexplored areas and emerging opportunities, setting a foundation for future research directions in this rapidly advancing field.

The key contributions of this review paper can be summarized as follows:

\begin{itemize}

    \item Reviews significant and recently published survey papers in the context of object detection in autonomous vehicles, summarizing their contents, strengths, and limitations. We highlight the key topics covered in each study, as well as the missing or underexplored areas that remain unaddressed

    \item Conducts an extensive review of state-of-the-art detection methods in AVs, including 2D camera-based approaches, 3D LiDAR-based techniques, and hybrid 2D–3D sensor fusion strategies. The review covers core methodologies, network architectures, performance benchmarks, and application contexts.

    \item Explores and evaluates emerging AI techniques, particularly Vision Transformers (ViTs), Large Language Models (LLMs), and Vision-Language Models (VLMs), as well as their integration into AV perception systems for enhanced object detection capabilities. We analyze how these models enable multimodal reasoning, spatial understanding, and instruction-following in autonomous driving tasks.

    \item Conducts a comparative evaluation of object detection algorithms across all four categories, including 2D (camera-based), 3D (LiDAR-based), 2D–3D fusion, and LLM/VLM-based methods, by analyzing their performance on widely used benchmark datasets. We also identify and discuss the top 10 representative algorithms in each category, examining their key innovations, architectures, strengths, and limitations.

    \item Examines various types of sensor technologies used in AVs and their applications in object detection. We outline the strengths and weaknesses of each sensor modality, including camera, ultrasonic, LiDAR, and Radar sensors, along with a discussion of their integration and utilization within AV perception systems.

    \item Proposes a unique and structured categorization of AV datasets developed in recent years, followed by a systematic analysis of each dataset. This includes ego-vehicle, roadside, and cooperative perception datasets (V2V, V2I, V2X, I2I). We highlight key characteristics such as sensor modalities, data volume, annotation types, collection environments, and application focus.

    \item Discuss the simulation platforms commonly used for synthetic AV dataset generation and testing. We compare these simulators based on key features such as supported sensor modalities (e.g., camera, LiDAR, Radar), environmental realism, weather and traffic control, annotation capabilities, computational requirements, and compatibility with AV development frameworks. 

    \item We highlight open problems and future directions at the end of each major section of our survey paper to guide future advancements in AV perception and object detection. This survey paper aims to to serve as a comprehensive and practical reference for researchers, practitioners, and developers in the field of autonomous driving, enabling better understanding, benchmarking, and design of AV detection systems.

\end{itemize}

\subsection{Organization of This Paper}
\begin{figure}[H]
   \centering
   \centerline{\includegraphics[width=0.92\textwidth]{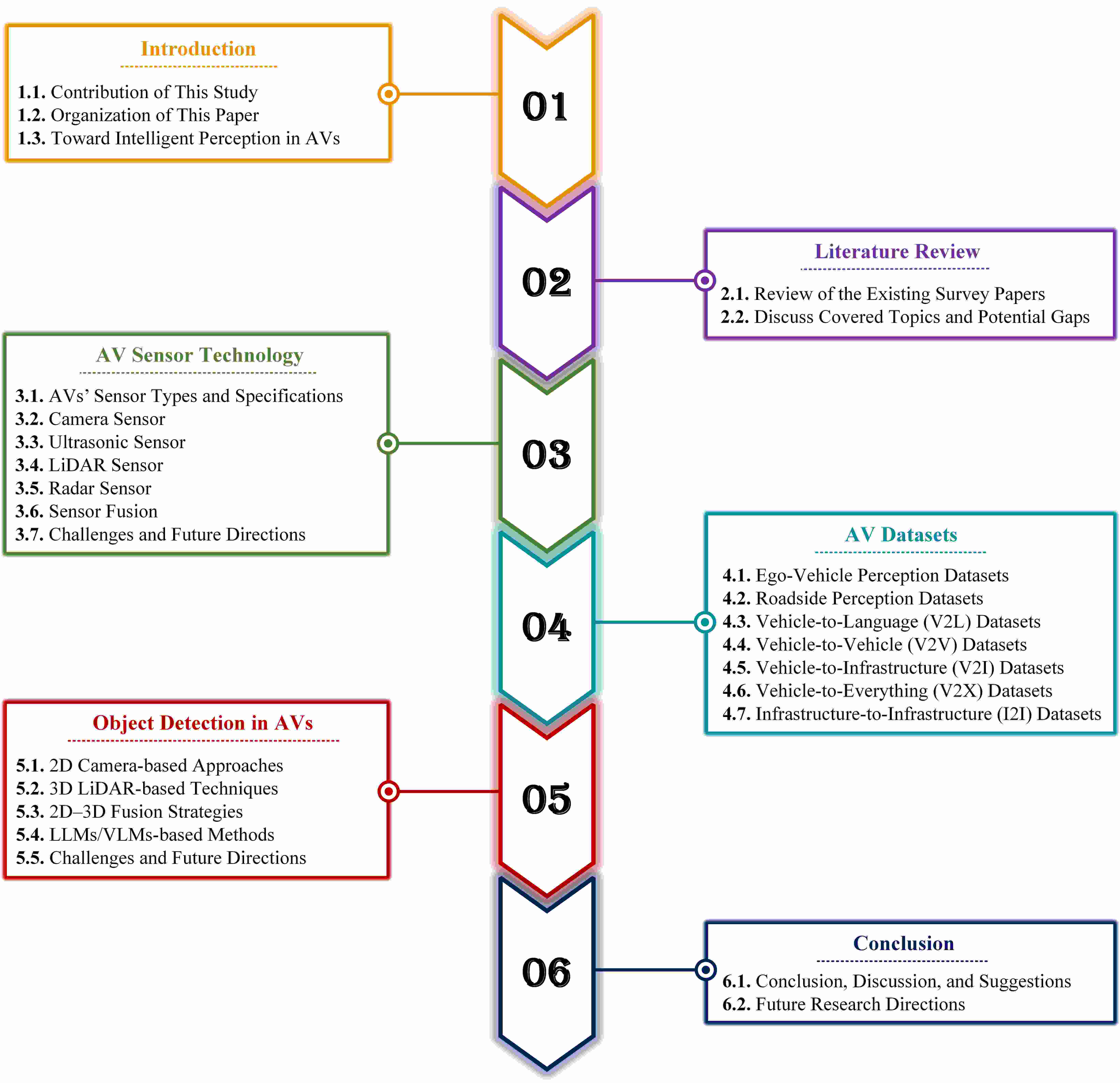}}
   \caption{The organization of this survey paper.}
   \label{fig: content}
\end{figure}

The general structure of the paper is illustrated in Figure \ref{fig: content}. Section \ref{sec: Related Literature} reviews existing survey literature on object detection in autonomous driving systems, highlighting their covered topics and identifying potential gaps. Section \ref{sec: AV Sensors} describes the details of AV technologies and sensor specifications used in autonomous driving applications. Section \ref{sec: AVs Simulator} then introduces autonomous vehicle simulators and discusses their role in development, testing, and benchmarking. The existing AV datasets are provided in section \ref{sec: AVs Dataset}, with a focus on their sensing modalities, annotation types, coverage diversity, and applicability to various perception tasks. Section \ref{sec: object detection} presents a comprehensive review of object detection methods in autonomous vehicles, covering 2D camera-based approaches, 3D LiDAR-based techniques, 2D–3D fusion strategies, and emerging methods based on LLMs and VLMs. Lastly, Section \ref{sec: Conclusion} contains conclusions and future directions.\\

\section{Survey Methodology and Related Literature}
\label{sec: Related Literature}

This section outlines the methodology used to construct our survey and summarizes the most relevant prior literature to contextualize our contributions. We first describe the search methodology used to identify and screen relevant studies, including the data sources consulted, search strategy, and selection criteria, to ensure transparency and reproducibility. We then review existing survey papers most closely related to object detection for autonomous vehicles, highlighting their scope and limitations, and clarifying how our work differs by offering a unified perspective across sensors, fusion strategies, datasets, detection paradigms, and recent vision–language and large language model developments.

\subsection{Search Methodology}

To ensure transparency and reproducibility, we followed a structured literature collection and screening protocol to construct the corpus analyzed in this survey. Our goal was to capture both (i) established object detection and multi-sensor fusion research for autonomous vehicles, and (ii) emerging directions enabled by LLMs and VLMs in autonomous driving perception.

\begin{figure}[H]
   \centering
   \centerline{\includegraphics[width=\textwidth]{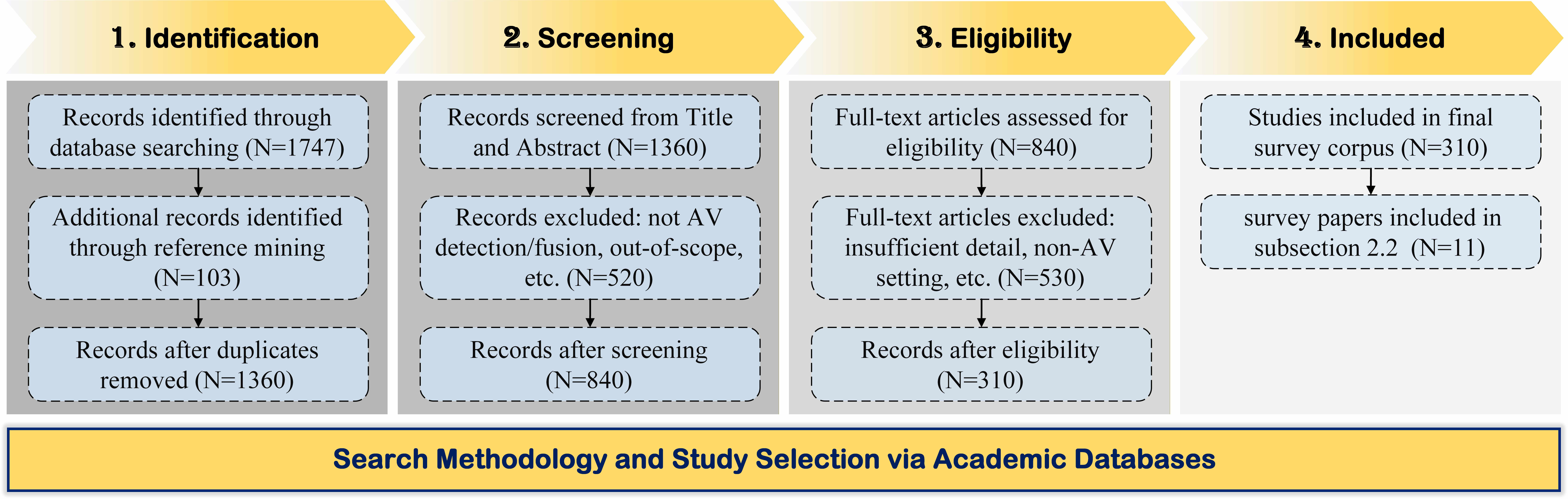}}
   \caption{PRISMA-style flow diagram of the literature search and study selection process used to construct the survey corpus, showing records identified across databases, duplicates removed, title/abstract screening, full-text eligibility assessment, exclusion reasons, and final included studies.}
   \label{fig:prisma}
\end{figure}

\begin{itemize}

\item \textbf{Data Sources:} We queried major academic and preprint repositories that broadly cover computer vision, robotics, and autonomous driving research, including \emph{IEEE Xplore}, \emph{ACM Digital Library}, \emph{Scopus}, \emph{Web of Science}, and \emph{arXiv}. To reduce the risk of missing highly cited or widely used works, We also conducted backward and forward snowballing (reference list checking and citation tracking) from key papers identified during the screening process.

\item \textbf{Time Window:} We focused on publications from 2018 to 2025 to reflect modern deep learning-based detection pipelines, recent multi-sensor fusion advances, and the rapid emergence of VLM/LLM-based perception and reasoning. Earlier seminal works were included only when necessary to provide historical context or widely adopted baselines.

\item \textbf{Search Queries:} We used a combination of keywords spanning autonomous driving, sensing, object detection, fusion, and multimodal foundation models. The final query set included combinations of ``autonomous driving'', ``autonomous vehicle'', ``self-driving'', ``object detection'', ``3D object detection'', ``BEV detection'', ``sensor fusion'', ``multimodal fusion'', ``multi-sensor'', ``LiDAR'', ``radar'', ``camera'', ``point cloud'', ``cooperative perception'', ``V2X'', ``vehicle-to-everything'', ``vision language model'', ``VLM'', ``large language model'', ``LLM'', ``open-vocabulary''. We applied these queries to titles, abstracts, and keywords (when supported by the database) and refined terms iteratively based on the terminology used in the most relevant retrieved papers.

\item \textbf{Inclusion and Exclusion Criteria:}
We included studies that met the following criteria: (i) focus on object detection for autonomous driving or closely related on-road perception settings; (ii) involve camera-based, LiDAR-based, radar-based, or multi-sensor fusion detection pipelines; (iii) present a clear technical contribution, comparative evaluation, or substantial survey perspective; (iv) are peer-reviewed publications or influential preprints with clear experimental evidence. We excluded works that: (i) do not address object detection or perception (e.g., papers focused only on planning/control); (ii) are not grounded in autonomous driving scenarios (e.g., generic robotics detection without driving relevance); (iii) provide insufficient technical detail (e.g., short abstracts, posters without methods); (iv) are duplicates, non-English, or redundant versions of the same work (in which case the most complete version was retained).

\item \textbf{Selection Summary:}
Figure \ref{fig:prisma} presents a PRISMA-style flow diagram summarizing our literature search, screening, eligibility assessment, and final inclusion. In total, we identified 1,850 records across databases. After removing 490 duplicates, 1360 papers remained for title and abstract screening, where 520 were excluded due to irrelevance to AV object detection and fusion. We assessed 840 full-text papers for eligibility and excluded 530 works for reasons such as lack of AV relevance, insufficient methodological detail, or redundancy. To ensure quality and relevance within the scope of the survey, we focused primarily on peer-reviewed articles published in top-tier conferences and reputable journals. The final corpus included 310 papers, which were then categorized into camera-based detection, LiDAR-based detection, radar and adverse-weather perception, multi-sensor fusion, cooperative perception, datasets/benchmarks, and emerging VLM/LLM-based directions.

\end{itemize}

\subsection{Review of the existing survey papers}
To identify the most relevant prior surveys, we conducted a systematic search across major academic databases to collect the most influential and relevant survey papers on Autonomous Vehicles (AVs) published between 2018 and 2025. Our search included keywords such as “autonomous vehicle,” “AV sensors,” “object detection,” “computer vision,” “Vision Language Models (VLMs),” “Large Language Models (LLMs),” “Cooperative Perception (CP),” “communication,” “multimodal fusion,” “deep learning,” “3D object detection,” “LiDAR,” “point cloud,” “camera,” “Radar,” “autonomous driving,” and “sensor fusion.” We then critically analyzed the identified papers based on their titles, abstracts, methodologies, findings, and citation counts. This approach allowed us to select survey papers that align closely with our focus on comprehensive AV research, including perception, decision-making, and multimodal sensor fusion. Table \ref{Table: Literature Review} summarizes the strengths and weaknesses of each selected survey, highlighting their limitations and open research challenges in the field.

\begin{sidewaystable*}[htbp]
\centering
\caption{Summary of existing survey papers on different aspects of autonomous vehicles.}
\vspace{1mm}
\label{Table: Literature Review}
\resizebox{\textwidth}{!}{
\setlength{\tabcolsep}{7pt}
\begin{tabular}{lllllllll}
\toprule
\textbf{Year}  &  \textbf{Survey} & \textbf{Content Included} & \textbf{Potential Gaps}  & \textbf{Application Domain} \\
\midrule

2025      
&\cite{wang2025developments}
&\tabitem 3D object detection methods that are based on LiDAR or fusion approaches.
&\tabitem Only considers 3D detection methods, missing 2D-based detection approaches.
&\makebox[0pt][l]{$\square$}\raisebox{.15ex}{\hspace{0.1em}$\checkmark$}
3D Object Detection\\
&&\tabitem Discusses practical deployment challenges in real-world AV driving applications.
&\tabitem Lack of detailed review of AV sensor hardware and configurations in AVs.
&\\[0.8mm]
&&\tabitem Provides research directions for temporal, predictive, and cooperative perceptions.
&\tabitem Does not explore LLMs, VLMs, and foundation models in autonomous driving.
& \\[2mm]

\myrowcolour
2025       
&\cite{wang2025survey}
&\tabitem Explores camera–LiDAR fusion algorithms for object detection tasks in AVs.
&\tabitem Most of the recent 2D and 3D object detection approaches in AVs are missing. 
&\makebox[0pt][l]{$\square$}\raisebox{.15ex}{\hspace{0.1em}$\checkmark$}
Sensor Fusion Detection\\
\myrowcolour
&&\tabitem Reviews sensor characteristics (camera, LiDAR, Radar) used in sensor fusion.
&\tabitem Lack of systematic categorization of AV datasets into meaningful taxonomies.
&\\[0.8mm]
\myrowcolour
&&\tabitem Compares 11 benchmark datasets (KITTI, nuScenes, Waymo, etc.) in AV driving.
&\tabitem Does not review and compare LLMs, VLMs, and foundation models in AVs.
&\\[2mm]

2025
&\cite{wang2025review}
&\tabitem Compares state-of-the-art point-based, voxel-based, and point-voxel methods.
&\tabitem Does not include sensor-level analysis and reviews of AV sensing hardware.
&\makebox[0pt][l]{$\square$}\raisebox{.15ex}{\hspace{0.1em}$\checkmark$}
3D Object Detection\\
&&\tabitem Reviews 7 popular datasets (e.g., KITTI, nuScenes, Waymo, ApolloScape).
&\tabitem There is no discussion of LLMs, VLMs, and foundation models used in AVs.
&\\[0.8mm]
&&\tabitem Highlights multi-modal fusion architectures (early, deep, and late fusion) in AVs.
&\tabitem Only mentions a limited number of AV datasets, most of them not considered.
&\\[2mm]

\myrowcolour
2024     
&\cite{song2024robustness} 
&\tabitem Summarizes 3D object detection methods in AVs categorized by sensor modality.
&\tabitem LLM-based, VLM-based, and foundation models are not addressed at all.
&\makebox[0pt][l]{$\square$}\raisebox{.15ex}{\hspace{0.1em}$\checkmark$} 
3D Object Detection\\
\myrowcolour
&&\tabitem Reviews robustness-aware methods in camera, LiDAR, and multi-modal detection.
&\tabitem Focuses solely on 3D object detection, leaving out 2D detection techniques.
&\\[0.8mm]
\myrowcolour
&&\tabitem Discusses transformer-based, graph-based, and point-voxel models in 3D detection.
&\tabitem Does not consider cooperative perception frameworks and datasets in AVs.
&\\[2mm]

2024      
&\cite{alaba2024emerging}
&\tabitem Provides a review of multimodal fusion techniques for 3D object detection in AVs.
&\tabitem Only considers a few number of AV datasets, CP and 2D datasets are missing.
&\makebox[0pt][l]{$\square$}\raisebox{.15ex}{\hspace{0.1em}$\checkmark$}
Multimodal 3D Detection\\
&&\tabitem Discusses sensor modalities (camera, LiDAR, Radar, thermal, ultrasonic, RGB-D).
&\tabitem does not address LLM-based or VLM-based models for autonomous vehicles.
&\\[0.8mm]
&&\tabitem Includes a comparative analysis across 15 widely used autonomous vehicle datasets.
&\tabitem Lack of 2D detection information, while focuses exclusively on 3D detection.
&\\[2mm]

\myrowcolour
2023      
&\cite{zou2023object} 
&\tabitem Provides a thorough historical and technical evolution of object detection methods.
&\tabitem Focuses on general-purpose object detection, not specialize in AV scenarios.
&\makebox[0pt][l]{$\square$}\raisebox{.15ex}{\hspace{0.1em}$\checkmark$} 
AV Object Detection\\
\myrowcolour 
&&\tabitem Analyzes detection challenges such as multiscale variation and object rotation.
&\tabitem Lacks exploration of Vision–Language Models or LLM‑based perception tasks.
&\\[0.8mm]
\myrowcolour
&&\tabitem Covers 8 object detection datasets and metrics (PASCAL VOC, MS-COCO, etc).
&\tabitem Does not include sensor-specific detection methods and hardware details. 
&\\[2mm]

2023      
&\cite{ma20233d}
&\tabitem Reviews image-based 3D object detection methods applied within AV scenarios.
&\tabitem There is no discussion of CP (V2V, V2X) frameworks, protocols, or datasets.
&\makebox[0pt][l]{$\square$}\raisebox{.15ex}{\hspace{0.1em}$\checkmark$}
3D Object Detection\\
&&\tabitem Introduces two taxonomies of the state-of-the-art for organizing existing methods.
&\tabitem Does not touch on LLMs, VLMs, or foundation models applied in AV driving. 
&\\[0.8mm]
&&\tabitem Reviews 11 AV datasets (KITTI, nuScenes, Waymo) and discusses their details.
&\tabitem Does not review LiDAR-only, Radar-only, or multi-modal fusion methods.
&\\[2mm]

\myrowcolour
2022    
&\cite{qian20223d}
&\tabitem Offers a taxonomy of 3D object detection methods, categorized by input modality.
&\tabitem Does not consider VLM, LLM, CP, and foundation models in object detection.
&\makebox[0pt][l]{$\square$}\raisebox{.15ex}{\hspace{0.1em}$\checkmark$}
3D Object Detection\\
\myrowcolour
&&\tabitem Presents a discussion on sensors (passive vs. active) in AV object detection.
&\tabitem Although a couple of datasets are reviewed, many AV datasets are missing.
&\\[0.8mm] 
\myrowcolour 
&&\tabitem Discusses diversity, annotation quality, and benchmarking details of 7 AV datasets.
&\tabitem Focuses mostly on 3D detection, with little to no coverage of 2D detection.
&\\[2mm]

2021     
&\cite{cui2021deep}     
&\tabitem Provides an extensive review of deep learning methods for camera-LiDAR fusion.
&\tabitem 2D object detection and traditional sensor reviews are not included in paper. 
&\makebox[0pt][l]{$\square$}\raisebox{.15ex}{\hspace{0.1em}$\checkmark$}
AV Multimodal Fusion\\
&&\tabitem Offers benchmarking and model comparisons for KITTI dataset for various tasks.
&\tabitem Does not cover cooperative perception frameworks and datasets such as V2V.
&\\[0.8mm]
&&\tabitem Identifies open challenges and future directions in autonomous vehicle systems.
&\tabitem Lacks a detailed review of AV datasets including scopes and characteristics.
&\\[2mm]

\myrowcolour
2020      
&\cite{feng2020deep}          
&\tabitem Offers a review of deep multi-modal object detection and semantic segmentation.
&\tabitem Despite the validity of the discussed approaches, they are quite outdated.
&\makebox[0pt][l]{$\square$}\raisebox{.15ex}{\hspace{0.1em}$\checkmark$}
Detection \& Segmentation\\
\myrowcolour
&&\tabitem Proposes a structured methodology for multi-modal fusion and fusion operations. 
&\tabitem Lack of detailed and technical information about AV datasets and sensors.
&\\[0.8mm] 
\myrowcolour 
&&\tabitem Discusses key open challenges such as sensor misalignment and label quality.
&\tabitem There is no information about CP, LLMs, VLMs, foundation models in AVs.
&\\[2mm]

2019     
&\cite{guo2019safe}          
&\tabitem Provides an overview for assessing drivability in autonomous vehicle driving.
&\tabitem Despite the validity of the discussed methods, they are quite outdated.
&\makebox[0pt][l]{$\square$}\raisebox{.15ex}{\hspace{0.1em}$\checkmark$}
AV Object Detection\\
&&\tabitem Reviews a taxonomy of 54 public driving datasets categorized by characteristics.
&\tabitem focuses heavily on dataset and risk assessment, but not sensors and hardware.
&\\[0.8mm] 
&&\tabitem Highlights open challenges in metric design and joint attention modeling in AVs.
&\tabitem Does not review or analyze object detection methods (2D, 3D, fusion-based).
&\\[2mm]

\midrule     
&&\tabitem \textbf{Reviews state-of-the-art 2D (Camera) object detection techniques in AVs.}
&&\\[0.8mm]
&&\tabitem \textbf{Reviews state-of-the-art 3D (LiDAR) object detection approaches in AVs.}
&&\\[0.8mm]
&&\tabitem \textbf{Surveys Camera–LiDAR fusion techniques for object detection in AVs.}
&&\\[0.8mm]
\textbf{Our Survey}
&&\tabitem \textbf{Provides an extensive overview of LLMs, VLMs, and ViTs in AV systems.}
&&\\[0.8mm]
&&\tabitem \textbf{Explores LLM/VLM-based multimodal methods for object detection in AVs.}
& \tabitem \textbf{N/A}
&\makebox[0pt][l]{$\square$}\raisebox{.15ex}{\hspace{0.1em}$\checkmark$}
\textbf{AV Object Detection}\\[0.8mm]
&&\tabitem \textbf{Analysis CP frameworks (V2V, V2I, V2X, I2I), as well as their datasets.}
&&\\[0.8mm]
&&\tabitem \textbf{Discusses AV sensor types, architectures, technologies, and modalities.}
&&\\[0.8mm]
&&\tabitem \textbf{Reviews AV dataset information, types, characteristics, and applications.}
&&\\[0.8mm]
&&\tabitem \textbf{Outlines existing simulation platforms used for AV dataset generation.}
&&\\[0.8mm]
&&\tabitem \textbf{Highlights challenges and open problems for each discussed topic in AVs.}
&&\\[0.8mm]

\bottomrule
 \end{tabular}}
\end{sidewaystable*}

In 2025, three recent survey papers \cite{wang2025developments, wang2025survey, wang2025review} provide a comprehensive review of advancements in 3D object detection and sensor fusion technologies for AVs. The first paper, titled “Developments in 3D Object Detection for Autonomous Driving: A Review”, presents an overview of 3D object detection methods using different sensor modalities such as camera, LiDAR, and multi-sensor fusion approaches. It categorizes detection architectures and discusses transformer-based models and real-time deployment challenges. However, it lacks a detailed review of AV datasets, sensor configurations, and hardware platforms, as well as the integration of foundation models, VLMs, or LLMs for advanced multimodal perception in AVs. The second paper, titled “A Survey of the Multi-Sensor Fusion Object Detection Task in Autonomous Driving”, focuses on fusion-based object detection architectures and compares several fusion strategies such as early, middle, and late fusion. It also discusses Transformer-based fusion techniques and popular datasets like KITTI and nuScenes. Nonetheless, the study fails to address recent advancements in 2D and 3D detection methods, Multimodal models (VLMs and LLMs), and a comprehensive review of AV datasets and sensor technologies. The third paper, titled “A Review of 3D Object Detection Based on Autonomous Driving”, investigates point-based, voxel-based, and projection-based 3D detection methods with their benchmarking. While it provides an architectural taxonomy, it does not review CP frameworks, omits deep analyses of sensors and datasets, and lacks exploration of large-scale language or vision-language models for AV perception.

In 2024, two review papers \cite{song2024robustness, alaba2024emerging} focus on breakthroughs in 3D object detection and the latest trends in multimodal fusion for autonomous driving perception tasks. The first paper, titled “Robustness-Aware 3D Object Detection in Autonomous Driving: A Review and Outlook”, provides an in-depth analysis of robustness-aware techniques across camera-only, LiDAR-only, and multi-sensor detection methods. It discusses adversarial vulnerabilities and surveys benchmark datasets such as KITTI and nuScenes. However, the paper does not explore all the existing datasets, CP frameworks, AV sensors, or VLMs and VLMs that are becoming increasingly relevant in AV systems. The second paper, titled “Emerging Trends in Autonomous Vehicle Perception: Multimodal Fusion for 3D Object Detection”, reviews fusion techniques across early, intermediate, and late fusion stages and highlights various sensor modalities, including LiDAR, cameras, Radar, and thermal imaging. However, not only are many existing AV 2D detection methods and significant AV datasets missing, but this paper also does not include any CP frameworks, VLM methods, or segmentation approaches.

In 2023, paper \cite{zou2023object} presents a broad historical survey of object detection methods over the past two decades, spanning traditional techniques like HOG and DPM to modern deep learning models such as YOLO, SSD, and Transformer-based detectors. However, this survey focuses on general-purpose object detection and lacks attention to autonomous vehicle-specific challenges such as sensor integration, cooperative perception, or communication-aware detection tasks. Similarly, the paper \cite{ma20233d} provides the first dedicated review of image-based 3D object detection for autonomous driving. Although this paper covers existing AV detection methods and some popular datasets, a few key techniques (multi-modal fusion) in AV technologies as well as sensor modalities such as LiDAR and Radar are missing. Additionally, it does not explore all the potential methods including cooperative and language-driven frameworks. A comprehensive review of 3D object detection methods for autonomous driving is provided in \cite{qian20223d}. This paper emphasizes various detection architectures categorized by input modalities, such as LiDAR-based, image-based, and multi-modal approaches, and includes detailed comparisons across major datasets like KITTI, nuScenes, and Waymo. It also highlights sensor characteristics, fusion strategies, and detection challenges under real-world driving conditions. Although this paper covers a wide range of detection algorithms and sensor fusion techniques, it lacks discussion on CP frameworks, a structured dataset taxonomy based on their types, and leaves out emerging topics such as VLMs, foundation models, and LLM-based perception frameworks in AV systems.

Paper \cite{cui2021deep} presents a exhaustive review of deep learning-based fusion techniques for combining image and point cloud data in autonomous driving. It categorizes fusion methods based on various stages (early, middle, and late), highlights sensor configurations, and the various possible fusion architectures in object detection tasks. Moreover, the paper provides valuable insights into the strengths and limitations of fusion strategies across multiple benchmarks. However, it does not cover recent advancements in multimodal foundation models, a detailed analysis of large-scale AV datasets, and segmentation frameworks. In 2020, a paper titled “Deep Multi-Modal Object Detection and Semantic Segmentation for Autonomous Driving” \cite{feng2020deep} provides a detailed review of multi-modal methods for object detection and semantic segmentation tasks in AVs. It highlights the advantages of combining data from different sensors such as cameras, LiDAR, and Radar, and presents an overview of datasets, fusion strategies. However, the paper does not sufficiently address recent advances in sensors modalities, all the existing datasets, foundation models, vision-language approaches, and cooperative driving. \ph{In 2019, the survey paper \cite{guo2019safe} provided a comprehensive review of object detection methods based on deep learning, focusing on the evolution of CNN-based models and benchmark datasets. It sheds light on both one-stage and two-stage detectors, as well as key improvements in detection performance and computational efficiency. It also reviews a taxonomy of 54 public driving datasets categorized by their types and characteristics. Nevertheless, multi-modal fusion techniques for object detection in autonomous driving are not addressed. Additionally, the integration of language-driven models for scene understanding and the discussion of large-scale AV-specific datasets and sensor analysis remain incomplete.}

\section{AV Technology and Sensor Specifications}
\label{sec: AV Sensors}

This section provides an overview of existing AV technologies, with a particular focus on recent advancements in AV sensors and device specifications relevant to autonomous driving. In the domain of autonomous vehicles, sensors play a critical role in ensuring robust perception and accurate environment understanding. Modern computer vision systems heavily rely on a wide range of sensors such as cameras, LiDAR, Radar, or ultrasonic devices to precisely detect and monitor surrounding objects. The subsequent subsection will particularly discuss the details of various sensors deployed in AVs for efficient perception, environment understanding, and robust decision-making.

\subsection{AVs' Sensor Types}

AV sensors are generally categorized into various groups based on various factors such as functionality, range, resolution, energy, design, and purpose. In this survey, we have categorized them into two major types based on their operational principles: proprioceptive and exteroceptive sensors. Proprioceptive sensors, also known as internal state sensors, capture the vehicle’s internal conditions such as velocity, acceleration, angular rate, tire pressure, or voltage levels. Common examples of sensors used in this category include IMUs (Inertial Measurement Units), GPSs (Global Positioning Systems), GNSSs (Global Navigation Satellite Systems), gyroscopes, and magnetometers \cite{hu2025security}. On the other hand, exteroceptive sensors, often referred to as external state sensors, collect data from the vehicle’s surrounding environment. This category includes cameras, LiDAR (Light Detection and Ranging), RADAR (Radio Detection and Ranging), and ultrasonic sensors, which are commonly used for perception tasks such as object detection, obstacle avoidance, and environmental mapping \cite{mohammadi2025economic, rasoulpour2026multi}.

\begin{figure}[h]
    \centering
    \centerline{\includegraphics[width=0.9\textwidth]{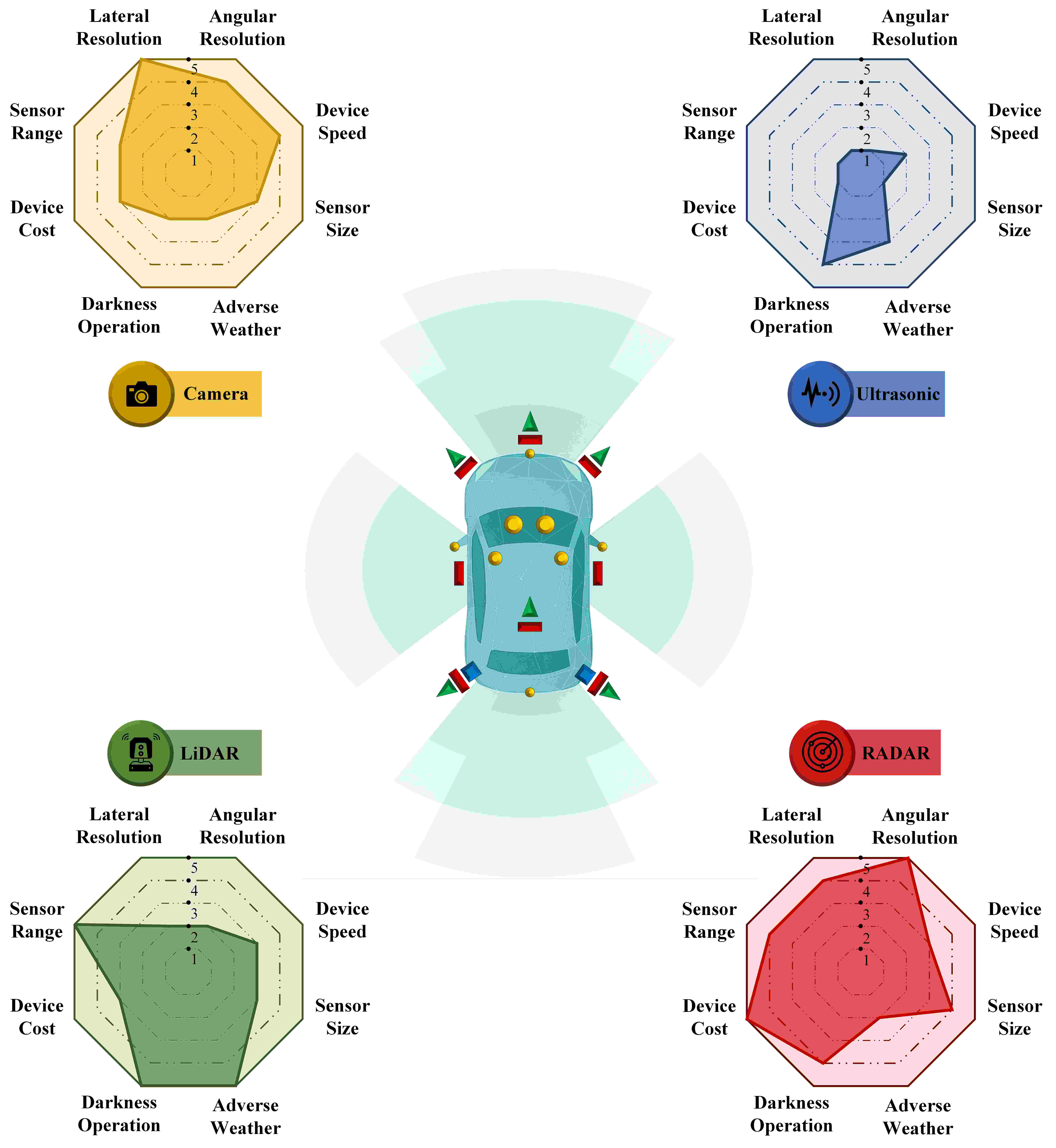}}
    \caption{Overview of major sensors used in AVs based on their types and perception performance. Sensor performance is evaluated on a scale from 1 to 5, where 1=Very Low, 2=Low, 3=Medium, 4=High, and 5=Very High}
    \label{fig: Sensor}
\end{figure}

In addition to this classification, sensors can be grouped as passive or active based on their signal emission behavior. Passive sensors rely only on surrounding signals without emitting any energy, such as RGB and Thermal cameras. However, active sensors emit energy into the environment and then measure the reflected signals to determine the output, such as LiDAR and Radar \cite{du2025advancements}. Figure \ref{fig: Sensor} provides an overall perspective and statistical analysis of various sensors used in AV. This figure visually presents the performance analysis of each AV sensor across multiple parameters, providing a clear assessment of the strengths and limitations of each sensor type. For convenience in comparison, each sensor attribute is ranked on a scale from 1 (Very Low) to 5 (Very High). Additionally, the possible positions of each sensor on the vehicle are illustrated using their corresponding colors (Yellow for the camera, Blue for the ultrasonic, Green for LiDAR, and Red for Radar).

To further elaborate on the characteristics of each AV sensor, the following Table \ref{Table: Sensor specification} provides a detailed specification on the four major exteroceptive sensors commonly integrated in autonomous vehicles: camera, ultrasonic, LiDAR, and Radar sensors. A thorough understanding of the specifications and capabilities of each sensor is crucial for more efficient and reliable  autonomous driving, enabling accurate perception and timely decision-making. Given that the primary focus of this survey is on the vision and perception aspects of AV systems, we exclude proprioceptive sensors and concentrate exclusively on exteroceptive sensing technologies. Each subsection will highlight the operational principles, technological foundations, and role of these sensors in AV systems, as well as their potential strengths and weaknesses.

\begin{table*}[htbp]
\caption{Detailed specifications of the most significant AV sensors employed in autonomous driving systems.}
\vspace{-0.3cm}
\label{Table: Sensor specification}
\begin{center}
\resizebox{\textwidth}{!}{
\setlength{\tabcolsep}{7pt}
\begin{tabular}{llccccp{3in}c}
\toprule
\textbf{Criteria}  & \textbf{Metric}  &  \textbf{Camera} & \textbf{Ultrasonic} & \textbf{LiDAR}  & \textbf{Radar}  & \textbf{Additional Note}\\
\midrule

\textbf{Lateral Resolution}        &Pixels        &Very High     &Very Low	          &High          &Low     &\makebox[0pt][l]{$\square$}\raisebox{.15ex}{\hspace{0.1em}$\checkmark$}
Indicates the level of detail the sensor captures across the image plane (X and Y).\\[7mm]

\myrowcolour
\textbf{Angular Resolution}       &Degrees ($^\circ$)         &High         &Very Low	            &Very High	        &Low       &\makebox[0pt][l]{$\square$}\raisebox{.15ex}{\hspace{0.1em}$\checkmark$}
Determines the minimum angle between two objects that a sensor can measure.\\[7mm]

\textbf{Device Speed}          &Latency          &High          &Low      &Moderate         &Moderate       &\makebox[0pt][l]{$\square$}\raisebox{.15ex}{\hspace{0.1em}$\checkmark$}
Defines how quickly sensor can capture and update the data from environment.\\[7mm]

\myrowcolour
\textbf{Sensor Dimension}        &Size ($\mathrm{cm}^3$)      &Medium    &Very Small	    &Large    &Medium   
&\makebox[0pt][l]{$\square$}\raisebox{.15ex}{\hspace{0.1em}$\checkmark$}
 Impacts sensor feasibility, aerodynamics, and design constraints in AVs.\\[7mm]

\textbf{Day Operation}        &Visibility      &Excellent      &Good      &Very Good	    &Good  
&\makebox[0pt][l]{$\square$}\raisebox{.15ex}{\hspace{0.1em}$\checkmark$}
Measures sensor performance under the bright lighting or sunlight glare.\\[7mm]

\myrowcolour
\textbf{Night Operation}    &Visibility      &Poor      &Good      &Very Good	    &Very Good  
&\makebox[0pt][l]{$\square$}\raisebox{.15ex}{\hspace{0.1em}$\checkmark$}
Indicates the sensor effectiveness in low-light or entirely darkness situations.\\[7mm] 

\textbf{Adverse Weather}         &Performance     &Poor      &Moderate      &Poor    &Very Good
     &\makebox[0pt][l]{$\square$}\raisebox{.15ex}{\hspace{0.1em}$\checkmark$}
Reflects the sensor robustness in heavy rain, fog, snow, or dust conditions.\\[7mm]

\myrowcolour
\textbf{Device Cost}        &USD (\$)    &Low      &Very Low	      &Very High	    &Moderate    
&\makebox[0pt][l]{$\square$}\raisebox{.15ex}{\hspace{0.1em}$\checkmark$}
One of the major factor for commercial deployment and large-scale AV integration.\\[7mm]

\textbf{Sensor Range}        &Meters ($m$)     &Moderate       &Very Short      &Long     &Very Long   
&\makebox[0pt][l]{$\square$}\raisebox{.15ex}{\hspace{0.1em}$\checkmark$}
Specifies the maximum range at which detection performance remains reliable.\\[7mm]

\myrowcolour
\textbf{Color Detection}    &RGB      &Yes       &No      &No    &No  
&\makebox[0pt][l]{$\square$}\raisebox{.15ex}{\hspace{0.1em}$\checkmark$}
A valuable ability for applications such as traffic sign and light recognition. \\[7mm]

\textbf{Distance Measurement}          &Accuracy (m)     &Moderate     &Low    &Very High   &High       
&\makebox[0pt][l]{$\square$}\raisebox{.15ex}{\hspace{0.1em}$\checkmark$}
Measures object distance which is crucial for collision avoidance and navigation.\\[7mm]

\myrowcolour
\textbf{Velocity Measurement}       &Speed ($m/s$)     &No      &No      &Limited    &Very High    
&\makebox[0pt][l]{$\square$}\raisebox{.15ex}{\hspace{0.1em}$\checkmark$}
Enables to estimate the speed of nearby moving objects surrounding the vehicle.\\[7mm]

\textbf{Depth Measurement}        &Resolution (mm)     &Low       &No      &Very High	    &Moderate   
&\makebox[0pt][l]{$\square$}\raisebox{.15ex}{\hspace{0.1em}$\checkmark$}
Describes the sensor’s capability to capture distance from the sensor to surfaces.\\[7mm]

\myrowcolour
\textbf{Field of View}    &Degrees ($^\circ$)      &90–120$^\circ$      &30$^\circ$      &360$^\circ$    &60-120$^\circ$ 
&\makebox[0pt][l]{$\square$}\raisebox{.15ex}{\hspace{0.1em}$\checkmark$}
Specifies the angular coverage the sensor can observe in a single frame.\\[7mm]

\textbf{Frame Rate}         &Rate ($Hz$)     &30–60 $Hz$      &$<$10 $Hz$      &10-20 $Hz$    &10-30 $Hz$ 
&\makebox[0pt][l]{$\square$}\raisebox{.15ex}{\hspace{0.1em}$\checkmark$}
Determines how frequently the sensor captures data over the time intervals.\\[7mm]

\myrowcolour
\textbf{Object Classification}        &Capability      &Very High	      &No      &Moderate    &Low  
&\makebox[0pt][l]{$\square$}\raisebox{.15ex}{\hspace{0.1em}$\checkmark$}
Indicates whether the sensor supports object classification task.\\[7mm]

\textbf{Noise Sensitivity}        &Level     &High      &Moderate      &High    &Low   
&\makebox[0pt][l]{$\square$}\raisebox{.15ex}{\hspace{0.1em}$\checkmark$}
Reflects sensor’s vulnerability to signal interference and false alarms.\\[7mm]

\myrowcolour
\textbf{Power Consumption}    &Watts ($w$)     &Low      &Very Low	      &High	    &Moderate   
&\makebox[0pt][l]{$\square$}\raisebox{.15ex}{\hspace{0.1em}$\checkmark$}
Indicates the amount of electrical power required to operate the sensor.\\[5mm]

\textbf{Computational Load}          &Load     &High      &Very Low	      &High    &Low      &\makebox[0pt][l]{$\square$}\raisebox{.15ex}{\hspace{0.1em}$\checkmark$}
Shows the required processing resources to handle sensor data.\\[7mm]

\myrowcolour
\textbf{Calibration Complexity}        &Level     &High      &Low      &High    &Low    
&\makebox[0pt][l]{$\square$}\raisebox{.15ex}{\hspace{0.1em}$\checkmark$}
Defines the complexity of setup and ongoing calibration tasks of the sensor.\\[7mm]

\textbf{Fusion Compatibility}        &Level     &High      &Low      &High    &Moderate  
&\makebox[0pt][l]{$\square$}\raisebox{.15ex}{\hspace{0.1em}$\checkmark$}
Shows how easily the sensor integrates into sensor fusion frameworks.\\[7mm]

\myrowcolour
\textbf{Maintenance Frequency}    &Frequency     &Low      &Moderate      &Moderate    &Low   
&\makebox[0pt][l]{$\square$}\raisebox{.15ex}{\hspace{0.1em}$\checkmark$}
Suggests how often the sensor needs cleaning, recalibration, or maintenance.\\[7mm]

\textbf{Installation Complexity}        &Level     &Low      &Very Low	      &High    &Low  
&\makebox[0pt][l]{$\square$}\raisebox{.15ex}{\hspace{0.1em}$\checkmark$}
Reflects effort and constraints involved in mounting the sensors on vehicle.\\[7mm]

\myrowcolour
\textbf{Sensor Durability}        &Lifespan     &Moderate      &Low      &High    &Very High 
&\makebox[0pt][l]{$\square$}\raisebox{.15ex}{\hspace{0.1em}$\checkmark$}
Indicates sensor robustness to vibration, impact, and environmental degradation.\\[7mm]
                            
\hline

\rotatebox{0}{\parbox[t][8cm][c]{2cm}{\textbf{Additional} \newline \textbf{Detail}}}   

&\rotatebox{270}{\parbox{9cm}{
\makebox[0pt][l]{$\square$}\raisebox{.15ex}{\hspace{0.1em}$\checkmark$}
Specifies how each parameter is measured (e.g., degrees, pixels, meters), enabling objective sensor comparison across various key factors such as resolution, range, cost, environmental robustness, and performance in AV perception tasks.}} 

&\rotatebox{270}{\parbox{9cm}{
\makebox[0pt][l]{$\square$}\raisebox{.15ex}{\hspace{0.1em}$\checkmark$}
Captures high-resolution 2D (RGB or grayscale) image data, enabling object classification, traffic sign recognition, and lane detection. It should be noted that the performance degrades in low-light, glare, fog, or rain.}}

&\rotatebox{270}{\parbox{9cm}{
\makebox[0pt][l]{$\square$}\raisebox{.15ex}{\hspace{0.1em}$\checkmark$}
Uses high-frequency sound waves to measure distance to nearby objects. It is suitable for parking assistance, blind spot detection, and short-range obstacle avoidance. Extremely low-cost and compact, but limited to close-range sensing.}}

&\rotatebox{270}{\parbox{9cm}{
\makebox[0pt][l]{$\square$}\raisebox{.15ex}{\hspace{0.1em}$\checkmark$}
Emits laser pulses to generate high-density 3D point clouds, enabling precise depth estimation, surface reconstruction, and shape recognition. It offers 360° scanning, however, it is expensive and performance degrades under adverse weather.}}

&\rotatebox{270}{\parbox{9cm}{
\makebox[0pt][l]{$\square$}\raisebox{.15ex}{\hspace{0.1em}$\checkmark$}
Transmits radio waves to detect object range, angle, and velocity. It is a valuable tool for tracking moving targets and operating in darkness, rain, or fog. However, it suffers from low spatial resolution and difficulty distinguishing object shapes.}}

&\rotatebox{0}{\parbox[t][9cm][c]{7.53cm}{
\makebox[0pt][l]{$\square$}\raisebox{.15ex}{\hspace{0.1em}$\checkmark$}
Here are the most potential applications of each sensor that align with their characteristics:\\[3mm]
\textbf{Camera:} Object Recognition, Pedestrian Detection, Traffic Sign Detection.\\[3mm] 
\textbf{Ultrasonic:} Parking assistance, blind Spot Detection, Obstacle Avoidance.\\[3mm]
\textbf{LiDAR:} Environment Mapping, Obstacle Detection, Distance Estimation.\\[3mm]
\textbf{Radar:} Lane Change Assistance, Moving Object Detection, Speed Measurement.\\[3mm]

}}\\[92mm]
\bottomrule
\end{tabular}}
\end{center}
\end{table*}


\subsection{Camera Sensor}

Autonomous driving systems rely mainly on cameras as the primary sensors, since they provide crucial visual data for a wide range of perception tasks. There are four major types of cameras used in AVs: monocular, stereo, depth, and thermal cameras. Monocular cameras use a single lens to capture 2D images of the environment, much like standard digital cameras. Their ability to produce high-resolution images, along with their low-cost design, makes them a favorable choice for many perception tasks in AVs. Additionally, they enable the extraction of detailed texture patterns, rich color information, and overall visual appearance. In modern systems, monocular vision is often paired with deep learning models such as CNNs and ViTs to infer depth information and extract semantic content from source images. However, since these systems rely on learned context and visual patterns to estimate depth, they require robust training data and well-designed algorithms to perform reliably under varying driving conditions. Monocular cameras suffer from several limitations, including vulnerability to light and weather conditions, as well as an inability to provide depth information. Despite these drawbacks, they are widely used in lane detection, traffic sign recognition, and object classification.

The stereo camera is another vision system widely used in autonomous driving. This sensor imitates human binocular vision through two horizontally spaced lenses to calculate depth measurements by analyzing disparities between the left and right image views. This disparity information enables the AV to create real-time 3D spatial reconstructions of the surrounding environment. However, precise stereo performance depends heavily on suitable calibration to ensure accurate alignment and consistent depth measurement. Moreover, stereo cameras demand significant computational resources, especially when operating in low-light conditions, on low-texture surfaces, and at long-range distances. By considering the range and lighting conditions, stereo systems are particularly effective for various tasks such as object detection, obstacle avoidance, and motion tracking \cite{wu2025fusion}. On the other hand, depth cameras provide distance information directly by using active sensing methods such as structured light or time-of-flight technology. Unlike stereo cameras that rely on visual disparity, depth cameras emit signals (infrared patterns or light pulses) and measure the time of the reflected signal to estimate depth. This direct measurement makes depth cameras more robust in low-texture environments and scenes with limited visual features. Since they need to operate at short ranges from objects, they are often used in low-speed driving and indoor navigation scenarios. Lower performance under strong sunlight, limited effective range, and sensitivity to reflective or transparent surfaces are some of the key limitations of depth camera sensors. Despite these drawbacks, they remain a valuable sensor in AV systems, especially when fused with other sensors.

The thermal camera is the last type of vision sensor commonly used in autonomous driving systems. It captures infrared radiation emitted by objects rather than relying on visible light for the perception of the surrounding environment. This makes them a reliable choice in situations where standard cameras often fail to perform very well, such as nighttime driving or conditions involving fog, heavy rain, smoke, or direct sunlight. They play a crucial role in enhancing safety by enabling pedestrian and wildlife detection, as well as emergency braking in low-visibility conditions. However, thermal cameras usually offer lower spatial resolution and are incapable of capturing color or fine texture details. They also require specialized processing algorithms, which make them more computationally expensive. Despite these limitations, thermal cameras provide an important complementary sensing modality, enhancing the overall robustness and safety of AV perception systems \cite{wu2024joint}. In summary, AVs can benefit from various camera sensors, as each camera type offers unique strengths and limitations depending on environmental conditions, object types, and specific driving tasks. Table \ref{Table: cameras} highlights the key advantages and disadvantages of each camera type, providing a comprehensive comparison to assist researchers in choosing the most appropriate camera for particular objectives.

\begin{table*}[htbp]
\centering
\caption{Summary of the advantages and disadvantages of the various types of camera sensors.}
\vspace{-0.1cm}
\label{Table: cameras}
\resizebox{\textwidth}{!}{
\setlength{\tabcolsep}{7pt}
\begin{tabular}{lllllllll}
\toprule
\textbf{Sensors}   & \textbf{Advantages} & \textbf{Disadvantages}  \\
\midrule

&\tabitem Low-cost solution for visual scene understanding.
&\tabitem Lacks natural depth perception for 3D estimation.

\\[1mm]
\textbf{Monocular} 
&\tabitem Lightweight and easy to integrate with AV system.
&\tabitem Performance notably degrades in low-light conditions.
\\[1mm]

\textbf{Cameras}
&\tabitem Captures rich texture and detailed color information.
&\tabitem Highly sensitive to weather and illumination changes.
\\[1mm]

&\tabitem Supports classification and object detection tasks.
&\tabitem Relies heavily on the high-quality training dataset.
\\[3mm]

\myrowcolour 
&\tabitem Provides real-time depth info from image disparity.
&\tabitem High computational cost for disparity matching.

\\[1mm]
\myrowcolour
\textbf{Stereo} 
&\tabitem Enables 3D reconstruction of nearby environment.
&\tabitem Struggles on texture-less and uniform surfaces.
\\[1mm]

\myrowcolour
\textbf{Cameras}
&\tabitem Passive sensing without emitting external signals.
&\tabitem Requires precise calibration between camera lenses.
\\[1mm]

\myrowcolour
&\tabitem Captures color and rich texture visual information.
&\tabitem Performance drops in adverse lighting conditions.
\\[3mm]

&\tabitem Directly captures depth from a single viewpoint.
&\tabitem Limited distance range compared to stereo or LiDAR.

\\[1mm]
\textbf{Depth} 
&\tabitem Offers precise short-range 3D scene understanding.
&\tabitem Frequently struggles in direct and bright sunlight.
\\[1mm]

\textbf{Cameras}
&\tabitem Compact and highly suitable for close-range tasks.
&\tabitem Performance affected by reflective surface materials.
\\[1mm]

&\tabitem Provides texture, color, and appearance information.
&\tabitem Less effective in consistent long-distance depth sensing.
\\[3mm]

\myrowcolour 
&\tabitem Detects heat signatures in complete darkness.
&\tabitem Expensive compared to standard vision cameras.

\\[1mm]
\myrowcolour
\textbf{Thermal} 
&\tabitem Performs very well in fog and glare environments.
&\tabitem Lacks color and texture visual information.
\\[1mm]

\myrowcolour
\textbf{Cameras}
&\tabitem Enhances safety in low-visibility conditions.
&\tabitem Lower spatial resolution than RGB cameras.
\\[1mm]

\myrowcolour
&\tabitem Useful for night-time pedestrian detection.
&\tabitem Difficult to interpret without contextual data.
\\[3mm]

\bottomrule
 \end{tabular}}
\end{table*}

\subsection{Ultrasonic Sensor}
Ultrasonic sensors are commonly used in modern vehicles for short-range proximity sensing. They measure distances by emitting high-frequency sound waves and analyzing their echoes. These sensors operate by generating sound waves ranging from 20 kHz to several hundred kHz (depending on the sensor type), which exceed the human hearing range. Then, they measure the distance between the vehicle and the object by calculating the time it takes for the emitted sound wave to reflect off the object and return as an echo. Generally, ultrasonic sensors can be categorized into one of the following three types: short-range, mid-range, and long-range. Short-range sensors ($<$2 m) provide high-resolution distance measurements, which are suitable for parking assist and blind-spot detection at low speeds. Mid-range sensors (2–5 m) offer moderate resolution and distance range, making them ideal for obstacle avoidance and curb detection. Lastly, long-range sensors ($>$5 m) extend detection capabilities for high-speed collision warning and closing-speed estimation on freeways or highways. 

Nowadays, the use of ultrasonic sensors in modern vehicles has increased, as they serve as essential components for Advanced Driver-Assistance Systems (ADAS). These sensors are typically embedded in the front and rear bumper corners for parking assistance and blind-spot detection, or in the vehicle’s front grille or upper bumper to enable high-speed collision warnings. Ultrasonic sensors offer several advantages for AVs by enhancing safety and reliability through close-range environmental monitoring. Additionally, unlike optical sensors, they perform very well in complete darkness, which makes them useful for night-time driving. They are also highly cost-effective compared to alternatives like LiDAR and Radar. However, their limited detection range makes them less effective not only for detecting long-distance objects but also in high-speed situations. Also, their performance can be decreased in the presence of intense traffic noise due to interference with high-frequency sound waves. Furthermore, environmental factors such as rain, humidity, and temperature fluctuations can affect their accuracy. Furthermore, they are highly sensitive to environmental factors such as rain, humidity, and temperature. As a result, ultrasonic sensors are well-suited for short-range tasks such as parking assistance and nearby object detection at low speeds.

\subsection{LiDAR Sensor}
LiDAR sensors are among the most fundamental components in modern AV perception systems. These devices calculate distances by transmitting laser pulses and measuring the time it takes for the reflected signals to return. As a result, the system is able to construct 3D point cloud data of the environment for accurate spatial awareness and understanding of the surroundings. Unlike 1D (one-dimensional) sensors that capture only distance information (x-axis) or 2D (two-dimensional) sensors that measure angular position (y-axis), 3D LiDAR sensors measure vertical coordinate (z-axis) and provide detailed three-dimensional representations of the environment. LiDAR technology is typically categorized into three main types: mechanical LiDAR, solid-state LiDAR, and hybrid LiDAR. Mechanical LiDAR, including rotating mirror and spinning sensor subtypes, uses rotating components to provide 360$^\circ$ scanning with high-resolution data. Fully Solid-State LiDAR includes two main types, where Flash LiDAR captures the entire scene with a single light pulse, and Optical Phased Array (OPA) uses electronic beam steering for rapid scanning \cite{li2023multi}. Hybrid LiDAR, including Micro-Electro-Mechanical Systems (MEMS) and Risley Prism subtypes, incorporates minimal moving parts like micro-mirrors or prisms to steer laser beams. Table \ref{Table: LiDAR} presents the major pros and cons of each LiDAR type to provide valuable information for researchers to choose the most suitable system for their specific application.

Autonomous driving systems significantly benefit from LiDAR technologies due to the several advantages they offer in AVs. First of all, they are able to provide exact distance measurements not only during the day but also at night. Additionally, they perform very accurately and reliably in adverse weather conditions, such as heavy rain, snow, and fog. They are able to capture detailed geometric information about the surroundings, which enhances object recognition and environmental mapping. In contrast, LiDAR sensors face several limitations in autonomous driving, including high cost and mechanical vulnerability. Moreover, the data generated by LiDAR requires substantial processing power and storage capacity due to their high volume and complexity.

\begin{table*}[htbp]
\centering
\caption{Summary of the advantages and disadvantages of the various types of LiDAR sensors.}
\vspace{-0.1cm}
\label{Table: LiDAR}
\resizebox{\textwidth}{!}{
\setlength{\tabcolsep}{7pt}
\begin{tabular}{lllllllll}
\toprule
\textbf{Sensors}   & \textbf{Advantages} & \textbf{Disadvantages}  \\
\midrule

&\tabitem Provides full 360$^\circ$ environmental coverage.
&\tabitem Limited vehicle integration due to bulky design.

\\[1mm]
\textbf{Mechanical} 
&\tabitem Offers high spatial scanning resolution.
&\tabitem High manufacturing and maintenance costs.
\\[1mm]

\textbf{LiDAR}
&\tabitem well-established in the automotive industry.
&\tabitem Mechanical components degrade over time.
\\[1mm]

&\tabitem Enables accurate long-range object detection.
&\tabitem Requires complex alignment for AV integration.
\\[3mm]

\myrowcolour 
&\tabitem High reliability due to non-mechanical parts.
&\tabitem Lower resolution than other LiDAR types.

\\[1mm]
\myrowcolour
\textbf{Solid-state} 
&\tabitem Offers high durability over the long term.
&\tabitem Limited detection range compared to mechanical.
\\[1mm]

\myrowcolour
\textbf{LiDAR}
&\tabitem Compact structure simplifies system design.
&\tabitem Limited availability in current market offerings.
\\[1mm]

\myrowcolour
&\tabitem Resistant to shock and vibration damage.
&\tabitem Less effective in complex terrain environments.
\\[3mm]

&\tabitem Compact size enables easier sensor placement.
&\tabitem Limited field of view coverage range.

\\[1mm]
\textbf{Hybrid} 
&\tabitem Lower cost than mechanical LiDAR systems.
&\tabitem The Calibration process may be complicated.
\\[1mm]

\textbf{LiDAR}
&\tabitem Faster scanning speed than other types.
&\tabitem Performance may degrade in bright sunlight.
\\[1mm]

&\tabitem Easier integration into modern AV designs.
&\tabitem Calibration process can be technically complex.
\\[3mm]

\bottomrule
 \end{tabular}}
\end{table*}

\subsection{Radar Sensor}
RADAR (Radio Detection and Ranging) sensors are another primary technology widely used in modern vehicles. These sensors emit radio waves to detect the distance and speed of surrounding objects. They measure object speed using the Doppler effect, which analyzes frequency shifts caused by moving objects. The two main types of Radar sensors broadly used in AVs are Impulse Radar and FMCW (Frequency-Modulated Continuous Wave) Radar. The Impulse transmits discrete radio pulses, while FMCW continuously emits modulated signals to provide superior depth perception and range. In AVs, Radar sensors are typically mounted on the front and rear bumpers to ensure full environmental awareness. Based on their range capabilities, they serve as an integral part of ADAS in modern vehicles. Short-Range Radar (SRR) is particularly effective for detecting objects during low-speed maneuvers, while Medium-Range Radar (MRR) and Long-Range Radar (LRR) are ideal for monitoring vehicles and obstacles at higher speeds on highways and busy roads \cite{wu2025fusion}. It is worth mentioning that the integration of Radar sensors with other sensor data has made it a rising trend in automotive safety.

Table \ref{Table: RADAR} outlines the key features of each Radar type, as well as providing major pros and cons of Radar sensors commonly used in autonomous vehicles to guide researchers in choosing the most appropriate Radar for particular objectives. Using Radar sensors in AV applications provides multiple benefits. These sensors operate reliably under adverse weather conditions and low-light environments, where optical sensors often struggle. These sensors also provide precise speed and distance measurements, which are highly crucial for collision avoidance and adaptive cruise control systems. Furthermore, Radar technologies are significantly more cost-effective than alternatives such as LiDAR sensors. However, the spatial resolution of Radar systems is lower than LiDAR and modern camera-based systems. The reflection of radar signals from metal objects such as road signs and guardrails sometimes results in incorrect detection. Despite these limitations, the reliable performance and cost-effectiveness of Radar sensors continue to support their important role in autonomous driving systems.

\begin{table}[H]
\centering
\caption{Summary of the key features, advantages, and disadvantages of various Radar sensor types.}
\vspace{-0.1cm}
\label{Table: RADAR}
\resizebox{\textwidth}{!}{
\setlength{\tabcolsep}{7pt}
\begin{tabular}{lccllllll}
\toprule
\textbf{Sensors}  & \textbf{Range} & \textbf{FOV}  & \textbf{Advantages} & \textbf{Disadvantages}  \\
\midrule

& 
&
&\tabitem Fast response in close range.
&\tabitem Lower resolution than cameras.

\\[1mm]
\textbf{Short-Range} 
& Short
& Wide
&\tabitem Wide-angle object detection.
&\tabitem Limited detection distance.
\\[1mm]

\textbf{Radar}
& $\sim$0.2\,-30\,m 
& $\sim$120$^\circ$
&\tabitem Effective in tight spaces.
&\tabitem Poor performance in heavy rain.
\\[1mm]

& 
&
&\tabitem Low power consumption.
&\tabitem Affected by surface interference.
\\[3mm]

\myrowcolour 
& 
&
&\tabitem Balanced range and FOV.
&\tabitem Not ideal for highway use.

\\[1mm]
\myrowcolour
\textbf{Medium-Range} 
& Medium
& Medium
&\tabitem Good for mid-speed driving.
&\tabitem May misclassify static objects.
\\[1mm]

\myrowcolour
\textbf{Radar}
& $\sim$30\,-80\,m 
& $\sim$60$^\circ$-90$^\circ$
&\tabitem Reliable in poor weather.
&\tabitem Resolution not very high.
\\[1mm]

\myrowcolour
& 
&
&\tabitem Works very well in traffic.
&\tabitem Limited vertical coverage.
\\[3mm]

& 
&
&\tabitem Perfect long-distance tracking.
&\tabitem Cannot detect nearby objects.

\\[1mm]
\textbf{Long-Range} 
&Long 
&Narrow 
&\tabitem Precise velocity measurement.
&\tabitem Expensive and larger size.
\\[1mm]

\textbf{Radar}
&$\sim$80\,-250\,m 
&$\sim$20$^\circ$-30$^\circ$
&\tabitem Robust in adverse weather.
&\tabitem Narrow field of view.
\\[1mm]

& 
&
&\tabitem Crucial for highway driving.
&\tabitem Complex to integrate.
\\[3mm]

\bottomrule
 \end{tabular}}
\end{table}

\subsection{Sensor Fusion}
Sensor fusion plays a foundational role in AVs' perception by combining data from multiple sensor types to enhance the accuracy and reliability of environmental understanding. It leverages the strengths of each sensor to address the limitations inherent in individual sensing modalities. AV systems require three essential factors for effective sensor fusion integration. First of all, a detailed understanding of each sensor's characteristics is necessary for determining its strengths, limitations, and optimal role within the fusion framework. Secondly, appropriate sensor calibration is required to ensure optimal performance by accurately aligning sensor placements and synchronizing timing data across all devices. Last but not least, selecting and evaluating the right sensor fusion algorithm is highly critical for effectively combining multi-sensor data to achieve robust and accurate environmental perception. Consequently, an effective sensor fusion system for AVs is built upon three key principles: comprehensive sensor understanding, precise calibration, and careful algorithm selection \cite{du2025advancements}. In addition to onboard sensors, modern AV stacks increasingly treat High-Definition (HD) maps as a redundant contextual prior that supports both perception and localization. HD maps provide high-precision representations of lane geometry and road topology (e.g., merges and intersections), drivable-space boundaries, and semantic landmarks such as crosswalks, stop lines, and traffic lights/signs. When integrated into a fusion framework, this prior can improve robustness under partial observability (e.g., occlusions), degraded sensing (e.g., heavy rain, fog, or glare), and GNSS-challenging environments by constraining perception to map-consistent hypotheses and providing an additional source of system-level redundancy \cite{elghazaly2023high}.

One of the most common and well-known applications of sensor fusion in AVs is object detection, where data from multiple sensors is combined to identify and classify objects in the driving environment, such as vehicles, pedestrians, and road signs. AVs require fast and precise object detection capability for more informed driving decisions to enhance road safety and reduce the risk of harm to people inside and outside the vehicle. Radar sensors perform reliably in adverse weather conditions and provide precise velocity measurements. Although they are valuable devices for dynamic environment monitoring, they face significant challenges in spatial resolution and object classification. In other words, Radar cannot accurately recognize and distinguish objects that are either non-moving or have complex shapes \cite{yao2025exploring}. Radar data is often combined with inputs from cameras and LiDAR to overcome these limitations. The camera sensor provides color and texture information, and LiDAR offers high-resolution 3D spatial data. Therefore, Radar improves detection performance in low-visibility conditions, while the camera enhances object recognition and spatial accuracy. Integrating multiple sensors results in a more robust object detection system, supporting safer and more reliable autonomous navigation.

Although Camera–LiDAR–Radar fusion mitigates many sensing weaknesses under challenging conditions (e.g., occlusions, severe weather, or GNSS-degraded localization), incorporating contextual priors such as HD maps can further improve perception. Accordingly, recent fusion pipelines treat HD maps as explicit, structured inputs that effectively serve as an additional modality, rather than using them only as planning context. In practice, map content is commonly encoded either as rasterized BEV tensors (e.g., lane/drivable masks, boundary distance fields, semantic layers) or as vectorized polylines/graphs (lane centerlines, boundaries, connectivity), and then fused with camera/LiDAR/radar features through attention-based cross-modal alignment, feature gating, or shared BEV encoders. This design is particularly effective in challenging scenes where sensor evidence is incomplete (e.g., long-range occlusions, glare, heavy precipitation), because the map prior can regularize the perception output toward topologically consistent and road-feasible hypotheses, improving stability of both object-level reasoning and road-structure understanding. Recent studies have demonstrated that explicitly integrating structured priors can strengthen robustness and coherence in road perception (e.g., PriorFusion) \cite{tang2025priorfusion}, while transformer-based frameworks for online vectorized map modeling provide practical mechanisms to represent and fuse HD-map elements in an end-to-end manner (e.g., MapTRv2 \cite{liao2025maptrv2}, GeMap \cite{zhang2024online}). Moreover, map fusion can be extended with map memory by retrieving a local map patch from a maintained historical/global map and combining it with current observations to mitigate short-term sensing failures (e.g., HRMapNet \cite{zhang2024enhancing}). From a deployment perspective, these benefits must be balanced against map freshness and localization/map-alignment sensitivity, motivating hybrid strategies that couple offline priors with online map perception and continual updates to maintain reliability in real-world operation.

Table \ref{Table: Fusion} presents a comprehensive overview of sensor fusion methods categorized by fusion level: low-level, mid-level, and high-level fusion. In low-level fusion, raw data from multiple sensors (e.g., pixel values, point clouds, or raw Radar signals) are combined to provide rich sensory information. This level requires precise calibration and high computational resources due to the complexity of processing raw sensor data. Mid-level fusion integrates intermediate features from individual sensors (e.g., edges, depth maps, or motion cues) to offer a balanced trade-off between accuracy and efficiency. High-level fusion combines the final decisions from separate processing modules, such as detected objects or classified regions. This approach makes the system more robust and modular, but may result in the loss of some finer details in the process.

Table \ref{Table: Fusion} presents a comprehensive overview of fusion strategies organized by processing level: low-level, mid-level, and high-level fusion. In low-level fusion, raw measurements from multiple sensors (e.g., pixel streams, point clouds, or raw Radar returns) are combined to maximize information content, but this setting demands strict calibration/synchronization and typically incurs high computational and bandwidth costs. Mid-level fusion merges intermediate representations (e.g., learned feature maps, depth/BEV features, motion cues), offering a practical accuracy--efficiency trade-off and enabling flexible cross-modal alignment. High-level fusion aggregates decisions from separate modules (e.g., detections, tracks, semantic regions), improving modularity and fault tolerance at the expense of some fine-grained detail. 
Importantly, this taxonomy can be extended beyond physical sensors: contextual priors such as HD maps can be fused at different levels depending on their representation---as rasterized BEV layers or vector polylines at low/mid levels, or as map-consistency constraints and hypothesis validation at high levels---thereby improving robustness when onboard sensing is partial or degraded.

\begin{table*}[htbp]
\centering
\caption{Summary of sensor fusion methods categorized by processing level, highlighting their sensor pairs, complementarity, applications, as well as their pros and cons.}
\vspace{-0.1cm}
\label{Table: Fusion}
\resizebox{\textwidth}{!}{
\setlength{\tabcolsep}{7pt}
\begin{tabular}{lllllllll}
\toprule
\textbf{Fusion Level}  & \textbf{Sensor Pair} & \textbf{Complementarity}  & \textbf{Use Case} & \textbf{Pros (\cmark) / Cons (\xmark)}  \\
\midrule

& Lidar + Camera 
& Depth + Texture 
&\tabitem 3D object detection
&\cmark \hspace{1mm} Detailed spatial information
\\[2mm]
\textbf{Low-Level} 
& Lidar + Radar
& Dense + Motion
&\tabitem Obstacle mapping
&\cmark \hspace{1mm} Rich sensory information
\\[2mm]
\textbf{Fusion}
& RGB + Thermal 
& Visual + Thermal 
&\tabitem Low-light vision
&\xmark \hspace{1mm} Requires precise calibration
\\[2mm]
& Stereo Cameras
& Dual Perspective
&\tabitem Thermal detection
&\xmark \hspace{1mm} High processing cost
\\[3mm]

\myrowcolour 
& Camera + Lidar
& Feature + Geometry
&\tabitem Obstacle detection
&\cmark \hspace{1mm} Accuracy-cost efficiency
\\[2mm]
\myrowcolour
\textbf{Mid-Level} 
& Camera + Radar
& Visual + Velocity
&\tabitem Lane assistance
&\cmark \hspace{1mm} Simpler than low-level
\\[2mm]
\myrowcolour
\textbf{Fusion}
& Camera + Ultrasonic
& Context + Proximity 
&\tabitem Scene classification
&\xmark \hspace{1mm} Feature alignment needed
\\[2mm]
\myrowcolour
& Lidar + Ultrasonic
& Shape + Proximity
&\tabitem Traffic sign recognition
&\xmark \hspace{1mm} Moderate latency
\\[3mm]

& All 4 Sensor Types
& Safety Redundancy
&\tabitem Trajectory planning
&\cmark \hspace{1mm} Easy to implement
\\[2mm]
\textbf{High-Level} 
& Camera+Lidar+Radar
& Decision Aggregation
&\tabitem Localization support
&\cmark \hspace{1mm} Robust to missing data
\\[2mm]
\textbf{Fusion}
& V2X + Camera/Lidar
& Communication
&\tabitem Object tracking
&\xmark \hspace{1mm} Inconsistency in decisions
\\[2mm]
& IMU + GPS
& Sensor + Localization
&\tabitem Mission planning
&\xmark \hspace{1mm} Low spatial precision
\\[3mm]

\bottomrule
 \end{tabular}}
\end{table*}

In summary, multi-sensor fusion technology offers significant advantages for AV driving systems by enhancing environmental perception and operational safety. Fusing data from various sensors, including camera, LiDAR, RADAR, and ultrasonic sensors, improves the system’s accuracy and reliability across diverse conditions. Using multiple sensors makes the system more secure and resilient by addressing individual sensor limitations and providing backup in case of sensor failure. On the other hand, there are several challenges in multi-sensor fusion that should be considered for achieving efficient autonomous driving systems. First, the system results in a high computational cost to process large volumes of heterogeneous data in real time, as well as maintaining synchronization and alignment. Additionally, the system's performance is highly dependent on accurate calibration and precise spatial and temporal alignment. Lastly, achieving effective sensor fusion requires developing more advanced algorithms to integrate and interpret data from different sensor types, which increases the overall system complexity. Despite these challenges, sensor fusion remains essential for building scalable and dependable autonomous driving systems that prioritize safety.

\subsection{Challenge, Discussion, and Future Directions}
Despite significant advancements in AV sensor technologies and fusion frameworks, several technical and practical challenges remain in real-world deployment. A primary challenge is integrating and calibrating heterogeneous sensors with different resolutions, response times, and data formats. Real-time processing of massive, multimodal sensory data imposes a high computational cost, affecting decision-making in time-critical scenarios. Additionally, individual sensor limitations, such as the inability of cameras to perform well in poor lighting or LiDAR’s performance degradation in adverse weather, directly affect overall system robustness. Sensor fusion, although powerful, introduces complexity in synchronization, feature alignment, and temporal consistency, making the system prone to drift or failure if not finely tuned. Furthermore, complex scenarios such as uncommon road geometries or unexpected human behavior are often not well represented in existing training datasets, which limits the ability of perception algorithms across unseen conditions.

Looking ahead, several novel and forward-thinking directions show great promise for the next generation of AV perception systems. One emerging idea is context-aware sensor fusion, where the vehicle intelligently adjusts which sensors to prioritize based on the current driving scene—for example, relying more on radar during heavy fog or emphasizing camera input in urban environments. Another exciting direction is using neuromorphic event-based cameras, which mimic the human retina and offer faster response times with lower data bandwidth. We also see potential in collaborative learning across AV fleets, where vehicles share knowledge about sensor calibration, unusual objects, or environmental conditions without transferring raw data to preserve privacy while improving the cooperative perception system. Lastly, using self-supervised cross-sensor learning, where one sensor helps teach another without labeled data, could dramatically reduce the cost and time needed to train reliable perception systems. These forward-looking ideas offer fresh possibilities for building smarter, safer, and more adaptable autonomous vehicles in the years ahead.

\section{Autonomous Vehicle Simulators}
\label{sec: AVs Simulator}
Autonomous vehicle simulators play a central role in modern AV research by enabling scalable data generation, controllable scenario design, and repeatable evaluation under diverse operational conditions that are difficult, costly, or unsafe to capture in real traffic. Simulators provide configurable sensor suites (e.g., multi-camera systems, LiDAR, radar, GNSS/IMU), precise ground-truth annotations (2D/3D boxes, segmentation, depth, trajectories), and programmable agents and traffic logic for studying perception, prediction, planning, and end-to-end driving policies. They are also essential for stress-testing robustness across weather conditions, illumination, and rare corner cases, and for supporting sim-to-real pipelines through domain randomization, synthetic-to-real adaptation, and closed-loop evaluation. In this section, we review representative AV simulators and summarize their capabilities and limitations, organizing the discussion into open-source and non-open-source platforms with attention to key practical dimensions such as realism and rendering quality, sensor fidelity, scenario coverage, annotation support, API extensibility, computational requirements, and integration with commonly used autonomy stacks.

\subsection{Open-Source Simulators}
Open-source simulators have become a cornerstone of autonomous driving research because they provide accessible, extensible, and reproducible testbeds for perception and decision-making under controlled yet diverse conditions. In addition to lowering the barrier to entry, these platforms enable the community to modify sensor models, traffic logic, map interfaces, and evaluation pipelines, which is particularly valuable for benchmarking object detection and multimodal fusion methods across consistent scenarios. However, open-source tools vary substantially in realism and rendering fidelity, sensor simulation accuracy, HD map support, and the ease of integration with autonomy stacks and learning frameworks. Table \ref{Table:opensource_sims} provides a consolidated comparison of representative open-source simulation platforms, summarizing their operating systems, underlying engines, programming interfaces, and the availability of key features and sensor modalities relevant to autonomous vehicle perception research.

\begin{table}[H]
\centering
\caption{Comparative overview of representative open-source autonomous driving simulators and their supported features and sensors.}
\vspace{-0.5cm}
\label{Table:opensource_sims}
\begin{center}
\resizebox{\textwidth}{!}{%
  \setlength{\tabcolsep}{4pt}%
  \begin{tabular}{l l c p{2.3cm} p{3cm} p{2.0cm} cc ccc l}
    \toprule
    \multicolumn{6}{c}{} 
      & \multicolumn{2}{c}{\textbf{Features}} 
      & \multicolumn{3}{c}{\textbf{Sensors}} 
      & \multicolumn{1}{c}{} \\[1mm]
    \cmidrule(l){7-8}\cmidrule(l){9-11}
    \textbf{Ref} & \textbf{Simulator} & \textbf{Release Year} & \textbf{OS} & \textbf{Engine} & \textbf{Language}
      & \textbf{HD Map} & \textbf{Depth}
      & \textbf{Camera} & \textbf{LiDAR} & \textbf{Radar}
      & \textbf{Access} \\
    \midrule

    \cite{gulino2023waymax} & Waymax            & 2023 & Windows, Linux, macOS & JAX-based & Python & No  & No  & \xmark & \xmark & \xmark & \href{https://waymo.com/research/waymax/}{Web} \\[5mm]
    
    \myrowcolour
    \cite{ahmed2024systemic} & NVIDIA Drive Sim  & 2021 & Windows, Linux                 & Omniverse           & C++, Python & Yes & Yes & \cmark & \cmark & \cmark & \href{https://developer.nvidia.com/drive/simulation}{Web} \\[5mm]
    
    \cite{rong2020lgsvl} & LGSVL             & 2019 & Windows, Linux        & Unity Engine        & C\#, Python  & Yes & Yes & \cmark & \cmark & \cmark & \href{https://hidetoshi-furukawa.github.io/post/lgsvl-simulator/}{Web} \\[5mm]
    
    \myrowcolour
    \cite{dosovitskiy2017carla} & CARLA             & 2017 & Windows, Linux, macOS & Unreal Engine       & C++, Python  & Yes & Yes & \cmark & \cmark & \cmark & \href{https://carla.org}{Web} \\[5mm]
    
    \cite{shah2017airsim} & AirSim            & 2017 & Windows, Linux, macOS & Unreal Engine       & C\#, Python, C++, Java & No & Yes & \cmark & \cmark & \xmark & \href{https://microsoft.github.io/AirSim/}{Web} \\[5mm]
    
    \myrowcolour
    \cite{feng2022application} & Baidu Apollo      & 2017 & Linux                 & Unity Engine        & C++          & Yes & No  & \cmark & \cmark & \cmark & \href{https://github.com/ApolloAuto/apollo}{Git} \\[5mm]
    
    \cite{smolyakov2018self} & Udacity           & 2016 & Windows, Linux, macOS & Unity Engine        & C\#          & No  & No  & \cmark & \xmark & \xmark & \href{https://github.com/udacity/self-driving-car-sim}{Git} \\[5mm]
    
    \myrowcolour
    \cite{kato2018autoware} & Autoware          & 2015 & Linux                 & Unity Engine        & C++, Python  & Yes & No  & \cmark & \cmark & \xmark & \href{https://autoware.org/awsim-end-to-end-digital-twin-simulation-platform/}{Web} \\[5mm]
    
    \cite{koenig2004design} & Gazebo            & 2002 & Linux, macOS          & Physics Engines & C++, Python & No & Yes & \cmark & \cmark & \cmark & \href{https://gazebosim.org/home}{Web} \\[5mm]
    
    \myrowcolour
    \cite{wymann2000torcs} & TORCS             & 1997 & cross-platform        & Self-developed      & C++          & No  & No  & \xmark & \xmark & \xmark & \href{https://sourceforge.net/projects/torcs/}{Web} \\[1mm]

    \bottomrule
  \end{tabular}%
}
\end{center}
\end{table}

\vspace{-0.5cm}
Waymax \cite{gulino2023waymax} is a data-driven simulation and evaluation library that instantiates driving scenarios from the Waymo Open Motion Dataset, with an explicit focus on behavior research rather than raw-sensor perception. Instead of producing camera/LiDAR streams, it represents agents as bounding boxes (the dataset’s minimal object representation), keeping experiments focused on planning, behavior prediction, and sim-agent interaction without requiring an in-sim perception stack. The library is implemented entirely in JAX to support accelerator execution and “in-graph” computation (e.g., JIT compilation and functional transforms), and it includes built-in behavior metrics (e.g., log divergence, collision, offroad, wrong-way, kinematic infeasibility) alongside vehicle simulation options such as direct state updates or a kinematic bicycle model (with a default 10 Hz step aligned to the dataset).

NVIDIA Drive Sim \cite{ahmed2024systemic} offers an Omniverse-centered workflow for creating rich 3D virtual worlds and synthetic data generation pipelines to support training, testing, and validation in autonomous-vehicle development. The developer page emphasizes integration of NVIDIA components for world generation and reconstruction, specifically Cosmos world foundation models for post-training/data generation and NuRec for neural reconstruction, positioning simulation as part of a broader data engine that can generate or curate multimodal AV training data and accelerate closed-loop development cycles. In this framing, the technical value is less “a single monolithic simulator” and more an ecosystem of tools and APIs intended to scale scenario creation and synthetic data production in support of AV stacks.

LGSVL \cite{rong2020lgsvl} is a unity-based autonomous-vehicle simulator designed to run end-to-end, closed-loop experiments with open-source driving stacks by generating realistic sensor streams from interactive 3D environments. In addition to photorealistic scene rendering, it is commonly used to emulate a configurable sensor suite (e.g., multi-camera setups, LiDAR, radar, GPS/IMU) with synchronized outputs, enabling repeatable perception-and-planning evaluation under controlled conditions. The simulator also supports scenario-driven testing (e.g., scripted traffic behaviors, route definitions, and event triggers) and integrates with HD map and traffic-rule context to better approximate urban driving dynamics. Documentation and project descriptions emphasize support for creating sensor inputs for autonomy stacks, manual driving for debugging, and populating scenes with other traffic participants.

CARLA \cite{dosovitskiy2017carla} is an open-source simulator built to support the development, training, and validation of autonomous driving systems, with an emphasis on controllable experiments and repeatable evaluation. The platform exposes programmatic control over static and dynamic actors and allows systematic variation of environmental conditions and maps, enabling scripted scenario generation rather than purely manual driving. For perception-centric work, CARLA provides a configurable sensor suite (e.g., RGB, depth, semantic segmentation, LiDAR, radar, GNSS, IMU, and event-style detectors such as collision and lane-invasion) and supports deterministic rollouts via synchronous mode with a fixed time step when strict reproducibility is required. Additionally, CARLA offers a client–server API (commonly via Python) that facilitates closed-loop integration with learning and planning pipelines, making large-scale data collection and benchmark-style testing practical.

AirSim \cite{shah2017airsim} is an open-source, Unreal Engine–based simulation platform that can be configured for autonomous driving research by running controllable vehicle scenarios while exposing a programmable interface for closed-loop autonomy stacks. For perception and object-detection studies, it provides a configurable sensor suite, including multi-camera setups and LiDAR, along with standard inertial and navigation sensors (e.g., IMU and GPS), and returns these signals via APIs that support dataset-style capture and online inference. This makes AirSim practical for prototyping and stress-testing camera- and LiDAR-based detection pipelines across varied viewpoints and operating conditions in a 3D environment, without requiring bespoke simulator instrumentation beyond the provided sensor models and configuration system.

Baidu Apollo \cite{feng2022application} is an open-source autonomous driving software platform that accelerates the development, testing, and deployment of autonomous vehicles through a modular, full-stack architecture. The project organizes core autonomy functionality into dedicated modules commonly including perception, localization, and planning, together with supporting toolchains intended to streamline development and integration. The public repository and release notes emphasize ongoing upgrades across these core modules and associated development workflows, positioning Apollo as a reusable baseline stack that is typically coupled with real vehicles or external simulators via middleware/bridges rather than being a simulator itself.

Udacity’s self-driving-car simulator \cite{kato2018autoware} is a Unity-based educational simulator released as a code and asset repository intended to support the Self-Driving Car Nanodegree, primarily for teaching learning-based lane-following and related end-to-end control exercises on curated road courses. The repository documents that the master branch is deprecated, directs users to maintained Unity-version branches, and provides precompiled builds for multiple course “terms,” along with setup notes such as using Git LFS and Unity to load the project assets.

Autoware \cite{kato2018autoware} is an open-source autonomous driving software stack (rather than a simulator) built on ROS, intended to provide an end-to-end pipeline that spans core autonomy functions. The project positions itself as a full-stack system, covering components from localization and perception through route planning and vehicle control, with the goal of enabling broad community contribution and reuse across research and engineering settings. In practice, Autoware is commonly used as the “driving brain” integrated with simulators or real vehicles, where its ROS-based modularity supports substituting datasets, sensors, and algorithmic modules while retaining a consistent autonomy interface and runtime architecture.

Gazebo \cite{koenig2004design} is an open-source robotics simulation framework that provides a high-fidelity simulation loop by combining physics, rendering, and sensor models, with multiple integration “entry points” including a GUI, plugins, and message/service interfaces. Its design emphasizes extensibility: physics backends can be selected at runtime via a plugin abstraction layer (with DART as the default), and sensor behavior is provided by a dedicated sensor library that generates realistic synthetic measurements. In practice, this architecture supports research and engineering workflows where custom robot models and world assets are paired with plugin-based controllers and ROS-facing interfaces for closed-loop testing under repeatable conditions.

TORCS \cite{wymann2000torcs} cross-platform 3D racing simulator that is explicitly described as serving three roles: a conventional racing game, an AI racing game, and a research platform. Its distribution highlights a “sophisticated physical model,” a modular architecture, and practical extensibility, community tracks/cars, add-ons, and “easy to modify” claims grounded in prior scientific and industrial use, making it a common lightweight environment for control and decision-making experiments where the driving task is simplified to racing dynamics rather than full urban autonomy. The project page also foregrounds performance and stability, as well as broad OS support, which partly explains its long-standing use in academic prototyping.

\vspace{-0.5cm}
\subsection{Non-Open-Source Simulators}
In addition to open-source platforms, a number of proprietary simulators are widely used in industry and selected research programs because they offer higher-fidelity rendering, more detailed sensor physics, and mature tooling for scenario authoring and closed-loop validation. These simulators typically provide production-grade assets and maps, scalable traffic and behavior models, and configurable sensor stacks that can emulate camera effects (e.g., motion blur, rolling shutter, lens artifacts), LiDAR characteristics (e.g., beam patterns, intensity and dropout), and radar returns (e.g., clutter and multipath). They also often support structured safety workflows, including scenario catalogs for corner cases, regression testing, and integration with autonomy stacks through standardized interfaces. However, proprietary simulators differ in accessibility, licensing costs, extensibility, and the level of transparency of their sensor and physics models, which can affect reproducibility and fair comparisons across studies. Table \ref{Table:non_opensource_sims} reviews representative non-open-source simulators and summarizes their capabilities, with emphasis on realism, sensor fidelity, scenario coverage, annotation support, API integration, and practical deployment considerations.

\begin{table}[H]
\centering
\caption{Comparative overview of representative non-open-source autonomous driving simulators and their supported features and sensors.}
\vspace{-0.5cm}
\label{Table:non_opensource_sims}
\begin{center}
\resizebox{\textwidth}{!}{%
  \setlength{\tabcolsep}{4pt}%
  \begin{tabular}{l l c p{2cm} p{2.9cm} p{2.2cm} cc ccc l}
    \toprule
    \multicolumn{6}{c}{} 
      & \multicolumn{2}{c}{\textbf{Features}} 
      & \multicolumn{3}{c}{\textbf{Sensors}} 
      & \multicolumn{1}{c}{} \\[1mm]
    \cmidrule(l){7-8}\cmidrule(l){9-11}
    \textbf{Ref} & \textbf{Simulator} & \textbf{Release Year} & \textbf{OS} & \textbf{Engine} & \textbf{Language}
      & \textbf{HD Map} & \textbf{Depth}
      & \textbf{Camera} & \textbf{LiDAR} & \textbf{Radar}
      & \textbf{Website} \\
    \midrule

    \cite{yang2024oasis} & Oasis Sim        & 2024 & Windows, Linux           & Unreal Engine        & C++, \hspace{2cm} Python           & Yes & Yes & \cmark & \cmark & \cmark & \href{https://www.synkrotron.ai/sim.html}{Web}  \\[5mm]
    
    \myrowcolour
    \cite{zhang2020development} & PanoSim          & 2022 & Windows                 & Unity Engine         & C++, \hspace{2cm} Python            & Yes & Yes & \cmark & \cmark & \cmark & \href{https://www.panosim.com/en/}{Web}  \\[5mm]
    
    \cite{hong2021system} & CarMaker        & 2021 & Windows, Linux           & Unigine Engine       & C++, Python C, Matlab & Yes & No & \cmark & \cmark & \cmark & \href{https://www.ipg-automotive.com/solutions/product-portfolio/carmaker}{Web}  \\[5mm]
    
    \myrowcolour
    \cite{faucher2022scaner} & SCANeR    & 2021 & Windows, Linux           & Unreal Engine        & C++, \hspace{2cm} Python            & Yes & No & \cmark & \cmark & \cmark & \href{https://www.avsimulation.com/en/scaner/}{Web}  \\[5mm]
    
    \cite{tideman2013simulation} & PreScan         & 2019 & Windows                 & Self-developed       & C++, \hspace{2cm} Python            & No & Yes & \cmark & \cmark & \cmark & \href{https://www.siemens.com/en-us/products/simcenter/autonomous-vehicle-solutions/prescan/}{Web}  \\[5mm]
    
    \myrowcolour
    \cite{sovani2017simulation} & Ansys  & 2017 & Windows, macOS    & Self-developed       & C++, \hspace{2cm} Python           & Yes & Yes & \cmark & \cmark & \cmark & \href{https://www.ansys.com/products/av-simulation/ansys-avxcelerate-autonomy}{Web}  \\[5mm]
    
    \cite{li2019aads} & CarCraft        & 2017 & N/A                      & Self-developed       & Not \hspace{2cm} Available                   & No & No & \cmark & \cmark & \cmark & \href{https://www.roblox.com/games/8116417963/Carcraft-Vehicle-Simulator}{Web} \\[5mm]
    
    \myrowcolour
    \cite{zhang2024virtual} & Cognata         & 2016 & MS Azure                 & Self-developed       & Not \hspace{2cm} Available                   & Yes & No & \xmark & \xmark & \xmark & \href{https://www.cognata.com/}{Web} \\[5mm]
    
    \cite{benekohal1988carsim} & CarSim          & 1996 & Windows                  & Self-developed       & C++, \hspace{2cm} Matlab            & Yes & Yes & \cmark & \xmark & \xmark & \href{https://www.carsim.com/products/carsim/}{Web} \\[5mm]

    \bottomrule
  \end{tabular}%
}
\end{center}
\end{table}

OASIS Sim \cite{yang2024oasis} is marketed as a high-fidelity autonomous-driving simulation platform with an explicit “vehicle + sensor” modeling focus, aiming to closely replicate real-world conditions and to scale beyond single runs. Synkrotron highlights scenario authoring via a GUI editor, with OpenSCENARIO-compatible definitions for environment conditions, ego tasks, and other-actor behaviors. The CARLA ecosystem documentation describes OASIS Sim as a CARLA-based system with capabilities spanning scenario import/editing, sensor configuration, distributed task management, and diagnosis via logs and rich simulation data. When it comes to AV object detection, CARLA is considered at the core, with an added layer for scenario lifecycle management and engineering workflow.

PanoSim \cite{zhang2020development} is presented as an integrated autonomous-driving simulation and testing system that combines vehicle dynamics, driving environment + traffic modeling, and sensor simulation within a single workflow intended for ADAS and autonomy R\&D. Its product description emphasizes a “one-stop” setup: high-precision dynamics, realistic environment/traffic, physics-based sensor models, and a scenario database for repeatable testing, i.e., the platform is framed around closed-loop evaluation of perception–planning–control under configurable scenarios rather than only offline dataset generation.

CarMaker \cite{hong2021system} is presented as an open integration and testing platform designed for end-to-end development and validation across classical verification stages (MIL, SIL, HIL, and VIL) covering passenger cars and light commercial vehicles. The product positioning is scenario-driven: it supports virtual test scenarios for ADAS/AV functions as well as powertrain and vehicle-dynamics development, and it explicitly includes a “ground truth sensor” as a centralized source of customizable ground-truth information for objects such as road elements and traffic participants. This makes CarMaker easy to place as a development-grade test environment that couples controllable scenarios with ground-truth instrumentation, rather than a dataset generator alone.

SCANeR Studio \cite{faucher2022scaner} is positioned as a comprehensive automotive simulation platform for prototyping, validation, and training of ADAS/AV functions, with an open, modular architecture that scales from conventional driving simulation setups to complex virtual test configurations. The vendor emphasizes configurability across different integration contexts (from traditional driving simulation to advanced virtual tests), implying a workflow where scenario execution, vehicle/traffic/world components, and external tool coupling can be assembled to match the intended verification stage. SCANeR naturally fits as a development-grade, scenario-based simulation environment for exercising ADAS/AV stacks under controlled, repeatable virtual conditions.

Simcenter PreScan \cite{tideman2013simulation} is marketed as a physics-based simulation platform for virtual verification of ADAS and autonomous vehicles, with particular emphasis on systematic scenario coverage and shortened verification loops. Siemens explicitly positions it as a development-and-test environment spanning common verification stages (e.g., MiL/SiL/DiL/HiL/ViL) and highlights advanced sensor simulation, flexible traffic/world modeling, and automated execution (including Monte Carlo studies) for large-scale scenario exploration. PreScan also advertises support for industry standards such as OpenDRIVE and OpenSCENARIO, and Siemens has publicly described added support for OpenSCENARIO 2.0 via a data-model API.

Ansys AVxcelerate Autonomy \cite{sovani2017simulation} is positioned as an end-to-end, safety-driven validation toolchain for L2+/L3 (and higher) ADAS/AD development, with an emphasis on scaling scenario-based testing through virtualization and cloud execution. The product description highlights a model-based systems engineering (MBSE) orientation and a cloud-native, modular workflow that combines statistical analysis with large-scale simulation, explicitly framing it as a way to automate scenario exploration and continuously validate safety-relevant software/system behavior. On the standards side, Ansys notes ASAM standards support and calls out full ASAM OpenSCENARIO 1.3 support (2025 R2), alongside analysis functions, such as sensitivity and reliability analysis, intended to support safety validation and sign-off discussions rather than one-off demos.

CarCraft \cite{li2019aads} is a Roblox-hosted vehicle experience centered on building and upgrading vehicles from parts, and it reads much more like a gameplay sandbox than an autonomy or robotics simulator. The public game listing emphasizes a large customization space, e.g., “over 200 unlockable parts and vehicles,” which suggests the core loop is progression through unlocking components and experimenting with configurations, not controlled scenario replay, sensor modeling, or ground-truth export.

Cognata \cite{zhang2024virtual} presents a simulation-first autonomy offering where OneSim is the training, testing, and validation foundation, and the platform supports closed-loop execution with common autonomy tooling in the loop (MATLAB/Simulink models, ROS components, or proprietary stacks), exposed via packaged interfaces and a real-time SDK with a RESTful API for configuration and control. The company materials also make concrete claims about sensor modeling, explicitly naming radar, LiDAR, camera, and GPS models in the closed-loop workflow and describing sensor simulation features, such as multi-sensor viewers, sensor-fusion simulation, and configurable effects like weather and lighting.

CarSim \cite{benekohal1988carsim} is a vehicle-dynamics simulation package targeted at passenger vehicles and light-duty trucks, marketed primarily as a high-fidelity tool for analyzing handling, stability, ride, and performance characteristics in a repeatable virtual setting. The vendor frames it as a validated engineering platform (citing decades of real-world validation) and emphasizes its use for vehicle-dynamics studies and active-control development, including work tied to active safety/ADAS. In practice, CarSim is best described as a physics-centric “truth model” for vehicle motion and response, useful when the research question depends on credible chassis/tire/powertrain dynamics rather than rich 3D world rendering or photorealistic sensing.

\vspace{0.5cm}
\section{Autonomous Vehicle Datasets}
\label{sec: AVs Dataset}
\begin{figure}[H]
    \centering
    \centerline{\includegraphics[width=0.9\textwidth]{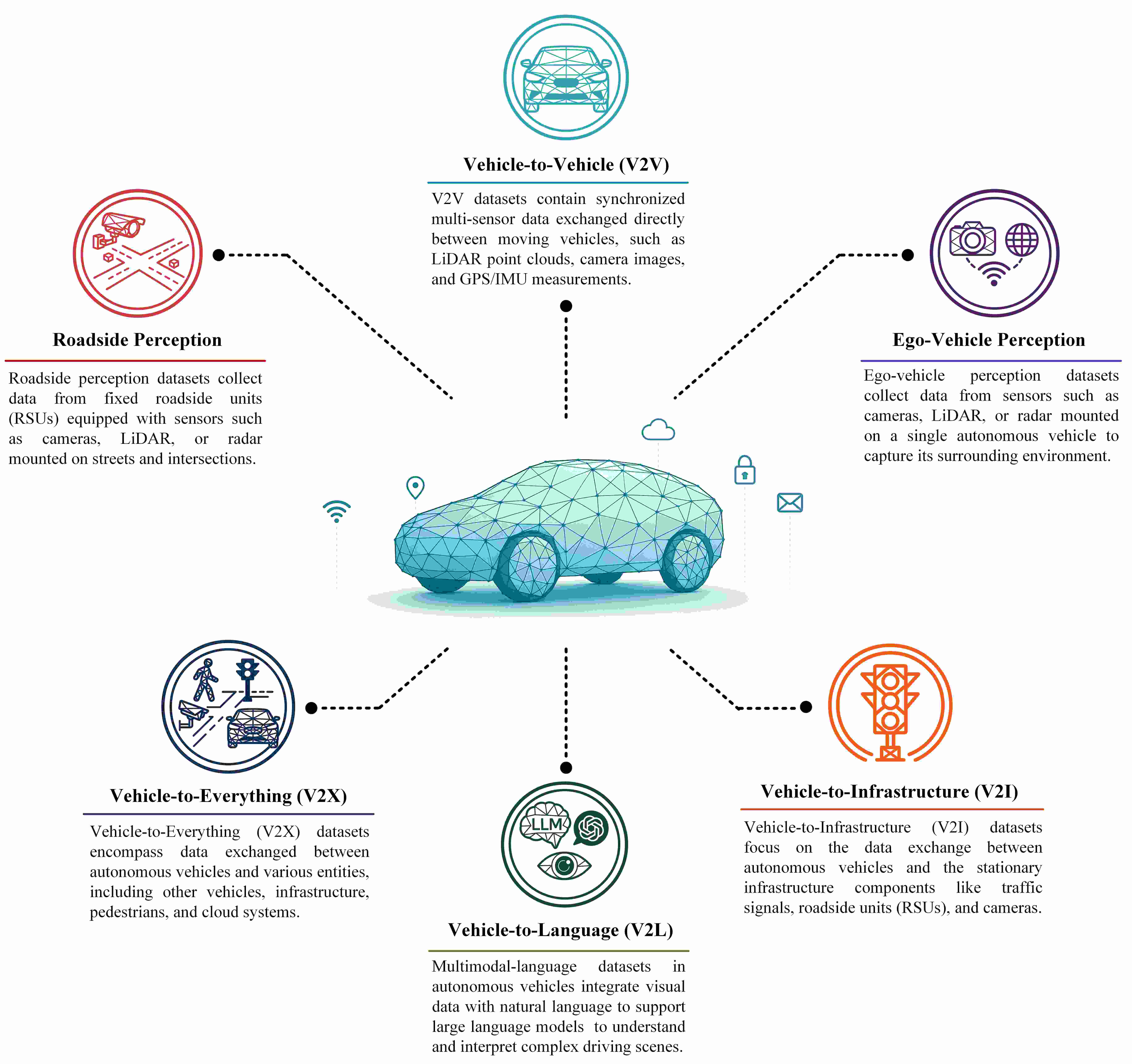}}
    \caption{Overview of major AV datasets based on their specifications and applications.}
    \label{fig: Dataset}
\end{figure}

\vspace{-0.4cm}
In the realm of autonomous vehicle research, the accessibility and diversity of high-quality datasets are fundamental to advancing perception systems, decision-making models, and overall driving safety. This section presents a comprehensive review of existing AV-related datasets, emphasizing their context, characteristics, applications, and strengths and limitations. These datasets differ significantly in terms of collection methods, sensor modalities, data granularity, and annotation depth, which can directly influence their effectiveness for various research objectives such as object detection, tracking, behavior prediction, and scene understanding. Recognizing these differences is critical for researchers seeking to identify suitable benchmarks for developing and evaluating AV algorithms. To facilitate a clear and systematic overview, we have categorized the existing AV datasets into seven major categories based on their specifications and purposes. Every AV dataset can be reasonably classified into one of the following types: Ego-Vehicle Perception, Roadside Perception, Vehicle-to-Language (V2L) Datasets, Vehicle-to-Vehicle (V2V), Vehicle-to-Infrastructure (V2I), Vehicle-to-Everything (V2X) and Infrastructure-to-Infrastructure (I2I). Each category represents a distinct communication and perception framework within the AV domain, shaped by specific design constraints and research objectives. Figure \ref{fig: Dataset} provides a comprehensive insight into the overview of various dataset categories used in AVs.

Ego-Vehicle Perception datasets are collected directly from sensors mounted on the AV itself, capturing the vehicle's first-person view of its surroundings. Roadside Perception datasets originate from fixed infrastructure, such as roadside cameras or LiDAR units, offering complementary, third-party observations of traffic environments. Vehicle-to-Language (V2L) Datasets combine visual inputs with natural language descriptions or queries to enable semantic understanding and reasoning within AV systems. Vehicle-to-Vehicle (V2V) datasets capture communication and interaction data exchanged between vehicles to enhance cooperative perception and coordination \cite{xiang2023multi}. Vehicle-to-Infrastructure (V2I) datasets focus on data shared between vehicles and static infrastructure elements, such as traffic lights or roadside units, to improve navigation and safety. Vehicle-to-Everything (V2X) encompasses broader connectivity scenarios, integrating V2V, V2I, and interactions with other road users, such as pedestrians and cyclists. Finally, Infrastructure-to-Infrastructure (I2I) datasets involve the communication and synchronization of data between multiple infrastructure nodes to create a cohesive traffic management framework.

The subsequent subsections will delve further into each AV dataset category, assessing key features such as the number of samples, data modalities, labeling schemes, and the specific methodologies used for data acquisition and annotation. Through a detailed analysis of these critical aspects, this work provides a comprehensive, structured resource for researchers and practitioners, enabling informed decisions when selecting datasets for training, testing, or benchmarking AV models. Ultimately, this collection contributes to the broader landscape of autonomous driving research by supporting reproducibility, robustness, and innovation in the development of intelligent transportation systems.

\subsection{Ego-Vehicle Perception Datasets}
Ego-vehicle perception datasets are based on sensor data collected directly from the autonomous vehicle itself, such as cameras, LiDARs, Radars, or GPS modules. These datasets capture the vehicle’s perspective as it navigates dynamic environments, and are crucial for developing core AV capabilities like object detection, semantic segmentation, motion planning, and trajectory prediction. Popular examples include KITTI, nuScenes, and Waymo Open Dataset, which enable benchmarking of perception and decision-making models based on real-time sensory input. Table \ref{Table: Ego-vehicle} provides a comprehensive summary and comparative analysis of the ego-vehicle perception datasets. In the following, we categorize the datasets discussed based on the perception tasks they support (detection, segmentation, and tracking). Specifically, we divide them into five categories: (1) datasets supporting all three tasks; (2) datasets that support detection and segmentation; (3) datasets that support detection and tracking; (4) datasets that support detection only; and (5) datasets that support tracking only.

\begin{table*}[t]
\caption{Summary and comparative analysis of the Ego-Vehicle Perception Datasets in autonomous vehicles.}
\vspace{-0.4cm}
\label{Table: Ego-vehicle}
\begin{center}
\resizebox{\textwidth}{!}{
\setlength{\tabcolsep}{4pt}
\begin{tabular}{lllllcccccccc}  
\toprule

\multicolumn{2}{c}{} & \multicolumn{4}{c}{} & \multicolumn{3}{c}{\textbf{Sensor}} &\multicolumn{3}{c}{\textbf{Application}} \\ [1.5mm]

\cmidrule(l){7-9}
\cmidrule(l){10-12}

\multirow{1}{*}{\textbf{Ref}}     & \textbf{Dataset}   & \textbf{Year}            &  \textbf{Region}   & \textbf{Frame}  & \textbf{Format} 
                                      & \textbf{Camera}         & \textbf{Lidar}            & \textbf{Radar}   & \textbf{Detection} & \textbf{Segmentation}  &  \textbf{Tracking}  & \textbf{Access}  \\  

\midrule

\multirow{1}{*}{\cite{kent2024msu} } &\ph{MSU-4S}  &\ph{2024}  &\ph{USA} &\ph{+100K}  &JPEG, PCD, YOLO    &\ph{Yes}   &\ph{Yes}    &\ph{Yes}   &\ph{\cmark} &\ph{\xmark} &\ph{\xmark} &  \href{https://www.egr.msu.edu/waves/msu4s/}{Web}\\[4mm]

\myrowcolour
\multirow{1}{*}{\cite{zheng2024omnihd} } &\ph{OmniHD-Scenes}  &\ph{2024}  &\ph{China} &\ph{+450K}  &RGB, LiDAR, Radar    &\ph{Yes}   &\ph{Yes}    &\ph{Yes}   &\ph{\cmark} &\ph{\cmark} &\ph{\cmark} &  \href{https://www.2077ai.com/OmniHD-Scenes}{Web}\\[4mm]

\multirow{1}{*}{\cite{alibeigi2023zenseact}} &\ph{Zenseact Open}   &\ph{2023}  &\ph{Europe} &\ph{+100K} &LiDAR, RGB, Radar, GPS    &\ph{Yes}   &\ph{Yes}    &\ph{Yes}   &\ph{\cmark} &\ph{\cmark} &\ph{\cmark} &\href{https://zod.zenseact.com/}{Web}  \\[4mm]

\myrowcolour
\multirow{1}{*}{\cite{diaz2022ithaca365}} &\ph{Ithaca365}   &\ph{2022}  &\ph{USA} &\ph{7K} &LiDAR, RGB, GPS   &\ph{Yes}   &\ph{Yes}    &\ph{No}   &\ph{\cmark} &\ph{\cmark} &\ph{\xmark} &\href{https://ithaca365.mae.cornell.edu/}{Web} \\[4mm]

\multirow{1}{*}{\cite{mao2021one}} &\ph{ONCE}   &\ph{2021}  &\ph{China} &\ph{8M}  &LiDAR, RGB   &\ph{Yes}   &\ph{Yes}    &\ph{No}   &\ph{\cmark} &\ph{\xmark} &\ph{\xmark} &\href{https://once-for-auto-driving.github.io/}{Git}  \\[4mm]

\myrowcolour
\multirow{1}{*}{\cite{deziel2021pixset} } &\ph{Leddar PixSet}  &\ph{2021}  &\ph{Canada} &\ph{29K}  &LiDAR, RGB, Radar, IMU    &\ph{Yes}   &\ph{Yes}    &\ph{Yes}   &\ph{\cmark} &\ph{\xmark} &\ph{\cmark}  &\href{https://leddartech.com/solutions/leddar-pixset-dataset/}{Web}\\[4mm]

\multirow{1}{*}{\cite{xiao2021pandaset}} &\ph{Pandaset}   &\ph{2020}  &\ph{USA} &\ph{64K}  &LiDAR, RGB    &\ph{Yes}   &\ph{Yes}    &\ph{No}   &\ph{\cmark} &\ph{\cmark} &\ph{\xmark} &\href{https://pandaset.org/}{Web}  \\[4mm]

\myrowcolour
\multirow{1}{*}{\cite{geyer2020a2d2}} &\ph{A2D2}   &\ph{2020}  &\ph{Germany} &\ph{+40K} &LiDAR, RGB   &\ph{Yes}   &\ph{Yes}    &\ph{No}   &\ph{\cmark} &\ph{\cmark} &\ph{\xmark} &\href{https://www.a2d2.audi/a2d2/en.html}{Web}\\[4mm]

\multirow{1}{*}{\cite{roboflow_self_driving}} &\ph{Udacity}   &\ph{2020}  &\ph{USA} &\ph{15K}  &RGB, GPS  &\ph{Yes}   &\ph{No}    &\ph{No}   &\ph{\cmark} &\ph{\xmark} &\ph{\xmark} & \href{https://github.com/udacity/self-driving-car}{Git} \\[4mm]

\myrowcolour
\multirow{1}{*}{\cite{sun2020scalability}} &Waymo   &2020  &USA  &+10M  &LiDAR, RGB, TFRecord    &Yes   &Yes    &No   &\cmark &\xmark &\cmark & \href{https://waymo.com/open/}{Web} \\[4mm]

\multirow{1}{*}{\cite{pham20203d}} &A*3D   &2020  &Singapore  &+20K   &LiDAR, RGB   &Yes   &Yes    &Yes   &\cmark &\xmark &\xmark &\href{https://github.com/I2RDL2/ASTAR-3D}{Git}  \\[4mm]

\myrowcolour
\multirow{1}{*}{\cite{inDdataset}} &InD   &2020    &Germany  &+50K &RGB, Trajectories    &Yes   &No    &No   &\xmark  &\xmark   &\cmark &\href{https://levelxdata.com/ind-dataset/}{Web} \\[4mm]

\multirow{1}{*}{\cite{exiDdataset}} &exiD  &2020  &Germany  &+30K &RGB, Trajectories     &Yes   &No    &No   &\xmark &\xmark &\cmark &\href{https://levelxdata.com/exid-dataset/}{Web}  \\[4mm]

\myrowcolour
\multirow{1}{*}{\cite{caesar2020nuscenes}} &nuScenes  &2019 &USA  &1.4M  &LiDAR, RGB, Radar, GPS   &Yes   &Yes    &Yes   &\cmark  &\cmark   &\xmark &\href{https://www.nuscenes.org/}{Web}    \\[4mm]

\multirow{1}{*}{\cite{chang2019argoverse}} &ArgoVerse  &2019  &USA &3M  &LiDAR, RGB, HD Maps   &Yes   &Yes    &No   &\cmark  &\xmark   &\cmark &\href{https://www.argoverse.org/}{Web}  \\[4mm]

\myrowcolour
\multirow{1}{*}{\cite{xue2019blvd}} &BLVD &2019 &USA  &+100K   &LiDAR, RGB   &Yes   &Yes    &No   &\cmark  &\xmark   &\cmark &\href{https://github.com/VCCIV/BLVD}{Git}  \\[4mm]

\multirow{1}{*}{\cite{schafer2018commute}} &Comma2K19  &2019  &USA  &+50K   &RGB, GPS, IMU  &Yes   &No    &No   &\cmark &\xmark &\xmark &\href{https://github.com/commaai/comma2k19}{Git}  \\[4mm]

\myrowcolour
\multirow{1}{*}{\cite{yu2018bdd100k}} &\ph{BDD100k}   &\ph{2018}  &\ph{USA} &\ph{120M}  &RGB, GPS  &\ph{Yes}   &\ph{No}    &\ph{No}   &\ph{\cmark} &\ph{\cmark} &\ph{\cmark} &\href{https://bair.berkeley.edu/blog/2018/05/30/bdd/}{Web} \\[4mm]

\multirow{1}{*}{\cite{huang2018apolloscape}} &ApolloScape   &2018  &China  &+140K  &LiDAR, RGB  &Yes   &Yes    &No   &\cmark &\cmark &\xmark &\href{https://apolloscape.auto/}{Web}  \\[4mm]

\myrowcolour
\multirow{1}{*}{\cite{Cordts2016Cityscapes}} &CityScape   &2016 &Germany   &+25K   &RGB  &Yes   &No    &No   &\cmark  &\cmark   &\xmark &\href{https://www.cityscapes-dataset.com/}{Web} \\[4mm]

\multirow{1}{*}{\cite{barnes2020oxford}} &\ph{Oxford RobotCar}   &\ph{2015}  &\ph{UK} &\ph{+20M} &LiDAR, RGB, GPS, IMU   &\ph{Yes}   &\ph{Yes}    &\ph{No} &\ph{\cmark} &\ph{\xmark} &\ph{\xmark} &\href{https://robotcar-dataset.robots.ox.ac.uk/}{Web}  \\[4mm]

\myrowcolour
\multirow{1}{*}{\cite{geiger2012we}} &KITTI  &2012  &Germany  &+40K   &LiDAR, RGB, GPS, IMU   &Yes   &Yes    &No   &\cmark &\xmark &\cmark &\href{https://www.cvlibs.net/datasets/kitti/}{Web}  \\[4mm]

\bottomrule
\end{tabular}}
\end{center}
\footnotesize{ 
\scriptsize $^*$ For the reader’s convenience, a direct hyperlink to each dataset’s original source—whether a website or GitHub repository—is provided in the 'Access' column to enable easy access and download.
}
\end{table*}

\subsubsection{Detection, Segmentation, and Tracking Datasets}
OmniHD-Scenes \cite{zheng2024omnihd}, Zenseact Open \cite{alibeigi2023zenseact}, and BDD100k \cite{yu2018bdd100k} datasets support all three major perception tasks, making them ideal for end-to-end autonomous driving pipelines. The OmniHD-Scenes dataset \cite{zheng2024omnihd} is a large-scale multimodal dataset developed and released by 2077AI in 2024 to provide complete omnidirectional high-definition data for autonomous driving. The dataset is collected from real-world roads across urban, suburban, and rural areas in China under varying weather and lighting conditions. OmniHD-Scenes integrates 360$^\circ$ high-definition LiDAR sensors, 4D imaging Radar sensors, and ultra-high-resolution RGB cameras. The 128-beam LiDAR sensor generates ultra-dense point clouds, providing precise geometric and depth information. Six high-resolution cameras capture front, rear, and surrounding views at resolutions of $3840\times2160$ and $1920\times1080$ pixels at 30 frames per second. Additionally, the vehicle is equipped with six 4D Radar sensors to enable omnidirectional environmental perception. Other onboard devices include a GNSS/IMU for measuring location and angular velocity, an ADCU for processing sensor data and making driving decisions, and an IPC for handling heavy data workloads. The advantages of this dataset are its extensive video clips, frames, and data points, combined with an advanced 4D annotation pipeline. It provides comprehensive omnidirectional high-definition coverage and a detailed multimodal annotation system. Moreover, the OmniHD-Scenes dataset offers strong generalization capabilities by capturing data from multiple cities under different weather conditions. This enables researchers to conduct long-range perception and robust dynamic object tracking under adverse conditions, as well as achieve complete 360° scene understanding without major occlusions or missing data regions. However, the OmniHD-Scenes dataset is limited by partial annotation coverage (only 200 of 1,501 clips are fully labeled) and requires substantial computational and storage resources due to the high resolution and large volume of its multimodal data. 

The Zensact Open Dataset (ZOD) \cite{alibeigi2023zenseact} was gathered over two years across fourteen European nations, from Sweden to Italy, to include diverse road conditions and environmental settings. It was designed to address existing dataset limitations by offering extensive sensing capabilities and diverse geographical coverage, as well as supporting multiple research applications such as detection, tracking, segmentation, sensor fusion, and weather classification. The Zensact Open Dataset used multiple vehicles with multiple sensor types to obtain high-resolution data across various modalities. The camera system contains an 8MP RGB sensor, which provides a 120$^\circ$ field of view at 10.1 Hz. Additionally, it provides point cloud data from one Velodyne VLS128 and two Velodyne VLP16 LiDAR units, which extend their maximum range to 245 meters. Meanwhile, the Continental ARS513 B1 Radar system provides medium-range spatial data at 60ms intervals, and GNSS/IMU units deliver high-precision vehicle localization. The ZOD dataset contains detailed annotations across different sensor modalities, including 2D and 3D bounding boxes for both dynamic and static objects, semantic segmentation of lanes, road surfaces, and ego-road labeling, as well as road surface conditions (e.g., wet, snowy) and classification of 156 traffic sign types. The ZOD dataset demonstrates strong potential for model generalization across different environments, as it covers fourteen countries with diverse weather conditions, road structures, and traffic patterns. It combines high-resolution camera, LiDAR, and Radar sensors with detailed annotations, supporting advanced perception, fusion, and autonomous driving challenges. On the other hand, the ZOD dataset has several weaknesses. Although the dataset covers many areas in Europe, it has a lower volume, lacks data from other continents, and includes only short sequences (up to a few minutes).

The BDD100K (Berkeley DeepDrive 100K) dataset \cite{yu2018bdd100k} is one of the largest and most diverse open datasets available for autonomous driving research. The video clips were captured from various cities across the United States, including San Francisco, Oakland, San Jose, and New York City. The video clips include over 1,000 hours of driving footage, 120 million frames, and GPS/IMU trajectory data for each video. This dataset offers several advantages, including coverage of a wide range of real-world driving scenarios such as different geographic locations, climate conditions, and daytime periods, which makes it well-suited for training robust and generalizable perception models. Additionally, it contains a broad range of dense annotations, including bounding boxes for object detection of cars, buses, trains, and pedestrians. Semantic segmentation and instance segmentation are also included to annotate every single pixel in the image. Notable drawbacks of this dataset are lacking Lidar or Radar information, which limits its applicability for sensor fusion tasks, 3D object detection, and perception system development.

\subsubsection{Detection and Segmentation Datasets}
Ithaca365 \cite{diaz2022ithaca365}, Pandaset \cite{diaz2022ithaca365}, A2D2 \cite{geyer2020a2d2}, nuScenes \cite{caesar2020nuscenes}, CityScape \cite{Cordts2016Cityscapes}, and ApolloScape \cite{huang2018apolloscape} datasets are useful for detection and segmentation tasks. The Ithaca365 \cite{diaz2022ithaca365} was collected through a 15km route in New York over the course of a year, focusing on temporal variability by recording daily scenes across all four seasons. This dataset used six RGB cameras, 3D LiDAR, and GNSS/INS to ensure accurate localization. These sensor modalities offer a detailed view of the environment suitable for multiple tasks such as 2D and 3D object detection, semantic segmentation, and depth estimation. A key strength of the Ithaca365 dataset is its ability to support strong modeling for long-term scene understanding and perception consistency at a low computational cost. In addition, it provides high-resolution RGB images, dense point clouds, and geo-referencing for both urban and suburban environments, accompanied by precise calibration files and metadata logs indicating the weather conditions at the time of capture. Regarding the limitations of the dataset, it contains only 7,000 frames, which is significantly fewer than other datasets and affects its generalizability. Furthermore, the dataset lacks Radar data collection, which restricts its applicability in sensor fusion tasks.

The PandaSet dataset \cite{diaz2022ithaca365} was collected using a Chrysler Pacifica minivan equipped with advanced sensors, capturing data across urban environments in California, USA. The sensor suite includes six cameras, a 360-degree spinning LiDAR (Pandar64), a forward-facing long-range LiDAR (PandarGT), a GPS unit, and IMU sensors. It contains 28 object classes (e.g., cars, cyclists, pedestrians) annotated with 3D bounding boxes, along with 37 semantic categories that provide point-level annotations. A key strength of this dataset is its high-resolution LiDAR point clouds from both the Pandar64 and PandarGT sensors, making it well-suited for sensor fusion tasks. However, the PandaSet dataset faces limited generalizability due to its lack of geographical diversity, as the data was collected from specific urban areas in California with less variation in road types, regions, and weather conditions. Furthermore, the camera-recorded scenes are relatively short (about 8 seconds per scene), which limits their effectiveness for some tasks such as long-term behavior modeling or route prediction. Lastly, the dataset does not include Radar data, which reduces its applicability for depth perception tasks. The A2D2 dataset \cite{geyer2020a2d2} was collected from various urban, suburban, and rural areas in southern Germany to provide a multimodal dataset suitable for semantic segmentation, 3D object detection, and sensor fusion. The data was recorded using six camera sensors and five LiDAR units, all equipped with various hardware and software to ensure time synchronization for supporting real-time processing, multi-sensor data fusion, and offline analysis. It includes semantic segmentation annotations across 38 classes, 12,497 frames with 3D bounding box annotations, and approximately 392,556 unannotated sequential frames for self-supervised learning. The combination of multiple cameras and LiDAR units is one of the major advantages of the A2D2 dataset, enabling multimodal support for various perception tasks. However, the annotation process was not sequential, which may pose challenges for researchers requiring temporal continuity. Additionally, generalizability may be limited since the data were collected exclusively in Germany.

The Cityscapes dataset \cite{Cordts2016Cityscapes} was developed to serve two main purposes: providing a benchmark and offering a large-scale dataset for training and testing various algorithms in pixel-level and instance-level semantic labeling. The data were collected over a period of seven months (spring, summer, and fall) in 50 cities across Germany and neighboring countries. The dataset contains 5,000 finely annotated and 20,000 coarsely annotated images, with annotations spanning 30 classes grouped into 8 categories. For example, the “vehicle” category includes classes such as car, truck, bus, on rails, motorcycle, bicycle, caravan, and trailer. The Cityscapes dataset also provides metadata that includes GPS coordinates, ego-motion data, and outside temperature. The strength of the dataset lies in its large scale and high-quality annotations. Additionally, its comprehensive metadata enhances its utility for context-aware modeling. However, a key limitation is that it focuses exclusively on urban scenes, which reduces its applicability to highway or rural environments. The nuScenes dataset \cite{caesar2020nuscenes}, released in March 2019, provides a comprehensive multimodal dataset for autonomous driving research. Data was collected from dense urban environments in Boston and Singapore, with 1,000 driving scenes, each lasting 20 seconds, totaling 5.5 hours of driving data. Each scene includes synchronized data from multiple sensors. Two Renault Zoe electric vehicles were equipped with six cameras (360$^\circ$ view), one LiDAR, five Radars, and a GPS/IMU. The dataset contains 1.4 million annotated 3D bounding boxes across 23 object classes, with additional attributes like visibility, pose, and activity. It also includes metadata such as velocities, accelerations, and tracking information.

Lastly, the ApolloScape dataset \cite{huang2018apolloscape}, released in 2018, was designed to support research in autonomous driving through extensive, semantically annotated data. It includes 143,906 video frames with pixel-level annotations for semantic segmentation collected from various urban environments in China under diverse traffic and weather conditions. A mid-size SUV was equipped with two VUX-1HA laser scanners, two front-facing VMX-CS6 cameras, and an IMU/GNSS unit for accurate positioning and orientation. ApolloScape comprises multiple sub-datasets covering scene parsing, car instances, lane segmentation, self-localization, trajectory, tracking, and stereo tasks. It offers high-resolution RGB images with pixel-wise annotations for 28 object classes, dense LiDAR scans with point-level labels, and lane marking annotations categorized by color and type. Manually annotated trajectories further support behavior prediction research. Data is divided into training, validation, and testing sets. ApolloScape's strengths lie in its diversity, detailed annotations, and real-world complexity. However, its geographic limitation to China may affect generalizability, and manual labeling introduces potential human error. The dataset's size also poses challenges for storage and processing.

\subsubsection{Detection and Tracking Datasets}
Leddar PixSet \cite{deziel2021pixset}, Waymo \cite{sun2020scalability}, ArgoVerse \cite{chang2019argoverse}, BLVD \cite{xue2019blvd}, and KITTI \cite{geiger2012we} datasets provide annotated sequences that support object detection, as well as short-term and long-term tracking, with a strong emphasis on object localization and temporal consistency. Leddar PixSet \cite{deziel2021pixset} is a publicly available dataset for autonomous driving research and development released in 2021. It includes full‐waveform LiDAR data to support developing and testing sensor‐fusion algorithms. Data were collected during the summer months across urban and suburban areas of Quebec City and Montreal, Canada. The PixSet dataset contains 29,000 frames distributed over 97 sequences, with 1.3 million annotated 3D bounding boxes. The sensor suite mounted on a Toyota RAV4 includes a Leddar Pixell LiDAR capable of full-waveform data acquisition, a mechanical scanning LiDAR, three cameras, a Radar unit, and an integrated IMU/GPS. The PixSet dataset offers several notable strengths. It is the first dataset to include full-waveform data from a flash LiDAR, offering richer information compared to conventional datasets. The combination of LiDAR, Radar, cameras, and IMU/GPS supports thorough sensor fusion research. Data collected across varying weather and lighting conditions contributes to the robustness of algorithms trained on it. Moreover, with over 1.3 million annotated 3D bounding boxes, the dataset is well-suited for developing advanced tracking algorithms. However, a key drawback is its limited geographic diversity.

The Waymo dataset \cite{sun2020scalability} is designed to offer a large-scale, diverse, and high-quality resource for object detection and tracking. It features 2D (camera-based) and 3D (LiDAR-based) bounding box annotations. Labeled object categories include vehicles, cyclists, pedestrians, and traffic signs. In the 2D data, objects are annotated with tightly fitting, axis-aligned bounding boxes with four degrees of freedom ($x,y,w,h$). In the 3D data, bounding boxes include seven degrees of freedom ($x,y,z,l,w,h, \theta$) to capture spatial information precisely. Each object in 2D and 3D data is given a unique ID maintained across frames to support tracking. Each object is assigned a consistent ID across frames in both 2D and 3D modalities, facilitating robust tracking. The Waymo dataset includes 1,150 scenes (6.4 hours) with synchronized camera and LiDAR data, split into 798 training, 202 validation, and 150 test scenes. Data was captured using five cameras and five LiDAR sensors, providing full 360$^\circ$ coverage in several cities, including San Francisco, Mountain View, and Phoenix, offering a variety of urban and suburban scenes. A primary benefit of the Waymo dataset is its comprehensive 2D and 3D annotations for object detection and tracking using synchronized camera and LiDAR data. It also enables research into camera-only systems through its 2D annotations. While the dataset covers several U.S. cities with urban and suburban scenes, it lacks representation of rural areas and adverse weather conditions. Thus, for AV applications focused on those scenarios, alternative datasets may be more appropriate. 

The AgroVerse dataset \cite{chang2019argoverse}, released in October 2019, supports 3D tracking and motion forecasting for autonomous driving, with data collected in Pittsburgh and Miami across different seasons and times. It includes 113 scenes for 3D tracking and over 324,000 five-second scenarios for motion forecasting, using LiDAR, 360$^\circ$ cameras, and 3D bounding box annotations. HD maps offer lane-level geometry and semantic data for map-based learning. AgroVerse2, released in 2021, expanded to six U.S. cities with diverse traffic, road layouts, and weather, adding stereo and ring cameras for a richer multimodal dataset. Despite its strengths, the dataset suffers from class imbalance, limited environmental diversity, and short scenario durations, hindering long-term behavior prediction. The BLVD dataset \cite{xue2019blvd}, released in 2019, supports dynamic 4D tracking (3D + time), 5D interactive event recognition (object interactions), and 5D intention prediction. It includes 3D bounding boxes with object IDs for temporal tracking and annotations for 28 predefined interactive events: 13 for vehicles, 8 for pedestrians, and 7 for riders. Future object states, such as location, orientation, and event type. Data was collected in Changshu, China, using multi-view cameras, a Velodyne HDL-64E LiDAR, and GPS/IMU under varied traffic, lighting, and road conditions. The dataset comprises 654 videos (120,000 frames), 250,000 3D boxes, 4,902 tracked objects, 6,004 event fragments, and 4,900 intention predictions. It is well-balanced across different conditions and object types. While segmentation data is not included, BLVD offers rich annotations for various tasks and evaluation metrics, focusing on relevant objects within a 50-meter range to support decision-making. Its dependence on both camera and LiDAR data, as well as its collection in China, should be considered when applying it to broader AV contexts.

KITTI dataset \cite{geiger2012we} aims to develop novel benchmarks that reflect the complexities of real-world scenarios for use in the development of AVs. The researchers equipped a Volkswagen Passat B6 with one inertial navigation system (OXTIS-RT-3003), one laser scanner (Velodyne-HDL-64E), two grayscale cameras (FL2-14S3M-C), two color cameras (FL2-14S3C-C) and four varifocal lenses (Edmund optics NT59-917). The strength of the KITTI dataset is that it provides data obtained from a variety of sensors and captures real-world driving scenarios, which can be applied to various practical algorithms in autonomous driving research. The downside is that all the data was collected in a single city, which reduces geographic diversity and limit the algorithms' generalizability.

\subsubsection{Detection-only Datasets}
MSU-4S \cite{kent2024msu}, ONCE \cite{mao2021one}, Udacity \cite{roboflow_self_driving}, A*3D \cite{pham20203d}, Comma2K19 \cite{schafer2018commute}, and Oxford RobotCar \cite{barnes2020oxford} datasets offer annotations solely for object detection tasks, typically in the form of 2D or 3D bounding boxes, without additional support for segmentation or tracking. The MSU-4S (Michigan State University Four Seasons) dataset \cite{kent2024msu}, released in 2024, provides multimodal data for evaluating autonomous driving performance across seasonal and environmental variations. The dataset is collected around the Michigan State campus using a modified 2017 Chevrolet Bolt EV. It contains over 100,000 high-resolution frames from cameras, LiDAR, and Radar, supporting tasks such as 2D/3D object detection, sensor fusion, and domain adaptation. The test vehicle is equipped with three FLIR cameras, two LiDARs (Ouster OS-1 and Velodyne VLP-32), six Continental Radar sensors, and an IMU, capturing robust data under a wide range of conditions, including heavy rain and snow. Annotations include 2D YOLO-format bounding boxes, 3D YAML-format boxes, and detailed metadata (e.g., weather, date, and time), with outputs in JPEG and PCD/YAML formats for easy integration. MSU-4S is particularly strong in representing real-world, seasonally diverse driving scenes suitable for cross-modal learning and sensor redundancy research. Despite its strengths, the dataset has several limitations: it is geographically constrained to a single region, lacks annotations for semantic segmentation and object tracking, and presents compatibility challenges for systems using different sensor configurations. Additionally, some sequences remain only partially annotated, although future updates are anticipated. Overall, it serves as a valuable resource, especially when combined with datasets from other regions.

The One Million Scenes (ONCE) dataset \cite{mao2021one}, released in 2021, supports 3D object detection and semi/self-supervised learning for autonomous driving. It includes one million LiDAR frames, seven million camera images, and 144 hours of driving data across 200 km² of urban and suburban areas in China under various weather and lighting conditions. Sixteen thousand scenes are fully annotated with 2D and 3D bounding boxes for five object classes. Data was collected using a vehicle equipped with a 40-beam LiDAR and seven cameras for full 360$^\circ$ coverage, with annotations provided in three coordinate systems. The dataset offers rich, synchronized multimodal data and extensive unlabeled samples, ideal for sensor fusion and addressing the long-tail problem. However, it is limited to only five annotated object classes. The Udacity dataset \cite{roboflow_self_driving}, released in 2016, was collected in Mountain View, California, during daylight to support autonomous driving tasks like object detection, tracking, and behavioral cloning. It includes around 404,916 training frames, 5,614 test frames, and 10 hours of driving data. The vehicle used a front-facing camera (1920×1200 at 2Hz), LiDAR, GPS, and IMU. Over 65,000 labels across 9,423 frames were annotated and split into training and testing sets. The dataset offers real-world multimodal sensor data for experimentation. However, the dataset lacks key labels (e.g., pedestrians, traffic lights), and being limited to daytime urban scenes with low temporal resolution, has reduced generalizability.

The A3D dataset \cite{pham20203d}, released in 2019, was designed to offer more diverse driving conditions than existing datasets, covering variations in weather, traffic density, and environments. It includes 39,000 frames with 230,000 3D object annotations for vehicles, cyclists, pedestrians, and blocked objects. Data was collected in Singapore using two PointGrey Chameleon3 cameras and a Velodyne HDL-64ES3 LiDAR, spanning urban, suburban, and diverse road types under various lighting and weather conditions. A3D supports 2D and 3D detection and is valuable for real-world algorithm development. However, its use of costly sensors, Singapore-specific scenes, and limited support for segmentation or tracking should be considered. The comma2k19 dataset \cite{schafer2018commute}, released in 2018, was designed to support mapping and localization algorithms using GNSS and commodity sensors. It includes 33 hours (2,019 one-minute segments) of video data collected along a 20 km stretch of California’s Highway 280 between San Jose and San Francisco, using a Sony IMX2984 camera. The dataset also provides raw GNSS data, CAN messages, and inertial measurements (gyroscope, accelerometer, magnetometer). It enables tasks like future lane estimation and global pose computation using low-cost sensors. However, it lacks annotations for object detection, segmentation, or tracking, limiting its use to localization-related applications or testing pre-trained models in highway scenarios.

The Oxford RobotCar dataset \cite{barnes2020oxford}, released in 2015 by the University of Oxford, supports long-term autonomous vehicle research in urban environments. It features over 100 traversals of a 10 km route in central Oxford, UK, recorded across different seasons, times of day, and weather conditions using a Nissan LEAF EV. The dataset spans 1,000 km and over 20 million frames. It enables research in SLAM, localization, map building, and sensor fusion, equipped with six cameras, 2D/3D LiDAR, GPS/INS, wheel odometry, and IMU. Its strength lies in capturing repeated, real-world environmental variations for long-term autonomy evaluation. While it lacks Radar data and semantic labels and is geographically limited to one route, it remains highly valuable for studying localization robustness, environmental drift, and sensor fusion. Due to its size, significant computational resources are required, but it is well-documented and widely used in the research community.

\subsubsection{Tracking-only Datasets}
InD \cite{inDdataset} and exiD \cite{exiDdataset} datasets focus exclusively on object tracking, often using trajectory-level annotations from top-down perspectives. The Intersection Drone (InD) dataset \cite{inDdataset}, collected in Germany in 2020 by RWTH Aachen University, captures detailed trajectories of road users at unsignalized urban intersections using drone-mounted cameras. Data was collected at four intersections in Aachen and Heinsberg, totaling over 50,000 annotated frames with 8,200 vehicles and 5,300 vulnerable road users (pedestrians and cyclists). Recorded at 25 FPS, the drone’s bird’s-eye view ensures high-accuracy, unobstructed trajectory data ideal for interaction modeling, behavior prediction, and safety validation in autonomous driving. The dataset includes semantic metadata (e.g., movement types, intersection maps) and tools for parsing and visualization. While its strength lies in capturing natural behavior in complex, mixed-traffic environments, limitations include the absence of LiDAR/Radar, its geographic specificity to Germany, a relatively small size, and a lack of support for perception tasks like detection or segmentation. However, InD remains a valuable resource for studying road user interactions and is best used alongside larger, multimodal datasets.

The Extended Interaction (exiD) dataset \cite{exiDdataset}, released in 2020 by RWTH Aachen University, captures trajectory data from drones observing highway driving in Germany. It focuses on behaviors like merging and lane changing across seven challenging 420-meter highway sections. The dataset includes 16 hours of footage, over 30,000 frames, and data on 69,430 road users covering 27,300 km. Drone-based bird’s-eye views provide accurate, time-stamped trajectories with annotations, semantic metadata (e.g., vehicle types, lane markings), and HD maps. Data is delivered in CSV format with Python tools for parsing and visualization. exiD excels in clean, high-resolution data for interaction modeling and planning in high-speed environments. However, it lacks LiDAR/Radar data, semantic segmentation, and object classification, and is limited in size and geographic diversity. Despite these limitations, it is a valuable resource for studying real-world highway interactions.

\subsection{Roadside Perception Datasets}
Roadside perception datasets are acquired from fixed infrastructure such as traffic surveillance cameras, LiDARs mounted on poles, and Radars installed at intersections. These static sensors provide an overhead or third-person view of traffic flow, enabling wide-area situational awareness and continuous monitoring of road segments. Such datasets are particularly valuable for augmenting ego-vehicle data, studying traffic behavior, and developing infrastructure-assisted safety systems. Table \ref{Table: Roadside} presents a detailed overview and comparison of the prominent roadside perception datasets. In the following, we categorize the datasets discussed in Table \ref{Table: Roadside} based on the perception tasks they support (detection, segmentation, and tracking). Specifically, we divide them into two categories: (1) datasets that support detection and tracking, (2) datasets that support detection using both camera and LiDAR sensors, and (3) detection datasets rely only on camera sensor.

\begin{table*}[htbp]
\caption{Summary and comparative analysis of the Roadside Perception Datasets in autonomous vehicles.}
\vspace{-0.4cm}
\label{Table: Roadside}
\begin{center}
\resizebox{\textwidth}{!}{
\setlength{\tabcolsep}{4pt}
\begin{tabular}{lllllcccccccc}  
\toprule

\multicolumn{2}{c}{} & \multicolumn{4}{c}{} & \multicolumn{3}{c}{\textbf{Sensor}} &\multicolumn{3}{c}{\textbf{Application}} \\ [1.5mm]

\cmidrule(l){7-9}
\cmidrule(l){10-12}

\multirow{1}{*}{\textbf{Ref}}     & \textbf{Dataset}   & \textbf{Year}            &  \textbf{Region}   & \textbf{Frame}  & \textbf{Format} 
                                      & \textbf{Camera}         & \textbf{Lidar}            & \textbf{Radar}   & \textbf{Detection} & \textbf{Segmentation}  &  \textbf{Tracking}  & \textbf{Access}  \\  

\midrule

\multirow{1}{*}{\cite{zhu2024roscenes}} &\ph{RoScenes}  &\ph{2024}  &\ph{China} &\ph{+4.8K}  &RGB, BEV, GPS   &\ph{Yes}   &\ph{No}    &\ph{No}   &\ph{\cmark} &\ph{\xmark} &\ph{\xmark} &  \href{https://github.com/roscenes/RoScenes}{Git}\\[4mm]

\myrowcolour
\multirow{1}{*}{\cite{zimmer2023tumtraf}} &\ph{TUMTraf}  &\ph{2023}  &\ph{Germany} &\ph{+4.8K}  &RGB, LiDAR   &\ph{Yes}   &\ph{Yes}    &\ph{No}   &\ph{\cmark} &\ph{\xmark} &\ph{\xmark} &  \href{https://innovation-mobility.com/en/project-providentia/a9-dataset/}{Web}\\[4mm]

\multirow{1}{*}{\cite{cress2022a9}} &\ph{A9-Dataset}  &\ph{2022}  &\ph{Germany} &\ph{+1K}  &RGB, LiDAR, GPS   &\ph{Yes}   &\ph{Yes}    &\ph{No}   &\ph{\cmark} &\ph{\xmark} &\ph{\xmark} &  \href{https://github.com/tum-traffic-dataset/tum-traffic-dataset-dev-kit}{Git}\\[4mm]

\myrowcolour
\multirow{1}{*}{\cite{wang2022ips300+}} &\ph{IPS300+}  &\ph{2022}  &\ph{China} &\ph{+14K}  &RGB, LiDAR   &\ph{Yes}   &\ph{Yes}    &\ph{No}   &\ph{\cmark} &\ph{\xmark} &\ph{\cmark} &  \href{http://www.openmpd.com/column/IPS300}{Web}\\[4mm]

\multirow{1}{*}{\cite{ye2022rope3d}} &\ph{Rope3D}  &\ph{2022}  &\ph{China} &\ph{+50K}  &RGB, LiDAR, GPS   &\ph{Yes}   &\ph{Yes}    &\ph{No}   &\ph{\cmark} &\ph{\xmark} &\ph{\xmark} &  \href{https://thudair.baai.ac.cn/rope}{Web}\\[4mm]

\myrowcolour
\multirow{1}{*}{\cite{busch2022lumpi}} &\ph{LUMPI}  &\ph{2022}  &\ph{Germany} &\ph{+200K}  &RGB, LiDAR, GPS   &\ph{Yes}   &\ph{Yes}    &\ph{No}   &\ph{\cmark} &\ph{\xmark} &\ph{\xmark} &  \href{https://data.uni-hannover.de/cs_CZ/dataset/lumpi}{Web}\\[4mm]

\multirow{1}{*}{\cite{howe2021weakly}} &\ph{WIBAM}  &\ph{2021}  &\ph{UK} &\ph{+33K}  &RGB, BEV  &\ph{Yes}   &\ph{No}    &\ph{No}   &\ph{\cmark} &\ph{\xmark} &\ph{\xmark} &  \href{https://github.com/MatthewHowe/WIBAM}{Git}\\[4mm]

\myrowcolour
\multirow{1}{*}{\cite{zhan2019interaction}} &\ph{Interaction}  &\ph{2019}  &\ph{Global$^*$} &\ph{+16Hr}  &Video, Trajectories   &\ph{Yes}   &\ph{Yes}    &\ph{No}   &\ph{\cmark} &\ph{\xmark} &\ph{\cmark} &  \href{https://interaction-dataset.com/}{Web}\\[4mm]

\multirow{1}{*}{\cite{tang2019cityflow}} &\ph{CityFlow}  &\ph{2019}  &\ph{USA} &\ph{+229K}  &MP4, GPS, Trajectories   &\ph{Yes}   &\ph{No}    &\ph{No}   &\ph{\cmark} &\ph{\xmark} &\ph{\cmark} &  \href{https://cityflow-project.github.io/}{Git}\\[4mm]

\bottomrule
\end{tabular}}
\end{center}
\footnotesize{ \scriptsize $*$ For the reader’s convenience, a direct hyperlink to each dataset’s original source—whether a website or GitHub repository—is provided in the 'Access' column to enable easy access and download.\\
$*$ `\textbf{Global}' indicates that the data was originally collected from different countries including USA, China, Germany, and Bulgaria.
}
\end{table*}

\subsubsection{Detection and Tracking Datasets}
IPS300+ \cite{wang2022ips300+}, Interaction \cite{zhan2019interaction}, and CityFlow \cite{tang2019cityflow} datasets enable both object detection and temporal tracking, making them ideal for trajectory prediction, multi-agent behavior modeling, and traffic flow analysis from a stationary infrastructure perspective. Data was collected at a single intersection in Beijing using an Intersection Perception Unit (IPU) with an 80-layer LiDAR, two 5.44 MP cameras, and GPS. It contains 14,198 frames with an average of 319.84 labeled objects per frame, making it ideal for dense traffic scene analysis. Annotations include 3D bounding boxes for seven classes (e.g., pedestrian, cyclist, car) in KITTI format. The top-down view offers unobstructed perception, which is impossible from vehicle-mounted sensors, and the fusion of LiDAR and camera data supports sensor fusion research. Limitations include data from only one fixed intersection, a stationary sensor setup, and a 5 Hz annotation rate, which may be insufficient for high-temporal-resolution tasks. Still, IPS300+ is a strong resource for intersection perception in dense urban traffic.

The Interaction dataset \cite{zhan2019interaction}, released in September 2019, captures over 100 hours of video from 11 locations across the U.S., China, Germany, and Bulgaria using drones and fixed cameras. It includes diverse driving scenarios, such as urban intersections, highway merges, roundabouts, lane changes, and U-turns, as well as various road users (cars, trucks, buses, bikes, and pedestrians). Each agent is annotated with position, velocity, heading, object type, and interaction type (e.g., cooperative merge, aggressive overtake). The dataset supports behavior prediction, trajectory forecasting, multi-agent motion planning, imitation learning, and reinforcement learning. Its strengths include rare interactive behaviors, international scene diversity, and precise bird’s-eye tracking. However, it lacks LiDAR/Radar data, has limited weather and lighting conditions, and requires manual map alignment for some planning frameworks. CityFlow dataset \cite{tang2019cityflow}, released in 2019, was designed to advance research in multi-agent multi-camera (MTMC) vehicle tracking and re-identification (ReID) in urban environments. It contains 3.25 hours of synchronized HD footage from 40 cameras across 10 intersections in a mid-sized U.S. city, covering up to 2.5km between cameras. The dataset includes 229,680 annotated bounding boxes for 666 unique vehicle identities and provides camera calibration for spatio-temporal analysis. Strengths include realistic traffic scenarios, high-quality annotations, and precise tracking. Limitations include a lack of long-term or rare-event coverage, geographic restriction, and fixed viewpoints that limit algorithm robustness.

\subsubsection{Detection-only Datasets (Camera + LiDAR)}
TUMTraf \cite{zimmer2023tumtraf}, A9-Dataset \cite{cress2022a9}, Rope3D \cite{ye2022rope3d}, and LUMPI \cite{busch2022lumpi} datasets support detection tasks using both RGB and LiDAR from roadside perspectives. TUMTraf dataset \cite{zimmer2023tumtraf}, released in 2023 by the Technical University of Munich, features data from a 7-meter-high gantry at a busy intersection near Munich, Germany. The setup includes two high-resolution RGB cameras and two LiDAR sensors, capturing 4,800 labeled LiDAR frames with 57,406 annotated 3D objects across 10 classes and 273,861 attributes. Manual annotations ensure accurate 3D bounding boxes and tracking. The dataset includes diverse traffic behaviors such as turns, overtaking, and U-turns, making it valuable for studying complex interactions and sensor fusion. Limitations include its single-location scope and fixed sensor setup, which may reduce generalizability and lack dynamic perspectives in vehicle-mounted systems. The A9-Dataset \cite{cress2022a9}, released in 2022, was collected from the A9 highway near Munich, Germany, as part of the Providentia++ project. It supports infrastructure-based research in perception, traffic analysis, and autonomous driving, with applications in 3D object detection, sensor fusion, and traffic prediction. The dataset includes 1,000 sensor frames and ~14,000 annotated objects, captured using a multi-sensor setup: four Basler cameras, an Ouster OS1-64 LiDAR, Doppler Radar, and event-based cameras. Each object is manually labeled with 3D bounding boxes, position, orientation, and size. Features include calibrated multi-sensor fusion and a bird’s-eye view of highway traffic. However, its scope is limited to a single highway setting with minimal geographic or scene diversity.

The Rope3D dataset \cite{ye2022rope3d} uses a unique perspective to advance monocular 3D perception for roadside applications. It was collected in China under varying weather and lighting, and combines data from fixed roadside cameras and LiDAR mounted on parked or moving vehicles. The dataset has 50,000 images and more than 1.5 million labeled objects in 13 classes. Although 2D annotations are more common due to occlusions and distance, the dataset provides detailed joint 2D-3D annotations. Its key strengths are the novel roadside viewpoint and environmental diversity. However, it has limited geographic diversity and lacks dynamic perspectives due to fixed sensor setups. The LUMPI (Leibniz University Multi-Perspective Intersection) dataset \cite{busch2022lumpi}, released in June 2022, was recorded at a busy intersection in Hanover, Germany, under various weather and lighting conditions. It includes 2D images and 3D point clouds captured from a multi-perspective setup: three surveillance cameras, four ego-perspective LiDARs, and a mobile mapping van with two RIEGL VQ-250 LiDARs and a GNSS/IMU system. Annotations were created through background subtraction, DBSCAN segmentation, tracking via Extended Kalman Filter and ICP, followed by 3D bounding box estimation, classification, and manual refinement. Its strengths lie in diverse viewpoints and environmental conditions for offering robust perception. However, its static sensor setup limits the dynamic perspective typically offered by vehicle-mounted systems.

\subsubsection{Detection-only Datasets (Camera-only)}
RoScenes \cite{zhu2024roscenes} and WIBAM \cite{howe2021weakly} datasets rely on monocular or BEV camera views, designed for object detection and scene understanding without additional 3D sensing or tracking. The RoScenes dataset, released in 2024, aims to enhance vision-based Bird’s Eye View (BEV) approaches using data from 14 highway scenes in China \cite{zhu2024roscenes}. It includes 1.21 million images and 21.3 million 3D bounding box labels covering 64,000m². The sensor setup features 6–12 roadside cameras mounted on 10m poles, two cameras (different zooms) on each side, and drones flying at 300m to capture full highway coverage without blind spots. Its strengths include a massive scale of data, diverse viewpoints that reduce occlusions, and an efficient BEV-to-3D annotation method. However, the dataset is limited to highway scenarios under sunny conditions, lacking scenes like tunnels or intersections. The WIBAM (Wide Baseline Multiview) dataset \cite{howe2021weakly}, released in October 2021, supports weakly supervised training of monocular 3D object detectors using traffic surveillance footage from elevated cameras at UK intersections. It addresses challenges in estimating 3D vehicle poses from elevated views. The dataset includes 33,000 images captured by four unsynchronized RGB cameras (originally 2560×1140, downsampled to 1920×1080), with 116,702 automatically generated 2D training annotations and 1,651 manually labeled 3D bounding boxes in the test set. It enables pose prediction and employs a weak supervision approach using multi-view reprojection loss. Strengths include reduced labeling costs and real-world applicability; however, its fixed elevated viewpoint limits use in autonomous driving, and auto-generated annotations may introduce noise.

\subsection{Vehicle-to-Language (V2L) Datasets}
V2L datasets or Multimodal-language datasets integrate visual data (images or video) with natural language inputs, including commands, descriptions, or queries. These datasets are designed to be compatible with large language models (LLMs), supporting tasks such as visual question answering, command interpretation, and semantic analysis in AV systems. They enable human-AI interaction and allow autonomous vehicles to interpret, respond, or learn from language-based cues. Table \ref{Table: LLM} below highlights the key features and differences among notable multimodal-language datasets used in AV research. To meaningfully organize the datasets discussed in Table \ref{Table: LLM}, we categorize them based on their underlying sensor modalities, which play a critical role in determining the complexity, realism, and multimodal learning potential of each dataset. Specifically, we divide the datasets into two groups: (1) those that incorporate both camera and LiDAR sensors, offering rich 3D spatial context and (2) those based on camera-only data, which focus on visual-language reasoning from camera inputs.

\begin{table*}[htbp]
\caption{Summary and comparative analysis of the Vehicle-to-Language (V2L) Datasets in autonomous vehicles. ``QA'' denotes Question Answering, ``VQA'' refers to the Visual Question Answering, ``PPR'' indicates Perception, Planning, Reasoning, and ``CPPR'' means Cooperative PPR.}
\vspace{-0.4cm}
\label{Table: LLM} 
\begin{center}
\resizebox{\textwidth}{!}{
\setlength{\tabcolsep}{4pt}
\begin{tabular}{llllllccccccc} 
\toprule

\multicolumn{4}{c}{} & \multicolumn{2}{c}{\textbf{Data}} & \multicolumn{2}{c}{\textbf{Sensor}} &\multicolumn{3}{c}{\textbf{Application}} \\ [1.5mm]

\cmidrule(l){5-6}
\cmidrule(l){7-8}
\cmidrule(l){9-12}

\multirow{1}{*}{\textbf{Ref}}     & \textbf{Dataset}   & \textbf{Year}            &  \textbf{Source}   & \textbf{\#Frame}  & \textbf{\#QA} 
                                      & \textbf{Camera}         & \textbf{Lidar}      & \textbf{VQA} & \textbf{PPR}  &  \textbf{CPPR}   &  \textbf{Detection} & \textbf{Access}  \\  

\midrule

\multirow{1}{*}{\cite{ishaq2025drivelmm}} &\ph{DriveLMM.o1}  &\ph{2025}  &\ph{NuScenes Dataset} &\ph{+1.9K}  &\ph{18K}    &\ph{Yes}   &\ph{Yes}    &\ph{\cmark}   &\ph{\cmark} &\ph{\xmark} &\ph{\xmark} &  \href{https://github.com/ayesha-ishaq/drivelmm-o1}{Git}\\[4mm]

\myrowcolour
\multirow{1}{*}{\cite{chen2025automated}} &\ph{CODA-LM}  &\ph{2025}  &\ph{CODA Dataset} &\ph{+9K}  &63K   &\ph{Yes}   &\ph{No}    &\ph{\cmark}   &\ph{\cmark} &\ph{\xmark} &\ph{\cmark} &  \href{https://github.com/DLUT-LYZ/CODA-LM}{Git}\\[4mm]

\multirow{1}{*}{\cite{chiu2025v2v}} &\ph{V2V-QA}  &\ph{2025}  &\ph{V2V4Real, V2X-Real} &\ph{+48K}  &1.45M   &\ph{Yes}   &\ph{Yes}    &\ph{\cmark}   &\ph{\cmark} &\ph{\cmark} &\ph{\xmark} &  \href{https://eddyhkchiu.github.io/v2vllm.github.io/}{Git}\\[4mm]

\myrowcolour
\multirow{1}{*}{\cite{sima2024drivelm}} &\ph{DriveLM}  &\ph{2024}  &\ph{NuScenes, Carla} &\ph{+74K}  &\ph{375K}   &Yes   &Yes   &\ph{\cmark}   &\ph{\cmark} &\ph{\xmark} &\ph{\xmark} &  \href{https://github.com/OpenDriveLab/DriveLM}{Git}\\[4mm]

\multirow{1}{*}{\cite{inoue2024nuscenes}} &\ph{NuScenes-MQA}  &\ph{2024}  &\ph{NuScenes Dataset} &\ph{+34K}  &1.5M   &\ph{Yes}   &\ph{No}    &\ph{\cmark}   &\ph{\cmark} &\ph{\xmark} &\ph{\cmark}  &\href{https://github.com/turingmotors/NuScenes-MQA}{Git}\\[4mm]

\myrowcolour
\multirow{1}{*}{\cite{wang2025omnidrive}} &\ph{OmniDrive}  &\ph{2024}  &\ph{NuScenes, OpenLane-v2} &\ph{+34K}  &450K   &\ph{Yes}   &\ph{No}    &\ph{\cmark}   &\ph{\cmark} &\ph{\xmark} &\ph{\xmark} &  \href{https://github.com/NVlabs/OmniDrive}{Git}\\[4mm]

\multirow{1}{*}{\cite{tian2024tokenize}} &\ph{TOKEN}  &\ph{2024}  &\ph{NuScenes, DriveLM} &\ph{+28K}  &434K   &\ph{Yes}   &\ph{No}    &\ph{\cmark}   &\ph{\cmark} &\ph{\xmark} &\ph{\cmark} & N/A \\[4mm]

\myrowcolour
\multirow{1}{*}{\cite{cao2024maplm}} &\ph{MAPLM-QA}  &\ph{2024}  &\ph{Original Dataset} &\ph{+14K}  &61K   &\ph{Yes}   &\ph{Yes}   &\ph{\cmark}   &\ph{\cmark} &\ph{\xmark} &\ph{\cmark}    &\href{https://github.com/LLVM-AD/MAPLM}{Git}\\[4mm]

\multirow{1}{*}{\cite{marcu2024lingoqa}} &\ph{Lingo-QA}  &\ph{2024}   &Original Dataset   &\ph{+28K} &\ph{420K}    &\ph{Yes}   &\ph{No}    &\ph{\cmark}   &\ph{\cmark} &\ph{\xmark} &\ph{\cmark} &  \href{https://github.com/wayveai/LingoQA}{Git}\\[4mm]

\myrowcolour
\multirow{1}{*}{\cite{qian2024nuscenes}} &\ph{NuScenes-QA}  &\ph{2024}  &\ph{NuScenes Dataset} &\ph{+34K}  &460K   &\ph{Yes}   &\ph{Yes}    &\ph{\cmark}   &\ph{\cmark} &\ph{\xmark} &\ph{\cmark}  &\href{https://github.com/qiantianwen/NuScenes-QA}{Git}\\[4mm]

\multirow{1}{*}{\cite{ding2024holistic}} &\ph{NuInstruct}  &\ph{2024}  &\ph{NuScenes Dataset} &\ph{+11K}  &91K   &\ph{Yes}   &\ph{No}    &\ph{\cmark}   &\ph{\cmark} &\ph{\cmark} &\ph{\cmark} &  \href{https://github.com/xmed-lab/NuInstruct}{Git}\\[4mm]

\myrowcolour
\multirow{1}{*}{\cite{malla2023drama}} &\ph{DARAMA}  &\ph{2023}  &\ph{Original Dataset} &\ph{+18K}  &103K   &\ph{Yes}   &\ph{No}    &\ph{\cmark}   &\ph{\cmark} &\ph{\xmark} &\ph{\cmark} &  \href{https://usa.honda-ri.com/drama}{Web}\\[4mm]

\bottomrule
\end{tabular}}
\end{center}
\footnotesize{ \scriptsize $*$ For the convenience of the reader, each dataset's original source—whether a website or GitHub repository hyperlink—is provided in the `Access' column to facilitate direct access and download.
}
\end{table*}

\subsubsection{Camera + LiDAR Datasets}
DriveLMM.o1 \cite{ishaq2025drivelmm}, V2V-QA \cite{chiu2025v2v}, MAPLM-QA \cite{cao2024maplm}, NuScenes-QA \cite{qian2024nuscenes}, and DriveLM \cite{sima2024drivelm} datasets provide both RGB imagery and point cloud data, making them suitable for spatially-grounded language tasks, 3D-aware VQA, and multi-sensor fusion in perception and reasoning. The DriveLMM.o1 dataset \cite{ishaq2025drivelmm}, released in March 2025 by Mohamed Bin Zayed University of AI, the DriveLMM-o1 dataset serves as a multimodal benchmark to evaluate LLM reasoning in autonomous driving. It includes over 18,000 visual question answering (VQA) samples across four subsets, covering diverse real-world driving scenarios that involve perception, prediction, and planning tasks. Each sample contains synchronized multiview images, LiDAR data, temporal cues, questions, ground-truth answers, and structured step-by-step reasoning. Designed to test LLMs’ interpretability and multi-step decision-making, the dataset also provides tools for data parsing and custom evaluation metrics. It supports research in visual reasoning, sensor fusion, and end-to-end autonomous decision-making. Strengths include detailed annotations, dynamic scene diversity, and structured explanations. Limitations include limited geographic and environmental diversity, as well as high computational demands due to data complexity. DriveLMM-o1 sets a new standard for evaluating multimodal LLM reasoning in Avs.

The V2V-QA dataset \cite{chiu2025v2v}, released in April 2025 by Carnegie Mellon University and NVIDIA, is a large-scale multimodal benchmark for enhancing collaborative autonomous driving via language-based reasoning. Built on real-world V2V4Real and V2X-Real datasets, it includes over 48,000 frames and 1.45 million question-answer (QA) pairs that simulate vehicle-to-vehicle communication. The QA system supports object grounding, identification, and driving planning tasks in multi-agent environments using synchronized LiDAR and camera data. Each frame contains five QA types generated through geometric templates and contextual rules. The dataset enables training and evaluation of vision-language models, like V2V-LLM, which outperform standard LLMs in cooperative perception and intention prediction. Strengths include its focus on explainability, situational awareness, and multi-agent interaction. Limitations include dependency on specific sensor setups, restricted geographic coverage, high computational demands, and uncertain usage permissions. NuScenes-QA dataset released in 2023 by Fudan University \cite{qian2024nuscenes}. It is a large-scale VQA benchmark built on the nuScenes dataset to enhance reasoning in autonomous driving. It contains 460,000 QA pairs derived from 34,000 driving scenes with synchronized RGB images and LiDAR point clouds. Using 3D scene graphs and logic-based templates adapted from the CLEVR benchmark, researchers generated diverse questions ranging from basic attributes to complex spatiotemporal inference. Temporal questions require multi-frame reasoning to track changes and object motion. The dataset evaluates vision-language models and includes baseline results from Transformer-based models. Although NuScenes-QA advances VQA for driving tasks, limitations include a lack of background scene understanding, limited linguistic diversity due to templated questions, and shallow semantic depth.

The MAPLM-QA dataset \cite{cao2024maplm}, introduced in 2024 through a collaboration between Tencent’s T Lab and several universities. It is a large-scale multimodal benchmark for vision-language reasoning in autonomous driving. It includes 61,000 QA pairs linked to 14,000 sensor-rich frames featuring panoramic RGB images, 3D LiDAR point clouds, and HD maps from diverse real-world environments like intersections, highways, and rural roads. To overcome limitations of prior VQA datasets, MAPLM-QA challenges LLMs to reason about road layouts, markings, and navigation cues. Data was collected using vehicles equipped with six cameras, LiDAR, and GPS/IMU, with QA annotations refined through a vision model and human verification. The dataset connects language prompts to spatial and environmental features and offers JSON-formatted tools for integration and evaluation. By evaluating LLMs like CLIP and LLaMA, this benchmark highlights their effectiveness in real-world grounding and exposes their limitations in spatial reasoning. Although it is limited by geographic scope and high computational requirements, MAPLM-QA is a pioneering resource for advancing LLM-based perception and planning. The DriveLM dataset \cite{sima2024drivelm}, released in January 2025 by OpenDriveLab, is a large-scale benchmark for evaluating LLMs and VLMs in autonomous driving via Graph VQA. Unlike standard perception-control datasets, DriveLM emphasizes multi-step reasoning across perception, prediction, planning, and behavior tasks. It includes over 74,000 annotated frames across two subsets: DriveLM-nuScenes (4,871 frames, $\sim$91.4 QAs/frame) and DriveLM-CARLA (70,006 frames, $\sim$20.5 QAs/frame). DriveLM-nuScenes uses keyframe-based manual annotation with automatic QA generation, while DriveLM-CARLA relies on expert-driven simulation and sensor data (RGB + LiDAR). DriveLM's strengths lie in explainability, zero-shot generalization, and structured decision flow. Limitations include reliance on simulation-based data and the high resource demands.

\subsubsection{Camera-only Datasets}
CODA-LM \cite{chen2025automated}, NuInstruct \cite{ding2024holistic}, OmniDrive \cite{wang2025omnidrive}, Lingo-QA \cite{marcu2024lingoqa}, and DARAMA \cite{malla2023drama} datasets rely solely on RGB visual input, often from front-facing or panoramic cameras, and are commonly used in visual-language reasoning, question answering, and instruction-following without explicit 3D spatial data. Dalian University of Technology introduced the CODA-LM dataset \cite{chen2025automated} in 2024 as a comprehensive benchmark for assessing Large Vision-Language Models (LVLMs) in practical autonomous driving systems. In order to evaluate multimodal understanding, it consists of 9,768 high-resolution scenes with uncommon driving events combined with textual prompts. CODA-LM supports three subtasks: general perception (object/event detection), regional perception (localized object description), and driving suggestion (context-based decision reasoning). It uses a hierarchical evaluation system with automated text-only LLM scoring, which offers better alignment with human judgment than multimodal evaluators. Additionally, it introduces CODA-VLM, a fine-tuned model that outperforms all open-sourced driving LVLMs and achieves GPT-4V-level results. Strengths include its focus on safety-critical corner cases, structured reasoning, and reduced reliance on human evaluation. Limitations include restricted coverage of all rare driving scenarios and high computational demands, which may limit accessibility.

In 2024, HKUST, Sun Yat-Sen University, and Huawei Noah's Ark Lab collaborated to release the NuInstruct dataset \cite{ding2024holistic}. It is a large-scale instruction-based multimodal benchmark for autonomous driving. It contains more than 91,000 instruction-response pairs derived from multiview driving videos, addressing limitations in prior datasets by incorporating spatial-temporal data and diverse perspectives. The dataset is organized into 17 subtasks across four categories, including perception, prediction, risk, and planning, with reasoning to support real-world driving decision-making processes. QA pairs are generated using an SQL-based method and require reasoning across frames and views. Strengths include comprehensive coverage of driving tasks and temporal Bird's-Eye-View cross-view reasoning. However, it lacks 3D object detection and traffic light tasks, requires high computational resources, and may need domain adaptation for generalization. OmniDrive dataset \cite{wang2025omnidrive}, released in April 2025 by Hong Kong Polytechnic University and Beijing Institute of Technology. It is a state-of-the-art multimodal dataset that links 3D driving tasks with LLMs using counterfactual reasoning. Built on nuScenes, it adds 34,000 annotated frames and 434,000 QAs focused on "what if" scenarios that bridge perception, planning, and reasoning. The dataset includes factual and counterfactual QA pairs, trajectory data, and keyframes in JSON format for transformer-based pipelines. Two evaluation agents, Omni-Q (trajectory supervision) and Omni-L (vision-language alignment), show improved scene understanding and planning response. OmniDrive's strengths lie in integrating perception and decision-making while enhancing interpretability and explainability for AVs. Limitations include synthetic scene generation that reduce realism and high computational demands.

The Lingo-QA dataset \cite{marcu2024lingoqa}, released in September 2024 by Wayve Technologies, is a large-scale multimodal benchmark for evaluating LLMs in autonomous driving. It includes 28,000 short video scenarios paired with 419,000 QA pairs covering tasks like hazard anticipation, traffic comprehension, and high-level route reasoning. Lingo-QA is built for fine-grained evaluation with QA annotations to evaluate visual perception and decision-making. Key strengths of the Lingo-QA dataset include robust annotation, multi-level reasoning, and standardized assessment. However, its limitations include reliance on short, front-facing video clips with no LiDAR or sensor data, limited scalability (up to 7B LLMs), and Lingo-Judge's dependence on the dataset and lack of support for diverse response formats or preference-based evaluation. DRAMA dataset (Driving Risk Assessment Mechanism with A captioning module) \cite{malla2023drama}, released in 2022 by Honda Research Institute USA, is a multimodal benchmark for developing risk-aware and explainable decision-making in autonomous vehicles. The DRAMA dataset combines video and object-level QA with free-form captions explaining driver reactions, linking risk localization to natural language reasoning. It features synchronized video from two front-facing cameras, along with CAN and IMU data. Annotations include open- and closed-ended questions about risk identification, causes, and reasoning, supported by natural language explanations. DRAMA's key strength lies in integrating multimodal risk perception with interpretability for LVLMs. However, its reliance on subjective human annotations, limited coverage of failure cases, and the risks of deploying incorrect models in safety-critical settings require careful use and additional validation for real-world applications.

\subsection{Vehicle-to-Vehicle (V2V) Datasets}
V2V datasets capture cooperative communication between vehicles, emphasizing shared perception, intent broadcasting, and collaborative planning. These datasets help address limitations of ego-centric perception by enabling vehicles to ``see'' beyond their own sensors, reducing latency in reaction times, and supporting safer maneuvering. Applications include joint trajectory planning, decentralized sensing, and early hazard detection. Table \ref{Table: V2V} outlines a comparative analysis of important V2V datasets for cooperative driving research. In the following, we categorize the datasets presented in Table \ref{Table: V2V} according to the data collection environment, distinguishing between simulation-based and real-world V2V datasets. We divide them into two categories: (1) simulation-based datasets, which are generated using platforms such as CARLA or AirSim and allow for controlled, repeatable experiments; and (2) real-world datasets, which are collected from physical vehicle deployments and offer higher environmental realism and sensor noise characteristics.

\begin{table*}[htbp]
\caption{Summary and comparative analysis of the Vehicle-to-Vehicle (V2V) Datasets in autonomous vehicles.}
\label{Table: V2V}
\vspace{-0.4cm}
\begin{center}
\resizebox{\textwidth}{!}{
\setlength{\tabcolsep}{4pt}
\begin{tabular}{lllllcccccccc}  
\toprule

\multicolumn{2}{c}{} & \multicolumn{4}{c}{} & \multicolumn{3}{c}{\textbf{Sensor}} &\multicolumn{3}{c}{\textbf{Application}} \\ [1.5mm]

\cmidrule(l){7-9}
\cmidrule(l){10-12}

\multirow{1}{*}{\textbf{Ref}}     & \textbf{Dataset}   & \textbf{Year}            &  \textbf{Region}   & \textbf{Frame}  & \textbf{Format} 
                                      & \textbf{Camera}         & \textbf{Lidar} &  \textbf{Radar}   & \textbf{Detection} & \textbf{Segmentation}   &  \textbf{CP} & \textbf{Access}  \\  

\midrule

\multirow{1}{*}{\cite{li2024multiagent}} &\ph{Open Mars}  &\ph{2024}  &\ph{USA} &\ph{+2M}  &RGB, LiDAR, Trajectories   &\ph{Yes}   &\ph{Yes}  &\ph{No}   &\ph{\cmark}  &\ph{\xmark} &\ph{\cmark} &  \href{https://ai4ce.github.io/MARS/}{Git}\\[4mm]

\myrowcolour
\multirow{1}{*}{\cite{lu2024extensible}} &\ph{OPV2V-H}  &\ph{2024}  &\ph{Carla} &\ph{+9K}  &RGB, LiDAR, Depth   &\ph{Yes}   &\ph{Yes}  &\ph{No}  &\ph{\cmark}   &\ph{\xmark} &\ph{\cmark} &  \href{https://github.com/yifanlu0227/HEAL}{Git}\\[4mm]

\multirow{1}{*}{\cite{hu2023collaboration}} &\ph{OPV2V+}  &\ph{2023}  &\ph{Carla} &\ph{11,464}  &RGB, LiDAR, Depth   &\ph{Yes}   &\ph{Yes}  &\ph{No}  &\ph{\cmark}   &\ph{\xmark} &\ph{\cmark} &  \href{https://mobility-lab.seas.ucla.edu/opv2v/}{Web}\\[4mm]

\myrowcolour
\multirow{1}{*}{\cite{wei2023asynchrony}} &\ph{IRV2V}  &\ph{2023}  &\ph{Carla} &\ph{42.5K${^*}^1$}  &RGB, LiDAR, BEW   &\ph{Yes}   &\ph{Yes}  &\ph{No}  &\ph{\cmark}   &\ph{\xmark} &\ph{\cmark} &  \href{https://github.com/MediaBrain-SJTU/CoBEVFlow}{Git}\\[4mm]

\multirow{1}{*}{\cite{xu2023v2v4real}} &\ph{V2V4Real}  &\ph{2023}  &\ph{USA} &\ph{60K}  &RGB, LiDAR, HDMaps   &\ph{Yes}   &\ph{Yes}  &\ph{No}  &\ph{\cmark}   &\ph{\xmark} &\ph{\cmark} &  \href{https://mobility-lab.seas.ucla.edu/v2v4real/}{Web}\\[4mm]

\myrowcolour
\multirow{1}{*}{\cite{axmann2023lucoop}} &\ph{LUCOOP}  &\ph{2023}  &\ph{Germany} &\ph{+54K}  &LiDAR, IMU, GNSS   &\ph{Yes}   &\ph{Yes}  &\ph{No}  &\ph{\cmark}   &\ph{\xmark} &\ph{\cmark} &  \href{https://data.uni-hannover.de/vault/icsens/axmann/lucoop-leibniz-university-cooperative-perception-and-urban-navigation-dataset/}{Web}\\[4mm]

\multirow{1}{*}{\cite{hu2022where2comm}} &\ph{CP-UAV}  &\ph{2022}  &\ph{AirSim} &\ph{131.9K}  &RGB, LiDAR, BEW   &\ph{Yes}   &\ph{Yes}  &\ph{No}  &\ph{\cmark}   &\ph{\xmark} &\ph{\cmark} &  \href{https://siheng-chen.github.io/dataset/coperception-uav/}{Git}\\[4mm]

\myrowcolour
\multirow{1}{*}{\cite{xu2022opv2v}} &\ph{OPV2V}  &\ph{2022}  &\ph{Carla} &\ph{11,464}  &RGB, LiDAR, BEW   &\ph{Yes}   &\ph{Yes}  &\ph{No}  &\ph{\cmark}   &\ph{\xmark} &\ph{\cmark} &  \href{https://mobility-lab.seas.ucla.edu/opv2v/}{Web}\\[4mm]

\multirow{1}{*}{\cite{arnold2021fast}} &\ph{CODD}  &\ph{2021}  &\ph{Carla} &\ph{8,783}  &LiDAR, Transform.txt   &\ph{No}   &\ph{Yes}  &\ph{No}  &\ph{\cmark}   &\ph{\cmark} &\ph{\xmark} &  \href{https://github.com/YuanYunshuang/cosense-simulation}{Git}\\[4mm]

\myrowcolour
\multirow{1}{*}{\cite{yuan2021comap}} &\ph{COMAP}  &\ph{2021}  &\ph{Carla} &\ph{+4.3K}  &RGB, LiDAR, Seg-Mask   &\ph{Yes}   &\ph{Yes}  &\ph{No}  &\ph{\cmark}   &\ph{\cmark} &\ph{\cmark} &  \href{https://github.com/eduardohenriquearnold/fastreg}{Git}\\[4mm]

\multirow{1}{*}{\cite{chen2019cooper}} &\ph{T\&J Cooper}  &\ph{2019}  &\ph{USA} &\ph{100}  &RGB, LiDAR, GPS   &\ph{Yes}   &\ph{Yes}  &\ph{Yes}  &\ph{\cmark}   &\ph{\xmark} &\ph{\cmark} &  N/A\\[4mm]

\bottomrule
\end{tabular}}
\end{center}
\footnotesize{ \scriptsize $*$ For the convenience of the reader, each dataset's original source—whether a website or GitHub repository hyperlink—is provided in the `Access' column to facilitate direct access and download.\\
${^*}^1$ IRV2V dataset includes 34K camera images and 8.5K LiDAR frames.
}
\end{table*}

\subsubsection{Simulation-based Datasets}
OPV2V \cite{xu2022opv2v}, OPV2V+ \cite{hu2023collaboration}, OPV2V-H \cite{lu2024extensible}, IRV2V \cite{wei2023asynchrony}, CP-UAV \cite{hu2022where2comm}, CODD \cite{arnold2021fast}, and COMAP \cite{yuan2021comap} datasets are generated using various simulators, offering controlled environments with customizable scenarios, ground-truth data, and ease of replication for CP tasks. The OPV2V dataset \cite{xu2022opv2v} is a large-scale, open simulated dataset for cooperative Vehicle-to-Vehicle perception and was released in 2021. It contains 70 scenes, 11,464 frames, and 232,913 annotated 3D vehicle bounding boxes. The data was collected from eight towns in CARLA and a digital replica of Culver City, Los Angeles, USA. A key strength of the dataset is that it provides 16 implemented models to evaluate different data fusion strategies. However, its main limitation is that it consists of synthetic data, which may not fully reflect the complexity of real-world environments. The OPV2V+ dataset \cite{hu2023collaboration} is the extension of OPV2V, released in 2023 and co-simulated using OpenCDA and CARLA. OpenCDA provides driving scenarios, and CARLA provides maps and weather conditions. It includes 73 Scenes, six road types, and nine cities. It contains 12,000 frames of LiDAR point clouds and RGB camera images, along with 230,000 annotated 3D bounding boxes. The benchmark includes four LiDARs and four fusion strategies, a total of 16 models. The limitation is that it is based on a simulated environment and not reflect real-world scenarios. It also depends on the original OPV2V dataset.

The OPV2V-H dataset \cite{lu2024extensible} is an extension of the OPV2V dataset, which was released in 2024 to address the open heterogeneous collaborative perception problem by introducing the HEterogeneous ALliance (HEAL) framework. HEAL supports the integration of new and heterogeneous agents into a collaborative perception system with minimal retraining. The sensor configuration includes two LiDARs and four cameras per agent. HEAL enhances collaborative performance through real-world scenarios in which vehicles utilize different sensor suites and reduce training costs. The primary constraint of this dataset is that it requires BEV features and depends on the OPV2V dataset. Researchers from Shanghai Jiao Tong University, the University of Southern California, and Shanghai AI Laboratory collaborated to develop the IRegular V2V (IRV2V) dataset in 2023 \cite{wei2023asynchrony}. It aims to help and promote research on temporal asynchrony for collaborative perception. In the real world, agents send and receive data at different times due to communication delays and interruptions, which can cause problems in collaborative perception. To solve this issue, they proposed CoBEVFLOW, a method that handles asynchronous collaborative messages sent at irregular, continuous time stamps without discretization. With a bird’s eye view (BEV) flow, it avoids additional noise by transporting the original perceptual features. The limitation is that it is a synthetic dataset, and real-world validations are limited to the DAIR-V2X dataset.

The CP-UAV (CoPerception) dataset \cite{hu2022where2comm}, released in 2022, supports UAV-based collaborative perception using co-simulated environments from AirSim and CARLA. It includes 131,900 synchronized images collected by five UAVs flying at three altitudes in three virtual towns. Each UAV is equipped with five RGB cameras, and the dataset provides pixel-wise semantic segmentation, 2D/3D bounding boxes, Bird’s Eye View maps, and occlusion labels. Strengths include multi-agent perspectives and comprehensive annotations. However, being fully simulated, it may not capture real-world complexities and is limited to RGB data without additional sensors like LiDAR. CODD (Cooperative Driving Dataset) \cite{arnold2021fast} is an open-source synthetic dataset released in 2021 and created using CARLA. The dataset contains 108 sequences used for training, validation, and testing. It is used in cooperative 3D object detection, cooperative object tracking, and multi-agent SLAM. CODD’s strength is in capturing interactions between multiple agents. Using CARLA allows for generating diverse and controlled scenarios and facilitates reproducibility and scalability, but may not reflect real-world environments. It focuses only on LIDAR, lacking sensor modalities such as cameras or Radars. COMAP (COllective Multi-Agent Perception) \cite{yuan2021comap} is a synthetic dataset generated using the CARLA and SUMO simulators, released in 2021. It supports training and testing of 3D object and BEV detectors under varying vehicle-to-vehicle communication ranges. CARLA handles sensor data generation, while SUMO manages traffic flow and vehicle navigation using maps transformed from CARLA. The dataset allows exploration of overlapping and non-overlapping field-of-view scenarios in a controlled environment. However, it lacks real-world sensor noise and environmental complexity, and is limited to LiDAR input, without camera or Radar data.

\subsubsection{Real-World Datasets}
Open Mars \cite{li2024multiagent}, V2V4Real \cite{xu2023v2v4real}, LUCOOP \cite{axmann2023lucoop}, and T\&J Cooper \cite{chen2019cooper} datasets are collected from physical vehicle deployments in real-world settings, making them valuable for evaluating the robustness and scalability of autonomous driving systems. The Open MARS Dataset \cite{li2024multiagent} is a large-scale multimodal dataset focused on real-world multiagent and multitraversal autonomous driving. It was collected in a 20 km² area in Ann Arbor, Michigan, from October 2023 to March 2024, which includes urban, highway, residential, and campus environments. It comprises two subsets: Multitraversal (5,757 traversals across 66 locations) and Multiagent (53 scenes involving 2–3 vehicles), totaling over 1.4 million frames and 15,000 multiagent image/LiDAR frames. The dataset uses a rich sensor suite: 128-channel LiDAR, IMU, GPS, three narrow-angle cameras, and three wide-angle fisheye cameras, all synchronized at 10 Hz. It provides full 3D ego-pose estimations and geometric calibration, with a structure compatible with nuScenes tools. Strengths include high-resolution, well-calibrated multimodal data and close-proximity multiagent interaction. However, it's geographically limited to a single city, lacks semantic labels for perception tasks, and is limited to a specific sensor setup, which affect generalizability.

The V2V4Real dataset \cite{xu2023v2v4real} was released in 2023 to facilitate research on V2V cooperative perception through a collaboration between UCLA and five other institutions. The data was collected by two vehicles equipped with multimodal sensors driving together through diverse scenarios. The dataset covers 410 km and comprises 20,000 LiDAR frames, 40,000 RGB frames, 240,000 annotated 3D bounding boxes across five classes, and high-definition maps (HDMaps). V2V4Real supports cooperative 3D object detection, cooperative 3D object tracking, and sim-to-real domain adaptation. It offers real-world, multimodal data but is limited to only two vehicles and may not fully capture vehicle interactions in complex traffic scenarios. The LUCOOP (Leibniz University Cooperative Perception and Urban Navigation) dataset \cite{axmann2023lucoop} is a time-synchronized, multimodal dataset collected by three interacting measurement vehicles. It was released in 2023 and includes approximately 54,000 LiDAR frames, around 700,000 IMU measurements, and more than 2.5 hours of 10 Hz GNSS raw data. The dataset provides 3D bounding box annotations for evaluating object detection approaches, as well as highly accurate ground-truth poses for each vehicle throughout the measurement campaign.

The Cooper (Cooperative Perception for Connected Autonomous Vehicles) system was introduced in 2019 \cite{chen2019cooper}. It supports the fusion and aggregation of 3D point clouds from multiple vehicles. Each vehicle collects raw LiDAR data independently and transmits it to nearby vehicles, after which a data fusion algorithm combines the information. The authors also introduced the T\&J dataset, which demonstrates higher performance for vehicle collaboration than the KITTI dataset. Their sensor setup includes cameras, GPS sensors, Radars, and LiDARs. Cooper is one of the pioneering works in cooperative perception.

\subsection{Vehicle-to-Infrastructure (V2I) Datasets}
V2I datasets focus on data exchange between AVs and stationary infrastructure components, such as traffic signals, roadside units (RSUs), and cameras. These interactions provide vehicles with vital contextual data like signal phase and timing (SPaT), road conditions, or environmental alerts. V2I communication enhances decision-making and supports intelligent transportation services like adaptive signal control and emergency vehicle prioritization. Table \ref{Table: V2I} summarizes the core attributes and comparative insights of major V2I datasets. To meaningfully organize the datasets discussed in Table \ref{Table: V2I}, we categorize them based on their simulation-based or real-world origin. We divide the datasets into two groups: (1) simulation-based datasets, which are generated in virtual environments; and (2) real-world datasets, which are captured from physical deployments across roads and intersections.

\begin{table*}[htbp]
\caption{Summary and comparative analysis of the Vehicle-to-Infrastructure (V2I) Datasets in autonomous vehicles.}
\vspace{-0.4cm}
\label{Table: V2I}
\begin{center}
\resizebox{\textwidth}{!}{
\setlength{\tabcolsep}{4pt}
\begin{tabular}{llllllccccccc}  
\toprule

\multicolumn{3}{c}{} & \multicolumn{1}{c}{} & \multicolumn{2}{c}{\textbf{Frame}} & \multicolumn{1}{c}{} & \multicolumn{3}{c}{\textbf{Sensor}}  &\multicolumn{2}{c}{\textbf{Application}} \\ [1.5mm]

\cmidrule(l){5-6}
\cmidrule(l){8-10}
\cmidrule(l){11-12}

\multirow{1}{*}{\textbf{Ref}}     & \textbf{Dataset}   & \textbf{Year}            &  \textbf{Region}   & \textbf{Image} & \textbf{LiDAR} & \textbf{Format} 
                                      & \textbf{Camera}         & \textbf{Lidar} &  \textbf{Radar}   & \textbf{Detection} & \textbf{Perception}   & \textbf{Access}  \\  

\midrule

\multirow{1}{*}{\cite{fan2025benchmark}} &\ph{V2I-HD}  &\ph{2025}  &\ph{China} &\ph{15,254}  &0        &RGB, HD Maps &\ph{Yes}   &\ph{No}  &\ph{No}   &\ph{\cmark}   &\ph{\cmark} &  \href{https://arxiv.org/abs/2503.23963}{Drive}\\[4mm]

\myrowcolour
\multirow{1}{*}{\cite{liu2024v2x}} &\ph{V2X-DSI}  &\ph{2024}  &\ph{Carla} &\ph{14K}  &57K  &RGB, LiDAR &\ph{Yes}   &\ph{Yes}  &\ph{No}   &\ph{\cmark}   &\ph{\cmark} &  N/A\\[4mm]

\multirow{1}{*}{\cite{yang2024v2x}} &\ph{V2X-Radar${^*}^1$}  &\ph{2024}  &\ph{China} &\ph{40K}     &20K   &RGB, LiDAR, BBox &\ph{Yes}   &\ph{Yes}  &\ph{Yes}   &\ph{\cmark}  &\ph{\cmark} &  \href{http://openmpd.com/column/V2X-Radar}{Git}\\[4mm]

\myrowcolour
\multirow{1}{*}{\cite{zimmer2024tumtraf}} &\ph{TUMTraf-V2X}  &\ph{2024}  &\ph{Germany} &\ph{5K}    & 2K   &RGB, LiDAR, BBox &\ph{Yes}   &\ph{Yes}  &\ph{No}   &\ph{\cmark}   &\ph{\cmark} &  \href{https://tum-traffic-dataset.github.io/tumtraf-v2x/}{Git}\\[4mm]

\multirow{1}{*}{\cite{zhu2024otvic}} &\ph{OTVIC}  &\ph{2024}  &\ph{China} &\ph{15,045}     &\ph{15,045}   &RGB, LiDAR, GPS  &\ph{Yes}   &\ph{Yes}  &\ph{No}   &\ph{\cmark}  &\ph{\cmark} &  N/A\\[4mm]

\myrowcolour
\multirow{1}{*}{\cite{wang2024dair}} &\ph{DAIR-V2XReid${^*}^2$}  &\ph{2024}  &\ph{China} &\ph{2,556}     &0    &RGB, Vehicle-ID &\ph{Yes}   &\ph{No}  &\ph{No}    &\ph{\xmark} &\ph{\cmark} &  \href{https://github.com/Niuyaqing/DAIR-V2XReid}{Git}\\[4mm]

\multirow{1}{*}{\cite{ma2024holovic}} &\ph{HoloVIC}  &\ph{2024}  &\ph{China} &\ph{+100K}     &\ph{+100K}    &RGB, LiDAR, BBox  &\ph{Yes}   &\ph{Yes}  &\ph{No}   &\ph{\cmark}  &\ph{\cmark} &  \href{https://holovic.net/}{Web}\\[4mm]

\myrowcolour
\multirow{1}{*}{\cite{yu2023v2x}} &\ph{V2X-Seq}  &\ph{2023}  &\ph{China} &\ph{15K}     &\ph{15K}    &RGB, BBox, Trajectory  &\ph{Yes}   &\ph{Yes}  &\ph{No}   &\ph{\cmark}  &\ph{\cmark} &  \href{https://github.com/AIR-THU/DAIR-V2X-Seq}{Git}\\[4mm]

\multirow{1}{*}{\cite{bai2022pillargrid}} &\ph{CARTI}  &\ph{2022}  &\ph{Carla} &0    &\ph{11K}        &LiDAR   &\ph{No}   &\ph{Yes}  &\ph{No}   &\ph{\cmark}  &\ph{\cmark} &  \href{https://github.com/zwbai/CARTI_Dataset}{Git}\\[4mm]

\myrowcolour
\multirow{1}{*}{\cite{yu2022dair}} &\ph{DAIR-V2X}  &\ph{2022}  &\ph{China} &\ph{71,254}     &\ph{71,254}    &RGB, LiDAR, BBox  &\ph{Yes}   &\ph{Yes}  &\ph{No}   &\ph{\cmark}  &\ph{\cmark} &  \href{https://github.com/AIR-THU/DAIR-V2X}{Git}\\[4mm]

\multirow{1}{*}{\cite{arnold2020cooperative}} &\ph{COOPER}  &\ph{2020}  &\ph{Carla} &\ph{5K}      &\ph{5K}   &RGB, LiDAR, Depth   &\ph{Yes}   &\ph{Yes}  &\ph{No}   &\ph{\cmark}   &\ph{\cmark} &  \href{https://github.com/eduardohenriquearnold/coop-3dod-infra?tab=readme-ov-file}{Git}\\[4mm]

\bottomrule
\end{tabular}}
\end{center}
\footnotesize{ \scriptsize $*$ For the convenience of the reader, each dataset's original source—whether a website or GitHub repository hyperlink—is provided in the `Access' column to facilitate direct access and download.\\
${^*}^1$ V2X-DSI dataset additionally comprises 20K 4D-Radar data.\\
${^*}^2$ DAIR-V2XReid dataset is typically used for vehicle matching and re-identification tasks.
}
\end{table*}

\vspace{-0.3cm}
\subsubsection{Simulation-based Datasets}
V2X-DSI \cite{liu2024v2x}, CARTI \cite{bai2022pillargrid}, and COOPER \cite{arnold2020cooperative} datasets are generated in synthetic environments, allowing for precise ground-truth annotations and scenario control. The V2X-DSI dataset \cite{liu2024v2x} was released in 2024 using the CARLA simulator. It contains 56,984 frames in 57 diverse urban scenarios. It evaluates LiDAR sensors with 16, 32, 64, and 128 beam densities on CP scenarios to determine cost-effective sensor configurations. The strength is that it addresses the practical concern of the high cost of dense beam LiDAR sensors. It also provides benchmarking across multiple LiDAR configurations, which offers valuable insights for researchers. The simulated data may not reflect the complexity of the real world. The CARTI dataset \cite{bai2022pillargrid} is a cooperative dataset developed using CARLA to evaluate a deep-fusion algorithm called Grid-wise Fusion (GFF). It was introduced along with PillarGrid, a feature-level cooperative 3D object detection method, in 2022. The sensor setup includes LiDAR mounted on the vehicle and roadside LiDAR. The COOPER dataset \cite{arnold2020cooperative} was collected from simulated urban road scenarios using CARLA and was released in 2020. The sensor setup included LiDAR and a camera on the vehicle and LiDAR on the roadside infrastructure. The authors used data to evaluate a cooperative perception system for 3D object detection using early and late fusion. However, using synthetic data limits real-world scenarios. Additionally, there is no radar in the sensor setup.

\subsubsection{Real-World Datasets}
V2I-HD \cite{fan2025benchmark}, V2X-Radar \cite{yang2024v2x}, TUMTraf-V2X \cite{zimmer2024tumtraf}, OTVIC \cite{zhu2024otvic}, DAIR-V2XReID \cite{wang2024dair}, HoloVIC \cite{ma2024holovic}, V2X-Seq \cite{yu2023v2x}, and DAIR-V2X \cite{yu2022dair} datasets are collected from physical roadside units and vehicles in real-world urban and highway environments. The V2I-HD dataset \cite{fan2025benchmark}, released in 2025, is a real-world dataset and benchmark to help the development of online HD mapping for Vehicle-Infrastructure Cooperative Autonomous Driving (VICAD) by promoting vision-centric V2I systems. The dataset includes collaborative and synchronized camera frames from vehicles and roadside infrastructures, as well as HD map elements annotated by humans. The strengths of this dataset are real-world data and the utilization of camera data, which is more cost-efficient than LiDAR. However, relying solely on cameras and excluding other sensors limits its applicability.

The V2X-Radar dataset \cite{yang2024v2x} was released in 2024 and was collected using a connected vehicle platform and a roadside unit equipped with 4D Radar, LiDAR, and multi-view cameras. It contains 20,000 LiDAR frames, 40,000 camera images, 20,000 4D Radar samples, and 350,000 annotated boxes across five categories: vehicles, buses, trucks, pedestrians, and bicycles. The strength of this dataset is that it integrates 4D radar to enhance perception in challenging weather conditions, such as sunny, rainy, dusk, and nighttime. The data is collected from real-world driving scenarios. However, its geographical scope is limited, and the dataset may lack exposure to rare events due to its collection from only a 15-hour driving log. The TUMTraf-V2X dataset \cite{zimmer2024tumtraf} was collected in Germany, and it consists of 2,000 annotated point clouds, 5,000 annotated images, and more than 3,000 3D bounding boxes. The annotations cover eight specific classes and represent complex driving scenarios. The sensor setup for the infrastructure includes one LiDAR and four cameras, while the vehicle consists of a LiDAR, a camera, a GPS, and an IMU sensor. The strength lies in offering data from both vehicle and roadside viewpoints. It also covers complex traffic conditions. However, the dataset does not have geographic diversity. 

The OTVIC dataset \cite{zhu2024otvic} is a multimodal, multi-view dataset featuring online transmission from real-world scenes for vehicle-to-infrastructure cooperative 3D object detection. The data was collected from a highway in China and released in 2024. The dataset includes 14,045 frames and 24,452 manually annotated vehicles. OTVIC provides a realistic benchmark but has limitations in terms of geographic coverage and sensor modalities. The HoloVIC dataset \cite{ma2024holovic} was released in 2024 and contains 100,000 synchronized data frames, as well as 11.47 million annotated 3D bounding boxes. The supported tasks are monocular 3D detection, LiDAR 3D detection, multi-object detection, and multi-sensor multi-object tracking. The strengths of this dataset are its diverse sensor layout, comprehensive 3D annotations, and real-world driving scenarios. However, it is geographically limited and focuses only on short-term interactions.

The V2X-Seq dataset \cite{yu2023v2x} was collected from 28 intersections with 672 hours of data and was released in 2023. The dataset contains two parts: the sequential perception dataset, which includes 15,000 frames from 95 scenarios, and the trajectory forecasting dataset, which contains 80,000 vehicle-view scenarios, 80,000 infrastructure-view scenarios, and 50,000 cooperative-view scenarios. The dataset comprises real-world sequential data. However, geographical diversity and sensor modality are limited, which can affect the generalizability and accuracy of the algorithms based on these data. The DAIR-V2X dataset \cite{yu2022dair} was introduced in 2022 to support research in autonomous driving systems. This dataset comprises 71,254 LiDAR frames, 71,254 camera frames, and 3D annotations. The data was collected from 28 intersections in Beijing, China. The strengths of the dataset are that it contains real-world data and facilitates sensor fusion between the infrastructure and the vehicle. However, the data was collected in one city, lacking scenario diversity. The data was also collected using only LiDARs, so it is limited in viewpoint diversity.

\subsection{Vehicle-to-Everything (V2X) Datasets}
V2X datasets provide a comprehensive view of inter-connected communication between vehicles, infrastructure, pedestrians, and cloud services. This category encompasses datasets that facilitate both V2I and V2V communications. These datasets enable AVs to function cohesively within complex urban ecosystems, addressing scenarios involving vulnerable road users, real-time environmental updates, and coordinated traffic operations. V2X datasets are pivotal for building fully cooperative, connected, and automated mobility systems. Table \ref{Table: V2X} presents a structured comparison of leading V2X datasets and their application domains. In the following, we categorize the datasets presented in Table \ref{Table: V2X} according to their simulation-based or real-world origin. We divide them into two categories: (1) simulation-based datasets, which are generated using simulation platforms; and (2) real-world datasets, which are collected from physical vehicle deployments.

\begin{table*}[htbp]
\caption{Summary and comparative analysis of the Vehicle-to-Everything (V2X) Datasets in autonomous vehicles.}
\vspace{-0.4cm}
\label{Table: V2X}
\begin{center}
\resizebox{\textwidth}{!}{
\setlength{\tabcolsep}{4pt}
\begin{tabular}{llllllccccccc}  
\toprule

\multicolumn{3}{c}{} & \multicolumn{1}{c}{} & \multicolumn{2}{c}{\textbf{Frame}} & \multicolumn{1}{c}{} & \multicolumn{3}{c}{\textbf{Sensor}}  &\multicolumn{2}{c}{\textbf{Application}} \\ [1.5mm]

\cmidrule(l){5-6}
\cmidrule(l){8-10}
\cmidrule(l){11-12}

\multirow{1}{*}{\textbf{Ref}}     & \textbf{Dataset}   & \textbf{Year}            &  \textbf{Region}   & \textbf{Image} & \textbf{LiDAR} & \textbf{Format} 
                                      & \textbf{Camera}         & \textbf{Lidar} &  \textbf{Radar}   & \textbf{Detection} & \textbf{Segmentation}   & \textbf{Access}  \\  

\midrule

\multirow{1}{*}{\cite{luo2025mixed}} &\ph{Mixed-Signals}  &\ph{2025}  &\ph{Australia} &\ph{0}  &\ph{45.1K}   &LiDAR, BBox  &\ph{No}   &\ph{Yes}  &\ph{No}   &\ph{\cmark}  &\ph{\xmark}  & \href{https://mixedsignalsdataset.cs.cornell.edu/}{Web}\\[4mm]

\myrowcolour
\multirow{1}{*}{\cite{xiang2025v2x}} &\ph{V2X-ReaLO}  &\ph{2025}  &\ph{China} &\ph{25,028}  &\ph{25,028}   &RGB, LiDAR, ROS-bags  &\ph{Yes}   &\ph{Yes}  &\ph{No}   &\ph{\cmark}  &\ph{\xmark}  & N/A\\[4mm]

\multirow{1}{*}{\cite{wang2024deepaccident}} &\ph{DeepAccident}  &\ph{2024}  &\ph{Carla} &\ph{57K}  &\ph{57K}    &RGB, LiDAR, BEW &\ph{Yes}   &\ph{Yes}  &\ph{No}   &\ph{\cmark}  &\ph{\cmark}  &  \href{https://github.com/tianqi-wang1996/DeepAccident}{Git}\\[4mm]

\myrowcolour
\multirow{1}{*}{\cite{karvat2024adver}} &\ph{Adver-City}  &\ph{2024}  &\ph{Carla} &\ph{24,087}  &\ph{24,087}   &RGB, LiDAR, IMU/GNSS &\ph{Yes}   &\ph{Yes}  &\ph{No}   &\ph{\cmark}  &\ph{\cmark}  &  \href{https://labs.cs.queensu.ca/quarrg/datasets/adver-city/}{Web}\\[4mm]

\multirow{1}{*}{\cite{li2024multi}} &\ph{Multi-V2X}  &\ph{2024}  &\ph{Carla} &\ph{549K}  &\ph{146K}   &RGB, LiDAR, BBox &\ph{Yes}   &\ph{Yes}  &\ph{No}   &\ph{\cmark}  &\ph{\xmark}  &  \href{https://github.com/RadetzkyLi/Multi-V2X}{Git}\\[4mm]

\multirow{1}{*}{\cite{chen2024whales}} &\ph{WHALES}  &\ph{2024}  &\ph{Carla} &\ph{70K}  &\ph{17K}   &RGB, LiDAR, Trajectory &\ph{Yes}   &\ph{Yes}  &\ph{No}   &\ph{\cmark}  &\ph{\xmark}  & \href{https://github.com/chensiweiTHU/WHALES}{Git}\\[4mm]

\myrowcolour
\multirow{1}{*}{\cite{huang2024v2x}} &\ph{V2X-R}  &\ph{2024}  &\ph{China} &\ph{150,908}  &\ph{37,727}   &RGB, LiDAR, BBox &\ph{Yes}   &\ph{Yes}  &\ph{Yes}   &\ph{\cmark}  &\ph{\xmark}  & \href{https://github.com/ylwhxht/V2X-R}{Git}\\[4mm]

\multirow{1}{*}{\cite{zhou2024v2xpnp}} &\ph{V2XPnP}  &\ph{2024}  &\ph{USA} &\ph{208K}  &\ph{40K}   &RGB, BBox, Trajectory &\ph{Yes}   &\ph{Yes}  &\ph{Yes}   &\ph{\cmark}  &\ph{\xmark}  & \href{https://mobility-lab.seas.ucla.edu/v2xpnp/}{Web}\\[4mm]

\myrowcolour
\multirow{1}{*}{\cite{gamerdinger2024scope}} &\ph{SCOPE}  &\ph{2024}  &\ph{Carla} &\ph{17,600}  &\ph{17,600}   &RGB, LiDAR, BBox  &\ph{Yes}   &\ph{Yes}  &\ph{No}   &\ph{\cmark}  &\ph{\cmark}  & \href{https://ekut-es.github.io/scope/}{Git}\\[4mm]

\multirow{1}{*}{\cite{wang2024rcdn}} &\ph{OPV2V-N}  &\ph{2024}  &\ph{Carla} &\ph{11,464}  &\ph{11,464}   &RGB, LiDAR, BBox  &\ph{Yes}   &\ph{Yes}  &\ph{No}   &\ph{\cmark}  &\ph{\xmark}  & N/A\\[4mm]

\myrowcolour
\multirow{1}{*}{\cite{li2022v2x}} &\ph{V2X-Sim}  &\ph{2022}  &\ph{Carla} &\ph{10K}  &\ph{10K}    &RGB, LiDAR, GPS, Depth &\ph{Yes}   &\ph{Yes}  &\ph{Yes}   &\ph{\cmark}  &\ph{\cmark}  &  \href{https://github.com/ai4ce/V2X-Sim}{Git}\\[4mm]

\multirow{1}{*}{\cite{xu2022v2x}} &\ph{V2X-ViT}  &\ph{2022}  &\ph{Carla} &\ph{11,447}   &11,447    &RGB, LiDAR, BBox &\ph{Yes}   &\ph{Yes}  &\ph{No}   &\ph{\cmark}  &\ph{\xmark}  &  \href{https://github.com/DerrickXuNu/v2x-vit}{Git}\\[4mm]

\myrowcolour
\multirow{1}{*}{\cite{mao2022dolphins}} &\ph{DOLPHINS}  &\ph{2022}  &\ph{Carla} &\ph{42,376}  &\ph{42,376}   &RGB, LiDAR &\ph{Yes}   &\ph{Yes}  &\ph{No}   &\ph{\cmark}  &\ph{\xmark}  &  \href{https://dolphins-dataset.net/}{Web}\\[4mm]

\multirow{1}{*}{\cite{xiang2024v2x}} &\ph{V2X-Real}  &\ph{2022}  &\ph{USA} &\ph{171K}  &\ph{34K}   &RGB, LiDAR, BBox &\ph{Yes}   &\ph{Yes}  &\ph{No}   &\ph{\cmark}  &\ph{\xmark}  &  \href{https://mobility-lab.seas.ucla.edu/v2x-real/}{Web}\\[4mm]

\bottomrule
\end{tabular}}
\end{center}
\footnotesize{ \scriptsize $*$ For the convenience of the reader, each dataset's original source—whether a website or GitHub repository hyperlink—is provided in the `Access' column to facilitate direct access and download.
}
\end{table*}

\subsubsection{Simulation-based Datasets} DeepAccident \cite{wang2024deepaccident}, Adver-City \cite{karvat2024adver}, Multi-V2X \cite{li2024multi}, WHALES \cite{chen2024whales}, SCOPE \cite{gamerdinger2024scope}, OPV2V-N \cite{wang2024rcdn}, V2X-Sim \cite{li2022v2x}, V2X-ViT \cite{xu2022v2x}, and DOLPHINS \cite{mao2022dolphins} datasets are created using simulation platforms, offering precise ground-truth labels, synchronized multi-agent data, and diverse cooperative scenarios. DeepAccident dataset \cite{wang2024deepaccident}, released in 2024 by HKU, Huawei, and Dalian University of Technology, is a safety-focused synthetic dataset for accident prediction in V2X environments. It was generated by CARLA and features 57,000 frames and 285,000 samples of multi-agent interactions in 12 major intersection accident types. Each scene includes data from RGB cameras and LiDAR sensors on four vehicles and one roadside unit. Tasks include 3D detection, tracking, BEV segmentation, motion forecasting, and full-scene accident prediction. While highly structured and valuable for risk-aware learning, it lacks real-world sensor noise and requires heavy computation.

The Adver-City dataset \cite{karvat2024adver}, released in March 2025 by Queen’s University, is a large-scale synthetic dataset for evaluating cooperative perception in adverse weather. Built using CARLA and OpenCDA, it includes 24,000 frames and 890,000 annotations across 110 scenarios featuring rain, fog, glare, and varying lighting and traffic conditions. Inspired by U.S. National Highway Traffic Safety Administration (NHTSA) crash data, it focuses on dangerous driving areas like intersections and rural curves. Data comes from five agents (ego, RSUs, and vehicles) equipped with RGB, segmentation cameras, LiDAR, and GNSS/IMU. It supports tasks like detection, tracking, and segmentation, with a label format matching OPV2V. Strengths include diverse agents, weather types (including glare), and realistic layouts. Limitations include a lack of real-world sensor data. Multi-V2X \cite{li2024multi} is a large-scale synthetic cooperative perception benchmark released in 2024 by Tsinghua University. It is the first dataset specifically designed to assess autonomous driving performance under varying CAV (Connected and Autonomous Vehicle) penetration rates. Built using CARLA and SUMO, it includes 549,000 RGB images, 146,000 LiDAR frames, and 4.2 million annotated 3D bounding boxes across six object classes. The dataset simulates both V2V and V2I scenarios across urban, suburban, and rural settings. Vehicles and RSUs are equipped with RGB cameras, LiDARs, and GNSS modules. The six included towns feature diverse road topologies and traffic conditions. Multi-V2X supports 3D object detection, tracking, and benchmarking using fusion strategies (early, late, intermediate). Meanwhile, its synthetic nature limits realism and weather variation.

WHALES (Wireless enHanced Autonomous vehicles with Large number of Engaged agentS) \cite{chen2024whales} is a large-scale synthetic dataset released in 2024 by Tsinghua University. It is generated using CARLA, and features 70,000 RGB images, 17,000 LiDAR frames, and over 2 million annotated 3D bounding boxes, with an average of 8.4 agents per scene. WHALES supports tasks like object detection and CP benchmarking under realistic communication constraints. While it offers rich multimodal annotations and novel scheduling tasks, its synthetic nature limits realism, and its high computational demands may restrict accessibility. SCOPE (Synthetic COllective PErception) \cite{gamerdinger2024scope} is a 2024 synthetic multi-agent dataset from the University of Tübingen and FZI Karlsruhe, designed to advance collective perception in AVs. It includes over 17,600 frames across 40+ scenarios with up to 24 agents, featuring realistic sensor models (RGB, solid-state LiDAR), detailed weather simulation, and urban environments. SCOPE offers annotations for object detection and semantic segmentation, supporting research on perception resilience, sensor alignment, and domain adaptation. Its strengths include high-fidelity environmental modeling and diverse road user types. Limitations include simulation-based constraints on real-world unpredictability and the high computational cost of processing multimodal data.

V2X-Sim \cite{li2022v2x} is a large-scale synthetic dataset released in 2022 by NYU, USC, and Shanghai Jiao Tong University. Built using CARLA and SUMO, it simulates V2X scenarios with synchronized multi-agent sensor data from LiDAR, RGB, semantic, and depth cameras on both vehicles and infrastructure. V2X-Sim’s strengths lie in its early contribution to multimodal, multi-agent CP benchmarks and its compatibility with open-source tools. However, its simulated nature limits realism in terms of sensor noise and unpredictable driving behavior. DOLPHINS (Dataset for cOLlaborative Perception enabled Harmonious and INterconnected Self-driving) \cite{mao2022dolphins} is a large synthetic benchmark released in 2022 by Tsinghua University to support both V2V and V2I collaborative perception. It features synchronized full-HD images and 64-line LiDAR point clouds collected from ego vehicles, auxiliary vehicles, and RSUs across six diverse scenarios, including urban, highway, mountain, and rural settings under weather conditions like rain and fog. While DOLPHINS excels in scope and detail, it is limited by its synthetic nature, low frame rate for high-speed scenes, and lack of radar data.

\vspace{-0.2cm}
\subsubsection{Real-World Datasets} Mixed-Signals \cite{luo2025mixed}, V2X-ReaLO \cite{xiang2025v2x}, V2X-R \cite{huang2024v2x}, V2XPnP \cite{zhou2024v2xpnp}, and V2X-Real \cite{xiang2024v2x} datasets are collected from physical vehicle deployments in real-world settings. Mixed-Signals \cite{luo2025mixed} is a real-world V2X dataset released in 2025 by Cornell and the University of Sydney. It features data from three vehicles and one RSU with different LiDAR setups, as well as 45,100 point clouds and 240,600 3D bounding boxes. Its main strengths are showcasing sensor heterogeneity and real-world multi-agent perception. Limitations include a single location, short recording duration, and lack of camera or radar data. V2X-ReaLO \cite{xiang2025v2x} is a real-world dataset released in March 2025 by UCLA. Built on V2X-Real, it offers 25,028 frames (6,850 annotated keyframes) from two connected vehicles and two RSUs, supporting real-time evaluation of early, late, and intermediate fusion strategies. Its key strength lies in practical and real-time V2X fusion under real-world conditions, with open-source support for modular experimentation. Limitations include a lack of radar data and limited geographic diversity.

V2X-R \cite{huang2024v2x} is a multimodal synthetic dataset released in March 2025 to advance robust autonomous driving in adverse weather. It includes 12,079 driving scenarios with 37,727 LiDAR and radar point clouds, 150,908 RGB images, and over 170,000 3D bounding boxes. Additionally, it enables enhanced 3D object detection under fog and snow. Despite being synthetic and resource-intensive, V2X-R offers a critical platform for developing weather-resilient, multi-agent, multimodal autonomous perception systems. V2XPnP \cite{zhou2024v2xpnp} is a large-scale real-world benchmark introduced by University of California researchers in December 2024. It includes around 40,000 frames across 100 vehicle-centric and 63 infrastructure-centric scenarios. Its advantages is combining temporal consistency with full V2X coverage, which is ideal for studying real-time multi-agent collaboration. However, it lacks detailed sensor specifications and demands high computational resources. 

\vspace{-0.2cm}
\subsection{Infrastructure-to-Infrastructure (I2I) Datasets}
I2I datasets encompass coordinated sensing and data sharing among multiple infrastructure nodes, such as inter-linked traffic cameras or RSUs. This communication paradigm supports large-scale traffic management, surveillance handoffs, and synchronized control across urban regions.I2I datasets help connect and coordinate multiple sensor systems that are spread out across different infrastructure components. This integration enhances redundancy, resilience, and scalability in smart city deployments. Table \ref{Table: I2I} provides an in-depth comparison of I2I datasets, highlighting their scope and infrastructure coordination capabilities.

\begin{table*}[htbp]
\caption{Summary and comparative analysis of the Infrastructure-to-Infrastructure (I2I) Datasets in autonomous vehicles.}
\vspace{-0.3cm}
\label{Table: I2I}
\begin{center}
\resizebox{\textwidth}{!}{
\setlength{\tabcolsep}{4pt}
\begin{tabular}{llllllccccccc}  
\toprule

\multicolumn{3}{c}{} & \multicolumn{1}{c}{} & \multicolumn{2}{c}{\textbf{Frame}} & \multicolumn{1}{c}{} & \multicolumn{3}{c}{\textbf{Sensor}}  &\multicolumn{2}{c}{\textbf{Application}} \\ [1.5mm]

\cmidrule(l){5-6}
\cmidrule(l){8-10}
\cmidrule(l){11-12}

\multirow{1}{*}{\textbf{Ref}}     & \textbf{Dataset}   & \textbf{Year}            &  \textbf{Region}   & \textbf{Image} & \textbf{LiDAR} & \textbf{Format} 
                                      & \textbf{Camera}         & \textbf{Lidar} &  \textbf{Radar}   & \textbf{Detection} & \textbf{Perception}   & \textbf{Access}  \\  

\midrule

\multirow{1}{*}{\cite{hao2024rcooper}} &\ph{RCooper}  &\ph{2024}  &\ph{China} &\ph{50K}  &\ph{30K}   &RGB, LiDAR, BBox  &\ph{Yes}   &\ph{Yes}  &\ph{No}   &\ph{\cmark}  &\ph{\xmark}  & \href{https://github.com/AIR-THU/DAIR-RCooper}{Git}\\[4mm]

\multirow{1}{*}{\cite{zhang2024inscope}} &\ph{InScope}  &\ph{2024}  &\ph{China} &\ph{0}  &\ph{21,317}   &LiDAR, BBox, Trajectory  &\ph{No}   &\ph{Yes}  &\ph{No}   &\ph{\cmark}  &\ph{\xmark}  & \href{https://github.com/xf-zh/InScope}{Git}\\[4mm]

\bottomrule
\end{tabular}}
\end{center}
\footnotesize{ \scriptsize $*$ For the convenience of the reader, each dataset's original source—whether a website or GitHub repository hyperlink—is provided in the `Access' column to facilitate direct access and download.\\
}
\end{table*}

The RCOOPER dataset \cite{hao2024rcooper} was collected from simulated urban scenarios using CARLA and was released in 2020. The sensor setup included LiDAR, a camera on the vehicle, and LiDAR on the roadside infrastructure. The authors used the data to evaluate a cooperative perception system for 3D object detection using early and late fusion. It is one of the premier studies on cooperative perception. However, using synthetic data limits real-world applicability. There is no radar in the sensor setup. The InScope dataset \cite{zhang2024inscope}, released in 2024, was developed to address occlusion challenges in autonomous driving. It includes 187,787 annotated 3D bounding boxes and 303 tracking trajectories. The dataset supports four benchmark tasks: collaborative 3D object detection, multi-source fusion, domain adaptation, and multi-object tracking. Its ability to mitigate occlusion through infrastructure-side sensing is one of its main advantages; however, its drawbacks include  limited geographic diversity and access requirements.

In addition to categorizing datasets by perception setting (ego-vehicle, roadside, and V2X) and task supervision (detection, segmentation, and tracking), it is important to consider whether datasets reflect the conditions encountered in real-world deployment. Real-world autonomous driving performance is often dictated by adverse operating conditions rather than nominal daytime scenes. In particular, nighttime scenarios with low illumination and rainy or wet-weather conditions can significantly degrade visual quality, introduce sensor noise and reflections, and expose failure modes that may not be evident under clear daytime conditions. To support practical dataset selection and more realistic benchmarking, we provide a consolidated summary of whether representative datasets include nighttime scenes and rain/wet-weather conditions. Table \ref{Table: weather} highlights datasets that contain such adverse conditions, enabling researchers to quickly identify suitable benchmarks for robustness evaluation and to reveal coverage gaps that remain underexplored in current AV perception research.

\begin{table*}[h]
\caption{Adverse weather and illumination coverage of representative autonomous driving datasets.}
\vspace{-0.4cm}
\label{Table: weather}
\begin{center}
\resizebox{\textwidth}{!}{
\setlength{\tabcolsep}{4pt}
\begin{tabular}{lllccccccccc}
\toprule

\multicolumn{3}{c}{} &
\multicolumn{6}{c}{\textbf{Condition}} &
\multicolumn{3}{c}{\textbf{Sensor}} \\ [1.5mm]

\cmidrule(l){4-9}
\cmidrule(l){10-12}

\textbf{Ref} & \textbf{Dataset} & \textbf{Setting} &
\textbf{Night} & \textbf{Rain} & \textbf{Fog} & \textbf{Snow} & \textbf{Glare} & \textbf{Wet} &
\textbf{Camera} & \textbf{Lidar} & \textbf{Radar} \\
\midrule

\cite{kent2024msu}          & MSU-4S            & Ego-vehicle & \cmark & \cmark & \xmark & \cmark & \xmark & \xmark & Yes & Yes & Yes \\
\cite{zheng2024omnihd}      & OmniHD-Scenes     & Ego-vehicle & \cmark & \cmark & \xmark & \xmark & \xmark & \xmark & Yes & Yes & Yes \\
\cite{alibeigi2023zenseact} & Zenseact Open (ZOD) & Ego-vehicle & \cmark & \cmark & \xmark & \cmark & \cmark & \xmark & Yes & Yes & Yes \\
\cite{diaz2022ithaca365}    & Ithaca365         & Ego-vehicle & \cmark & \cmark & \xmark & \cmark & \xmark & \xmark & Yes & Yes & No  \\
\cite{mao2021one}           & ONCE              & Ego-vehicle & \cmark & \cmark & \xmark & \xmark & \xmark & \xmark & Yes & Yes & No  \\
\cite{deziel2021pixset}     & Leddar PixSet     & Ego-vehicle & \cmark & \cmark & \xmark & \xmark & \xmark & \xmark & Yes & Yes & Yes \\
\cite{xiao2021pandaset}     & PandaSet          & Ego-vehicle & \cmark & \xmark & \xmark & \xmark & \xmark & \xmark & Yes & Yes & No  \\
\cite{geyer2020a2d2}        & A2D2              & Ego-vehicle & \xmark & \cmark & \xmark & \xmark & \xmark & \xmark & Yes & Yes & No  \\
\cite{sun2020scalability}   & Waymo Open        & Ego-vehicle & \cmark & \cmark & \xmark & \xmark & \xmark & \xmark & Yes & Yes & No  \\
\cite{pham20203d}           & A*3D              & Ego-vehicle & \cmark & \cmark & \xmark & \xmark & \xmark & \xmark & Yes & Yes & Yes \\
\cite{caesar2020nuscenes}   & nuScenes          & Ego-vehicle & \cmark & \cmark & \xmark & \xmark & \xmark & \xmark & Yes & Yes & Yes \\
\cite{xue2019blvd}          & BLVD              & Ego-vehicle & \cmark & \xmark & \xmark & \xmark & \xmark & \xmark & Yes & Yes & No  \\
\cite{schafer2018commute}   & Comma2K19         & Ego-vehicle & \cmark & \cmark & \xmark & \xmark & \xmark & \xmark & Yes & No  & No  \\
\cite{yu2018bdd100k}        & BDD100K           & Ego-vehicle & \cmark & \cmark & \cmark & \xmark & \xmark & \xmark & Yes & No  & No  \\
\cite{barnes2020oxford}     & Oxford RobotCar   & Ego-vehicle & \cmark & \cmark & \xmark & \cmark & \xmark & \cmark & Yes & Yes & No  \\

\cite{zhu2024roscenes}        & RoScenes       & Roadside & \cmark & \cmark & \xmark & \xmark & \xmark & \xmark & Yes & No  & No \\
\cite{zimmer2023tumtraf}      & TUMTraf        & Roadside & \cmark & \cmark & \xmark & \xmark & \xmark & \xmark & Yes & Yes & No \\
\cite{cress2022a9}            & A9-Dataset     & Roadside & \cmark & \cmark & \xmark & \xmark & \xmark & \xmark & Yes & Yes & No \\
\cite{ye2022rope3d}           & Rope3D         & Roadside & \cmark & \cmark & \xmark & \xmark & \xmark & \xmark & Yes & Yes & No \\

\cite{li2024multiagent}      & Open MARS     & V2V & \cmark & \cmark & \xmark & \xmark & \cmark & \xmark & Yes & Yes & No \\

\cite{chen2025automated}      & CODA-LM           & V2L & \cmark & \cmark & \xmark & \xmark & \xmark & \xmark & Yes & No  & No \\
\cite{sima2024drivelm}       & DriveLM         & V2L & \cmark & \cmark & \xmark & \xmark & \xmark & \xmark & Yes & Yes & No \\
\cite{inoue2024nuscenes}     & NuScenes-MQA    & V2L & \cmark & \cmark & \xmark & \xmark & \xmark & \xmark & Yes & No  & No \\
\cite{wang2025omnidrive}     & OmniDrive       & V2L & \cmark & \cmark & \xmark & \xmark & \xmark & \xmark & Yes & No  & No \\
\cite{tian2024tokenize}      & TOKEN          & V2L & \cmark & \cmark & \xmark & \xmark & \xmark & \xmark & Yes & No  & No \\
\cite{qian2024nuscenes}      & NuScenes-QA     & V2L & \cmark & \cmark & \xmark & \xmark & \xmark & \xmark & Yes & Yes & No \\
\cite{ding2024holistic}      & NuInstruct      & V2L & \cmark & \cmark & \xmark & \xmark & \xmark & \xmark & Yes & No  & No \\

\cite{yang2024v2x}           & V2X-Radar     & V2I & \cmark & \cmark & \cmark & \cmark & \xmark & \cmark & Yes & Yes & Yes \\
\cite{zimmer2024tumtraf}     & TUMTraf-V2X   & V2I & \cmark & \xmark & \xmark & \xmark & \xmark & \xmark & Yes & Yes & No  \\
\cite{yu2023v2x}             & V2X-Seq       & V2I & \cmark & \cmark & \xmark & \xmark & \xmark & \xmark & Yes & Yes & No  \\
\cite{yu2022dair}            & DAIR-V2X      & V2I & \cmark & \cmark & \xmark & \xmark & \xmark & \xmark & Yes & Yes & No  \\

\cite{wang2024deepaccident}  & DeepAccident   & V2X & \cmark & \cmark & \xmark & \xmark & \xmark & \cmark & Yes & Yes & No  \\
\cite{karvat2024adver}       & Adver-City     & V2X & \cmark & \cmark & \cmark & \xmark & \cmark & \cmark & Yes & Yes & No  \\
\cite{huang2024v2x}          & V2X-R          & V2X & \xmark & \xmark & \cmark & \cmark & \xmark & \xmark & Yes & Yes & Yes \\
\cite{gamerdinger2024scope}  & SCOPE          & V2X & \cmark & \cmark & \cmark & \xmark & \xmark & \cmark & Yes & Yes & No  \\
\cite{mao2022dolphins}       & DOLPHINS       & V2X & \xmark & \cmark & \cmark & \xmark & \xmark & \xmark & Yes & Yes & No  \\

\bottomrule
\end{tabular}}
\end{center}
\footnotesize{ \scriptsize $*$ Glare includes low-sun and strong backlighting.\\
$*$ Wet denotes wet roads/reflections, with or without active rainfall.\\
}
\end{table*}

Beyond acting as a quick reference, Table \ref{Table: weather} enables a condition-driven view of dataset suitability that is often missing from conventional summaries focused on sensors, labels, and tasks. By explicitly marking adverse factors—night, rain, fog, snow, glare, and wet-road reflections—across ego-vehicle, roadside, and V2X settings, the table helps researchers align dataset choice with the intended deployment regime (e.g., low-illumination intersection reasoning, rain-induced visibility degradation, or fog/snow attenuation effects). This is particularly valuable for safety-critical evaluation because many perception and reasoning models exhibit strong performance under nominal conditions yet fail disproportionately under illumination shifts, specular reflections, and precipitation-related artifacts. As a result, the consolidated view in Table \ref{Table: weather} supports more defensible experimental design: it clarifies which benchmarks can legitimately support robust claims and which should be complemented by additional datasets or controlled stress tests.

The table also reveals important coverage gaps and imbalance patterns that motivate future dataset development. While several widely used benchmarks include night and rain, fewer datasets explicitly document rarer but high-impact conditions such as fog, snow, glare, or wet-road reflectance, especially in cooperative/V2X regimes where communication and multi-view fusion introduce additional failure modes. This indicates that current evaluations may underestimate real-world risk in precisely the scenarios where perception uncertainty and sensing artifacts are most severe. Consequently, Table \ref{Table: weather} highlights opportunities for more comprehensive benchmarking, including (i) cross-dataset generalization under matched adverse conditions, (ii) condition-specific ablations (e.g., day-to-night transfer, clear-to-rain transfer), and (iii) the creation of cooperative datasets with explicit annotation of adverse weather and illumination factors to better reflect operational driving environments.

\section{Object Detection in Autonomous Vehicles}
\label{sec: object detection}
\begin{figure}[H]
    \centering
    \centerline{\includegraphics[width=1\textwidth]{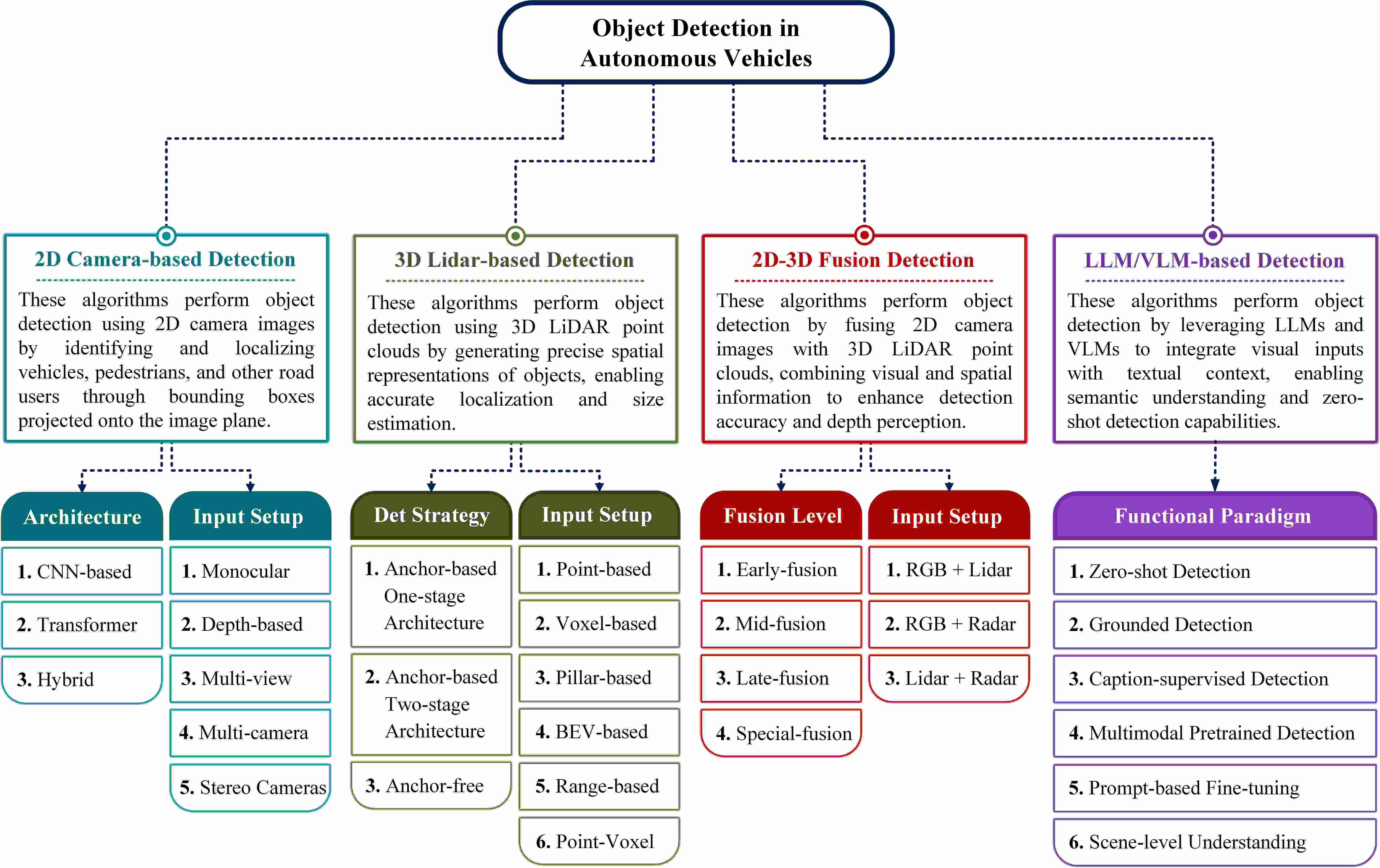}}
    \caption{A comprehensive taxonomy of object detection methods in AVs, categorized into four primary types. Each category includes representative subtypes based on sensor configuration, data representation, fusion strategy, and model architecture.}
    \label{fig: AV algorithms}
\end{figure}

Object detection plays a pivotal role in AV perception systems by enabling real-time understanding of the driving environment. Accurate detection of surrounding vehicles, pedestrians, cyclists, and static obstacles is essential for safe navigation, decision-making, and Path planning. Over the past decade, significant advances have been made in object detection techniques, driven by developments in sensor technology, deep learning, transformer-based architectures, and more recently, methods based on Large Language Models (LLMs) and Vision-Language Models (VLMs). In AV systems, object detection pipelines are typically built upon a variety of sensor modalities, such as RGB cameras, LiDAR, and radar, with details discussed in Section \ref{sec: AV Sensors}. Building on this foundation, modern object detection approaches can be broadly categorized based on the primary input modality or through the use of multi-modal fusion strategies. Generally, object detection methods in AVs can be broadly categorized into one of the following four types: 2D camera-based detection, 3D LiDAR-based detection, 2D–3D fusion detection, and LLM/VLM-based detection. Figure \ref{fig: AV algorithms} provides a comprehensive overview of these detection approaches along with their sensor modalities, learning paradigms, and model architectures.

\vspace{-0.4cm}
In 2D camera-based detection, algorithms rely on RGB imagery to identify and localize objects using bounding boxes projected onto the image plane \cite{chen2025object}. These methods can be further classified by their underlying architecture into three main groups: CNN-based (e.g., YOLOv4, Faster R-CNN), Transformer-based (e.g., DETR, Deformable DETR), and Hybrid models that integrate convolutional backbones with transformer modules (e.g., Swin Transformer + Faster R-CNN). At the same time, 2D detection approaches can also be categorized by their input configuration, including monocular (single-view RGB), depth-based monocular, multi-view geometry-based, multi-camera system-level, and stereo camera setups, each tailored to specific perception requirements and environmental conditions. In contrast, 3D LiDAR-based detection methods operate on raw or preprocessed point clouds to generate precise spatial representations of objects. These techniques encompass different detection strategies, including anchor-based one-stage detectors (e.g., SECOND, YOLO3D), anchor-based two-stage detectors (e.g., PV-RCNN, Voxel R-CNN), and anchor-free detectors (e.g., 3DSSD, YOLOv8-3D), each offering trade-offs between speed, localization accuracy, and robustness to sparse data. 

In parallel, 3D detection approaches can also be categorized by their input setup, including point-based, voxel-based, pillar-based, BEV-based, point-voxel-based, and range-based. 2D–3D fusion detection combines visual and geometric information from both cameras and LiDAR sensors to enhance object recognition under challenging conditions. Fusion strategies are typically classified as early-fusion, mid-fusion, late-fusion, or special-fusion (not strictly fitting into the previous three). Additionally, they can be categorized based on their input modalities to three different categories. Lastly, LLM/VLM-based detection approaches utilize large language and vision-language models to incorporate textual context into visual reasoning, enabling semantic-level understanding and generalization across unseen categories. These techniques include zero-shot detection, grounded detection (region-text alignment), caption-supervised Detection (language-augmented supervision), multimodal pretrained detection (Joint vision-language models), prompt-based fine-tuning (domain-adapted queries), and scene-level understanding. This categorization provides a unique framework and valuable insight into the current landscape of object detection in AVs, guiding future research toward more robust and generalizable perception systems \cite{zha2025real}.

The following subsections highlight recent advancements across diverse sensor modalities and model architectures. Subsection \ref{sec: 2D} reviews 2D camera-based object detection methods, while Subsection \ref{sec: 3D} explores 3D object detection techniques using LiDAR (point-cloud). Subsection \ref{sec: 2D-3D} examines 2D–3D fusion strategies that combine camera (image) and LiDAR (point-cloud) data, categorized by early, mid, and late fusion stages. Finally, Subsection \ref{sec: LLMs/VLMs} focuses on emerging LLMs/VLMs-based object detection methods that enable open-vocabulary and prompt-driven detection through multimodal understanding.

\subsection{2D Camera-based Approaches}
\label{sec: 2D}
\begin{figure}[H]
    \centering
    \centerline{\includegraphics[width=0.8\textwidth]{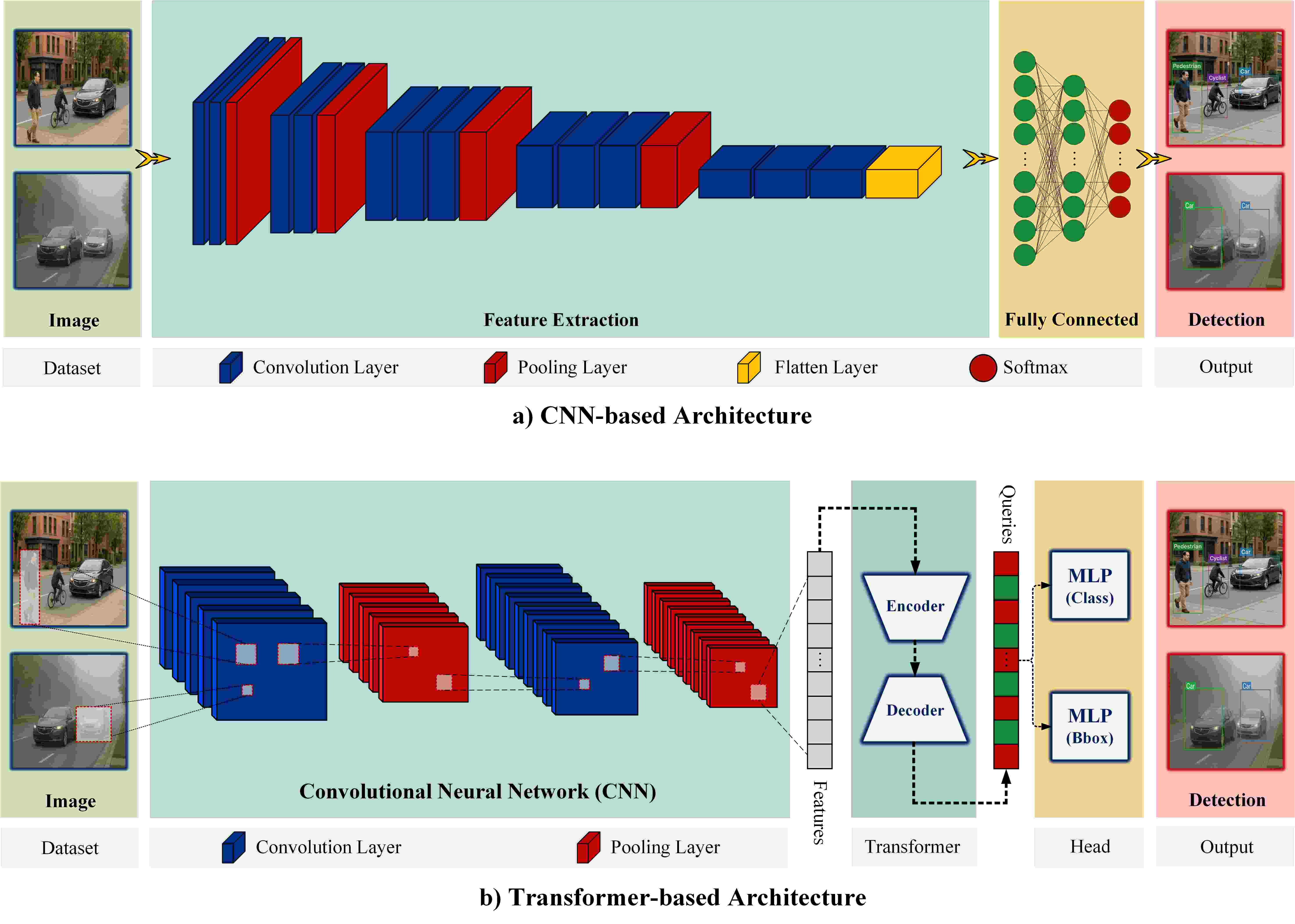}}
    \caption{Overall framework of the 2D camera-based approaches in the context of autonomous driving systems. \textbf{a)} general architecture of CNN-based models, \textbf{b)} general architecture of Transformer-based models.}
    \label{fig: 2D architecture}
\end{figure}

Identifying and localizing key objects in the driving scene, including vehicles, pedestrians, and cyclists, is essential for the safe operation of autonomous driving systems. In the context of 2D camera-based object detection, existing algorithms can generally be classified into three categories based on their architectures: CNN-based, Transformer-based, and hybrid approaches that integrate both CNN and Transformer components. As illustrated in Figure \ref{fig: 2D architecture}, these approaches follow an overall framework where camera-captured RGB images are processed through feature extraction, region proposal, and object classification/localization stages. In CNN-based models, convolutional layers learn hierarchical spatial features, gradually refining them for detection tasks. In contrast, Transformer-based models leverage self-attention mechanisms to capture long-range dependencies and global context across the image. Hybrid approaches aim to combine the localized feature learning of CNNs with the reasoning capability of Transformers, enhancing detection accuracy and robustness in driving scenes.

\vspace{-0.3cm}
This section presents all existing 2D camera-based object-detection algorithms for autonomous vehicles published from 2016 to 2025. To ensure a consistent and fair comparison, we organize the results into four separate tables, as these methods report performance on different datasets and use varying evaluation metrics. Table \ref{table: 2D_Kitti} provides a comparative analysis of 2D object detection methods using camera-based data from the KITTI dataset, where both the input and output are in 2D (RGB images to 2D bounding boxes). Table \ref{table: 3D_Kitti} focuses on 3D object detection methods that use camera-based KITTI data, but generate 3D bounding boxes from 2D image inputs. Table \ref{table: 3D_NuScenes} presents a comparative analysis of 2D/3D object detection methods using camera-based data from the NuScenes dataset, where the input is 2D images and the output is either 2D or 3D bounding boxes. Finally, Table \ref{table: yolo_detection} summarizes YOLO-based methods evaluated on the camera-based KITTI dataset for 2D object detection. This structured categorization enables consistent benchmarking and highlights performance trends across architectures, datasets, and detection tasks. It should be noted that the values highlighted in red represent the best performance achieved for each evaluation metric across all listed methods.

\begin{table}[htbp]
\centering
\caption{Comparative analysis of 2D object detection methods using the camera-based KITTI dataset.}
\vspace{-0.4cm}
\label{table: 2D_Kitti}
\begin{center}
\resizebox{\textwidth}{!}{%
  \setlength{\tabcolsep}{4pt}%
  \begin{tabular}{lllll ccc ccc ccc c}
    \toprule
    \multicolumn{5}{c}{} 
      & \multicolumn{3}{c}{\textbf{Car}} 
      & \multicolumn{3}{c}{\textbf{Pedestrian}} 
      & \multicolumn{3}{c}{\textbf{Cyclist}} \\[1mm]
    \cmidrule(l){6-8}\cmidrule(l){9-11}\cmidrule(l){12-14}
    \textbf{Ref} & \textbf{Method} & \textbf{Year} & \textbf{Backbone} & \textbf{Modality}
      & \textbf{Easy} & \textbf{Medium} & \textbf{Hard}
      & \textbf{Easy} & \textbf{Medium} & \textbf{Hard}
      & \textbf{Easy} & \textbf{Medium} & \textbf{Hard} 
     & \textbf{Runtime}
      \\
    \midrule

    \multirow{1}{*}{\cite{yu2025vikienet}} 
      & ViKIENet & 2025 & BiSeNet V2 & Mono
      & 98.63 & \red{\textbf{98.06}} & \red{\textbf{93.21}}
      & $-$ & $-$ & $-$
      & $-$ & $-$ & $-$  & 0.04 Sec \\[2mm]
      \myrowcolour
      & \multicolumn{1}{l}{\textbf{Innovation:}}
        & \multicolumn{13}{l}{\parbox[t]{1\linewidth}{Selects virtual key instances for efficient instance-aware LiDAR–camera fusion with reduced redundancy.}}\\[2mm]

    \multirow{1}{*}{\cite{yu2025vikienet}} 
      & ViKIENet-R & 2025 & BiSeNet V2 & Mono
      & 98.42 & 97.57 & 93.21
      & $-$ & $-$ & $-$
      & $-$ & $-$ & $-$ & 0.06 Sec \\[2mm]
      \myrowcolour
      & \multicolumn{1}{l}{\textbf{Innovation:}}
        & \multicolumn{13}{l}{\parbox[t]{1\linewidth}{Applies rotation-equivariant sparse convolutions on virtual key instances for efficient robust detection.}} \\[2mm] 
        

    \multirow{1}{*}{\cite{liang2022aspectnet}} 
      & AspectNet & 2022 & ResNet-50 & Mono
      & 96.16 & 93.87  & 88.17
      & 87.17 & 76.57 & 68.42
      & 86.15 & 77.11 & 71.24 & \textbf{$-$} \\[2mm]
      \myrowcolour
      & \multicolumn{1}{l}{\textbf{Innovation:}}
        & \multicolumn{13}{l}{\parbox[t]{1\linewidth}{Adds an aspect ratio prediction head with improved sampling and loss to handle imbalance.}} \\[2mm]

    \multirow{1}{*}{\cite{choi2019gaussian}} 
      & Gaussian YOLOv3 & 2019 & Darknet-53 & Mono
      & 90.61 & 90.20 & 81.19
      & 87.84 & 79.57 & 72.30
      & 89.31 & 81.30 & 80.20 &  0.02 Sec \\[2mm]
      \myrowcolour
      & \multicolumn{1}{l}{\textbf{Innovation:}}
        & \multicolumn{13}{l}{\parbox[t]{1\linewidth}{Models bounding boxes as Gaussians with predicted localization uncertainty.}} \\[2mm]

    \multirow{1}{*}{\cite{choi2019gaussian}} 
      & Gaussian YOLOv3\textsuperscript{†} & 2019 & Darknet-53 & Mono
      & 98.74 & 90.48 & 89.47
      & 87.85 & \red{\textbf{79.96}}  & \red{\textbf{76.81}} 
      & 90.08 & 86.59 & \red{\textbf{81.09}}   &  0.02 Sec  \\[2mm]
      \myrowcolour
      & \multicolumn{1}{l}{\textbf{Innovation:}}
        & \multicolumn{13}{l}{\parbox[t]{1\linewidth}{Enhanced NMS and loss weighting for Gaussian outputs.}} \\[2mm]

    \multirow{1}{*}{\cite{hu2018sinet}} 
      & SINet & 2019 & PVA-Net & Mono
      & \red{\textbf{99.11}} & 90.59 & 79.77
      & \red{\textbf{88.09}}  & 79.22 & 70.30
      & \red{\textbf{94.41}}   & \red{\textbf{86.61}}  & 80.68 &  0.3 Sec \\[2mm]
      \myrowcolour
      & \multicolumn{1}{l}{\textbf{Innovation:}}
        & \multicolumn{13}{l}{\parbox[t]{1\linewidth}{Context-aware RoI pooling with multi-branch heads to reduce scale variance.}} \\[2mm]

    \multirow{1}{*}{\cite{yi2019assd}} 
      & ASSD & 2019 & ResNet-101 & Mono
      & 89.28 & 89.95 & 82.11
      & 69.07 & 62.49 & 60.18
      & 75.23 & 76.16 & 72.83 &  0.01 Sec \\[2mm]
      \myrowcolour
      & \multicolumn{1}{l}{\textbf{Innovation:}}
      & \multicolumn{13}{l}{\parbox[t]{1\linewidth}{In-place self-attention for global context modeling in SSD.}} \\[2mm]

    \multirow{1}{*}{\cite{zhang2018single}} 
      & RefineDet & 2018 & VGG-16 & Mono
      & 98.96 & 90.44 & 88.82
      & 84.40 & 77.44 & 73.52
      & 86.33 & 80.22 & 79.15 &  0.15 Sec  \\[2mm]
      \myrowcolour
      & \multicolumn{1}{l}{\textbf{Innovation:}}
        & \multicolumn{13}{l}{\parbox[t]{1\linewidth}{Two-step anchor refinement with transfer connection blocks.}} \\[2mm]

    \multirow{1}{*}{\cite{liu2018receptive}} 
      & RFBNet & 2018 & VGG-16 & Mono
      & 87.31 & 87.27 & 84.44
      & 66.16 & 61.77 & 58.04
      & 74.89 & 72.05 & 71.01 &  0.01 Sec \\[2mm]
      \myrowcolour
      & \multicolumn{1}{l}{\textbf{Innovation:}}
        & \multicolumn{13}{l}{\parbox[t]{1\linewidth}{Introduces the Receptive Field Block (RFB), a module that enhances feature discriminability.}} \\[2mm]

    \multirow{1}{*}{\cite{wu2017squeezedet}} 
      & SqueezeDet & 2017 & SqueezeNet & Mono
      & 90.20 & 84.70 & 73.90
      & 77.10 & 68.30 & 65.80
      & 82.90 & 75.40 & 72.10 &  0.01 Sec  \\[2mm]
      \myrowcolour
      & \multicolumn{1}{l}{\textbf{Innovation:}}
        & \multicolumn{13}{l}{\parbox[t]{1\linewidth}{Fully convolutional single-stage detection with ConvDet, optimized for speed and size.}} \\[2mm]

    \multirow{1}{*}{\cite{cai2016unified}} 
      & MS-CNN & 2016 & VGG-Net & Mono
      & 90.03 & 89.02 & 76.11
      & 83.92 & 73.70 & 68.31
      & 84.06 & 75.46 & 66.07 &  0.4 Sec  \\[2mm]
      \myrowcolour
      & \multicolumn{1}{l}{\textbf{Innovation:}}
        & \multicolumn{13}{l}{\parbox[t]{1\linewidth}{Multi-scale detection with proposal and detection sub-networks deconvolutional upsampling.}} \\[2mm]

    \multirow{1}{*}{\cite{yang2016exploit}} 
      & SDP + RPN & 2016 & VGG-16 & Mono
      & 95.16 & 92.03 & 79.16
      & 80.09 & 70.16 & 64.82
      & 81.37 & 73.74 & 65.31 &  0.4 Sec  \\[2mm]
      \myrowcolour
      & \multicolumn{1}{l}{\textbf{Innovation:}}
        & \multicolumn{13}{l}{\parbox[t]{1\linewidth}{Selects features by object scale and rejects negatives layer-wise to enhance detection accuracy.}} \\[2mm]

\bottomrule
\end{tabular}}
\end{center}
\end{table}

From Table \ref{table: 2D_Kitti}, we provide an in-depth discussion of two selected models: AspectNet \cite{liang2022aspectnet} and SINet \cite{hu2018sinet}, which achieved the highest performance among all evaluated methods, and Gaussian YOLOv3 \cite{choi2019gaussian}, one of the most widely used models in the context of autonomous driving. AspectNet \cite{liang2022aspectnet} approaches object detection tasks uniquely with an anchor-free detection scheme. It sets itself apart from comparable models by adding a special head that predicts object aspect ratios. Rather than treating tackline classification and regression features separately, the network combines them before the aspect module, allowing geometric details to influence both. This additional branch is designed with only a few convolutional layers, ensuring it does not significantly slow inference time. Another notable change is how positive and negative training samples are defined: the criteria for matching are tweaked to address class imbalance, which can hurt performance in single-stage models. These architectural decisions enhance the detector's ability to recognize elongated or skewed objects, as the receptive field is shaped better to fit the actual geometry of the detected objects.

SINet is a two-stage detector that uses a conventional CNN backbone (PVA) to generate proposals from multiple feature levels. Per proposal, a Context-Aware RoI pooling (CARoI) layer upsamples only small RoIs via bilinear deconvolution, with kernel size determined by the ratio of the target pooled size to the proposal size, and then applies max pooling, preserving small-object structure without upscaling entire feature maps. The pooled features from several backbone layers are concatenated and fed to a scale-aware decision head that partitions proposals by size; each branch (one conv + one fully connected layer with classification and box-regression heads) shares early features but is trained with a Gaussian-jittered threshold around the dataset’s median object scale to reduce intra-class variation. At inference, outputs from all branches are merged using a lightweight “soft-NMS” that averages the coordinates of highly overlapping, high-confidence boxes. Because deconvolution is applied only to small RoIs and the branching does not increase the number of processed proposals, these additions maintain end-to-end training while introducing essentially no extra runtime.

Gaussian YOLOv3 \cite{choi2019gaussian} maintains the overall YOLOv3 layout, including three detection scales, feature aggregation from the backbone, and dense predictions. However, it changes the way bounding boxes are parameterized. Each coordinate is modeled as a Gaussian variable with its own variance, meaning the network outputs not only position estimates but also an uncertainty term for each coordinate. These variances are fed into a modified regression loss based on negative log-likelihood, and they also influence the suppression of overlapping boxes at inference. The result is a system that can flag low-confidence localizations without discarding high-confidence detections that fall near ambiguous boundaries. Compared with AspectNet’s emphasis on encoding geometric priors into a deterministic pipeline, Gaussian YOLOv3 takes a probabilistic route to preserve the speed of a one-stage detector while offering a measure of confidence that can be applied to downstream tasks.

Table \ref{table: 3D_Kitti} focuses on 3D object detection methods that use camera-based KITTI data, but generate 3D bounding boxes from 2D image inputs. From Table \ref{table: 3D_Kitti}, we provide a detailed discussion of four selected models: M3D-RPN \cite{brazil2019m3d}, recognized as one of the most widely adopted models in this field; MonoDGP \cite{Pu_2025_CVPR}, which is the latest released method; MonoTAKD \cite{Liu_2025_CVPR}, which achieved the highest results in $AP_{BEV}$ $Car$ and $Pedestrian$ evaluations; and MonoDiff \cite{Ranasinghe_2024_CVPR} which achieved the highest results in $AP_{3D}$ $Car$ evaluation on the KITTI dataset. It should be mentioned that the values highlighted in red represent the best performance achieved for each evaluation metric across all listed methods.

\begin{table}[H]
\centering
\caption{Comparative analysis of 3D object detection methods using the camera-based KITTI dataset.}
\vspace{-0.4cm}
\label{table: 3D_Kitti}
\begin{center}
\resizebox{\textwidth}{!}{%
  \setlength{\tabcolsep}{4pt}%
  \begin{tabular}{llllll ccc ccc ccc ccc c}
    \toprule
    \multicolumn{6}{c}{} 
      & \multicolumn{3}{c}{$\textbf{AP}_{\textbf{BEV}}$ \hspace{1mm}\textbf{Car}} 
      & \multicolumn{3}{c}{$\textbf{AP}_{\textbf{3D}}$\hspace{1mm}\textbf{Car}}
      & \multicolumn{3}{c}{$\textbf{AP}_{\textbf{3D}}$\hspace{1mm}\textbf{Pedestrian}} 
      & \multicolumn{3}{c}{$\textbf{AP}_{\textbf{3D}}$\hspace{1mm}\textbf{Cyclist}} \\[1mm]
    \cmidrule(l){7-9}\cmidrule(l){10-12}\cmidrule(l){13-15}\cmidrule(l){16-18}
    \textbf{Ref} & \textbf{Method} & \textbf{Year} & \textbf{Venue} & \textbf{Backbone} & \textbf{Modality}
      & \textbf{Easy} & \textbf{Moderate} & \textbf{Hard}
      & \textbf{Easy} & \textbf{Moderate} & \textbf{Hard}
      & \textbf{Easy} & \textbf{Moderate} & \textbf{Hard}
      & \textbf{Easy} & \textbf{Moderate} & \textbf{Hard} 
      & \textbf{Runtime}
      \\
    \midrule

    \multirow{1}{*}{\cite{Pu_2025_CVPR}} 
      & MonoDGP  & 2025 & CVPR & Res50 & Mono
      & 35.24 & 25.23 & 22.02
      & 26.35 & 18.72 & 15.97
      & $-$     & $-$     & $-$
      & $-$     & $-$     & $-$ & 0.03 Sec\\
      \myrowcolour
      & \multicolumn{18}{l}{\textbf{Innovation:} Proposes geometry error prediction with decoupled 2D/3D queries and region segmentation to advance monocular 3D detection.} \\[2mm]

    \multirow{1}{*}{\cite{Liu_2025_CVPR}} 
      & MonoTAKD  & 2025 & CVPR & Res50 & Mono
      & \red{\textbf{38.75}} & \red{\textbf{27.76}}  & \red{\textbf{24.14}} 
      & 27.91 & 19.43 & 16.51
      & \red{\textbf{16.15}}   & \red{\textbf{10.41}}  & \red{\textbf{9.68}}
      & \red{\textbf{13.54}}  & \red{\textbf{7.23}}   & \red{\textbf{6.86}} & 0.08 Sec\\
      \myrowcolour
      &  \multicolumn{18}{l}{\textbf{Innovation:} Presents teaching assistant distillation uniting intra-modal visual knowledge transfer and cross-modal residual learning to enhance monocular 3D detection.} \\[2mm]

    \multirow{1}{*}{\cite{wu2024fd3d}} 
      & FD3D     & 2024 & AAAI & MonoDLE                & Mono
      & 34.20 & 23.72 & 20.76
      & 25.38 & 17.12 & 14.50
      & $-$     & $-$     & $-$
      & $-$     & $-$     & $-$ & \textbf{$-$}\\
      \myrowcolour
      &  \multicolumn{18}{l}{\textbf{Innovation:} Introduces AFOD-based feature supervision with pixel-level and distribution-level guidance to enhance monocular 3D object detection accuracy.} \\[2mm]

    \multirow{1}{*}{\cite{Yan_2024_CVPR}} 
      & MonoCD      & 2024 & CVPR  & DLA-34           & Mono
      & 33.41 & 22.81 & 19.57
      & 25.53 & 16.59 & 14.53
      & $-$     & $-$     & $-$
      & $-$     & $-$     & $-$ & 0.03 Sec\\
      \myrowcolour
      &  \multicolumn{18}{l}{\textbf{Innovation:} Introduces complementary depth prediction using global clues and geometric relations to reduce coupling and improve monocular 3D detection.} \\[2mm]
    \multirow{1}{*}{\cite{Ranasinghe_2024_CVPR}} 
      & MonoDiff    & 2024 & CVPR  & Res50-FPN  & Mono
      & $-$     & $-$     & $-$
      & \red{\textbf{30.18}} & \red{\textbf{21.02}}  & \red{\textbf{18.16}} 
      & 13.51 & 8.94  & 7.28
      & 8.52  & 4.35  & 3.78 & 0.08 Sec\\
      \myrowcolour
      &  \multicolumn{18}{l}{\textbf{Innovation:} Employs diffusion models for monocular 3D detection and pose estimation, refining predictions through iterative denoising of depth and pose.} \\[2mm]

    \multirow{1}{*}{\cite{lu2024gupnet++}} 
      & GUPNet++     & 2024 & TPAMI & DLA-34           & Depth
      & $-$     & $-$     & $-$
      & 24.99 & 16.48 & 14.58
      & 12.45 & 8.13  & 6.91
      & 6.71  & 3.91  & 3.80 & \textbf{$-$}\\
      \myrowcolour
      &  \multicolumn{18}{l}{\textbf{Innovation:} Introduces probabilistic geometry uncertainty propagation with IoU-guided confidence to mitigate projection error amplification in monocular 3D detection.} \\[2mm]

    \multirow{1}{*}{\cite{Zhang_2023_ICCV}} 
      & MonoDETR    & 2023 & ICCV & Res50                & Mono
      & 33.60 & 22.11 & 18.60
      & 25.00 & 16.47 & 13.58
      & $-$     & $-$     & $-$
      & $-$     & $-$     & $-$   & 0.04 Sec\\
      \myrowcolour
      &  \multicolumn{18}{l}{\textbf{Innovation:} Introduces depth-guided Transformer with global reasoning and depth-aware queries to enhance monocular 3D object detection accuracy.} \\[2mm]

    \multirow{1}{*}{\cite{jinrang2023monouni}} 
      & MonoUNI      & 2023 & NeurIPS & Res50                & Mono
      & $-$     & $-$     & $-$
      & 24.75 & 16.73 & 13.49
      & 15.78 & 1.34  & 8.74
      & 7.34  & 4.28  & 3.78 & 0.04 Sec\\
      \myrowcolour
      &  \multicolumn{18}{l}{\textbf{Innovation:} Proposes unified vehicle/infrastructure-side monocular 3D detection network leveraging sufficient depth clues for robust cross-perspective object localization.} \\[2mm]

    \multirow{1}{*}{\cite{kumar2022deviant}} 
      & DEVIANT      & 2023 & CVPR & Res50                & Mono
      & 29.65 & 20.44 & 17.43
      & 21.88 & 14.46 & 11.89
      & $-$     & $-$     & $-$
      & $-$     & $-$     & $-$ & 0.04 Sec\\
      \myrowcolour
      &  \multicolumn{18}{l}{\textbf{Innovation:} Introduces depth-guided dynamic deformable attention to adaptively refine features for improved monocular 3D object detection accuracy.} \\[2mm]


    \multirow{1}{*}{\cite{wu2023monopgc}} 
      & MonoPGC      & 2023 & CVPR & DLA34                & Mono
      & 32.50 & 23.14 & 20.30
      & 24.68 & 17.17 & 14.14
      & $-$     & $-$     & $-$
      & $-$     & $-$     & $-$ & 0.04 Sec\\
      \myrowcolour
      &  \multicolumn{18}{l}{\textbf{Innovation:} Incorporates pixel-level geometry via depth cross-attention, depth-space-aware transformer, and depth-gradient encoding for improved 3D detection.} \\[2mm]

    \multirow{1}{*}{\cite{Zhou_2023_CVPR}} 
      & MonoATT      & 2023 & CVPR & DLA-34           & Mono
      & 36.87 & 24.42 & 21.88
      & 24.72 & 17.37 & 15.00
      & 10.55 & 6.66  & 5.43
      & 5.74  & 3.68  & 2.94 & 0.05 Sec\\
      \myrowcolour
      &  \multicolumn{18}{l}{\textbf{Innovation:} Proposes adaptive token Transformer that dynamically adjusts token count to improve efficiency and accuracy in monocular 3D detection.} \\[2mm]


    \multirow{1}{*}{\cite{li2022densely}} 
      & DCD          & 2023 &ECCV & DLA34                & Depth
      & $-$     & $-$     & $-$
      & 23.81 & 15.90 & 13.21
      & 10.37 & 6.73  & 6.28
      & 4.72  & 2.74  & 2.41 & 0.03 Sec\\
      \myrowcolour
      &  \multicolumn{18}{l}{\textbf{Innovation:} Introduces depth-consistent distillation to transfer depth awareness from LiDAR-based teacher to monocular student for improved detection.} \\[2mm]


    \multirow{1}{*}{\cite{Li_2022_CVPR}} 
      & MonoDDE      & 2022 & CVPR & DLA34        & Mono
      & 33.58 & 23.46 & 20.37
      & 24.93 & 17.14 & 15.10
      & 11.13 & 7.32  & 6.67
      & 5.94  & 3.78  & 3.33 & 0.04 Sec\\
      \myrowcolour
      &  \multicolumn{18}{l}{\textbf{Innovation:} Develops a depth solving system producing 20 diverse estimations and robustly combining them for reliable monocular 3D detection.} \\[2mm]

    \multirow{1}{*}{\cite{Qin_2022_CVPR}} 
      & MonoGround   & 2022 & CVPR & DLA-34           & Mono
      & 30.07 & 20.47 & 17.74
      & 21.37 & 14.36 & 12.62
      & 12.37 & 7.89  & 7.13
      & 4.62  & 2.68  & 2.53 & 0.03 Sec\\
      \myrowcolour
      &  \multicolumn{18}{l}{\textbf{Innovation:} Introduces a learnable ground plane prior with dense depth supervision, depth-align training, and two-stage depth inference.} \\[2mm]

    \multirow{1}{*}{\cite{Lian_2022_CVPR}} 
      & MonoJSG      & 2022 & CVPR & DLA-34           & Mono
      & 32.59 & 21.26 & 18.18
      & 24.69 & 16.14 & 13.64
      & 11.94 & 7.36  & 6.03
      & 8.03  & 3.87  & 3.33 & 0.04 Sec\\
      \myrowcolour
      &  \multicolumn{18}{l}{\textbf{Innovation:} Proposes joint semantic and geometric cost volume with adaptive depth sampling for refined monocular 3D detection.} \\[2mm]

    \multirow{1}{*}{\cite{brazil2020kinematic}} 
      & Kinematic3D  & 2022 & ECCV & Dense121      & Mono
      & 26.69 & 17.52 & 13.10
      & 19.07 & 12.72 & 9.17
      & $-$     & $-$     & $-$
      & $-$     & $-$     & $-$ & 0.12 Sec\\
      \myrowcolour
      &  \multicolumn{18}{l}{\textbf{Innovation:} Incorporates kinematic motion constraints to refine monocular 3D object detection in dynamic driving scenes.} \\[2mm]

    \multirow{1}{*}{\cite{Rukhovich_2022_WACV}} 
      & ImVoxelNet   & 2022 & WACV & Res50                & Mono
      & $-$     & $-$     & $-$
      & 17.15 & 10.97 & 9.15
      & $-$     & $-$     & $-$
      & $-$     & $-$     & $-$  & 0.2 Sec\\
      \myrowcolour
      &  \multicolumn{18}{l}{\textbf{Innovation:} Projects posed images into a unified voxel representation enabling general-purpose 3D detection from monocular or multi-view input.} \\[2mm]

    \multirow{1}{*}{\cite{Gu_2022_CVPR}} 
      & ImVoxelNet & 2022 & CVPR & DLA34           & Mono
      & $-$     & $-$     & $-$
      & 20.10 & 12.99 & 10.50
      & 12.47 & 7.62  & 6.72
      & 1.52  & 0.85  & 0.94 & 0.2 Sec\\
      \myrowcolour
      &  \multicolumn{18}{l}{\textbf{Innovation:} Introduces homography loss aligning projected 3D boxes to ground plane to improve monocular 3D detection accuracy.} \\[2mm]

    \multirow{1}{*}{\cite{Gu_2022_CVPR}} 
      & MonoFlex & 2022 & CVPR & DLA34            & Mono
      & $-$     & $-$     & $-$
      & 21.75 & 14.94 & 13.07
      & 11.87 & 7.66  & 6.82
      & 5.48  & 3.50  & 2.99 & 0.03 Sec\\
      \myrowcolour
      &  \multicolumn{18}{l}{\textbf{Innovation:} Introduces homography loss aligning projected 3D boxes to ground plane to improve monocular 3D detection accuracy.} \\[2mm]

    \multirow{1}{*}{\cite{hong2022cross}} 
      & CMKD         & 2022 & ECCV & Res50                & Depth
      & $-$     & $-$     & $-$
      & 25.09 & 16.99 & 15.30
      & 17.79 & 11.69 & 10.09
      & 9.60  & 5.24  & 4.50  & 0.1 Sec\\
      \myrowcolour
      &  \multicolumn{18}{l}{\textbf{Innovation:} Transfers depth-aware features from a LiDAR-based teacher to a monocular student using cross-modal knowledge distillation framework.} \\[2mm]

    \multirow{1}{*}{\cite{peng2022did}} 
      & DID-M3D      & 2022 & ECCV & DLA34                & Depth
      & 32.95     & 22.76     & 19.83
      & 24.40 & 16.29 & 13.75
      & $-$     & $-$     & $-$
      & $-$     & $-$     & $-$ & 0.04 Sec\\
      \myrowcolour
      &  \multicolumn{18}{l}{\textbf{Innovation:} Decouples instance depth into visual and attribute depths with uncertainty modeling and enables effective affine augmentation.} \\[2mm]

    \multirow{1}{*}{\cite{Lu_2021_ICCV}} 
      & GUPNet       & 2021 & ICCV & DLA34                & Mono
      & $-$     & $-$     & $-$
      & 20.11 & 14.20 & 11.77
      & 14.72 & 9.53  & 7.87
      & 4.18  & 2.65  & 2.09 & \textbf{$-$}\\
      \myrowcolour
      &  \multicolumn{18}{l}{\textbf{Innovation:} Proposes Geometry Uncertainty Projection and Hierarchical Task Learning to mitigate depth error amplification in monocular 3D detection.} \\[2mm]

    \multirow{1}{*}{\cite{Ma_2021_CVPR}} 
      & MonoDLE      & 2021 & CVPR & DLA34                & Mono
      & 24.79 & 18.89 & 16.00
      & 17.23 & 12.26 & 10.29
      & 5.34  & 3.28  & 2.83
      & 4.59  & 2.66  & 2.45 & 0.04 Sec\\
      \myrowcolour
      &  \multicolumn{18}{l}{\textbf{Innovation:} Diagnoses localization error in monocular 3D detection and proposes center supervision, distance filtering, and IoU-based size loss.} \\[2mm]

    \multirow{1}{*}{\cite{Shi_2021_ICCV}} 
      & MonoRCNN     & 2021 & ICCV & Res50                & Mono
      & 25.48 & 18.11 & 14.10
      & 18.36 & 12.65 & 10.03
      & 11.21 & 7.28  & 5.85
      & 2.89  & 1.67  & 1.54 & 0.07 Sec\\
      \myrowcolour
      &  \multicolumn{18}{l}{\textbf{Innovation:} Introduces geometry-based distance decomposition with uncertainty-aware regression to improve interpretability, robustness, and accuracy in detection.} \\[2mm]

    \multirow{1}{*}{\cite{Zhang_2021_CVPR}} 
      & MonoFlex     & 2021 & CVPR & DLA34                & Mono
      & 28.23 & 19.75 & 16.89
      & 19.94 & 13.89 & 12.07
      & 9.43  & 6.31  & 5.26
      & 4.17  & 2.35  & 2.04 & 0.03 Sec\\
      \myrowcolour
      &  \multicolumn{18}{l}{\textbf{Innovation:} Decouples truncated object prediction and adaptively ensembles multiple depth estimators using uncertainty for robust monocular 3D detection.} \\[2mm]

    \multirow{1}{*}{\cite{Chen_2021_CVPR}} 
      & MonoRUn      & 2021 & CVPR & Res101                & Mono
      & 27.94 & 17.34 & 15.24
      & 19.65 & 12.30 & 10.58
      & 10.88 & 6.78  & 5.83
      & 1.01  & 0.61  & 0.48 & 0.07 Sec\\
      \myrowcolour
      &  \multicolumn{18}{l}{\textbf{Innovation:} Employs self-supervised dense 3D reconstruction with uncertainty propagation and Robust KL loss for monocular 3D detection.} \\[2mm]

    \multirow{1}{*}{\cite{Wang_2021_CVPR}} 
      & DDMP-3D      & 2021 & CVPR & Res101       & Depth
      & 28.08 & 17.89 & 13.44
      & 19.71 & 12.78 & 9.80
      & 4.93  & 3.55  & 3.01
      & 4.18  & 2.35  & 2.04 & 0.18 Sec\\
      \myrowcolour
      &  \multicolumn{18}{l}{\textbf{Innovation:} Introduces depth-conditioned dynamic message propagation to enhance monocular 3D object detection through adaptive multi-scale feature interaction.} \\[2mm]



    \multirow{1}{*}{\cite{chen2020monopair}} 
      & MonoPair     & 2020 & CVPR & DLA34                & Mono
      & $-$     & $-$     & $-$
      & 13.04 & 9.99  & 8.65
      & 10.02 & 6.68  & 5.53
      & 3.79  & 2.12  & 1.83 & 0.06 Sec\\
      \myrowcolour
      &  \multicolumn{18}{l}{\textbf{Innovation:} Leverages pairwise spatial constraints with uncertainty-aware optimization to improve detection of occluded objects.} \\[2mm]

    \multirow{1}{*}{\cite{li2020rtm3d}} 
      & RTM3D        & 2020 & ECCV & DLA34                & Mono
      & 19.17     & 14.20     & 11.99
      & 14.41 & 10.34 & 8.77
      & $-$     & $-$     & $-$
      & $-$     & $-$     & $-$ &0.05 Sec\\
      \myrowcolour
      &  \multicolumn{18}{l}{\textbf{Innovation:} Reformulates monocular 3D detection as keypoint detection of nine projected 3D box vertices plus center, using geometric constraints for real-time inference.} \\[2mm]

    \multirow{1}{*}{\cite{shi2020distance}} 
      & UR3D         & 2020 & ECCV & Res34                & Mono
      & 21.85     & 12.51     & 9.20
      & 15.58 & 8.61  & 6.00
      & $-$     & $-$     & $-$
      & $-$     & $-$     & $-$  & \textbf{$-$}\\
      \myrowcolour
      &  \multicolumn{18}{l}{\textbf{Innovation:} Learns a distance-normalized unified representation with distance-guided NMS and cascaded point regression for compact, accurate 3D detection.} \\[2mm]

    \multirow{1}{*}{\cite{simonelli2020towards}} 
      & MoVi-3D      & 2020 & ECCV & Res34                & Mono
      & $-$     & $-$     & $-$
      & 15.19 & 10.90 & 9.26
      & 8.99  & 5.44  & 4.57
      & 1.08  & 0.63  & 0.70 & \textbf{$-$}\\
      \myrowcolour
      &  \multicolumn{18}{l}{\textbf{Innovation:} Leverages a virtual view-based training strategy to normalize object appearances across different depths, enabling a single-stage network to generalize.} \\[2mm]

    \multirow{1}{*}{\cite{ye2020monocular}} 
      & DA-3Ddet     & 2020 & ECCV & Res101                & Mono
      & $-$     & $-$     & $-$
      & 16.77 & 11.50 & 8.93
      & 8.7   & 7.1   & 6.7
      & 14.5  & 11.5  & 11.5 &0.03 Sec\\
      \myrowcolour
      &  \multicolumn{18}{l}{\textbf{Innovation:} Introduces feature domain adaptation and context-aware segmentation to align pseudo-LiDAR features with real-LiDAR representations.} \\[2mm]


    \multirow{1}{*}{\cite{simonelli2019disentangling}} 
      & MonoDIS      & 2019 & ICCV & Res34                & Mono
      & 17.23 & 13.19 & 11.12
      & 10.37 & 7.94  & 6.40
      & $-$     & $-$     & $-$
      & $-$     & $-$     & $-$ &0.06 Sec\\
      \myrowcolour
      &  \multicolumn{18}{l}{\textbf{Innovation:} Proposes loss disentanglement for 2D/3D detection and a self-supervised 3D-box confidence, enabling stable end-to-end monocular training.} \\[2mm]

    \multirow{1}{*}{\cite{brazil2019m3d}} 
      & M3D-RPN      & 2019 & ICCV & Dense121                & Mono
      & $-$     & $-$     & $-$
      & 14.76 & 9.71  & 7.42
      & 4.92  & 3.48  & 2.94
      & 0.94  & 0.65  & 0.47  &0.16 Sec\\
      \myrowcolour
      &  \multicolumn{18}{l}{\textbf{Innovation:} Introduces a single-shot monocular 3D RPN with depth-aware convolution and shared 2D–3D anchors for accurate localization.} \\[2mm]

\bottomrule
\end{tabular}}
\end{center}
\end{table}

\vspace{-0.5cm}
The Monocular 3D Region Proposal Network (M3D-RPN) \cite{brazil2019m3d} presents an integrated approach for generating 2D and 3D object proposals from a single RGB image. Built upon a DenseNet-121 backbone modified with dilated convolutions to preserve a stride of 16, the network branches into two complementary pathways: one employing standard spatially invariant convolutions to capture global context, and another utilizing depth-aware convolutions to encode row-dependent geometric cues. In the latter, feature maps are divided into horizontal bins, with each bin processed by distinct convolutional kernels to reflect perspective variations inherent to urban driving scenes. Anchors are designed to carry 2D and 3D parameters, initialized from dataset-derived statistics to provide strong geometric priors. Predictions from the two branches are combined to regress full 3D bounding box parameters, followed by a lightweight post-optimization step that refines object orientation through projection consistency between 3D boxes and their 2D counterparts. M3D-RPN reduces complexity while maintaining accuracy by consolidating proposal generation into a single end-to-end stage.

MonoDGP \cite{Pu_2025_CVPR} employs a transformer-based architecture that unites geometric priors with a novel perspective-invariant error formulation for depth refinement. A ResNet-50 backbone extracts multi-scale features, which are enhanced by a Region Segmentation Head (RSH) to emphasize foreground content and embed segment labels distinguishing object regions from background. This network's detection is decoupled into two parallel decoding processes: a 2D visual decoder for initializing queries and reference points from appearance cues and a 3D depth-guided decoder that integrates visual and depth features to model spatial relationships. Rather than relying on direct depth prediction or multi-branch fusion, MonoDGP computes geometric depth using the projection formula and introduces an explicit geometry error term to correct systematic offsets caused by perspective and object surface effects. With its stable and compact statistical distribution, this error term reduces training complexity while improving depth accuracy. Combining query decoupling allows the method to achieve improved convergence stability and precise 3D localization without requiring auxiliary data sources.

The MonoTAKD \cite{Liu_2025_CVPR} framework introduces a three-part distillation strategy to enhance monocular 3D detection, comprising a LiDAR-based teacher, a camera-based teaching assistant (TA), and a camera-based student. Using a ResNet-50 backbone, the TA network integrates accurate ground-truth depth maps with visual features to produce high-fidelity BEV representations, referred to as 3D visual knowledge. These representations guide the student via intra-modal distillation, mitigating errors caused by monocular depth estimation. The student, architecturally aligned with the TA but relying on predicted depth, is structured with two feature branches: one dedicated to learning visual cues from the TA, and another designed to absorb spatial information from residual BEV features obtained by subtracting the TA's output from that of the LiDAR-based teacher. Cross-modal residual distillation transfers only these spatial cues, narrowing the modality gap without requiring the student to replicate the entire LiDAR feature space. A Spatial Alignment Module, combining dilated and deformable convolutions with channel attention, further refines spatial branch outputs. At the same time, a Feature Fusion Module merges the visual and spatial streams for final detection. This targeted multi-stage guidance enables the student to capture a richer geometric context than conventional monocular models.

Table \ref{table: 3D_NuScenes} presents a comparative analysis of 3D object detection methods using camera-based data from the NuScenes dataset, where the input is 2D images and the output is 3D bounding boxes. From Table \ref{table: 3D_NuScenes}, we provide a detailed discussion of three selected models: DETR3D \cite{wang2022detr3d}, recognized as a well-known and widely adopted model in this field; CorrBEV \cite{Xue_2025_CVPR}, one of the most recent methods evaluated on the NuScenes benchmark; and DualViewDistill \cite{kappeler2025bridging} that achieved the highest results on the NuScenes dataset. It should be noted that the values highlighted in red represent the best performance achieved for each evaluation metric across all listed methods.

DETR3D \cite{wang2022detr3d} addresses multi-view 3D object detection by establishing a direct mapping between a sparse set of learned 3D object queries and multi-camera image features, bypassing any intermediate dense geometric reconstruction. Feature extraction is performed by a shared ResNet–FPN backbone, producing multi-scale image representations for each view. Each object query contains a latent 3D hypothesis, decoded into a reference point in the scene. Through calibrated camera intrinsics and extrinsics, these points are back-projected into each image, where bilinear interpolation retrieves the corresponding feature vectors from all pyramid levels. Aggregated features are merged into the queries and fine-tuned using multi-head self-attention. This projection–refinement cycle repeats across multiple transformer decoder layers, with each stage producing updated classifications and 3D box regressions. DETR3D maintains an end-to-end pipeline that improves efficiency without sacrificing competitive detection accuracy by avoiding dense depth estimation, voxelization, and non-maximum suppression.

\begin{table*}[htbp]
\centering
\caption{Comparative analysis of 2D/3D object detection methods using the camera-based NuScenes dataset.}
\vspace{-0.5cm}
\label{table: 3D_NuScenes}
\begin{center}
\resizebox{\textwidth}{!}{%
  \setlength{\tabcolsep}{4pt}%
  \begin{tabular}{llllll ccccccc}
    \toprule
    \textbf{Ref} & \textbf{Method} & \textbf{Year} \hspace{5mm}   & \textbf{Venue} & \textbf{Backbone} \hspace{3mm} & \textbf{Modality}
      & \textbf{mAP} & \textbf{NDS} & \textbf{mATE} & \textbf{mASE} & \textbf{mAOE} & \textbf{mAVE} & \textbf{mAAE} \\
    \midrule

    \multirow{1}{*}{\cite{kappeler2025bridging}} 
      & DualViewDistill & 2025 & arXiv & ViT-L & Multi-View
      & \red{\textbf{62.1}} & \red{\textbf{69.5}} & \red{\textbf{0.360}} & \red{\textbf{0.227}} & \red{\textbf{0.272}} & \red{\textbf{0.168}} & \red{\textbf{0.126}} \\[1mm]
      \myrowcolour
      & \multicolumn{1}{l}{\textbf{Innovation:}}
        & \multicolumn{11}{l}{Combines dual-view PV–BEV reasoning with distilled neural BEV maps for robust camera-based 3D detection.} \\[2mm]

    \multirow{1}{*}{\cite{Xue_2025_CVPR}} 
      & CorrBEV-fm & 2025 & CVPR & Res101 & Multi-View
      & 41.30 & 50.70 & 0.646 & 0.265 & 0.482 & 0.473 & 0.134 \\[1mm]
      \myrowcolour
      & \multicolumn{1}{l}{\textbf{Innovation:}}
        & \multicolumn{11}{l}{Introduces multimodal prototypes and correlation learning to improve occluded detection.} \\[2mm]

    \multirow{1}{*}{\cite{Xue_2025_CVPR}} 
      & CorrBEV-sp & 2025 & CVPR & Res101 & Multi-View
      & \red{\textbf{56.20}} & \red{\textbf{64.00}}  & \red{\textbf{0.488 }} & \red{\textbf{0.241}}  & \red{\textbf{0.303}}  & \red{\textbf{0.243}}  & \red{\textbf{0.123}}  \\[1mm]
      \myrowcolour
      & \multicolumn{1}{l}{\textbf{Innovation:}}
        & \multicolumn{11}{l}{\parbox[t]{1\linewidth}{Enhances detection of occluded objects in multi-view 3D perception.}} \\[2mm]

    \multirow{1}{*}{\cite{ji2025ropetr}} 
      & RoPETR & 2025 & ArXiv & V2-99 & Multi-View
      & 52.90 & 61.40 & 0.537 & 0.255 & 0.289 & 0.229 & 0.195 \\[1mm]
      \myrowcolour
      & \multicolumn{1}{l}{\textbf{Innovation:}}
        & \multicolumn{11}{l}{Enhances camera-only detection with multimodal rotary position embedding for better velocity estimation.} \\[2mm]

    \multirow{1}{*}{\cite{xia2024henet}} 
      & HENet & 2024 & ECCV &ResNet50 & Multi-View
      & \red{\textbf{57.80}} & \red{\textbf{63.8}} & \red{\textbf{0.432}} & \red{\textbf{0.242}} & \red{\textbf{0.368}} & \red{\textbf{0.320}} & \red{\textbf{0.129}} \\[1mm]
      \myrowcolour
      & \multicolumn{1}{l}{\textbf{Innovation:}}
        & \multicolumn{11}{l}{End-to-end multi-task 3D perception using hybrid image encoding and attention-based temporal BEV fusion.} \\[2mm]

    \multirow{1}{*}{\cite{li2024bevformer}} 
      & BEVFormer & 2024 & IEEE & Res101 & Multi-View
      & 44.50 & 53.50 & 0.631 & 0.257 & 0.405 & 0.435 & 0.143 \\[1mm]
      \myrowcolour
      & \multicolumn{1}{l}{\textbf{Innovation:}}
        & \multicolumn{11}{l}{Uses spatiotemporal transformers with deformable attention to learn BEV from LiDAR-camera inputs.} \\[2mm]

    \multirow{1}{*}{\cite{liu2024ray}} 
      & RayDN & 2024 & ECCV & Res50 & Multi-View
      & 46.90 & 56.30 & 0.579 & 0.264 & 0.433 & 0.256 & 0.187 \\[1mm]
      \myrowcolour
      & \multicolumn{1}{l}{\textbf{Innovation:}}
        & \multicolumn{11}{l}{Introduces depth-aware hard negative sampling along camera rays to reduce false positives in 3D detection.} \\[2mm]

    \multirow{1}{*}{\cite{Wang_2023_ICCV}} 
      & StreamPETR & 2023 & ICCV & Res50 & Multi-View
      & 45.00 & 55.00 & 0.569 & 0.267 & 0.413 & 0.265 & 0.196 \\[1mm]
      \myrowcolour
      & \multicolumn{1}{l}{\textbf{Innovation:}}
        & \multicolumn{11}{l}{Uses object-centric temporal modeling with motion-aware layer normalization for efficient 3D detection.} \\[2mm]

    \multirow{1}{*}{\cite{Liu_2023_ICCV}} 
      & PETRv2 & 2023 & ICCV & Res50 & Multi-View
      & 34.90 & 45.60 & 0.700 & 0.275 & 0.580 & 0.437 & 0.187 \\[1mm]
      \myrowcolour
      & \multicolumn{1}{l}{\textbf{Innovation:}}
        & \multicolumn{11}{l}{Extends PETR with temporal 3D position embedding alignment and task-specific queries for 3D perception.} \\[2mm]

    \multirow{1}{*}{\cite{li2023bevdepth}} 
      & BEVDepth & 2023 & AAAI & Res50 & Multi-View
      & 35.10 & 47.50 & 0.639 & 0.267 & 0.479 & 0.428 & 0.198 \\[1mm]
      \myrowcolour
      & \multicolumn{1}{l}{\textbf{Innovation:}}
        & \multicolumn{11}{l}{Improves 3D detection via explicit depth supervision, camera prediction, and depth refinement modules.} \\[2mm]

    \multirow{1}{*}{\cite{Liu_2023_ICCV2}} 
      & SparseBEV & 2023 & ICCV & Res50 & Multi-View
      & 44.80 & 55.80 & 0.581 & 0.271 & 0.373 & 0.247 & 0.190 \\[1mm]
      \myrowcolour
      & \multicolumn{1}{l}{\textbf{Innovation:}}
        & \multicolumn{11}{l}{Proposes a sparse BEV detector with adaptive attention, spatio-temporal sampling, and adaptive mixing.} \\[2mm]

    \multirow{1}{*}{\cite{Zong_2023_ICCV}} 
      & HoP & 2023 & ICCV & Res50 & Multi-View
      & 40.00 & 51.00 & 0.607 & 0.275 & 0.515 & 0.286 & 0.216 \\[1mm]
      \myrowcolour
      & \multicolumn{1}{l}{\textbf{Innovation:}}
        & \multicolumn{11}{l}{Introduces method with short- and long-term temporal decoders to improve BEV feature learning.} \\[2mm]

    \multirow{1}{*}{\cite{jiang2023polarformer}} 
      & PolarFormer & 2023 & AAAI & Res101 & Multi-View
      & 41.50 & 47.00 & 0.657 & 0.263 & 0.405 & 0.911 & 0.139 \\[1mm]
      \myrowcolour
      & \multicolumn{1}{l}{\textbf{Innovation:}}
        & \multicolumn{11}{l}{Introduces Polar coordinate BEV representation with cross-attention decoding and multi-scale learning.} \\[2mm]

    \multirow{1}{*}{\cite{Wang_2023_CVPR}} 
      & FrustumFormer & 2023 & CVPR & Res101 & Multi-View
      & 47.80 & 56.10 & 0.575 & 0.257 & 0.402 & 0.411 & 0.132 \\[1mm]
      \myrowcolour
      & \multicolumn{1}{l}{\textbf{Innovation:}}
        & \multicolumn{11}{l}{Introduces adaptive instance-aware resampling with temporal frustum fusion to enhance object localization.} \\[2mm]

    \multirow{1}{*}{\cite{Li_2023_ICCV}} 
      & FB-BEV & 2023 & ICCV & V2-99 & Multi-View
      & 53.70 & 62.40 & 0.439 & 0.250 & 0.358 & 0.270 & 0.128 \\[1mm]
      \myrowcolour
      & \multicolumn{1}{l}{\textbf{Innovation:}}
        & \multicolumn{11}{l}{Combines forward and depth-aware backward projection to densify BEV features.} \\[2mm]


    \multirow{1}{*}{\cite{Wang_2023_ICCV2}} 
      & MV2D & 2023 & ICCV & V2-99 & Multi-View
      & 46.30 & 51.40 & 0.542 & 0.247 & 0.403 & 0.854 & 0.127 \\[1mm]
      \myrowcolour
      & \multicolumn{1}{l}{\textbf{Innovation:}}
        & \multicolumn{11}{l}{Generates dynamic 3D object queries from 2D detections with sparse cross-attention.} \\[2mm]

    \multirow{1}{*}{\cite{Xiong_2023_CVPR}} 
      & CAPE & 2023 & CVPR & V2-99 & Multi-View
      & 52.50 & 61.00 & 0.503 & 0.242 & 0.361 & 0.306 & 0.114 \\[1mm]
      \myrowcolour
      & \multicolumn{1}{l}{\textbf{Innovation:}}
        & \multicolumn{11}{l}{Introduces camera-view position embeddings with bilateral attention to decouple extrinsic variations.} \\[2mm]

    \multirow{1}{*}{\cite{Shu_2023_ICCV}} 
      & 3DPPE & 2023 & ICCV & VoV-99 & Multi-View
      & 51.40 & 56.60 & 0.569 & 0.255 & 0.394 & 0.796 & 0.138 \\[1mm]
      \myrowcolour
      & \multicolumn{1}{l}{\textbf{Innovation:}}
        & \multicolumn{11}{l}{Introduces depth-guided 3D point positional encoding with a hybrid-depth module for object localization.} \\[2mm]

    \multirow{1}{*}{\cite{liu2022petr}} 
      & PETR & 2022 & ECCV & Res101 & Multi-View
      & 37.00 & 45.50 & 0.647 & 0.251 & 0.433 & 0.933 & 0.143 \\[1mm]
      \myrowcolour
      & \multicolumn{1}{l}{\textbf{Innovation:}}
        & \multicolumn{11}{l}{Encodes 3D coordinates into multi-view image features, producing position-aware representations detection.} \\[2mm]

    \multirow{1}{*}{\cite{li2022unifying}} 
      & UVTR & 2022 & NeurIPS  & Res101 & Multi-View
      & 37.90 & 48.30 & 0.731 & 0.267 & 0.350 & 0.510 & 0.200 \\[1mm]
      \myrowcolour
      & \multicolumn{1}{l}{\textbf{Innovation:}}
        & \multicolumn{11}{l}{Unifies voxel-based representation across modalities, enabling cross-modality fusion and knowledge transfer.} \\[2mm]


  \multirow{1}{*}{\cite{li2023bevstereo}} 
      & BEVStereo & 2022 & AAAI & Res50 & Multi-View
      & 37.20 & 50.00 & 0.597 & 0.270 & 0.438 & 0.367 & 0.190 \\[1mm]
      \myrowcolour
      & \multicolumn{1}{l}{\textbf{Innovation:}}
        & \multicolumn{11}{l}{Proposes dynamic temporal stereo with EM-updated depth sampling and size-aware circle NMS.} \\[2mm]


    \multirow{1}{*}{\cite{wang2022probabilistic}} 
      & PGD & 2022 & CoRL & Res101 & Multi-View
      & 38.60 & 44.80 & 0.626 & 0.245 & 0.451 & 1.509 & 0.127 \\[1mm]
      \myrowcolour
      & \multicolumn{1}{l}{\textbf{Innovation:}}
        & \multicolumn{11}{l}{Combines probabilistic depth uncertainty modeling with graph-based geometric depth propagation.} \\[2mm]

    \multirow{1}{*}{\cite{wang2022detr3d}} 
      & DETR3D & 2022 & CoRL & Res101+FPN & Multi-View
      & 41.20 & 47.90 & 0.641 & 0.255 & 0.394 & 0.845 & 0.133 \\[1mm]
      \myrowcolour
      & \multicolumn{1}{l}{\textbf{Innovation:}}
        & \multicolumn{11}{l}{Uses sparse 3D object queries with geometric back-projection to fuse multi-view features.} \\[2mm]

    \multirow{1}{*}{\cite{chen2022graph}} 
      & Graph-DETR3D & 2022 & ACM & Res101 & Multi-View
      & 41.80 & 47.20 & 0.668 & 0.250 & 0.440 & 0.876 & 0.139 \\[1mm]
      \myrowcolour
      & \multicolumn{1}{l}{\textbf{Innovation:}}
        & \multicolumn{11}{l}{Leverages dynamic 3D graph feature aggregation and depth-invariant multi-scale training.} \\[2mm]

    \multirow{1}{*}{\cite{Wang_2021_ICCV2}} 
      & FCOS3D & 2021 & ICCV & Res101 & Multi-View
      & 35.80 & 42.80 & 0.690 & 0.249 & 0.452 & 1.434 & 0.124 \\[1mm]
      \myrowcolour
      & \multicolumn{1}{l}{\textbf{Innovation:}}
        & \multicolumn{11}{l}{Adapts anchor-free FCOS with 3D target reformulation and 2D-guided multi-level prediction.} \\[2mm]

\bottomrule
\end{tabular}}
\end{center}
\end{table*}

CorrBEV \cite{Xue_2025_CVPR} improves upon transformer-based multi-view detection frameworks like BEVFormer. The task is completed by integrating a correlation-driven multi-modal approach to counteract the loss of discriminative information under occlusion. Its "Multi-modal Prototype Generator" forms category-specific visual prototypes from cropped ground-truth object patches and derives complementary language prototypes from class names processed through a lightweight BERT encoder. These visual–linguistic embeddings are combined and introduced into the backbone features using a depth-wise correlation mechanism. The Correlation-guided Query Learner leverages these correlation features in two complementary ways: first, to initialize object queries using high-confidence spatial locations, improving recall for partially occluded objects; and second, to perform dual-path mixed sampling in which 3D queries aggregate both backbone and correlation-derived features for richer semantic–geometric representation. An occlusion-aware trainer further augments the system via pseudo-occlusion, randomly masking pixels in high-visibility samples, then a contrastive alignment loss that brings occluded and obvious instances of the same category closer in the embedding space. This modular design yields consistent gains across all occlusion levels and integrates seamlessly into diverse baseline architectures with minimal computational cost.

DualViewDistill \cite{kappeler2025bridging} uses a transformer-based camera-only setup. It combines perspective view image features with a bird’s-eye view representation guided by a foundation model. A convolutional image backbone pulls multi-scale features from synchronized surround view images. These features are projected into a 3D frustum space with a depth-aware projection module and then organized into a structured BEV grid. The system includes two complementary learning streams. The first is a BEV feature distillation branch that aligns intermediate BEV features with DINOv2 foundation model features projected from sparse LiDAR data during training. The second is a transformer detection head that handles object queries defined as 3D anchors. Instead of calculating depth through direct regression, DualViewDistill supervises spatial reasoning using cosine similarity alignment in the BEV space. This approach enhances the semantic richness of map features. A deformable attention module switches between BEV and perspective view aggregation to improve object queries across different modalities and over time. Additionally, an instance memory module connects queries between frames to support stable identity tracking and 3D localization.

Table \ref{table: yolo_detection} summarizes YOLO-based methods evaluated on the camera-based KITTI dataset for 2D object detection. Monocular 2D object detection methods have evolved to advance efficiency, representational depth, and domain-specific tuning. Early designs, such as YOLOV3-tiny \cite{adarsh2020yolo}, prioritized minimizing computation cost to improve efficiency with a precision trade-off. Later variants, including YOLOv5n \cite{gao2025algorithm} with its scaled CSP-based backbone (C2f) derived from CSPDarknet53 and YOLOv6n \cite{li2022yolov6} employing the EfficientRep backbone, coupled backbone revisions with refined neck structures and decoupled heads to raise accuracy without undermining real-time operation. YOLOv7 \cite{wang2023yolov7} extended these gains by introducing re-parameterized convolution, a broadened trainable bag-of-freebies, and coarse-to-fine hierarchical label assignment.

\begin{table}[H]
\centering
\caption{Comparative analysis of Yolo-based detection methods using the camera-based KITTI dataset.}
\vspace{-0.4cm}
\label{table: yolo_detection}
\begin{center}
\resizebox{\linewidth}{!}{%
  \setlength{\tabcolsep}{4pt}%
  \begin{tabular}{llllll lll lll}
    \toprule
    \textbf{Ref}
      & \textbf{Model} & \textbf{Year} & \textbf{Venue} & \textbf{Backbone} & \textbf{Modality}
      & \textbf{Precision} & \textbf{Recall} & \textbf{mAP@50}
      & \textbf{Parameters} & \textbf{Model Size} & \textbf{FPS} \\
    \midrule

    \multirow{2}{*}{\cite{an2025improved}} 
      & SwinYOLOv5s & 2025 & MDPI & MobileNetV3 & Mono
      & \red{\textbf{96.00\%}} & \red{\textbf{90.20\%}} & \red{\textbf{95.70\%}}
      & 50.30M & $-$ & $-$ \\[2mm]
      \myrowcolour
      & \multicolumn{1}{l}{\textbf{Innovation:}}
        & \multicolumn{10}{l}{Integrates swin transformer attention and adaptive fusion to improve global feature capture and reduce false detections.} \\[2mm]

    \multirow{2}{*}{\cite{gao2025algorithm}} 
      & YOLOv8-RTDAV & 2025 & Scientific Reports & C2f & Mono
      & 86.90\% & 81.10\% & 88.00\%
      & 2.81M & 5.78MB & 102.00 \\[2mm]
      \myrowcolour
      & \multicolumn{1}{l}{\textbf{Innovation:}}
        & \multicolumn{10}{l}{Enhances YOLOv8n with ECA-based C2f, P2 small-object head, SPPELAN, and EIoU loss for better road-target accuracy.} \\[2mm]

    \multirow{2}{*}{\cite{li2024yolo}} 
      & YOLO-Vehicle-v1s & 2024 & ArXiv & CSPDarknet53 & Mono
      & $-$ & $-$ & 93.50\%
      & $-$ & 45.40MB & 226.00 \\[2mm]
      \myrowcolour
      & \multicolumn{1}{l}{\textbf{Innovation:}}
        & \multicolumn{10}{l}{Introduces vehicle-focused YOLO variant with image dehazing and multimodal image-text fusion to improve  detection.} \\[2mm]

    \multirow{2}{*}{\cite{yang2024yolov8}} 
      & YOLOv8-Lite & 2024 & ICCK  & C2f & Mono
      & 76.72\% & $-$ & 75.32\%
      & 4.39M & $-$ & 85.00 \\[2mm]
      \myrowcolour
      & \multicolumn{1}{l}{\textbf{Innovation:}}
        & \multicolumn{10}{l}{Employs FastDet backbone and CBAM attention to achieve lightweight, and high detection accuracy..} \\[2mm]

    \multirow{2}{*}{\cite{he2023object}} 
      & ShuffYOLOX & 2023 & MDPI & ShuffDet & Mono
      & $-$ & $-$ & 92.20\%
      & 35.45M & 113.90MB & $-$ \\[2mm]
      \myrowcolour
      & \multicolumn{1}{l}{\textbf{Innovation:}}
        & \multicolumn{10}{l}{Uses lightweight ShuffDet backbone and ECA-enhanced PAFPN to increase speed and maintain accuracy in AVs.}\\[2mm]

    \multirow{2}{*}{\cite{jia2023fast}} 
      & improved YOLOv5 & 2023 & Scientific Reports & RepNAS & Mono
      & 92.90\% & 90.10\% & 90.10\%
      & $-$ & $-$ & 202.00 \\[2mm]
      \myrowcolour
      & \multicolumn{1}{l}{\textbf{Innovation:}}
        & \multicolumn{10}{l}{Adds structural re-parameterization, NAS-pruned branches, small-object head, and coordinate attention for  detection.} \\[2mm]

    \multirow{2}{*}{\cite{wang2024yolov10}} 
      & YOLOv10n & 2024 & NeurIPS & CSPNet & Mono
      & 87.20\% & 72.10\% & 80.20\%
      & 27.02M & 5.50MB & 103.00 \\[2mm]
      \myrowcolour
      & \multicolumn{1}{l}{\textbf{Innovation:}}
        & \multicolumn{10}{l}{Redesigns YOLO with efficiency--accuracy balance and NMS-free training via consistent dual assignment strategy.} \\[2mm]

    \multirow{2}{*}{\cite{wang2024yolov9}} 
      & YOLOv9t & 2024 & ECCV & GELAN & Mono
      & 82.10\% & 69.30\% & 77.50\%
      & 2.01M & 4.43MB & 48.00 \\[2mm]
      \myrowcolour
      & \multicolumn{1}{l}{\textbf{Innovation:}}
        & \multicolumn{10}{l}{Introduces GELAN backbone and programmable gradient information to improve training stability and detection accuracy.} \\[2mm]

    \multirow{2}{*}{\cite{gao2025algorithm}} 
      & YOLOv8n & 2023 & N/A & C2f & Mono
      & 88.00\% & 76.30\% & 85.20\%
      & 3.01M & 5.97MB & 148.00 \\[2mm]
      \myrowcolour
      & \multicolumn{1}{l}{\textbf{Innovation:}}
        & \multicolumn{10}{l}{Employs C2f backbone and decoupled anchor-free head for stronger features and simpler inference.} \\[2mm]

    \multirow{2}{*}{\cite{wang2023yolov7}} 
      & YOLOv7 & 2023 & CVPR & ELAN & Mono
      & 92.30\% & 97.10\% & 95.20\%
      & $-$ & $-$ & 132.00 \\[2mm]
      \myrowcolour
      & \multicolumn{1}{l}{\textbf{Innovation:}}
        & \multicolumn{10}{l}{Adds trainable bag-of-freebies, re-parameterization, and coarse-to-fine lead-guided label assignment for better accuracy.}\\[2mm]

    \multirow{2}{*}{\cite{li2022yolov6}} 
      & YOLOv6n & 2022 & ArXiv & EfficientRep & Mono
      & 85.30\% & 67.10\% & 76.00\%
      & 4.23M & 8.30MB & 189.00 \\[2mm]
      \myrowcolour
      & \multicolumn{1}{l}{\textbf{Innovation:}}
        & \multicolumn{10}{l}{Employs EfficientRep backbone, Rep-PAN neck, and efficient decoupled head for speed with maintained accuracy.}\\[2mm]

    \multirow{2}{*}{\cite{gao2025algorithm}} 
      & YOLOv5n & 2020 & N/A & CSPDarknet53 & Mono
      & 84.90\% & 70.00\% & 78.60\%
      & 2.51M & 5.04MB & 127.00 \\[2mm]
      \myrowcolour
      & \multicolumn{1}{l}{\textbf{Innovation:}}
        & \multicolumn{10}{l}{Nano-scale YOLOv5 variant with CSPDarknet backbone for lightweight real-time detection.} \\[2mm]

    \multirow{2}{*}{\cite{adarsh2020yolo}} 
      & YOLOv3-tiny & 2020 & ICACCS & Tiny Darknet & Mono
      & 82.90\% & 64.80\% & 72.50\%
      & 1.21M & 23.20MB & 276.00 \\[2mm]
      \myrowcolour
      & \multicolumn{1}{l}{\textbf{Innovation:}}
        & \multicolumn{10}{l}{Compact YOLOv3 variant with reduced Darknet backbone trading accuracy for faster inference.}\\[2mm]

\bottomrule
\end{tabular}}
\end{center}
\end{table}

\vspace{-0.5cm}
In turn, YOLOv8n \cite{gao2025algorithm} replaced C3 modules with C2f blocks. It adopted an anchor-free head, simplifying the pipeline while preserving detection strength, whereas YOLOv8-Lite \cite{yang2024yolov8} targeted resource-limited environments through compact backbone choices, lightweight feature aggregation, and attention mechanisms. ShuffYOLOX \cite{he2023object} modified the YOLOX framework by utilizing the lightweight ShuffDet backbone and integrating ECA attention within the PAFPN, which reduces architectural complexity without negatively affecting detection capabilities. Domain-specific optimization has emerged as a critical trend within recent iterations. YOLO-Vehicle-v1s \cite{li2024yolo} was designed for vehicle detection in adverse visual conditions, incorporating image dehazing and multimodal image-text fusion to enhance robustness. YOLOv8-RTDAV \cite{gao2025algorithm} targeted road-target detection by embedding ECA-based C2f modules, DySample-based upsampling, and EIoU loss, yielding notable improvements in small-object accuracy. Transformer-CNN hybridization appeared in SwinYOLOv5s \cite{an2025improved}, which leveraged swin transformer attention with adaptive self-concat fusion to capture global context while mitigating false detections. Similarly, improved YOLOv5 \cite{jia2023fast} employed NAS-pruned branches, coordinate attention, and structural re-parameterization to achieve both speed and high precision for small-object detection. The latest designs, YOLOv9t \cite{wang2024yolov9} and YOLOv10n \cite{wang2024yolov10}, represent a shift toward more fundamental architectural changes: the introduction of the GELAN backbone and programmable gradient optimization in YOLOv9t improves training stability, while YOLOv10n adopts an NMS-free consistent dual-assignment strategy to harmonize efficiency with accuracy. Collectively, these methods illustrate a clear trajectory from speed-oriented architectures to hybrid designs that balance inference efficiency with enhanced feature representation.

\subsection{3D LiDAR-based Techniques}
\label{sec: 3D}
LiDAR sensors play a pivotal role in AV perception by providing accurate 3D geometric information of the surrounding environment, enabling precise object localization and size estimation under varying lighting and weather conditions. Unlike camera-based approaches that operate in the 2D image plane, LiDAR-based methods directly process point clouds, offering richer spatial representations and improved robustness to illumination changes. Over the past decade, a wide range of 3D LiDAR-based object detection architectures have emerged, each employing different spatial representations and learning strategies to balance accuracy, efficiency, and robustness \cite{cao2025kptr}. In this survey, we categorize existing approaches into six main groups, including BEV-based, point-based, voxel-based, pillar-based, range-based, and point-voxel hybrid methods, providing a detailed analysis of their core principles, representative algorithms, strengths, and limitations.

BEV-based methods take LiDAR point clouds and project them onto a top-down Bird's-Eye View grid, where each cell collects information from the points that fall within it. This BEV representation gets treated like a regular 2D image and processed through convolutional neural networks. The approach is computationally efficient and works well for understanding spatial relationships like road layouts, lane structures, and where vehicles are positioned. Since it shows the scene from above, it naturally fits with the mapping and path planning systems that autonomous cars use. However, flattening everything into this top-down view can reduce depth detail and hide information about what's stacked vertically, making it harder to detect objects hidden behind others or to capture fine height variations. The output is typically a set of 3D bounding boxes in the global coordinate frame, along with class labels and confidence scores.

\begin{figure}[H]
    \centering
    \centerline{\includegraphics[width=1\textwidth]{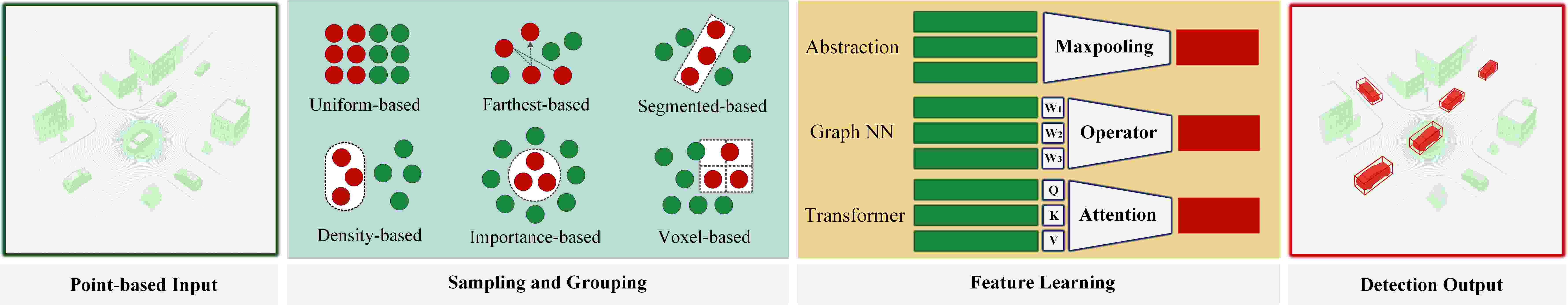}}
    \caption{Overall framework of the Point-based approaches in 3D Lidar object detection.}
    \label{fig: point-based}
\end{figure}

Figure \ref{fig: point-based} illustrates the overall architecture of point-based approaches in 3D LiDAR-based object detection. Point-based methods directly operate on raw, unordered LiDAR point clouds by applying feature sampling, grouping, and point-wise feature learning to extract meaningful representations from the data. Then, 3D bounding boxes are predicted based on the downsampled points and features. Sampling is an important step in the architecture, and several strategies are commonly employed in 3D LiDAR-based detection: Random Sampling, which selects points without geometric bias; Farthest Point Sampling (FPS), which iteratively selects the farthest points to ensure spatial coverage; Voxel Grid Sampling, which divides the space into uniform voxels and selects representative points from each; Uniform Sampling, which ensures evenly spaced points within a defined extent; Importance Sampling, which prioritizes points with higher learned or predefined significance; and Density-based Sampling, which reduces over-representation of dense areas while preserving points in sparse regions. By processing each sampled point individually while grouping local neighborhoods, point-based methods preserve complete geometric details of the scene, making them highly suitable for fine-grained shape modeling. However, the need to handle large point sets in dense urban environments leads to high computational costs, which can limit real-time deployment in autonomous driving.

\begin{figure}[H]
    \centering
    \centerline{\includegraphics[width=1\textwidth]{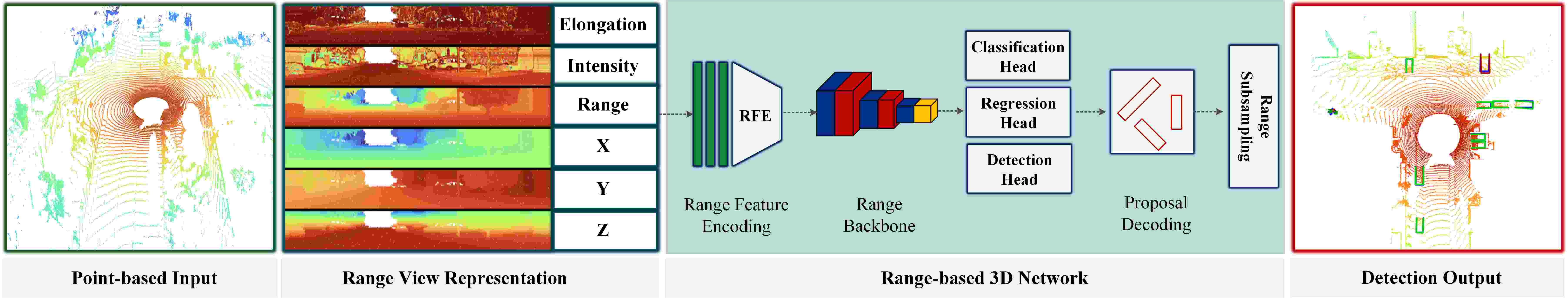}}
    \caption{Overall framework of the Range-based approaches in 3D Lidar object detection.}
    \label{fig: range-based}
\end{figure}

\vspace{-0.3cm}
Figure \ref{fig: range-based} illustrates the overall architecture of range-based LiDAR object detection methods. Range-based methods project LiDAR point clouds into a Range View representation using spherical coordinates, where each pixel corresponds to a single LiDAR return. Typically, the input combines geometric features, such as range, X, Y, and Z coordinates, with auxiliary features like elongation and intensity. For instance, in the Waymo dataset, each channel is remapped so that warmer colors represent the smallest values and cooler colors represent the largest values within their respective domains. This representation preserves accurate depth information and provides full 360° coverage of the environment. By treating the range image as a dense 2D grid, 2D convolutional neural networks can efficiently process the data while leveraging well-established image-based techniques. Range-based methods are particularly effective for long-range perception, as they maintain precise range and angular relationships. However, they can introduce spatial distortions, especially for far-away or steeply angled objects. 

\begin{figure}[H]
    \centering
    \centerline{\includegraphics[width=1\textwidth]{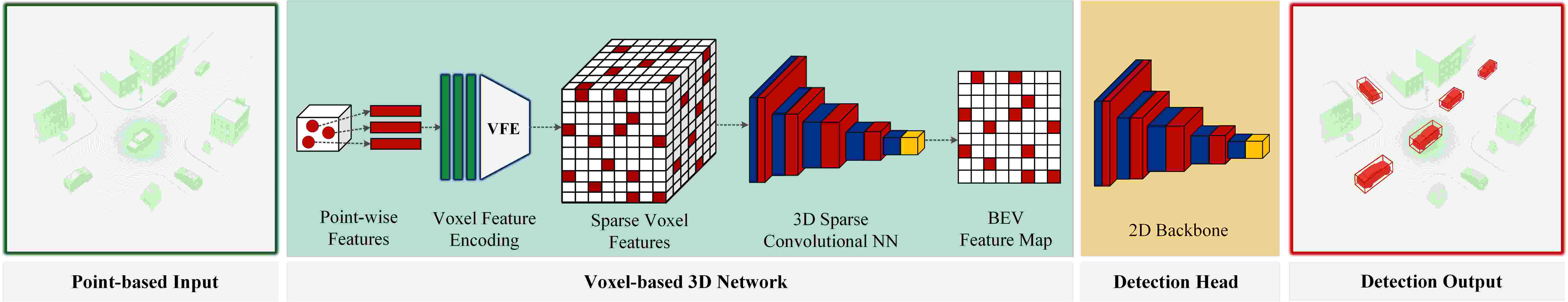}}
    \caption{Overall framework of the Voxel-based approaches in 3D Lidar object detection.}
    \label{fig: Voxel-based}
\end{figure}

\vspace{-0.3cm}
Figure \ref{fig: Voxel-based} presents the overall architecture of voxel-based 3D LiDAR object detection methods, which includes the voxelization process, feature extraction, and detection head. Voxel-based methods transform raw LiDAR point clouds into structured volumetric representations by dividing the 3D space into spaced voxels. In order to extract spatial features, 3D convolutional neural networks (3D CNNs) aggregate the points within each voxel. This representation makes Convolutional efficiency possible while maintaining the significance of geometric context. Voxelization, transforming point clouds into fixed-size voxel grids, is a crucial stage in voxel-based architectures. The resolution of these voxels significantly impacts performance: finer resolutions capture more geometric detail but increase memory consumption and computational cost. In comparison, rougher resolutions reduce resource demands but may lose critical structural information. Some methods also employ sparse convolution techniques to process only occupied voxels, improving efficiency. By applying convolutions over the voxelized space, these methods effectively capture local and global spatial patterns, making them suitable for large-scale autonomous driving scenes. However, voxel-based approaches suffer from quantization errors due to discretization, and high-resolution grids can lead to substantial memory overhead.

\begin{figure}[H]
    \centering
    \centerline{\includegraphics[width=1\textwidth]{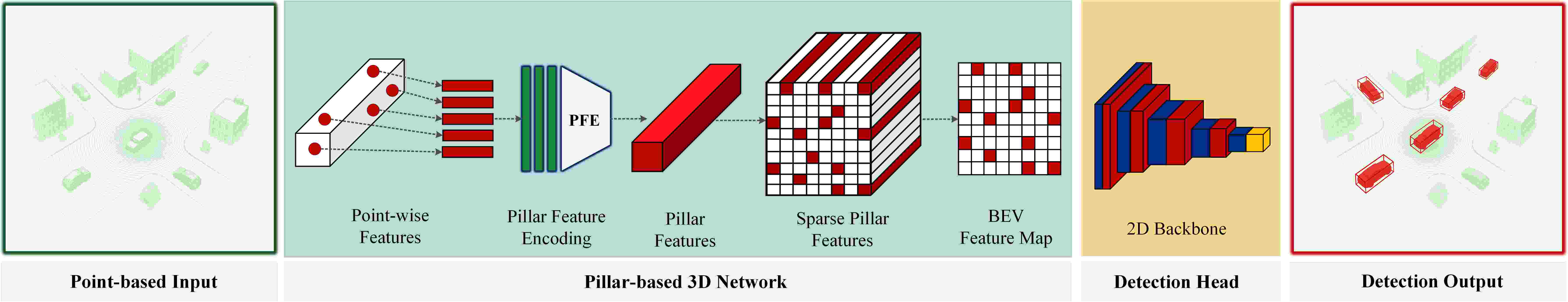}}
    \caption{Overall framework of the Pillar-based approaches in 3D Lidar object detection.}
    \label{fig: Pillar-based}
\end{figure}

\vspace{-0.3cm}
Figure \ref{fig: Pillar-based} outlines the overall architecture of pillar-based LiDAR object detection methods, including the pillar transformation process, pillar feature encoding, and detection head. Pillar-based methods organize LiDAR point cloud data by dividing the scene into vertical columns, or “pillars,” and merging the height dimension. Instead of keeping the whole 3D structure, the data is flattened into a 2D layout, with each pillar storing the features from the inside points. These features are often aggregated using methods such as mean pooling or learned encoders before being passed to the detection network. This design makes the 2D convolutional networks faster and less memory-intensive, enabling high throughput on embedded automotive hardware. The advantage of this approach lies in its efficiency, making it well-suited for real-time detection in autonomous driving scenarios. However, by discarding vertical resolution, pillar-based methods may struggle with tall or vertically complex objects, where fine height details are important for accurate recognition. The final output is typically a set of 3D bounding boxes, each with associated class labels and confidence scores.

\begin{figure}[H]
    \centering
    \centerline{\includegraphics[width=1\textwidth]{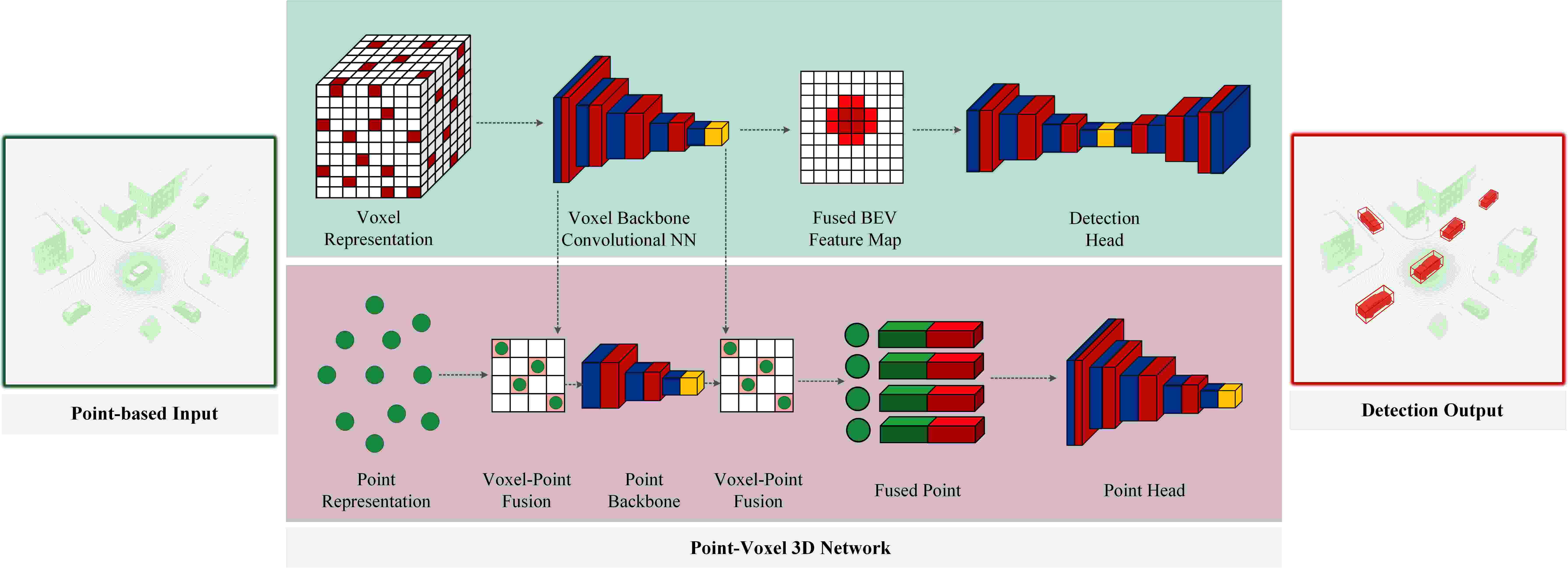}}
    \caption{Overall framework of the Voxel-Point approaches in 3D Lidar object detection.}
    \label{fig: voxel-point}
\end{figure}

\vspace{-0.3cm}
Figure \ref{fig: voxel-point} presents the overall architecture of point–voxel hybrid LiDAR object detection methods, including the voxelization stage, point-level feature extraction, feature fusion, and detection head. Point–voxel hybrid methods combine the strengths of point-based and voxel-based methods to achieve greater accuracy and efficiency. The key idea is to learn fine-grained point features that capture local geometric details, while extracting structured voxel features that enable efficient convolutional operations on regular grids. Typically, raw point clouds are first voxelized to produce coarse spatial features, and representative points are sampled (e.g., using farthest point sampling) to extract high-resolution features. These two feature streams are then fused, either through concatenation or attention-based aggregation, to provide a richer representation for the detection head. This strategy preserves detailed geometry while maintaining the computational benefits of voxel processing. However, the architecture and training process are generally more complex, often requiring careful balancing of the contributions from each branch. The final output typically includes accurate 3D bounding boxes, along with what each object is and how confident the system is in those predictions.\\

\vspace{-0.3cm}
In the following, we present all existing 3D LiDAR-based algorithms for object detection in autonomous vehicles from 2018 to 2025. To ensure a consistent and fair comparison, we organize the results into three separate tables, as the performance of these methods is evaluated on different datasets. Table \ref{table: 3D-Kitti} provides a comparative analysis of 3D object detection methods using LiDAR-based data from the KITTI dataset, while Table \ref{table: 3D-Nuscenes} summarizes 3D object detection methods evaluated on the LiDAR-based NuScenes dataset. Finally, Table \ref{table:3D-VOD} presents a comparative analysis of 3D object detection methods based on LiDAR–Radar data from the VOD dataset.

\begin{table}[H]
\centering
\caption{Comparative analysis of 3D object detection methods using the LiDAR-based KITTI dataset.}
\vspace{-0.4cm}
\label{table: 3D-Kitti}
\begin{center}
\resizebox{\textwidth}{!}{%
  \setlength{\tabcolsep}{4pt}%
  \begin{tabular}{llllll ccc ccc ccc ccc c}
    \toprule
    \multicolumn{6}{c}{} 
      & \multicolumn{3}{c}{$\textbf{AP}_{\textbf{BEV}}$ \hspace{1mm}\textbf{Car}} 
      & \multicolumn{3}{c}{$\textbf{AP}_{\textbf{3D}}$\hspace{1mm}\textbf{Car}}
      & \multicolumn{3}{c}{$\textbf{AP}_{\textbf{3D}}$\hspace{1mm}\textbf{Pedestrian}} 
      & \multicolumn{3}{c}{$\textbf{AP}_{\textbf{3D}}$\hspace{1mm}\textbf{Cyclist}} \\[1mm]
    \cmidrule(l){7-9}\cmidrule(l){10-12}\cmidrule(l){13-15}\cmidrule(l){16-18}
    \textbf{Ref} & \textbf{Method} & \textbf{Year} & \textbf{Venue} & \textbf{Backbone} & \textbf{Modality}
      & \textbf{Easy} & \textbf{Moderate} & \textbf{Hard}
      & \textbf{Easy} & \textbf{Moderate} & \textbf{Hard}
      & \textbf{Easy} & \textbf{Moderate} & \textbf{Hard}
      & \textbf{Easy} & \textbf{Moderate} & \textbf{Hard} 
      & \textbf{Runtime} \\
    \midrule

\multirow{1}{*}{\cite{ye2025fade3d}} 
  & Fade3D & 2025 & T-ITS & SFENet & Point
  & 94.86 & 89.28 & 86.55
  & 90.92 & 82.00 & 77.49
  & $-$ & $-$ & $-$
  & $-$ & $-$ & $-$ & 0.01 Sec\\
\myrowcolour
  & \multicolumn{18}{l}{\textbf{Innovation:} Lightweight input encoder, spatially enhanced BEV backbone, and IoU-aware re-weighting for deployable 3D detection.} \\[2mm]

\multirow{1}{*}{\cite{ding2025det}} 
  & AS-Det & 2025 & AAAI & AS & Point
  & 95.22 & 91.33 & 86.25
  & 90.55 & 82.41 & 77.08
  & $-$ & $-$ & $-$
  & $-$ & $-$ & $-$ & \textbf{$-$}\\
\myrowcolour
  & \multicolumn{18}{l}{\textbf{Innovation:} Active sampling and multi-scale center-feature aggregation for adaptable single-stage 3D detection (LiDAR/4D Radar).} \\[2mm]

\multirow{1}{*}{\cite{liu2024lion}} 
  & LION & 2024 & NeurIPS & LION & Voxel
  & 91.40 & 89.29 & 87.26
  & 89.25 & 80.97 & 78.28
  & $-$ & $-$ & $-$
  & $-$ & $-$ & $-$ & \textbf{$-$}\\
\myrowcolour
  & \multicolumn{18}{l}{\textbf{Innovation:} Linear Group RNN models long-range voxel dependencies efficiently.} \\[2mm]

\multirow{1}{*}{\cite{hoang2024tsstdet}} 
  & TSSTDet & 2024 & IEEE & 3D CNN  & Voxel
  & \red{\textbf{95.80}} & \red{\textbf{92.11}} & \red{\textbf{89.23}}
  & 91.84 & 85.47 & 80.65
  & 75.13 & 69.38 & 64.31
  & \red{\textbf{95.16}} & \red{\textbf{76.24}}  & \red{\textbf{71.62}}  & 0.02 Sec\\
\myrowcolour
  & \multicolumn{18}{l}{\textbf{Innovation:} Models spatial shape features with a transformer inside the voxel pipeline.} \\[2mm]

\multirow{1}{*}{\cite{jin2024swiftpillars}} 
  & SwiftPillars & 2024 & AAAI & SPE & Pillar
  & $-$ & $-$ & $-$
  & 88.07 & 79.83 & 77.50
  & 55.34 & 50.21 & 45.93
  & 86.06 & 64.92 & 60.75 & 0.013 Sec\\
\myrowcolour
  & \multicolumn{18}{l}{\textbf{Innovation:} Dual-attention pillar encoding with lightweight operators and multi-scale aggregation for high-speed deployment.} \\[2mm]


\multirow{1}{*}{\cite{yang2023pvt}} 
  & PVT-SSD & 2023 & CVPR & Sparse CNN & Point-Voxel
  & 95.23 & 91.63 & 86.43
  & 90.66 & 82.29 & 76.85
  & $-$ & $-$ & $-$
  & $-$ & $-$ & $-$ & 0.04 Sec\\
\myrowcolour
  & \multicolumn{18}{l}{\textbf{Innovation:} Point-voxel transformer with query initialization and virtual range-image acceleration.} \\[2mm]


\multirow{1}{*}{\cite{wang2023dsvt}} 
  & DSVT & 2023 & CVPR &Transformer & Voxel
  & 92.04 & 88.33 & 87.81
  & 87.89 & 80.90 & 78.66
  & $-$ & $-$ & $-$
  & $-$ & $-$ & $-$ & 0.037 Sec\\
\myrowcolour
  & \multicolumn{18}{l}{\textbf{Innovation:} Dynamic sparse window attention and rotated set partitioning; no custom CUDA.} \\[2mm]

\multirow{1}{*}{\cite{zhao2023ada3d}} 
  & Ada3D & 2023 & ICCV & 3D/2D CNN & Voxel
  & 93.51 & 89.42 & 88.19
  & 89.70 & 85.29 & 82.86
  & $-$ & $-$ & $-$
  & $-$ & $-$ & $-$ & \textbf{$-$}\\
\myrowcolour
  & \multicolumn{18}{l}{\textbf{Innovation:} Adaptive inference via importance prediction, density guidance, and sparsity-preserving normalization.} \\[2mm]

\multirow{1}{*}{\cite{chen2023voxelnext}} 
  & VoxelNeXt & 2023 & CVPR & Sparse CNN & Voxel
  & 92.66 & 87.87 & 87.04
  & 87.86 & 80.02 & 77.34
  & $-$ & $-$ & $-$
  & $-$ & $-$ & $-$ & 0.066 Sec\\
\myrowcolour
  & \multicolumn{18}{l}{\textbf{Innovation:} Dense heads removed; direct prediction from sparse voxel features for efficient detection/tracking.} \\[2mm]

\multirow{1}{*}{\cite{xia20233}} 
  & 3D HANet & 2023 & TGRS & VFE  & Point-Voxel
  & 94.33 & 91.63 & 86.33
  & 90.79 & 84.18 & 77.57
  & $-$ & $-$ & $-$
  & $-$ & $-$ & $-$ & 0.06 Sec\\
\myrowcolour
  & \multicolumn{18}{l}{\textbf{Innovation:} Plug-and-play 3D heatmap auxiliary network enhancing spatial perception without added latency.} \\[2mm]


\multirow{1}{*}{\cite{wu2023transformation}} 
  & TED-S & 2023 & AAAI & TESP CNN & Voxel
  & $-$ & $-$ & $-$
  & \red{\textbf{93.05}}  & \red{\textbf{87.91}}   & \red{\textbf{85.81}} 
  & 72.38 & 67.81 & 63.54
  & 93.09 & 75.77 & 71.20 & \textbf{$-$}\\
\myrowcolour
  & \multicolumn{18}{l}{\textbf{Innovation:} Transformation-equivariant voxel features with TeSpConv; TeBEV and TiVoxel pooling.} \\[2mm]

\multirow{1}{*}{\cite{hoang20233onet}} 
  & 3ONet & 2023 & IEEE  & RPN & Point
  & \red{\textbf{95.87}} & \red{\textbf{90.07}} & \red{\textbf{85.09}}
  & 92.03 & 85.47 & 78.64
  & 52.81 & 43.45 & 39.74
  & 82.34 & 68.37 & 60.20 & 0.03 Sec\\
\myrowcolour
& \multicolumn{1}{l}{\textbf{Innovation:}} 
  & \multicolumn{17}{l}{Proposes a structure-aware two-stage 3D detector that learns to recover occluded object geometry.} \\[2mm]

\multirow{1}{*}{\cite{sheng2022rethinking}} 
  & RDIoU & 2022 & ECCV & CT-stacked & Voxel
  & 94.90 & 89.75 & 84.67
  & 90.65 & 82.30 & 77.26
  & $-$ & $-$ & $-$
  & $-$ & $-$ & $-$ & 0.03 Sec\\
\myrowcolour
  & \multicolumn{18}{l}{\textbf{Innovation:} Rotation-Decoupled IoU stabilizes training by mitigating rotation coupling in 3D IoU optimization.} \\[2mm]

\multirow{1}{*}{\cite{zhang2022not}} 
  & IA-SSD & 2022 & CVPR & PointNet++ & Point
  & 93.14 & 89.48 & 84.42
  & 88.87 & 80.32 & 75.10
  & 46.51 & 39.03 & 35.60
  & 78.35 & 61.94 & 55.70 & 0.05 Sec\\
\myrowcolour
  & \multicolumn{18}{l}{\textbf{Innovation:} Learnable instance-aware downsampling and contextual centroid perception for efficient point-based detection.} \\[2mm]

\multirow{1}{*}{\cite{shi2022pillarnet}} 
  & PillarNet & 2022 & ECCV & VGG-18/34 & Pillar
  & 93.66 & 87.85 & 86.64
  & 89.94 & 79.01 & 77.31
  & $-$ & $-$ & $-$
  & $-$ & $-$ & $-$ & 0.053 Sec\\
\myrowcolour
  & \multicolumn{18}{l}{\textbf{Innovation:} Powerful sparse 2D CNN encoder with optimized neck for pillar-based detection.} \\[2mm]

\multirow{1}{*}{\cite{he2022voxel}} 
  & VoxSeT & 2022 & CVPR & VoxSeT & Voxel
  & 92.70 & 89.07 & 86.29
  & 88.53 & 82.06 & 77.46
  & $-$ & $-$ & $-$
  & $-$ & $-$ & $-$ & 0.033 Sec\\
\myrowcolour
  & \multicolumn{18}{l}{\textbf{Innovation:} Voxel-based set attention enabling transformer self-attention on variable-size clusters.} \\[2mm]


\multirow{1}{*}{\cite{xu2022behind}} 
  & BtcDet & 2022 & AAAI & RPN & Voxel
  & 92.81 & 89.34 & 84.53
  & 90.60 & 83.02 & 77.36
  & 69.39 & 61.19 & 55.86
  & 91.45 & 74.70 & 70.08 & 0.10 Sec\\
\myrowcolour
  & \multicolumn{18}{l}{\textbf{Innovation:} Learns occluded shape priors and integrates shape occupancy into proposal generation/refinement.} \\[2mm]


\multirow{1}{*}{\cite{deng2021voxel}} 
  & Voxel-RCNN & 2021 & AAAI & RPN & Voxel
  & 93.54 & 91.18 & 88.91
  & 92.38 & 85.29 & 82.86
  & 64.88 & 57.32 & 52.11
  & 89.25 & 72.52 & 67.03 & 0.09 Sec\\
\myrowcolour
  & \multicolumn{18}{l}{\textbf{Innovation:} Voxel RoI pooling to directly extract 3D voxel features for refinement.} \\[2mm]

\multirow{1}{*}{\cite{zheng2021cia}} 
  & CIA-SSD & 2021 & AAAI & SPConvNet & Voxel
  & 93.74 & 89.84 & 82.39
  & 89.59 & 80.28 & 72.87
  & $-$ & $-$ & $-$
  & $-$ & $-$ & $-$ & 0.09 Sec\\
\myrowcolour
  & \multicolumn{18}{l}{\textbf{Innovation:} Aligns confidence and localization via feature aggregation with IoU-aware confidence rectification.} \\[2mm]

\multirow{1}{*}{\cite{sheng2021improving}} 
  & CT3D & 2021 & ICCV & RPN & Voxel
  & 92.36 & 88.83 & 84.07
  & 87.83 & 81.77 & 77.16
  & 65.73 & 58.56 & 53.04
  & 91.99 & 71.60 & 67.34 & 0.11 Sec\\
\myrowcolour
  & \multicolumn{18}{l}{\textbf{Innovation:} Channel-wise Transformer with proposal-to-point embedding and extended re-weighting.} \\[2mm]

\multirow{1}{*}{\cite{xu2021spg}} 
  & SPG & 2021 & ICCV & VFE & Pillar
  & 92.80 & 89.12 & 86.27
  & 90.64 & 82.66 & 77.91
  & $-$ & $-$ & $-$
  & $-$ & $-$ & $-$ & 0.09 Sec\\
\myrowcolour
  & \multicolumn{18}{l}{\textbf{Innovation:} Generates semantic points in missing regions to recover degraded LiDAR data.} \\[2mm]

\multirow{1}{*}{\cite{zheng2021se}} 
  & SE-SSD & 2021 & CVPR & SPConvNet & Voxel
  & 95.68 & 91.84 & 87.62
  & 91.49 & 82.54 & 77.15
  & 67.98 & 59.72 & 54.83
  & 91.77 & 72.54 & 68.78 & 0.08 Sec\\
\myrowcolour
  & \multicolumn{18}{l}{\textbf{Innovation:} Self-ensembling with consistency constraints, orientation-aware IoU loss, and shape-aware augmentation.} \\[2mm]


\multirow{1}{*}{\cite{fan2021rangedet}} 
  & RangeDet & 2021 & ICCV &  FPN & Range
  & 87.96 & 69.03 & 48.88
  & $-$ & $-$ & $-$
  & \red{\textbf{82.20}} & \red{\textbf{75.39}} & \red{\textbf{65.74}}
  & $-$ & $-$ & $-$ & \textbf{$-$}\\
\myrowcolour
  & \multicolumn{18}{l}{\textbf{Innovation:} Pure range-view detector with Meta-Kernel, range-conditioned FPN, and weighted NMS.} \\[2mm]

\multirow{1}{*}{\cite{he2020structure}} 
  & SA-SSD & 2020 & CVPR & RPN & Voxel
  & 95.03 & 91.03 & 85.96
  & 88.75 & 79.79 & 74.16
  & 61.72 & 55.01 & 49.94
  & 88.82 & 71.25 & 65.88 & 0.04 Sec\\
\myrowcolour
  & \multicolumn{18}{l}{\textbf{Innovation:} Improves single-stage detection with a detachable auxiliary network, point-level supervision, and part-sensitive warping.} \\[2mm]

\multirow{1}{*}{\cite{shi2020points}} 
  &Part-A$^2$  & 2020 & IEEE & SPConvNet  & Point-Voxel
  &84.76 &89.52 &81.47
  &77.86 &85.94 &72.00
  &44.50 &54.49 &42.36
  &62.73 &75.58 &57.74 & 0.08 Sec\\
\myrowcolour
  & \multicolumn{18}{l}{\textbf{Innovation:} Part-aware segmentation and part-aggregation RoI pooling for proposal refinement.} \\[2mm]

\multirow{1}{*}{\cite{shi2020point}} 
  & Point-GNN & 2020 & CVPR & GNN & Point
  & 93.11 & 89.17 & 83.90
  & 88.33 & 79.47 & 72.29
  & $-$ & $-$ & $-$
  & $-$ & $-$ & $-$ & 0.60 Sec\\
\myrowcolour
  & \multicolumn{18}{l}{\textbf{Innovation:} Fixed-radius graph with auto-registration reduces translation variance; box-merging improves accuracy.} \\[2mm]

\multirow{1}{*}{\cite{yang20203dssd}} 
  & 3DSSD & 2020 & CVPR & SA & Point
  & 92.66 & 89.02 & 85.86
  & 88.36 & 79.57 & 74.55
  & 35.03 & 27.76 & 26.08
  & 66.69 & 59.00 & 55.62 & 0.04 Sec\\
\myrowcolour
  & \multicolumn{18}{l}{\textbf{Innovation:} Removes FP layers and refinement using fusion sampling, anchor-free regression, and 3D center-ness labels.} \\[2mm]

\multirow{1}{*}{\cite{liu2020tanet}} 
  & TANet & 2020 & AAAI & Attention & Voxel
  & $-$ & $-$ & $-$
  & 83.81 &75.38 &67.66
  & 54.92 &46.67 &42.42 
  & 73.84 &59.86 &53.46 & \textbf{$-$}\\
\myrowcolour
  & \multicolumn{18}{l}{\textbf{Innovation:} Triple attention (channel/point/voxel) with coarse-to-fine regression.} \\[2mm]

\multirow{1}{*}{\cite{tang2020searching}} 
  & SPVCNN & 2020 & ECCV & SPVCNN & Point-Voxel
  & $-$ & $-$ & $-$
  & $-$ & $-$ & $-$
  & 49.20 & 41.40 & 38.40
  & 80.10  & 63.70  & 56.20 & \textbf{$-$}\\
\myrowcolour
  & \multicolumn{18}{l}{\textbf{Innovation:} Sparse Point-Voxel Convolution with high-resolution point branch and NAS-optimized design.} \\[2mm]

\multirow{1}{*}{\cite{shi2020pv}} 
  & PV-RCNN & 2020 & CVPR & 3D CNN & Point-Voxel
  & 94.98 & 90.65 & 86.14
  & 90.25 & 81.43 & 76.82
  & 64.26 & 56.67 & 51.91
  & 88.88 & 71.95 & 66.78 & 0.08 Sec\\
\myrowcolour
  & \multicolumn{18}{l}{\textbf{Innovation:} Voxel-to-keypoint encoding and keypoint-to-grid RoI abstraction.} \\[2mm]

\multirow{1}{*}{\cite{shi2019pointrcnn}} 
  & PointRCNN & 2019 & CVPR & PointNet++ & Point
  & 92.13 & 87.39 & 82.72
  & 86.96 & 75.64 & 70.70
  & 47.98 & 39.37 & 36.01
  & 74.96 & 58.82 & 52.53 & 0.10 Sec\\
\myrowcolour
  & \multicolumn{18}{l}{\textbf{Innovation:} Bottom-up 3D proposals via point-cloud segmentation, refined in canonical coordinates using bin-based regression.} \\[2mm]

\multirow{1}{*}{\cite{yang2019std}} 
  & STD & 2019 & ICCV & PointNet++ & Point-Voxel
  & 89.66 & 65.32 & 76.06
  & 86.61 & 77.63 & 76.06
  & 53.08 & 44.24 & 41.97
  & 78.89 & 62.53 & 55.77 & \textbf{$-$}\\
\myrowcolour
  & \multicolumn{18}{l}{\textbf{Innovation:} Transformer backbone on sparse voxels with Local/Dilated Attention and GPU hash--based Fast Voxel Query.} \\[2mm]

\multirow{1}{*}{\cite{yan2018second}} 
  & SECOND & 2018 & MDPI & Sparse CNN & Voxel
  & 88.07 & 79.37 & 77.95
  & 83.13 & 73.66 & 66.20
  & 45.31 & 35.52 & 33.14
  & 75.83 & 60.82 & 53.67 & 0.04 Sec\\
\myrowcolour
  & \multicolumn{18}{l}{\textbf{Innovation:} Introduces sparse convolution with GPU-optimized rule generation, novel angle loss, and ground-truth sampling for faster LiDAR 3D detection.} \\[2mm]

\multirow{1}{*}{\cite{zhou2018voxelnet}} 
  & VoxelNet & 2018 & CVPR & VFE & Voxel
  & 89.60 & 84.81 & 87.57
  & 81.97 & 65.46 & 62.85
  & 65.95 & 61.05 & 56.98
  & 74.41 & 52.18 & 50.49 & 0.23 Sec\\
\myrowcolour
  & \multicolumn{18}{l}{\textbf{Innovation:} End-to-end voxel feature encoding learning local geometry and global context.} \\[2mm]

\bottomrule
\end{tabular}
}
\end{center}
\end{table}

\vspace{-0.1cm}
From Table \ref{table: 3D-Kitti}, we have selected four models for detailed discussion, focusing on their distinctive contributions for 3D object detection in KITTI dataset: TED-S \cite{wu2023transformation}, obtained the highest results in $AP_{3D}$ $Car$ evaluations; TSSTDet \cite{hoang2024tsstdet}, which achieved the highest results in $AP_{BEV}$ $Car$ and $AP_{3D}$ $Cyclist$ evaluations; SECOND \cite{yan2018second}, noted as one of the most widely adopted models in this field; and Fade3D \cite{ye2025fade3d}, representing the latest advancement in this domain. It should be mentioned that the values highlighted in red represent the best performance achieved for each evaluation metric across all listed methods.

TED-S \cite{wu2023transformation} is a two-stage LiDAR 3D object detector that introduces transformation-equivariant processing into the Voxel-RCNN framework. The network first applies a Transformation-equivariant Sparse Convolution (TeSpConv) backbone, in which the input point cloud is transformed into multiple discrete yaw-rotated and mirrored copies, voxelized, and processed with shared sparse convolutional weights to produce multi-channel equivariant voxel features. These are aggregated at the scene level through TeBEV pooling, which collapses each transformation channel into a BEV map, aligns them via bilinear sampling, and max-pools across channels to form a compact, transformation-consistent representation for the 2D RPN proposal stage. Proposals are then refined using TiVoxel pooling, where instance-level grids are projected into each transformation channel, voxel features are sampled via Voxel Set Abstraction, concatenated across channels, and fused using cross-grid attention for the final detection head. This design preserves geometric consistency under predefined rigid transformations while leveraging an efficient RPN-refinement pipeline for high-accuracy 3D detection.

TSSTDet framework \citep{hoang2024tsstdet} utilizes a multistage design that addresses two challenges in LiDAR-based object detection: the considerable variation in object orientation and occlusion affecting geometry. In the first stage, a Rotational-Transformation Convolutional backbone (RTConv) processes multiple rotated and reflected variants of the input point cloud through a shared sparse convolutional network, producing sets of voxel features that remain equivariant to the applied transformations. These features are aligned and merged into a unified bird's-eye-view representation via a custom RT BEV pooling process, which feeds a region proposal network to generate initial 3D candidate boxes. Building on these proposals, the second stage introduces the Voxel-Point Shape Transformer (VPST). This autoregressive transformer module reconstructs the object geometry by predicting a sequence of quantized voxel features from the partial observation, thereby enriching the proposal with occlusion-compensated detail. The final stage uses a Fusion and Refinement network, which combines the original rotation equivariant features with the completed shape features for enabling high-confidence bounding box predictions.

SECOND \cite{yan2018second} extends voxel-based 3D detection by introducing a sparse convolutional backbone that dramatically reduces the computational footprint of earlier dense 3D convolution designs. The network begins by grouping points into voxels and applying a Voxel Feature Encoding (VFE) module to produce learned descriptors for each occupied cell. Those features enter the Sparse Convolutional Middle Extractor, a series of 3D sparse convolutional layers that operate on non-empty locations to maintain spatial fidelity without unnecessary computation in space. The vertical resolution is reduced as processing progresses, enabling the feature map to be reshaped into a dense BEV representation suitable for efficient 2D convolutional processing. This representation is then handled by a region proposal network with an SSD-like multi-scale design, combining standard and transposed convolutions to generate proposals across object sizes. This design also introduces refinements such as a sine-error formulation for orientation regression, which further enhances geometric precision and detection stability while maintaining the backbone's streamlined nature.

Fade3D framework \cite{ye2025fade3d} aims to minimize computational latency while maintaining the essential spatial cues for 3D object detection. The processing begins with a  Lightweight Input Encoder (LIE), which converts the raw point cloud into a compact Birds Eye View (BEV) image. Rather than employing computationally intensive point-based encoders, the system relies on a pilot study showing that recording only the maximum height and maximum reflectance per grid cell achieves an optimal balance between accuracy and runtime. This encoded BEV representation is passed to SFENet, a specialized 2D convolutional backbone constructed in three stages: an initial convolutional block for base feature extraction, a depth-wise shuffle block augmented with channel attention for efficient mid-level processing, and a re-parameterized block that simplifies to a single convolution at inference time to accelerate deployment. The final stage integrates an IoU aware loss re-weighting strategy toward more challenging samples without impacting runtime, which ensures an uncompromised inference speed.

\begin{table*}[htbp]
\centering
\caption{Comparative analysis of 3D object detection methods using the LiDAR-based NuScenes dataset.}
\vspace{-0.4cm}
\label{table: 3D-Nuscenes}
\begin{center}
\resizebox{\textwidth}{!}{%
  \setlength{\tabcolsep}{4pt}%
  \begin{tabular}{llllll cc cccccccccc}
    \toprule
    \textbf{Ref} & \textbf{Method} & \textbf{Year} & \textbf{Venue} & \textbf{Backbone} & \textbf{Modality}
      & \textbf{NDS} & \textbf{mAP} \hspace{2mm}  & \textbf{Car} \hspace{2mm} & \textbf{Truck} \hspace{2mm} & \textbf{Bus} \hspace{2mm} & \textbf{Trailer} &  \textbf{Pedestrian} & \textbf{Motorcycle} & \textbf{Bicycle}\\
    \midrule

    \multirow{1}{*}{\cite{zhang2024voxel}}
      & Voxel Mamba & 2024 & NeurIPS & SSMs & Voxel
      & \red{\textbf{73.0}} & \red{\textbf{69.0}} & 86.8 & 57.1 & 68.0 & \red{\textbf{63.2}}  & 89.5 & 74.7 & 50.8  \\[1mm]
      \myrowcolour
      & \multicolumn{1}{l}{\textbf{Innovation:}}
      & \multicolumn{16}{l}{Replaces attention with group‑free state‑space operators on voxels to model long‑range dependencies efficiently.} \\[2mm]

    \multirow{1}{*}{\cite{liu2024lion}}
      & LION\mbox{-}RetNet & 2024 & NeurIPS & LION3D & Point
      & 71.9 & 67.3 & \red{\textbf{89.6}}  & 64.3 & \red{\textbf{78.7}} & 44.6  & \red{\textbf{89.8}} & 73.5 & 56.6  \\[1mm]
      \myrowcolour
      & \multicolumn{1}{l}{\textbf{Innovation:}}
      & \multicolumn{16}{l}{Enhances BEV fusion via LiDAR–image multi-scale interaction with cross-attention and dynamic modality weighting.} \\[2mm]

    \multirow{1}{*}{\cite{liu2024lion}}
      & LION\mbox{-}RWKV & 2024 & NeurIPS & LION3D & Voxel
      & 71.7 & 67.6 & 89.4 & 59.0 & 77.6 & 37.1  & 89.7 & 74.3 & 56.2  \\[1mm]
      \myrowcolour
      & \multicolumn{1}{l}{\textbf{Innovation:}}
      & \multicolumn{16}{l}{Employs the same BEV fusion strategy with an RWKV backbone for recurrent state-based spatio-temporal feature modeling.} \\[2mm]

    \multirow{1}{*}{\cite{liu2024lion}}
      & LION\mbox{-}Mamba & 2024 & NeurIPS & LION3D & Voxel
      & 72.1 & 68.0 & 87.9 & \red{\textbf{64.9}} & 77.6 & 44.4  & 89.6 & \red{\textbf{75.6}} & \red{\textbf{59.4}}  \\[1mm]
      \myrowcolour
      & \multicolumn{1}{l}{\textbf{Innovation:}}
      & \multicolumn{16}{l}{Employs the same BEV fusion strategy with a Mamba backbone for efficient long-range dependency modeling.} \\[2mm]

    \multirow{1}{*}{\cite{mao2024pillarnest}}
      & PillarNeSt\mbox{-}Base & 2024 & T-IV & ConNeXt & Pillar
      & 71.3 & 66.9 & 87.4 & 56.4 & 64.0 & 63.0  & 86.6 & 69.4 & 46.8 \\[1mm]
      \myrowcolour
      & \multicolumn{1}{l}{\textbf{Innovation:}}
      & \multicolumn{16}{l}{Scales pillar-based backbones with large-scale pretraining to enhance spatial feature learning for LiDAR 3D object detection.} \\[2mm]

    \multirow{1}{*}{\cite{mao2024pillarnest}}
      & PillarNeSt\mbox{-}Large & 2024 & T-IV & ConvNeXt & Pillar
      & 71.6 & 66.9 & 87.4 & 56.4 & 64.0 & 63.0  & 86.6 & 69.4 & 51.5  \\[1mm]
      \myrowcolour
      & \multicolumn{1}{l}{\textbf{Innovation:}}
      & \multicolumn{16}{l}{Enlarges model capacity within the same architecture to further enhance feature representation and detection performance.} \\[2mm]

    \multirow{1}{*}{\cite{zhang2024safdnet}}
      & SAFDNet & 2024 & CVPR & Sparse CNN & Voxel
      & 72.3 & 68.3 & 87.3 & 57.3 & 68.0 & 61.7 & 89.0 & 71.1 & 44.8  \\[1mm]
      \myrowcolour
      & \multicolumn{1}{l}{\textbf{Innovation:}}
      & \multicolumn{16}{l}{Introduces adaptive feature diffusion to mitigate center feature missing while preserving efficiency in fully sparse LiDAR detection.} \\[2mm]

    \multirow{1}{*}{\cite{fan2024fsd}}
      & FSDv2 & 2024 & TPAMI & SparseUNet & Voxel
      & 71.7 & 66.5 & 86.1 & 53.0 & 66.5 & 61.1 5 & 87.1 & 71.1 & 51.7  \\[1mm]
      \myrowcolour
      & \multicolumn{1}{l}{\textbf{Innovation:}}
      & \multicolumn{16}{l}{Replaces instance clustering with virtual voxels, simplifying FSD and addressing center‑feature missing in fully sparse detectors.} \\[2mm]

    \multirow{1}{*}{\cite{chen2023voxelnext}}
      & VoxelNeXt & 2023 & CVPR & Sparse CNN & Voxel
      & 70.0 & 64.5 & 84.6 & 53.0 & 64.7 & 55.8  & 85.8 & 73.2 & 45.7 \\[1mm]
      \myrowcolour
      & \multicolumn{1}{l}{\textbf{Innovation:}}
      & \multicolumn{16}{l}{Eliminates dense heads by directly predicting objects from sparse voxel features, enabling efficient fully 3D detection and tracking.} \\[2mm]

    \multirow{1}{*}{\cite{lu2023link}}
      & LinK & 2023 & CVPR & LinK-based & Voxel
      & 71.0 & 66.3 & 86.7 & 55.7 & 65.7 & 62.1  & 85.5 & 73.5 & 47.5  \\[1mm]
      \myrowcolour
      & \multicolumn{1}{l}{\textbf{Innovation:}}
      & \multicolumn{16}{l}{Introduces a linear kernel that assigns weights only to non‑empty voxels, improving efficiency in sparse 3D CNNs.} \\[2mm]

    \multirow{1}{*}{\cite{zhang2023hednet}}
      & HEDNet & 2023 & NeurIPS & Sparse CNN  & Point-Voxel
      & 72.0 & 67.7 & 86.5 & 57.9 & 70.4 & 61.2  & 87.9 & 70.4 & 46.9 \\[1mm]
      \myrowcolour
      & \multicolumn{1}{l}{\textbf{Innovation:}}
      & \multicolumn{16}{l}{Builds hierarchical encoder–decoder with semantic‑ and depth‑guided blocks to exploit multi‑scale 3D context efficiently.} \\[2mm]

    \multirow{1}{*}{\cite{chen2023largekernel3d}}
      & LargeKernel3D & 2023 & CVPR & 3D CNN & Voxel
      & 70.9 & 65.4 & 85.5 & 53.8 & 64.4 & 59.5  & 85.9 & 72.7 & 46.8  \\[1mm]
      \myrowcolour
      & \multicolumn{1}{l}{\textbf{Innovation:}}
      & \multicolumn{16}{l}{Uses spatial‑wise partition convolution and position embedding to realize efficient, very large 3D kernels in sparse CNNs.} \\[2mm]

    \multirow{1}{*}{\cite{wang2023dsvt}}
      & DSVT & 2023 & CVPR & BEV & Voxel
      & 72.7 & 68.0 & 86.8 & 54.8 & 67.4 & 63.1  & 88.0 & 73.0 & 47.2  \\[1mm]
      \myrowcolour
      & \multicolumn{1}{l}{\textbf{Innovation:}}
      & \multicolumn{16}{l}{Efficient transformer backbone using dynamic sparse attention and rotated set partitioning without custom CUDA operations.} \\[2mm]

    \multirow{1}{*}{\cite{zhang2023fully}}
      & FSTR & 2023 & TGRS & VoxelNeXt & Point
      & 71.5 & 67.2 & 86.5 & 54.1 & 66.4 & 58.4  & 88.6 & 73.7 & 48.1 \\[1mm]
      \myrowcolour
      & \multicolumn{1}{l}{\textbf{Innovation:}}
      & \multicolumn{16}{l}{Introduces a fully sparse transformer with dynamic queries and Gaussian denoising for efficient long-range LiDAR 3D detection.} \\[2mm]

    \multirow{1}{*}{\cite{wang2023uni3detr}}
      & Uni3DETR & 2023 & NeurIPS & Sparse CNN & Point-Voxel
      & 68.5 & 57.3 & 86.1 & 57.8 & 63.5 & 38.2  & 88.7 & 74.6 & 42.2 \\[1mm]
      \myrowcolour
      & \multicolumn{1}{l}{\textbf{Innovation:}}
      & \multicolumn{16}{l}{A unified DETR‑style framework using point–voxel interaction to handle indoor and outdoor scenes consistently.} \\[2mm]

    \multirow{1}{*}{\cite{tian2022fully}}
      & FCOS\mbox{-}LiDAR & 2022 & NeurIPS & LiDAR-Net & Range
      & 65.7 & 60.2 & 82.2 & 47.7 & 52.9 & 48.8  & 84.5 & 68.0 & 39.0  \\[1mm]
      \myrowcolour
      & \multicolumn{1}{l}{\textbf{Innovation:}}
      & \multicolumn{16}{l}{Standard 2D convolutions and multi‑frame MRV projection.} \\[2mm]

    \multirow{1}{*}{\cite{hu2022afdetv2}}
      & AFDetV2 & 2022 & AAAI & RPN & Voxel
      & 68.5 & 62.2 & 86.3 & 52.6 & 62.8 & 59.9  & 85.4 & 68.3 & 34.8  \\[1mm]
      \myrowcolour
      & \multicolumn{1}{l}{\textbf{Innovation:}}
      & \multicolumn{16}{l}{Single-stage anchor-free LiDAR detector with SC-Conv backbone, IoU-aware rescoring, and keypoint auxiliary supervision.} \\[2mm]

    \multirow{1}{*}{\cite{li2022unifying}}
      & UVTR\mbox{-}L & 2022 & NeurIPS & ResNet-101 & Voxel
      & 69.7 & 63.9 & 86.3 & 52.2 & 62.8 & 59.7  & 84.5 & 68.8 & 41.1 \\[1mm]
      \myrowcolour
      & \multicolumn{1}{l}{\textbf{Innovation:}}
      & \multicolumn{16}{l}{Cross‑modality interaction and a transformer decoder.} \\[2mm]


    \multirow{1}{*}{\cite{chen2022focal}}
      & Focals Conv & 2022 & CVPR & Sparse CNN & Voxel
      & 70.0 & 63.8 & 86.7 & 56.3 & 67.7 & 59.5  & 87.5 & 64.5 & 36.3 \\[1mm]
      \myrowcolour
      & \multicolumn{1}{l}{\textbf{Innovation:}}
      & \multicolumn{16}{l}{Makes sparsity learnable via position‑wise importance, improving sparse CNNs for LiDAR‑only and multi‑modal detection.} \\[2mm]

    \multirow{1}{*}{\cite{shi2022pillarnet}}
      & PillarNet & 2022 & ECCV & VGG-18/34 & Pillar
      & 71.4 & 66.4 & 86.1 & 53.1 & 65.3 & 63.1  & 87.3 & 70.1 & 42.3  \\[1mm]
      \myrowcolour
      & \multicolumn{1}{l}{\textbf{Innovation:}}
      & \multicolumn{16}{l}{Introduces a powerful sparse 2D CNN encoder and optimized neck for efficient pillar-based 3D detection.} \\[2mm]

    \multirow{1}{*}{\cite{bai2022transfusion}}
      & TransFusion & 2022 & CVPR & 3D CNN & Point
      & 70.2 & 65.5 & 86.2 & 56.7 & 66.3 & 58.8  & 86.1 & 68.3 & 44.2  \\[1mm]
      \myrowcolour
      & \multicolumn{1}{l}{\textbf{Innovation:}}
      & \multicolumn{16}{l}{Uses a transformer decoder to fuse LiDAR BEV features.} \\[2mm]

    \multirow{1}{*}{\cite{fazlali2022versatile}}
      & VMVF & 2022 & CVPR & 3D CNN & Range
      & 66.3 & 58.3 & 84.4 & 51.5 & 58.3 & 53.9  & 85.3 & 63.0 & 32.5  \\[1mm]
      \myrowcolour
      & \multicolumn{1}{l}{\textbf{Innovation:}}
      & \multicolumn{16}{l}{Enhances BEV-based 3D detection by fusing RV panoptic segmentation features with semantic and instance guidance.} \\[2mm]

    \multirow{1}{*}{\cite{yin2021center}}
      & CenterPoint & 2021 & CVPR & VoxelNet & Point
      & 65.5 & 58.0 & 84.6 & 51.0 & 60.2 & 53.2  & 83.4 & 53.7 & 28.7 \\[1mm]
      \myrowcolour
      & \multicolumn{1}{l}{\textbf{Innovation:}}
      & \multicolumn{16}{l}{Reformulates 3D detection and tracking as center keypoint estimation, enabling efficient two-stage refinement.} \\[2mm]

    \multirow{1}{*}{\cite{pan20213d}}
      & Pointformer & 2021 & CVPR & Pointformer & Point
      & 63.8 & 53.8 & 82.3 & 45.1 & 48.3 & 45.4  & 85.1 & 55.7 & 25.8  \\[1mm]
      \myrowcolour
      & \multicolumn{1}{l}{\textbf{Innovation:}}
      & \multicolumn{16}{l}{Pure Transformer backbone capturing local and global context for point cloud detection.} \\[2mm]

    \multirow{1}{*}{\cite{chen2020every}}
      & CVCNET & 2020 & NeurIPS & 3D CNN & Voxel
      & 66.6 & 58.2 & 82.6 & 49.5 & 59.4 & 51.1  & 83.0 & 61.8 & 38.8 \\[1mm]
      \myrowcolour
      & \multicolumn{1}{l}{\textbf{Innovation:}}
      & \multicolumn{16}{l}{Introduces hybrid cylindrical–spherical voxelization and cross‑view transformers to enforce BEV/RV consistency.} \\[2mm]

    \multirow{1}{*}{\cite{yang20203dssd}}
      & 3DSSD & 2020 & CVPR & SA & Point
      & 56.4 & 42.6 & 81.2 & 47.2 & 61.4 & 30.5  & 70.2 & 36.0 & 8.6  \\[1mm]
      \myrowcolour
      & \multicolumn{1}{l}{\textbf{Innovation:}}
      & \multicolumn{16}{l}{Removes FP layers and refinement stage via fusion sampling, candidate generation, and anchor-free regression.} \\[2mm]

    \multirow{1}{*}{\cite{yin2020lidar}}
      & 3DVID & 2020 & CVPR & PMPNet & Pillar
      & 53.1 & 45.4 & 79.7 & 33.6 & 47.1 & 43.1  & 76.5 & 40.7 & 19.9  \\[1mm]
      \myrowcolour
      & \multicolumn{1}{l}{\textbf{Innovation:}}
      & \multicolumn{16}{l}{Introduces PMPNet and AST-GRU for spatial/temporal coherence in online 3D video detection.} \\[2mm]


    \multirow{1}{*}{\cite{zhu2019class}}
      & CBGS & 2019 & CVPR & Sparse CNN & Voxel
      & 63.3 & 52.8 & 81.1 & 48.5 & 54.9 & 42.9  & 80.1 & 51.5 & 22.3  \\[1mm]
      \myrowcolour
      & \multicolumn{1}{l}{\textbf{Innovation:}}
      & \multicolumn{16}{l}{Class‑balanced sampling/augmentation and a balanced grouping head for severe class imbalance.} \\[2mm]

    \multirow{1}{*}{\cite{lang2019pointpillars}}
      & PointPillars & 2019 & CVPR & 2D CNN & Pillar
      & 45.3 & 30.5 & 68.4 & 23.0 & 28.2 & 23.4  & 59.7 & 27.4 & 1.1  \\[1mm]
      \myrowcolour
      & \multicolumn{1}{l}{\textbf{Innovation:}}
      & \multicolumn{16}{l}{Pillar-based PointNet encoder converting point clouds to dense pseudo-images for 2D CNN detection.} \\

    \bottomrule
  \end{tabular}}
\end{center}
\end{table*}

Table \ref{table: 3D-Nuscenes} presents a comparative analysis of 3D object detection methods using the LiDAR-based NuScenes dataset. From Table \ref{table: 3D-Nuscenes}, we provide a detailed discussion of three selected models: PointPillars \cite{lang2019pointpillars}, recognized as one of the pioneer model in this field; Voxel Mamba \cite{zhang2024voxel}, which achieved the highest results in terms of NDS (NuScenes Detection Score), mAP, and trailer detection on the NuScenes dataset; and LION \cite{liu2024lion}, obtained the highest results for car, truck, bus, pedestrian, and cyclist detection on NuScenes dataset. PointPillars \cite{lang2019pointpillars} adopts a pillar-based representation to enable a purely 2D convolutional pipeline for real-time 3D object detection. The approach first discretizes the point cloud into vertical columns (pillars) with infinite extent along the z-axis, then uses a simplified PointNet to extract point-wise features within each pillar. These are aggregated and scattered into a dense pseudo-image in the bird’s-eye-view plane. A 2D CNN backbone, composed of a top-down path and a multi-scale upsampling path, processes this pseudo-image to extract high-level spatial features. Finally, an SSD-style detection head performs classification and oriented 3D bounding box regression directly in BEV space, allowing end-to-end training without any 3D convolutions.

Voxel Mamba \cite{zhang2024voxel} introduces a group-free state-space backbone for 3D object detection, exploiting the linear-time sequence modeling of State Space Models (SSMs) to bypass the computational bottlenecks of Transformer-based voxel encoders. Instead of partitioning the voxel grid into smaller local groups, it serializes all non-empty voxels into a single sequence using a Hilbert Input Layer, which maps 3D voxel coordinates into a 1D order while preserving spatial locality. The backbone employs a Dual-scale SSM Block (DSB), wherein a forward branch processes the high-resolution sequence. In contrast, a backward branch operates on a down-sampled BEV representation, extending the receptive field without quadratic complexity. An Implicit Window Partition (IWP) injects relative position embeddings for intra- and inter-window offsets, enabling the model to recover local context cues without explicit grouping. LION \cite{liu2024lion} replaces transformer-based voxel backbones with linear recurrent neural operators to model long-range dependencies across large voxel regions while avoiding quadratic complexity. Point clouds are voxelized and organized into fixed-size 3D windows, within which LION Blocks process voxel features hierarchically. Each block contains a pair of linear group RNN operators, applied along orthogonal spatial axes, to propagate information efficiently across extended spatial extents. Before this recurrent mixing, a 3D spatial descriptor based on submanifold convolutions enriches local geometric context to offset the positional information loss during voxel flattening. Within its hierarchy, the model reduces voxel resolution at selected stages, representing features at multiple scale, and giving them higher-level layers access.

Table \ref{table:3D-VOD} presents a comparative analysis of 3D object detection methods using LiDAR-based data from the VOD dataset. From Table \ref{table:3D-VOD}, we provide a detailed discussion of three selected models: ELMAR \cite{peng2025elmar} and MoRAL \cite{peng2025moral}, which achieved the highest results in terms of $AP\%$ for entire area and driving corridor on VOD dataset; and L4DAR \cite{huang2025l4dr}, recognized as one of the notable model in the context of 3D object detection.

\begin{table*}[htbp]
\centering
\caption{Comparative analysis of 3D object detection methods using the LiDAR-based VOD dataset.}
\vspace{-0.4cm}
\label{table:3D-VOD}
\begin{center}
\resizebox{\textwidth}{!}{%
  \setlength{\tabcolsep}{4pt}%
  \begin{tabular}{llllll cccc cccc}
\toprule
\multicolumn{6}{c}{} 
  & \multicolumn{4}{c}{\textbf{AP in the Entire Annotated Area (\%)}} 
  & \multicolumn{4}{c}{\textbf{AP in the Driving Corridor (\%)}} \\[1mm]
\cmidrule(l){7-10} \cmidrule(l){11-14}
\textbf{Ref} & \textbf{Method} & \textbf{Year} & \textbf{Venue} & \textbf{Backbone} & \textbf{Modality}
  & \textbf{Car} & \textbf{Pedestrian} & \textbf{Cyclist} & \textbf{mAP$_{\textbf{3D}}$}
  & \textbf{Car} & \textbf{Pedestrian} & \textbf{Cyclist} & \textbf{mAP$_{\textbf{3D}}$} \\
\midrule

    \multirow{1}{*}{\cite{peng2025elmar}}
      & ELMAR & 2025 & ArXiv & DS+SA & Point
      & \red{\textbf{76.41}} & 69.34 & 78.91 & \red{\textbf{74.89}}
      & 91.74 & \red{\textbf{81.35}}& 93.01 & \red{\textbf{88.70}} \\[1mm]
      \myrowcolour
      & \multicolumn{1}{l}{\textbf{Innovation:}} &
      \multicolumn{12}{l}{Encodes radar-derived motion and aligns LiDAR–radar predictions via uncertainty modeling for cross-modal misalignment.} \\[2mm]

    \multirow{1}{*}{\cite{peng2025moral}}
      & MoRAL & 2025 & ArXiv & Sparse CNN & Pillar
      & 71.23 & \red{\textbf{69.67}} & 79.01 & 73.30
      & 90.91 & 78.90 & \red{\textbf{96.25}} & 88.68 \\[1mm]
      \myrowcolour
      & \multicolumn{1}{l}{\textbf{Innovation:}} &
      \multicolumn{12}{l}{Motion-compensated multi-frame LiDAR-radar enhancement via motion-aware gating to mitigate misalignment artifacts.} \\[2mm]

    \multirow{1}{*}{\cite{peng2025mutualforce}}
      & MutualForce & 2025 & ICASSP & CNN+RPN & Pillar
      & 71.67 & 66.26 & 77.35 & 71.76
      & \red{\textbf{92.31}} & 76.79 & 89.97 & 86.36 \\[1mm]
      \myrowcolour
      & \multicolumn{1}{l}{\textbf{Innovation:}} &
      \multicolumn{12}{l}{Bidirectional radar–LiDAR feature guidance with velocity cues and LiDAR-radar enhancement for improved fusion detection.} \\[2mm]

    \multirow{1}{*}{\cite{ding2025det}}
      & AS\mbox{-}Det & 2025 & AAAI & AS & Point
      & 42.46 & 49.02 & 78.03 & 56.50
      & 77.45 & 59.41 & 96.12 & 77.73 \\[1mm]
      \myrowcolour
      & \multicolumn{1}{l}{\textbf{Innovation:}} &
      \multicolumn{12}{l}{Introduces active sampling and multi-scale center feature aggregation for adaptable single-stage 3D detection.} \\[2mm]

    \multirow{1}{*}{\cite{huang2025l4dr}}
      & L4DR & 2024 & AAAI & PointNet++ & Pillar
      & 69.10 & 66.20 & \red{\textbf{82.80}} & 72.70
      & 90.80 & 76.10 & 95.50 & 87.47 \\[1mm]
      \myrowcolour
      & \multicolumn{1}{l}{\textbf{Innovation:}} &
      \multicolumn{12}{l}{Two-stage LiDAR–radar fusion with denoising, pillar-based feature sharing, and multi-scale fusion for weather robustness.} \\[2mm]

    \multirow{1}{*}{\cite{deng2024robust}}
      & CM\mbox{-}FA & 2024 & ICRA & SA & Point
      & 71.39 & 68.54 & 76.60 & 72.18
      & 90.91 & 80.78 & 87.80 & 86.50 \\[1mm]
      \myrowcolour
      & \multicolumn{1}{l}{\textbf{Innovation:}} &
      \multicolumn{12}{l}{Cross-modal hallucination with spatial/feature alignment enabling modality-agnostic training and single-sensor inference.} \\[2mm]

    \multirow{1}{*}{\cite{xu2024rlnet}}
      & RLNet & 2024 & ECCV & 2D CNN & Voxel
      & 70.88 & 69.43 & 78.12 & 72.81
      & 90.82 & 78.71 & 91.67 & 87.07 \\[1mm]
      \myrowcolour
      & \multicolumn{1}{l}{\textbf{Innovation:}} &
      \multicolumn{12}{l}{LiDAR–radar fusion with adaptive weighting, Doppler speed compensation, and modality dropout for robust detection.} \\[2mm]


    \multirow{1}{*}{\cite{wang2022interfusion}}
      & InterFusion & 2022 & IROS & CNN+RPN & Pillar
      & 67.50 & 63.21 & 78.79 & 69.83
      & 88.11 & 74.80 & 87.50 & 83.47 \\[1mm]
      \myrowcolour
      & \multicolumn{1}{l}{\textbf{Innovation:}} &
      \multicolumn{12}{l}{Self-attention interaction aligning LiDAR/radar pillar features with radar noise reduction via pitch-angle correction.} \\[2mm]

    \multirow{1}{*}{\cite{yin2021center}}
      & CenterPoint & 2021 & CVPR & VoxelNet & Point
      & 32.74 & 38.00 & 65.42 & 45.42
      & 62.01 & 48.18 & 84.98 & 65.66 \\[1mm]
      \myrowcolour
      & \multicolumn{1}{l}{\textbf{Innovation:}} &
      \multicolumn{12}{l}{Reformulates 3D detection as center keypoint estimation, enabling anchor-free predictions with two-stage refinement.} \\[2mm]

    \multirow{1}{*}{\cite{lang2019pointpillars}}
      & PointPillars & 2019 & CVPR & 2D\mbox{-}CNN & Pillar
      & 65.55 & 55.71 & 72.96 & 64.74
      & 81.10 & 67.92 & 88.96 & 79.33 \\[1mm]
      \myrowcolour
      & \multicolumn{1}{l}{\textbf{Innovation:}} &
      \multicolumn{12}{l}{Pillar-based PointNet encoder converting point clouds to dense pseudo-images for efficient end-to-end 3D detection.} \\

    \bottomrule
  \end{tabular}}
\end{center}
\end{table*}

\vspace{-0.3cm}
The ELMAR algorithm\cite{peng2025elmar} is designed for single-frame operation, pairing LiDAR with 4D radar without any temporal accumulation. Its radar branch incorporates a Dynamic Motion-Aware Encoding module. It learns to classify entire objects as moving or static via a dedicated motion-aware loss, avoiding the unreliability of fixed velocity cutoffs. This produces motion descriptors that are stable enough to be used directly in detection. After each branch produces its initial set of bounding boxes, the Cross-Modal Uncertainty Alignment module attempts to match boxes between modalities, using geometric differences between matched pairs as a fine-grained measure of uncertainty. Considerable disagreements often cause predictions to automatically downweigh during training, directing the model towards cases where both sensors agree. This means the detector only captures the advantages of the radar's motion sensitivity but also uses the radar's consistency to overcome the LiDAR's weaknesses in challenging conditions.

The MoRAL framework \cite{peng2025moral} was developed with the goal of tackling the distortions that often result from combining multiple 4D radar frames, especially the misalignment of fast objects. Conventionally, points from earlier frames can stretch along the direction of motion, which creates elongated "tails" that distort shape and affect detection efficiency. MoRAL avoids this by embedding a Motion-aware Radar Encoder (MRE) at the front of the radar branch. This module uses a segmentation-based approach to distinguish moving points from static ones, then shifts only the moving points into alignment with a chosen reference frame using their absolute radial velocity and capture time. While the radar branch undergoes this correction, the LiDAR input is processed independently through a sparse encoder. The two streams meet in the Motion Attention Gated Fusion (MAGF) module, which injects motion cues into the LiDAR feature space and applies a gating mask. That mask strengthens the signal from dynamic foreground regions while naturally dampening static background responses before the network predicts final bounding boxes. Meanwhile, L4DR pipeline \cite{huang2025l4dr} addresses a different challenge: the steep quality gap between LiDAR and radar data, and the fact that each sensor degrades differently in poor weather. Its processing begins with Foreground-Aware Denoising, which uses learned segmentation to strip radar scans of spurious returns before any fusion occurs. Cleaned radar data is fused early with raw LiDAR through the Multi-Modal Encoder. In this step, points are grouped into shared pillars, and each pillar allows a two-way exchange of information—radar's velocity and RCS values inform the LiDAR points in that cell. Meanwhile, LiDAR contributes its high-precision geometry and reflectivity back to radar. The fused data then passes into the Inter-modal and Intra-modal backbone, which maintains three streams: one for each modality and one combined. To prevent all three from carrying redundant patterns, a Multi-Scale Gated Fusion block filters each intra-modal stream based on cues from the inter-modal branch, enabling the network to emphasize whichever sensor is more trustworthy at that moment.

\subsection{2D–3D Fusion Strategies}
\label{sec: 2D-3D}
2D–3D fusion methods combine complementary information from camera images and LiDAR point clouds to improve detection accuracy and robustness in autonomous driving scenarios. Camera data provides rich texture and color cues, which are valuable for recognizing objects’ appearance and fine details. Meanwhile, LiDAR delivers precise geometric structure and depth information, enabling accurate distance estimation and spatial localization. By integrating these two modalities, fusion strategies can overcome the limitations of single-sensor approaches, such as poor depth perception in monocular vision or reduced semantic detail in LiDAR-only systems. These methods vary in how and when they combine features, with common approaches including early fusion at the raw data level, mid-level fusion of intermediate features, and late fusion at the decision stage. The choice of fusion scheme significantly impacts performance, balancing computational cost, information richness, and real-time feasibility. Recent advances in deep learning-based fusion architectures show that well-designed pipelines outperform single-modality methods, particularly under occlusion, low lighting, and adverse weather.

\begin{figure}[H]
    \centering
    \centerline{\includegraphics[width=0.9\textwidth]{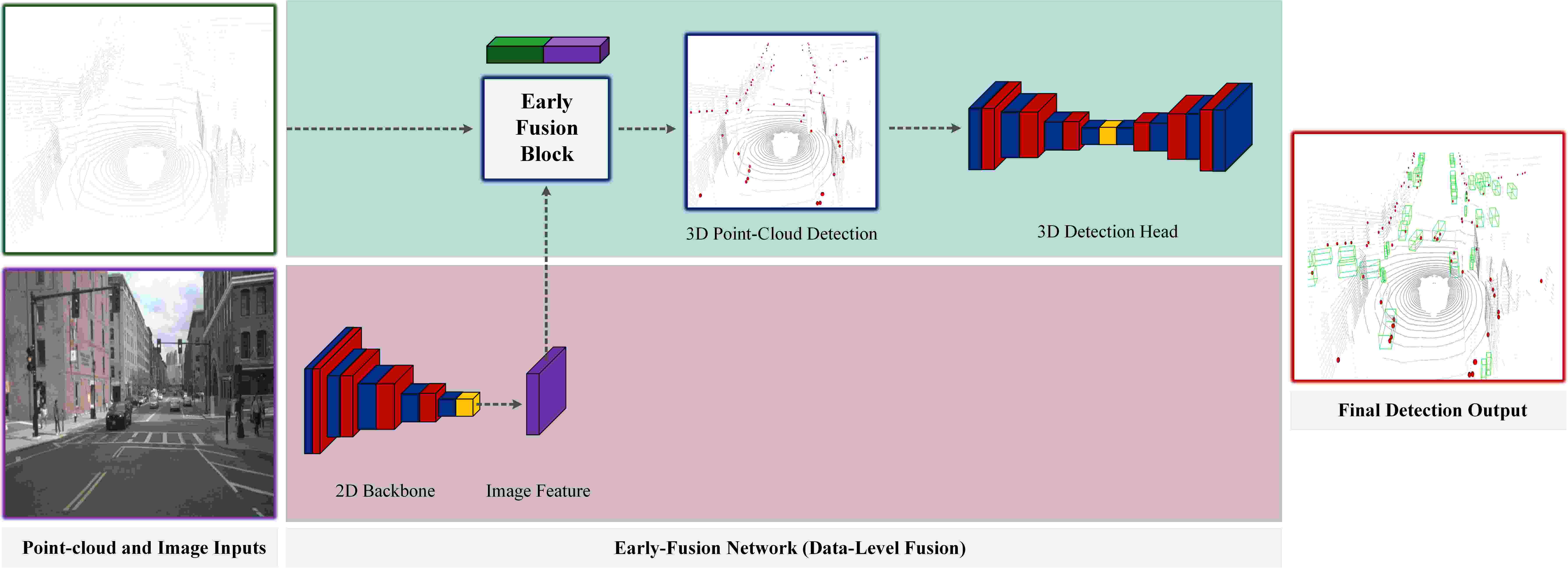}}
    \caption{Overall framework of Early-Fusion 3D object detection in the context of autonomous driving systems.}
    \label{figure: early-fusion}
\end{figure}

Figure \ref{figure: early-fusion} illustrates the general architecture of an early-fusion network, where multi-sensor data (LiDAR point clouds and RGB images) are combined at the raw data stage. This approach, also called data-level fusion, aligns sensor outputs into a shared representation space, by projecting LiDAR points onto the image plane or generating dense depth maps from point clouds. By fusing early information, this method preserves the full richness of the input data. It enables the network to learn low-level correlations between geometric structure and visual appearance. However, early fusion has some limitations, including high computational cost due to large input dimensionality, sensitivity to spatial and temporal calibration errors, and reduced robustness when one modality is degraded or missing. Despite these challenges, it is widely adopted in applications such as dense 3D object detection, semantic segmentation, and bird’s-eye view mapping.

\begin{figure}[H]
    \centering
    \centerline{\includegraphics[width=0.89\textwidth]{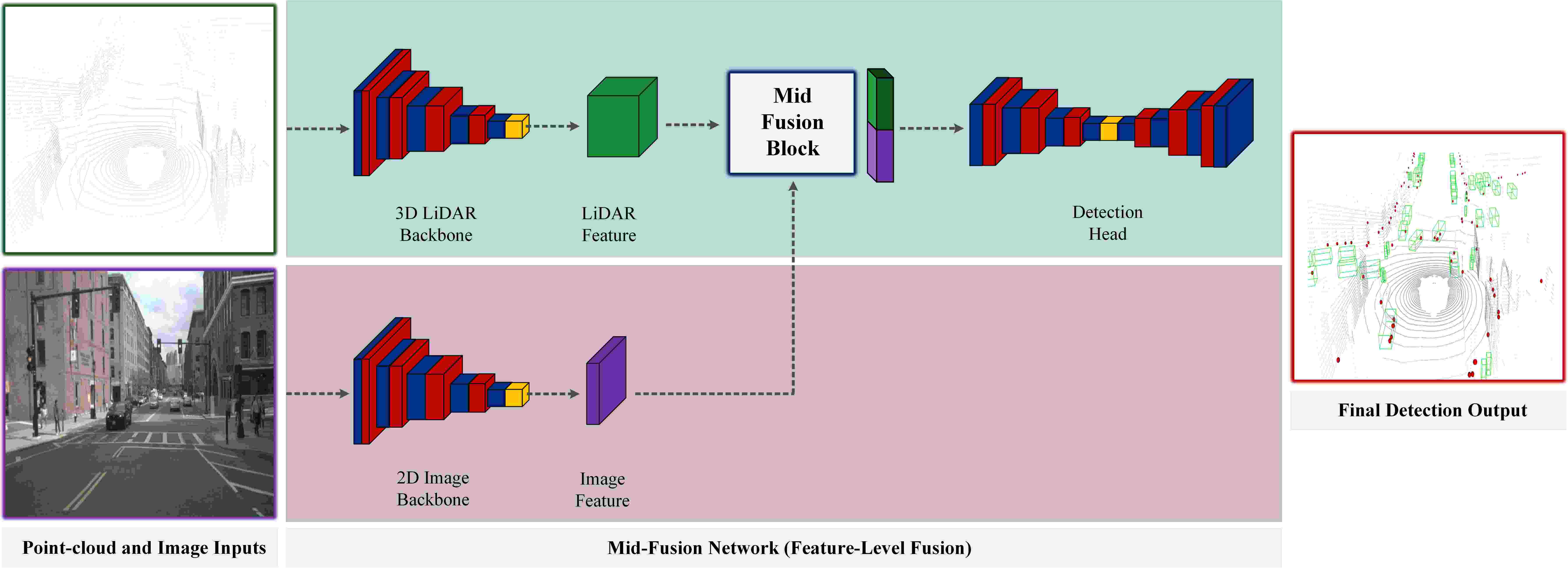}}
    \caption{Overall framework of Mid-Fusion 3D object detection in the context of autonomous driving systems.}
    \label{figure: mid-fusion}
\end{figure}

\vspace{-0.4cm}
Figure \ref{figure: mid-fusion} represents the general architecture of a mid-fusion network, where LiDAR point clouds and RGB images are first processed independently through modality-specific backbones to extract intermediate feature representations before being combined. This approach, also known as feature-level fusion, merges modality-specific features, often through concatenation, element-wise operations, or attention-based mechanisms. Mid-fusion allows each backbone to specialize in processing its own modality, capturing modality-specific characteristics such as the geometric precision of LiDAR and the rich semantic content of images. Compared to early fusion, mid-fusion is less sensitive to precise sensor calibration since the fusion occurs at a higher abstraction level, and it can be more computationally efficient by reducing the data dimensionality prior to fusion. However, it may lose some fine-grained cross-modal correlations in raw data and requires careful design to ensure feature compatibility across modalities. This strategy is widely used in AVs applications, such as 3D object detection, lane and road boundary detection, and BEV mapping.

\begin{figure}[H]
    \centering
    \centerline{\includegraphics[width=0.89\textwidth]{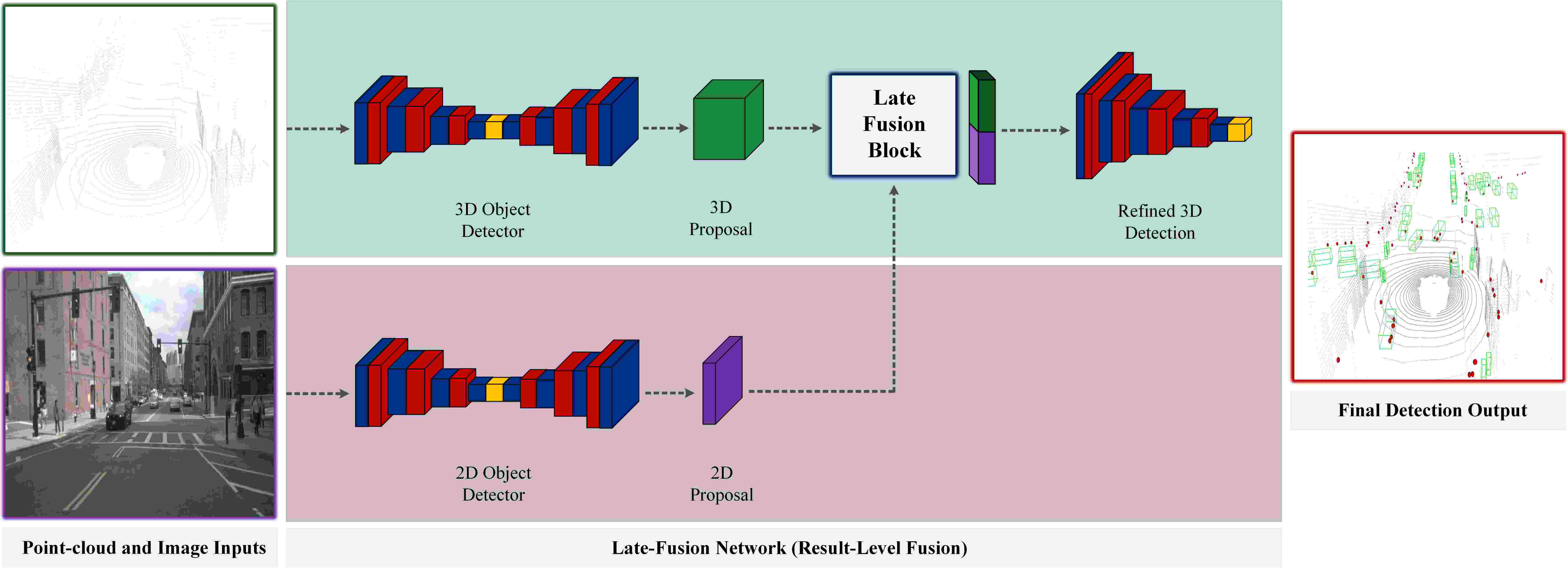}}
    \caption{Overall framework of Late-Fusion 3D object detection in the context of autonomous driving systems.}
    \label{figure: late-fusion}
\end{figure}

\vspace{-0.3cm}
Figure \ref{figure: late-fusion} shows the general architecture of a late-fusion network, where LiDAR point clouds and RGB images are processed entirely separately through independent detection pipelines. The resulting outputs, such as bounding boxes, classification scores, or segmentation maps, are combined at the decision stage. This approach, also referred to as result-level fusion, integrates high-level predictions rather than raw data or intermediate features, typically using strategies such as weighted averaging, confidence-based selection, non-maximum suppression, across modalities, or rule-based decision logic. Late-fusion offers several advantages, including high modularity, robustness to partial sensor failure, and computational efficiency. However, late-fusion cannot leverage low- or mid-level cross-modal correlations, potentially limiting performance gains compared to early or mid-fusion approaches, and it relies heavily on the accuracy of individual modality-specific detectors. This strategy is often applied in safety-critical AV systems, where redundancy and fault tolerance are essential, such as in collision avoidance, multi-sensor tracking, and decision-making modules.

\vspace{-0.3cm}
In the following, we present all existing 2D–3D fusion algorithms for object detection in autonomous vehicles from 2017 to 2025. To ensure a consistent and fair comparison, we organize the results into two separate tables, as the performance of these methods is evaluated on different datasets. Table \ref{table: fusion-kitti} provides a comparative analysis of fusion-based object detection methods using paired image–LiDAR data from the KITTI dataset, while Table \ref{table: fusion-nuscense} summarizes those evaluated on the NuScenes dataset. It should be mentioned that the values highlighted in red represent the best performance achieved for each evaluation metric across all listed methods.

\begin{table}[H]
\centering
\caption{Comparative analysis of 2D–3D fusion detection methods on the paired image–LiDAR KITTI dataset.}
\vspace{-0.3cm}
\label{table: fusion-kitti}
\begin{center}
\resizebox{\textwidth}{!}{%
  \setlength{\tabcolsep}{4pt}%
  \begin{tabular}{llll ll ccc ccc ccc ccc c}
    \toprule
    \multicolumn{4}{c}{}
      & \multicolumn{2}{c}{\textbf{Backbone}} 
      & \multicolumn{3}{c}{$\textbf{AP}_{\textbf{BEV}}$ \hspace{1mm}\textbf{Car}} 
      & \multicolumn{3}{c}{$\textbf{AP}_{\textbf{3D}}$\hspace{1mm}\textbf{Car}}
      & \multicolumn{3}{c}{$\textbf{AP}_{\textbf{3D}}$\hspace{1mm}\textbf{Pedestrian}} 
      & \multicolumn{3}{c}{$\textbf{AP}_{\textbf{3D}}$\hspace{1mm}\textbf{Cyclist}}
 \\[1mm]
    \cmidrule(l){5-6}\cmidrule(l){7-9}\cmidrule(l){10-12}\cmidrule(l){13-15}\cmidrule(l){16-18}
    \textbf{Ref} & \textbf{Method} & \textbf{Year} & \textbf{Venue} \hspace{2mm} & \textbf{Image} & \textbf{LiDAR}
      & \textbf{Easy} & \textbf{Moderate} & \textbf{Hard}
      & \textbf{Easy} & \textbf{Moderate} & \textbf{Hard}
      & \textbf{Easy} & \textbf{Moderate} & \textbf{Hard}
      & \textbf{Easy} & \textbf{Moderate} & \textbf{Hard}
      & \textbf{Runtime} \\
    \midrule

\multirow{1}{*}{\cite{yu2025vikienet}} 
  & ViKIENet & 2025 & CVPR & 2D CNN & 3D CNN
  & \red{\textbf{96.31}} & \red{\textbf{93.73}} & \red{\textbf{91.64}}
  & 91.79 & 84.96 & 80.20
  & $-$ & $-$ & $-$
  & $-$ & $-$ & $-$
  & 0.04 Sec \\
\myrowcolour
      & \multicolumn{1}{l}{\textbf{Innovation:}} 
  & \multicolumn{17}{l}{Introduces VKI-based LiDAR–camera fusion using SKIS, VIFF, and VIRA to enhance efficiency and detection accuracy.} \\[2mm]

\multirow{1}{*}{\cite{yu2025vikienet}} 
  & ViKIENet-R & 2025 & CVPR & 2D CNN & 3D CNN
  & 95.45 & 93.58 & 91.33
  & 91.20 & 86.04 & 81.18
  & $-$ & $-$ & $-$
  & $-$ & $-$ & $-$
  & 0.06 Sec \\
\myrowcolour
& \multicolumn{1}{l}{\textbf{Innovation:}} 
  & \multicolumn{17}{l}{Adds VIFF-R which includes rotationally equivariant features} \\[2mm]

\multirow{1}{*}{\cite{mushtaq2025cls}} 
  & CLS-3D & 2025 & IEEE & DeepLabv3+ & 3D CNN
  & 94.08 & 89.10 & 85.92
  & 89.52 & 81.23 & 76.32
  & 53.18 & 44.79 & 42.08
  & 78.43 & 64.17 & 57.24
  & $-$ \\
\myrowcolour
& \multicolumn{1}{l}{\textbf{Innovation:}} 
  & \multicolumn{17}{l}{Fuses LiDAR and image features via point-wise semantic augmentation with slot reweighting transformer for improved 3D detection.} \\[2mm]

\multirow{1}{*}{\cite{shen2025point}} 
  & Point-Level & 2025 & MDPI & CNN+ECA & CNN+ECA
  & $-$ & $-$ & $-$
  & 88.34 & 78.08 & 75.38
  & 59.67 & 54.67 & 50.03
  & 82.49 & 61.43 & 57.38
  & $-$ \\
\myrowcolour
& \multicolumn{1}{l}{\textbf{Innovation:}} 
  & \multicolumn{17}{l}{ECA-enhanced PointPillars for LiDAR–camera fusion improving small-object detection and orientation.} \\[2mm]

\multirow{1}{*}{\cite{li2025tinyfusiondet}} 
  & TinyFusionDet & 2025 & TCSVT & TBN & TBN
  & 92.32 & 84.77 & 79.74
  & 89.29 & 79.40 & 74.99
  & 56.71 & 51.66 & 47.81
  & 86.50 & 66.43 & 62.58
  & $-$ \\
\myrowcolour
& \multicolumn{1}{l}{\textbf{Innovation:}} 
  & \multicolumn{17}{l}{HSPS, CMHA, and CMFI for hardware-efficient FPGA LiDAR–camera 3D detection.} \\[2mm]

\multirow{1}{*}{\cite{uzair2024channel}} 
  & CSMNET & 2024 & TGRS & DeepLabv3+ & C-FPS
  & 90.16 & 85.18 & 80.51
  & 82.05 & 72.64 & 67.10
  & 48.95 & 42.15 & 39.44
  & 82.86 & 67.86 & 60.90
  & $-$ \\
\myrowcolour
& \multicolumn{1}{l}{\textbf{Innovation:}} 
  & \multicolumn{17}{l}{Class-guided sampling with PAT-based fusion to enhance small/distant object detection.} \\[2mm]

\multirow{1}{*}{\cite{zhang2023unleash}} 
  & UPIDet & 2023 & NeurIPS & ResNet18 & RPN
  & $-$ &  $-$ & $-$
  & 89.13 & 82.97 & 80.05
  & 55.59 & 48.77 & 46.12
  & 86.74 & 74.32 & 67.45
  & 0.11 Sec \\
\myrowcolour
& \multicolumn{1}{l}{\textbf{Innovation:}} 
  & \multicolumn{17}{l}{Transfers image-branch supervision to point backbones, guided by auxiliary 2D tasks and DOF analysis.} \\[2mm]

\multirow{1}{*}{\cite{li2023logonet}} 
  & LoGoNet & 2023 & CVPR & FPN & Voxel-RCNN
  & $-$ & $-$ & $-$
  & 91.80 & 85.06 & 80.74
  & 70.20 & 63.72 & 59.46
  & 84.47 & 71.70 & 64.67
  & 0.1 Sec \\
\myrowcolour
& \multicolumn{1}{l}{\textbf{Innovation:}} 
  & \multicolumn{17}{l}{Local-to-global fusion (GoF/LoF/FDA) for stronger multi-modal feature representation.} \\[2mm]

\multirow{1}{*}{\cite{wu2023virtual}} 
  & VirConv-L & 2023 & CVPR & VirConvNet & VirConvNet
  & 95.53 & 91.95 & 87.07
  & 91.41 & 85.05 & 80.22
  & $-$ & $-$ & $-$
  & $-$ & $-$ & $-$
  & $-$ \\
\myrowcolour
& \multicolumn{1}{l}{\textbf{Innovation:}} 
  & \multicolumn{17}{l}{ StVD and NRConv reduce virtual point redundancy and suppress noise for efficient fusion.} \\[2mm]

\multirow{1}{*}{\cite{wu2023virtual}} 
  & VirConv-T & 2023 & CVPR & VirConvNet & VirConvNet
  & 96.11 & 92.65 & 89.69
  & \red{\textbf{92.54}} & 86.25 & 81.24
  & 73.32 & 66.93 & 60.38
  & 90.04 & 73.90 & 69.06
  & 0.09 s \\
\myrowcolour
& \multicolumn{1}{l}{\textbf{Innovation:}} 
  & \multicolumn{17}{l}{based on a transformed refinement scheme} \\[2mm]

\multirow{1}{*}{\cite{wu2023virtual}} 
  & VirConv-S & 2023 & CVPR & VirConvNet & VirConvNet
  & 95.99 & 93.52 & 90.38
  & 92.48 & \red{\textbf{87.20}} & \red{\textbf{82.45}}
  & $-$ & $-$ & $-$
  & $-$ & $-$ & $-$
  & 0.09 Sec \\
\myrowcolour
& \multicolumn{1}{l}{\textbf{Innovation:}} 
  & \multicolumn{17}{l}{based on a pseudo-label framework} \\[2mm]

\multirow{1}{*}{\cite{wu2023transformation}} 
  & TED-M & 2023 & AAAI & TeSpConv & TeSpConv
  & 55.25 & 79.21 & 46.52
  & 91.61 & 85.28 & 80.68
  & 55.85 & 49.21 & 46.52
  & 88.82 & 74.12 & 66.84
  & $-$ \\
\myrowcolour
& \multicolumn{1}{l}{\textbf{Innovation:}} 
  & \multicolumn{17}{l}{Transformation-equivariant voxel features (TeSpConv), TeBEV and TiVoxel pooling.} \\[2mm]

\multirow{1}{*}{\cite{chen2023lidar}} 
  & DFIM & 2023 & Elsevier & 2D CNN & SA
  & 92.61 & 88.69 & 85.77
  & 88.36 & 81.37 & 76.71
  & 49.91 & 41.97 & 39.27
  & 76.99 & 59.12 & 52.97
  & $-$ \\
\myrowcolour
& \multicolumn{1}{l}{\textbf{Innovation:}} 
  & \multicolumn{17}{l}{Dual-feature interaction with uncertainty-aware IoU optimization.} \\[2mm]

\multirow{1}{*}{\cite{li2023mvmm}} 
  & MVMM & 2023 & IEEE & ResNet-50 & 3D CNN
  & 92.17 & 88.70 & 85.47
  & 87.59 & 78.87 & 73.78
  & 47.54 & 40.49 & 38.36
  & 77.82 & 64.81 & 58.79
  & $-$ \\
\myrowcolour
& \multicolumn{1}{l}{\textbf{Innovation:}} 
  & \multicolumn{17}{l}{Point-cloud coloring with encoder–decoder and multi-view fusion for efficient single-stage 3D detection.} \\[2mm]

\multirow{1}{*}{\cite{qin2023supfusion}} 
  & SupFusion-v1 & 2023 & ICCV & MVXNet & MVXNet
  & $-$ & $-$ & $-$
  & 88.20 & 78.46 & 74.01
  & 63.11 & 58.13 & 53.73
  & 74.79 & 57.60 & 53.82
  & $-$ \\
\myrowcolour
& \multicolumn{1}{l}{\textbf{Innovation:}} 
  & \multicolumn{17}{l}{Supervised fusion with Polar Sampling and deep fusion module for robust LiDAR–camera 3D detection.} \\[2mm]

\multirow{1}{*}{\cite{qin2023supfusion}} 
  & SupFusion-v2 & 2023 & ICCV & PV-RCNN & PV-RCNN
  & $-$ & $-$ & $-$
  & 92.02 & 85.20 & 82.94
  & 74.29 & 66.67 & 61.38
  & 91.00 & \red{\textbf{76.13}} & \red{\textbf{71.66}}
  & $-$ \\
\myrowcolour
& \multicolumn{1}{l}{\textbf{Innovation:}} 
  & \multicolumn{17}{l}{Implements the same supervised fusion strategy within a PV-RCNN backbone for hybrid point–voxel 3D detection.} \\[2mm]

\multirow{1}{*}{\cite{zhu2022vpfnet}} 
  & VPFNet & 2022 & IEEE & 2D Backbone & Voxel-RCNN
  & 93.02 & 91.86 & 86.94
  & 91.02 & 83.21 & 78.20
  & 71.47 & 63.59 & 56.85
  & 87.77 & 71.99 & 66.14
  & 0.1 Sec \\
\myrowcolour
& \multicolumn{1}{l}{\textbf{Innovation:}} 
  & \multicolumn{17}{l}{Two-stage voxel-level stereo–LiDAR fusion with mitigation of resolution mismatch and modality imbalance.} \\[2mm]

\multirow{1}{*}{\cite{wu2022sparse}} 
  & SFD & 2022 & CVPR & CPConvs & Voxel-RCNN
  & 95.64 & 91.85 & 86.83
  & 91.73 & 84.76 & 77.92
  & 72.94 & 66.69 & 61.59
  & \red{\textbf{93.39}} & 72.95 & 67.26
  & 0.1 Sec \\
\myrowcolour
& \multicolumn{1}{l}{\textbf{Innovation:}} 
  & \multicolumn{17}{l}{3D-GAF for fine-grained RoI fusion of raw/pseudo points; SynAugment and CPConv feature extraction.} \\[2mm]

\multirow{1}{*}{\cite{chen2022focal}} 
  & Focals Conv & 2022 & CVPR & 2D CNN & 3D CNN
  & 95.45 & 91.51 & 91.21
  & 90.55 & 82.28 & 77.59
  & $-$ & $-$ & $-$
  & $-$ & $-$ & $-$
  & 0.1 Sec \\
\myrowcolour
& \multicolumn{1}{l}{\textbf{Innovation:}} 
  & \multicolumn{17}{l}{Learnable position-wise importance for sparse CNNs in LiDAR-only and multi-modal detection.} \\[2mm]

\multirow{1}{*}{\cite{li2022voxel}} 
  & VFF & 2022 & CVPR & ResNet-50 & PV-RCNN
  & 95.43 & 91.40 & 90.66
  & 89.58 & 81.97 & 79.17
  & $-$ & $-$ & $-$
  & $-$ & $-$ & $-$
  & $-$ \\
\myrowcolour
& \multicolumn{1}{l}{\textbf{Innovation:}} 
  & \multicolumn{17}{l}{Point-to-ray projection with ray-wise voxel fusion and aligned cross-modal augmentation.} \\[2mm]

\multirow{1}{*}{\cite{li2022voxel}} 
  & VFF & 2022 & CVPR & ResNet-50 & Voxel-RCNN
  & 95.65 & 91.75 & 91.39
  & 89.50 & 82.09 & 79.29
  & 73.26 & 65.11 & 60.03
  & 89.40 & 73.12 & 69.86
  & $-$ \\
\myrowcolour
& \multicolumn{1}{l}{\textbf{Innovation:}} 
  & \multicolumn{17}{l}{Optimizes fully voxel-based detection by embedding point-to-ray voxel field fusion to strengthen dense spatial reasoning and feature alignment.} \\[2mm]

\multirow{1}{*}{\cite{yang2022graph}} 
  & Graph-Vol & 2022 & ECCV & DLA-34 & RPN
  & 95.69 & 90.10 & 86.85
  & 91.89 & 83.27 & 77.78
  & $-$ & $-$ & $-$
  & $-$ & $-$ & $-$
  & $-$ \\
\myrowcolour
& \multicolumn{1}{l}{\textbf{Innovation:}} 
  & \multicolumn{17}{l}{Proposes DPA for adaptive point aggregation, RGP for robust RoI feature learning, and VFA for refinement-stage image–point fusion.} \\[2mm]

\multirow{1}{*}{\cite{zhang2022cat}} 
  & CAT-Det & 2022 & CVPR & IT+CMT & PT+CMT
  & $-$ & $-$ & $-$
  & 89.87 & 81.32 & 76.68
  & 54.26 & 45.44 & 41.94
  & 83.68 & 68.81 & 61.45
  & 0.1 Sec \\
\myrowcolour
& \multicolumn{1}{l}{\textbf{Innovation:}} 
  & \multicolumn{17}{l}{Cross-Modal Transformer backbone with contrastive augmentation for 3D detection.} \\[2mm]

\multirow{1}{*}{\cite{chen2022msl3d}} 
  & MSL3D & 2022 & Elsevier & ResNet-18 & RPN
  & $-$ & $-$ & $-$
  & 87.27 & 81.15 & 76.56
  & 52.16 & 43.38 & 38.80
  & 76.74 & 62.27 & 56.20
  & $-$ \\
\myrowcolour
& \multicolumn{1}{l}{\textbf{Innovation:}} 
  & \multicolumn{17}{l}{Aligns multi-modal receptive fields via 2D set abstraction; unified augmentation with auxiliary multi-sensor training.} \\[2mm]

\multirow{1}{*}{\cite{an2022deep}} 
  & StructuralIF & 2022 & CVIU & SA-SSD & SA-SSD
  & 91.78 & 88.38 & 85.67
  & 87.15 & 80.69 & 76.26
  & 50.80 & 42.42 & 38.35
  & 72.54 & 56.39 & 49.28
  & 0.02 Sec \\
\myrowcolour
& \multicolumn{1}{l}{\textbf{Innovation:}} 
  & \multicolumn{17}{l}{3D–2D consistent features and Lambertian MCR for attentional structural voxels.} \\[2mm]

\multirow{1}{*}{\cite{liu2022epnet++}} 
  & EPNet++ & 2022 & IEEE & PointRCNN & PointRCNN
  & $-$ & $-$ & $-$
  & 91.37 & 81.96 & 76.71
  & 52.79 & 44.38 & 41.29
  & 76.15 & 59.71 & 53.67
  & 0.1 Sec \\
\myrowcolour
& \multicolumn{1}{l}{\textbf{Innovation:}} 
  & \multicolumn{17}{l}{CB-Fusion for cascade bidirectional interaction and MC loss for cross-modal confidence consistency.} \\[2mm]

\multirow{1}{*}{\cite{pang2020clocs}} 
  & CLOCs-v1 & 2020 & IROS & 2D CNN & 2D CNN
  & 91.16 & 88.23 & 82.63
  & 86.38 & 78.45 & 72.45
  & 62.54 & 56.76 & 52.26
  & $-$ & $-$ & $-$
  & 0.1 Sec \\
\myrowcolour
& \multicolumn{1}{l}{\textbf{Innovation:}} 
  & \multicolumn{17}{l}{Late fusion leveraging geometric/semantic consistency for efficient combination of pre-trained 2D/3D detectors.} \\[2mm]

\multirow{1}{*}{\cite{pang2020clocs}} 
  & CLOCs-v2 & 2020 & IROS & 2D CNN & 2D CNN
  & 92.60 & 88.99 & 81.74
  & 87.50 & 76.68 & 71.20
  & 60.33 & 54.17 & 46.42
  & $-$ & $-$ & $-$
  & 0.1 Sec \\
\myrowcolour
& \multicolumn{1}{l}{\textbf{Innovation:}} 
  & \multicolumn{17}{l}{Replaces SECOND with PointRCNN to leverage point-based 3D proposals within the late-fusion consistency framework.} \\[2mm]

\multirow{1}{*}{\cite{pang2020clocs}} 
  & CLOCs-PVCas & 2020 & IROS & 2D CNN & 2D CNN
  & 93.05 & 89.80 & 86.57
  & 88.94 & 80.67 & 77.15
  & $-$ & $-$ & $-$
  & $-$ & $-$ & $-$
  & 0.1 Sec \\
\myrowcolour
& \multicolumn{1}{l}{\textbf{Innovation:}} 
  & \multicolumn{17}{l}{Replaces SECOND with PV-RCNN to incorporate hybrid point–voxel 3D proposals within the late-fusion consistency framework.} \\[2mm]

\multirow{1}{*}{\cite{huang2020epnet}} 
  & EPNet & 2020 & ECCV & 2D CNN & SA + FP
  & 94.22 & 88.47 & 83.69
  & 89.81 & 79.28 & 74.59
  & 52.79 & 44.38 & 41.29
  & 76.15 & 59.71 & 53.67
  & 0.1 Sec \\
\myrowcolour
& \multicolumn{1}{l}{\textbf{Innovation:}} 
  & \multicolumn{17}{l}{Point-wise correspondence weighting with CE loss enforcing classification–localization consistency.} \\[2mm]

\multirow{1}{*}{\cite{vora2020pointpainting}} 
  & PointPainting-v1 & 2020 & CVPR & DeepLabv3+ & PointPillars
  & $-$ & $-$ & $-$
  & 90.01 & 87.65 & 85.65
  & \red{\textbf{77.25}} & \red{\textbf{72.41}} & \red{\textbf{68.53}}
  & 81.72 & 68.76 & 63.99
  & 0.4 Sec \\
\myrowcolour
& \multicolumn{1}{l}{\textbf{Innovation:}} 
  & \multicolumn{17}{l}{Sequentially fuses semantic image segmentation with LiDAR by appending per-point semantic scores before PointPillars detection.} \\[2mm]

\multirow{1}{*}{\cite{vora2020pointpainting}} 
  & PointPainting-v2 & 2020 & CVPR & DeepLabv3+ & VoxelNet
  & $-$ & $-$ & $-$
  & 90.05 & 87.51 & 86.66
  & 73.16 & 65.05 & 57.33
  & 87.46 & 68.08 & 65.59
  & 0.4 Sec \\
\myrowcolour
& \multicolumn{1}{l}{\textbf{Innovation:}} 
  & \multicolumn{17}{l}{Applies sequential semantic point painting prior to VoxelNet-based 3D detection.} \\[2mm]

\multirow{1}{*}{\cite{vora2020pointpainting}} 
  & PointPainting-v3 & 2020 & CVPR & DeepLabv3+ & PointRCNN
  & 92.13 & 87.39 & 82.72
  & 90.19 & 87.64 & 86.71
  & 72.65 & 66.06 & 61.24
  & 86.33 & 73.69 & 70.17
  & 0.4 Sec \\
\myrowcolour
& \multicolumn{1}{l}{\textbf{Innovation:}} 
  & \multicolumn{17}{l}{Applies sequential semantic point painting prior to PointRCNN-based 3D detection.} \\[2mm]

\multirow{1}{*}{\cite{ku2018joint}} 
  & AVOD-FPN & 2018 & IROS & VGG-16 & VGG-16
  & 88.53 & 83.79 & 77.90
  & 81.94 & 71.88 & 66.38
  & 50.80 & 42.81 & 40.88
  & 64.00 & 52.18 & 46.61
  & 0.1 Sec \\
\myrowcolour
& \multicolumn{1}{l}{\textbf{Innovation:}} 
  & \multicolumn{17}{l}{BEV–RGB feature pyramid with multimodal fusion RPN and geometry-constrained 3D encoding.} \\[2mm]

\multirow{1}{*}{\cite{xu2018pointfusion}} 
  & PointFusion & 2018 & CVPR & ResNet-50 & PointNet
  & $-$ & $-$ & $-$
  & 77.92 & 63.00 & 53.27
  & 33.36 & 28.04 & 23.38
  & 49.34 & 29.42 & 26.98
  & $-$ \\
\myrowcolour
& \multicolumn{1}{l}{\textbf{Innovation:}} 
  & \multicolumn{17}{l}{Dense fusion using 3D points as spatial anchors with learned scoring.} \\[2mm]

\multirow{1}{*}{\cite{qi2018frustum}} 
  & F-PointNet -v1 & 2018 & CVPR & FPN-Based & PointNet
  & 87.28 & 77.09 & 67.90
  & 80.62 & 64.70 & 56.07
  & 50.88 & 41.55 & 38.04
  & 69.36 & 53.50 & 52.88
  & 0.17 Sec \\
\myrowcolour
& \multicolumn{1}{l}{\textbf{Innovation:}} 
  & \multicolumn{17}{l}{Frustum proposals from 2D detections enable efficient 3D PointNet processing.} \\[2mm]

\multirow{1}{*}{\cite{qi2018frustum}} 
  & F-PointNet -v2 & 2018 & CVPR & FPN-Based & PointNet++
  & 88.70 & 84.00 & 75.33
  & 81.20 & 70.39 & 62.19
  & 51.21 & 44.89 & 40.23
  & 71.96 & 56.77 & 50.39
  & 0.17 Sec \\
\myrowcolour
& \multicolumn{1}{l}{\textbf{Innovation:}} 
  & \multicolumn{17}{l}{Extends F-PointNet with PointNet++ hierarchical feature learning to improve geometric detail capture in 3D detection.} \\[2mm]

\multirow{1}{*}{\cite{chen2017multi}} 
  & MV3D & 2017 & CVPR & VGG-16 & VGG-16
  & 86.02 & 76.90 & 68.49
  & 71.29 & 62.68 & 56.56
  & 39.48 & 33.69 & 31.51
  & $-$ & $-$ & $-$
  & 0.36 Sec \\
\myrowcolour
& \multicolumn{1}{l}{\textbf{Innovation:}} 
  & \multicolumn{17}{l}{Multi-view region-based fusion combining BEV, FV, and RGB features for oriented 3D detection.} \\[2mm]

    \bottomrule
  \end{tabular}
}
\end{center}
\end{table}

\vspace{-0.5cm}
From Table \ref{table: fusion-kitti}, we provide a detailed discussion of four selected models: VirConv \cite{wu2023virtual}, obtained the highest results in $AP_{3D}$ $Car$ evaluations; ViKIENet \cite{yu2025vikienet}, which achieved the highest results in $AP_{BEV}$ $Car$ evaluations; PointPainting \cite{vora2020pointpainting}, got the highest results in $AP_{3D}$ $Pedestrain$ evaluations; SupFusion \cite{qin2023supfusion}, which attained the highest results in $AP_{3D}$ $Cyclist$, and UPIDet \cite{zhang2023unleash} achieved competitive results in evaluations on the KITTI dataset. VirConv \cite{wu2023virtual} is a multimodal 3D object detection framework that uses virtual points to strengthen LiDAR–RGB fusion while tackling two persistent issues in depth-completion-based augmentation: excessive point density and noise from inaccurate depth estimates. Its core component, the VirConv operator, combines Stochastic Voxel Discard (StVD) to filter out redundant near-range voxels while preserving geometry-rich distant points, and Noise-Resistant Submanifold Convolution (NRConv) to encode features jointly in 3D space and their 2D image projections, enabling effective suppression of boundary noise without eroding useful edge detail. From this foundation, three detector configurations were developed. VirConv-L uses an early-fusion scheme to minimize processing time. VirConv-T follows a late-fusion route and adds multi-transformation refinement to raise detection precision. VirConv-S extends the approach to a semi-supervised framework that uses high-confidence pseudo-labels.

ViKIENet \cite{yu2025vikienet} is a multi-modal 3D object detection framework that replaces the dense, noise-prone virtual points of traditional LiDAR–camera fusion with a compact set of semantically enriched Virtual Key Instances (VKIs). Its architecture comprises three tightly integrated modules. The Semantic Key Instance Selection (SKIS) module generates VKIs by combining depth completion with semantic segmentation, converting only object-relevant pixels into 3D points augmented with class scores. These VKIs and raw LiDAR points are voxelized in separate branches, processed by sparse 3D convolutions, and aligned through the Virtual-Instance-to-Real Attention (VIRA) module, which refines VKI features using precise LiDAR depth cues. The Virtual Instance Focused Fusion (VIFF) module then performs two-stage fusion: BEV-level integration via channel and spatial attention to enhance proposal generation, and RoI-level bi-directional cross-attention to combine VKI semantic richness with LiDAR geometric accuracy for final bounding box refinement. This design reduces computation, suppresses depth-induced noise, and preserves discriminative spatial–semantic cues. 

PointPainting \cite{vora2020pointpainting} is a sequential fusion mechanism that decorates every LiDAR point with image semantics before any LiDAR-only detector processes it. An image segmentation network produces per-pixel class scores; LiDAR points are projected into that score map and “painted” by appending the corresponding class-probability vector, yielding an augmented point cloud that remains compatible with arbitrary LiDAR backbones (e.g., PointPillars, VoxelNet/SECOND, PointRCNN). The method’s key property is modularity: fusion is realized as feature augmentation at the point level, enabling independent advancement of the segmentation and detection stacks and practical pipelining for low-latency deployment. SupFusion \cite{qin2023supfusion} is a training strategy for LiDAR–camera detectors that introduces auxiliary feature-level supervision from an assistant model trained on densified LiDAR via Polar Sampling. During training, standard 3D/2D backbones and a deep fusion module (stacked MLPs with dynamic fusion blocks) produce fusion features that are optimized not only by the detection loss but also to mimic high-quality features generated by the assistant on enhanced data; this supervision strengthens fusion layers and the upstream encoders without incurring inference overhead. The pipeline thus pairs decision-level supervision with auxiliary alignment of fusion representations, thereby improving the robustness of multimodal feature formation and downstream detection.

UPIDet \cite{zhang2023unleash} introduces a LiDAR–image object detection platform that strengthens both geometric representation in the point branch and spatial awareness in the image branch through tightly coupled training objectives. A point cloud backbone extracts multi-scale point features for proposal generation while a Resnet-18 backbone predicts auxiliary outputs to shape the learned 2D representation. The model's key mechanism is bidirectional feature propagation that is performed stage-wise. A pixel-to-point module projects 3D points into the image plane and samples corresponding image features via interpolation, while a point-to-pixel module aggregates point embeddings into a dense 2D grid by pooling features of points falling into each pixel bin and refining the result with shallow convolutions. Beyond 2D/3D semantic segmentation and 3D center estimation, the image branch is supervised by normalized local coordinate map estimation to encode per-pixel relative position within objects, providing spatial cues that complement sparse LiDAR observations, as well as supplying gradients that improve the point backbone during joint optimization.

Table \ref{table: fusion-nuscense} presents an analysis of 2D–3D fusion detection methods on the paired image–LiDAR NuScenes dataset. We provide a detailed discussion of four selected models: MV2DFusion \cite{wang2024mv2dfusion}, CLS-3D \cite{mushtaq2025cls}, and ECL3D \cite{chen2025multi}, which achieved the highest results among all the methods in terms of mAP and NDS metrics on the NuScenes dataset; and TransFusion \cite{bai2022transfusion}, recognized as one of the popular model in the context of object detection. MV2DFusion \cite{wang2024mv2dfusion} employs separate 2D and 3D detection backbones, ResNet-50 with a conventional image detector and FSDv2 for LiDAR, to independently produce candidate object proposals for each modality. These proposals are transformed into modality-specific object queries by dedicated query generation modules. A six-layer transformer-based fusion decoder then processes the queries, first modeling relationships within each set via self-attention, and subsequently enabling feature exchange across modalities through cross-attention. A query calibration stage refines the fused representations before final prediction. By focusing on proposal-level fusion, this design maintains the semantic integrity to avoid the pitfalls of dense feature concatenations.

\begin{table}[H]
\centering
\caption{Comparative analysis of 2D–3D fusion detection methods on the paired image–LiDAR NuScenes dataset.}
\vspace{-0.4cm}
\label{table: fusion-nuscense}
\begin{center}
\resizebox{\textwidth}{!}{%
  \setlength{\tabcolsep}{4pt}%
  \begin{tabular}{llllll ccccccc}
    \toprule
    \textbf{Ref} & \textbf{Method} & \textbf{Year} \hspace{5mm}   & \textbf{Venue} & \textbf{Image Backbone} \hspace{3mm} & \textbf{LiDAR Backbone}
      & \textbf{mAP} & \textbf{NDS} & \textbf{mATE} & \textbf{mASE} & \textbf{mAOE} & \textbf{mAVE} & \textbf{mAAE} \\
    \midrule

    \multirow{1}{*}{\cite{chen2025multi}}
      & ECL3D & 2025 & T-ITS & ResNet-50 & VoxelNet+
      & \red{\textbf{72.8}} & \red{\textbf{75.2}} & \red{\textbf{0.252}}  & \red{\textbf{0.235}} & \red{\textbf{0.326}} & \red{\textbf{0.184}} & \red{\textbf{0.127}} \\[1mm]
      \myrowcolour
      & \multicolumn{1}{l}{\textbf{Innovation:}}
      & \multicolumn{11}{l}{Introduces dual-scale depth supervision and deformable cross-attention for enhanced BEV feature fusion without added complexity.} \\[2mm]


    \multirow{1}{*}{\cite{mushtaq2025cls}}
      & CLS-3D & 2025 & IEEE & DeepLabv3+ & 3D backbone
      & \red{\textbf{75.1}} & \red{\textbf{76.8}} & \red{\textbf{0.273}}  & \red{\textbf{0.244}} & \red{\textbf{0.302}} & \red{\textbf{0.214}} & \red{\textbf{0.108}} \\[1mm]
      \myrowcolour
      & \multicolumn{1}{l}{\textbf{Innovation:}}
      & \multicolumn{11}{l}{Fuses LiDAR and image features via point-wise semantic augmentation with reweighted transformer for enhanced 3D detection.} \\[2mm]



    \multirow{1}{*}{\cite{zhang2024sparselif}}
      & SparseLIF-T & 2024 & ECCV & VoV-99 & VoxelNet
      & 75.9 & 77.7 & 0.244 & 0.231 & 0.284 & 0.152 & 0.117 \\[1mm]
      \myrowcolour
      & \multicolumn{1}{l}{\textbf{Innovation:}}
      & \multicolumn{11}{l}{Fully sparse LiDAR-camera detection via perspective-aware queries and uncertainty-guided fusion.}\\[2mm]

    \multirow{1}{*}{\cite{zhao2024simplebev}}
      & SimpleBEV & 2024 & arXiv & ConvXt-Tiny & 3D Sparse CNN
      & 75.7 & 77.6 & 0.236 & 0.235 & 0.283 & 0.143 & 0.126 \\[1mm]
      \myrowcolour
      & \multicolumn{1}{l}{\textbf{Innovation:}}
      & \multicolumn{11}{l}{Enhances BEV fusion via LiDAR-rectified depth estimation and multi-scale sparse LiDAR features.} \\[2mm]

    \multirow{1}{*}{\cite{yin2024fusion}}
      & IS-Fusion & 2024 & CVPR & Swin-T & VoxelNet
      & 76.5 & 77.4 & 0.230 & 0.231 & 0.267 & 0.223 & 0.774 \\[1mm]
      \myrowcolour
      & \multicolumn{1}{l}{\textbf{Innovation:}}
      & \multicolumn{11}{l}{Hybrid instance- and scene-level fusion enabling structured multimodal interaction for robust BEV detection.} \\[2mm]

    \multirow{1}{*}{\cite{huang2024detecting}}
      & DAL & 2024 & ECCV & ResNet-50 & VoxelNet+
      & 71.5 & 74.0 & 0.253 & 0.239 & 0.334 & 0.174 & 0.120 \\[1mm]
      \myrowcolour
      & \multicolumn{1}{l}{\textbf{Innovation:}}
      & \multicolumn{11}{l}{``Detecting As Labeling'' paradigm with elegant training pipeline and instance-level velocity augmentation for robust fusion.} \\[2mm]


    \multirow{1}{*}{\cite{wang2024mv2dfusion}}
      & MV2DFusion-v1 & 2024 & ArXiv & ResNet-50 & FSDv2
      & 74.5 & 76.7 & 0.245 & 0.229 & 0.269 & 0.119 & 0.115 \\[1mm]
      \myrowcolour
      & \multicolumn{1}{l}{\textbf{Innovation:}}
      & \multicolumn{11}{l}{Depth-guided multi-view 2D fusion aligning image features across views before BEV projection for robust image--LiDAR fusion.} \\[2mm]

    \multirow{1}{*}{\cite{wang2024mv2dfusion}}
      & MV2DFusion-v2 & 2024 & ArXiv & ResNet-50 & FSDv2
      & \red{\textbf{77.9}} & \red{\textbf{78.8}} & \red{\textbf{0.237}}  & \red{\textbf{0.226}} & \red{\textbf{0.247}} & \red{\textbf{0.192}} & \red{\textbf{0.119}} \\[1mm]
      \myrowcolour
      & \multicolumn{1}{l}{\textbf{Innovation:}}
      & \multicolumn{11}{l}{MV2DFusion with model ensemble and test-time augmentation for further accuracy gains.} \\[2mm]


    \multirow{1}{*}{\cite{liu2022bevfusion}}
      & BEVFusion & 2023 & ICRA & Swin-Tiny & VoxelNet
      & 70.2 & 72.9 & 0.261 & 0.239 & 0.329 & 0.260 & 0.134 \\[1mm]
      \myrowcolour
      & \multicolumn{1}{l}{\textbf{Innovation:}}
      & \multicolumn{11}{l}{Performs sensor fusion in BEV space instead of LiDAR space, offering a simpler yet stronger fusion paradigm.} \\[2mm]



    \multirow{1}{*}{\cite{xie2023sparsefusion}}
      & SparseFusion & 2023 & ICCV & ResNet-50 & VoxelNet+
      & 72.0 & 73.8 & 0.258 & 0.243 & 0.329 & 0.265 & 0.131 \\[1mm]
      \myrowcolour
      & \multicolumn{1}{l}{\textbf{Innovation:}}
      & \multicolumn{11}{l}{Instance-level sparse feature fusion with cross-modality transfer for fast, accurate LiDAR--camera 3D detection.} \\[2mm]

    \multirow{1}{*}{\cite{yan2023cross}}
      & CMT & 2023 & ICCV & VOVNet & VoxelNet
      & 72.0 & 74.1 & 0.279 & 0.235 & 0.308 & 0.259 & 0.112 \\[1mm]
      \myrowcolour
      & \multicolumn{1}{l}{\textbf{Innovation:}}
      & \multicolumn{11}{l}{End-to-end multi-modal 3D detection with direct 3D position encoding into tokens, avoiding grid sampling or voxel pooling.} \\[2mm]

    \multirow{1}{*}{\cite{chen2023futr3d}}
      & FUTR3D & 2023 & CVPR & VoVNet & VoxelNet
      & 69.4 & 72.1 & 0.284 & 0.241 & 0.310 & 0.300 & 0.120 \\[1mm]
      \myrowcolour
      & \multicolumn{1}{l}{\textbf{Innovation:}}
      & \multicolumn{11}{l}{Modality-agnostic feature sampler enabling end-to-end fusion across arbitrary sensor configurations and combinations.} \\[2mm]

    \multirow{1}{*}{\cite{chen2023focalformer3d}}
      & FocalFormer3D & 2023 & ICCV & ResNet-50 & VoxelNet
      & 72.9 & 75.0 & 0.250 & 0.242 & 0.328 & 0.226 & 0.126 \\[1mm]
      \myrowcolour
      & \multicolumn{1}{l}{\textbf{Innovation:}}
      & \multicolumn{11}{l}{Hard Instance Probing to identify false negatives and enhance BEV features for improved 3D detection recall.} \\[2mm]

    \multirow{1}{*}{\cite{chen2023largekernel3d}}
      & LargeKernel3D-F & 2023 & CVPR & SW-LK & SW-LK
      & 71.1 & 74.2 & 0.236 & 0.228 & 0.298 & 0.241 & 0.131 \\[1mm]
      \myrowcolour
      & \multicolumn{1}{l}{\textbf{Innovation:}}
      & \multicolumn{11}{l}{Uses spatial-wise partition convolution and position embedding to realize efficient, very large 3D kernels in sparse CNNs.} \\[2mm]


    \multirow{1}{*}{\cite{hu2023fusionformer}}
      & FusionFormer & 2023 & ArXiv & VoV-99 & VoxelNet
      & 72.6 & 75.1 & 0.267 & 0.236 & 0.328 & 0.226 & 0.105 \\[1mm]
      \myrowcolour
      & \multicolumn{1}{l}{\textbf{Innovation:}}
      & \multicolumn{11}{l}{Transformer framework with uniform sampling to fuse modalities without pre-BEV voxel compression.} \\[2mm]

    \multirow{1}{*}{\cite{hu2023ea}}
      & EA-LSS & 2023 & ArXiv & 2D Backbone & N/A
      & 72.2 & 74.4 & 0.247 & 0.237 & 0.304 & 0.250 & 0.133 \\[1mm]
      \myrowcolour
      & \multicolumn{1}{l}{\textbf{Innovation:}}
      & \multicolumn{11}{l}{Edge-aware, fine-grained depth modules improving accuracy and alignment in LSS-based BEV detection.} \\[2mm]

    \multirow{1}{*}{\cite{li2022unifying}}
      & UVTR-M & 2022 & NeurIPS  & ResNet-101 & VoxelNet+
      & 67.1 & 71.1 & 0.306 & 0.245 & 0.351 & 0.225 & 0.124 \\[1mm]
      \myrowcolour
      & \multicolumn{1}{l}{\textbf{Innovation:}}
      & \multicolumn{11}{l}{Cross-modality interaction and a transformer decoder} \\[2mm]

    \multirow{1}{*}{\cite{bai2022transfusion}}
      & TrasFusion & 2022 & CVPR & ResNet-50 & VoxelNet
      & 68.9 & 71.7 & 0.259 & 0.243 & 0.329 & 0.288 & 0.127 \\[1mm]
      \myrowcolour
      & \multicolumn{1}{l}{\textbf{Innovation:}}
      & \multicolumn{11}{l}{Uses a transformer decoder to fuse LiDAR BEV features.} \\[2mm]




    \multirow{1}{*}{\cite{xu2021fusionpainting}}
      & FusionPainting & 2021 & ITSC & ResNet-50 & VoxelNet+
      & 66.3 & 70.4 & 0.256 & 0.236 & 0.346 & 0.274 & 0.132 \\[1mm]
      \myrowcolour
      & \multicolumn{1}{l}{\textbf{Innovation:}}
      & \multicolumn{11}{l}{Semantic-level fusion with voxel-level attention to integrate multi-modal context features for improved 3D detection.} \\[2mm]

    \multirow{1}{*}{\cite{yin2021multimodal}}
      & MVP & 2021 & NeurIPS  & DLA-34 & VoxelNet+
      & 66.4 & 70.5 & 0.263 & 0.238 & 0.231 & 0.313 & 0.134 \\[1mm]
      \myrowcolour
      & \multicolumn{1}{l}{\textbf{Innovation:}}
      & \multicolumn{11}{l}{Virtual points from images to balance LiDAR density across distances and enhance multi-modal 3D detection accuracy.} \\[2mm]

    \multirow{1}{*}{\cite{wang2021pointaugmenting}}
      & PointAugmenting & 2021 & CVPR & DLA-34 & VoxelNet+
      & 66.8 & 71.1 & 0.253 & 0.234 & 0.345 & 0.266 & 0.123 \\[1mm]
      \myrowcolour
      & \multicolumn{1}{l}{\textbf{Innovation:}}
      & \multicolumn{11}{l}{Uses 2D detector features with cross-modal data augmentation to enhance LiDAR--camera 3D object detection.} \\[2mm]

    \multirow{1}{*}{\cite{yoo20203d}}
      & 3D-CVF & 2020 & ECCV & ResNet-18 & 3D Sparse CNN
      & 67.5 & 57.8 & 0.280 & 0.246 & 0.367 & 0.239 & 0.129 \\[1mm]
      \myrowcolour
      & \multicolumn{1}{l}{\textbf{Innovation:}}
      & \multicolumn{11}{l}{Auto-calibrated cross-view projection with adaptive gated fusion and 3D RoI-based refinement for robust camera--LiDAR detection.} \\[2mm]

    \bottomrule
  \end{tabular}}
\end{center}
\end{table}

\vspace{-0.5cm}
CLS-3D \cite{mushtaq2025cls} integrates LiDAR and RGB information through a unified backbone in which each modality has its encoder but contributes to a shared feature space for joint reasoning. The image pathway applies DeepLabv3+ to obtain pixel-wise semantic probability maps geometrically aligned to the LiDAR frame. Then, each 3D point is annotated with the semantic probabilities of its corresponding image pixel, creating an enriched point set that combines geometric and class-level cues. This representation is processed by a transformer module equipped with a slot-based reweighting mechanism, allowing the network to emphasise channels that carry the most relevant spatial or temporal information. For final detection, the network employs an I3C-IoU loss considering box overlap, centre displacement, and scale, producing more precise localisation in crowded or visually challenging driving environments.

ECL3D \cite{chen2025multi} follows a dense-to-sparse fusion pipeline with parallel image and LiDAR streams producing BEV features via ResNet-50+FPN with an LSS view transformer and a VoxelNet+SECOND backbone, respectively. The dense stage fuses BEV features to generate a category heatmap, from which top-K proposals are selected. In the sparse stage, dual-modal proposal features are extracted for final regression and classification. Two core enhancements improve accuracy: a dual-scale depth-semantic supervision module in the image branch augments LSS with an auxiliary high-resolution depth prediction branch without increasing inference cost, and an Instance BEV Feature Enhancement (IBFE) module applies deformable cross-attention between LiDAR and image BEVs to align semantic and geometric cues before proposal refinement. TransFusion \cite{bai2022transfusion} extracts LiDAR BEV features via a VoxelNet backbone and image features via a ResNet-50 image backbone, then applies a two-stage transformer decoder as the detection head. The first stage decodes sparse, category-aware object queries from LiDAR BEV features into initial bounding boxes. The second stage performs LiDAR–camera fusion using Spatially Modulated Cross-Attention (SMCA), which imposes a locality bias by focusing on relevant image regions for each query. An image-guided query initialization pathway seeds queries using 2D features to help detect small objects under sparse LiDAR sampling. This strategy improves robustness against sensor misalignment and degraded image conditions.

\vspace{-0.5cm}
\subsection{Comparative Analysis of 2D, 3D, and Fusion}

In this subsection, we provide a comparative analysis of 2D, 3D, and fusion-based object detection paradigms in autonomous driving, highlighting their complementary strengths and practical trade-offs. While 2D methods benefit from rich appearance cues and mature deployment pipelines, 3D approaches offer stronger geometric consistency for localization and scene understanding, and fusion frameworks aim to combine the advantages of multiple sensors to improve reliability under challenging conditions. Beyond reporting accuracy alone, a meaningful comparison must reflect both benchmark performance and real-world deployment constraints, where perception modules are required to operate reliably under strict real-time and hardware limitations. In the remainder of this subsection, we discuss these factors along six dimensions: (1) accuracy and robustness, (2) computational cost, (3) inference latency, (4) energy efficiency, (5) hardware requirements, and (6) calibration and synchronization.

\vspace{-0.5cm}
\subsubsection{Accuracy and Robustness}
\begin{figure}[htbp]
    \centering
    \centerline{\includegraphics[width=1\textwidth]{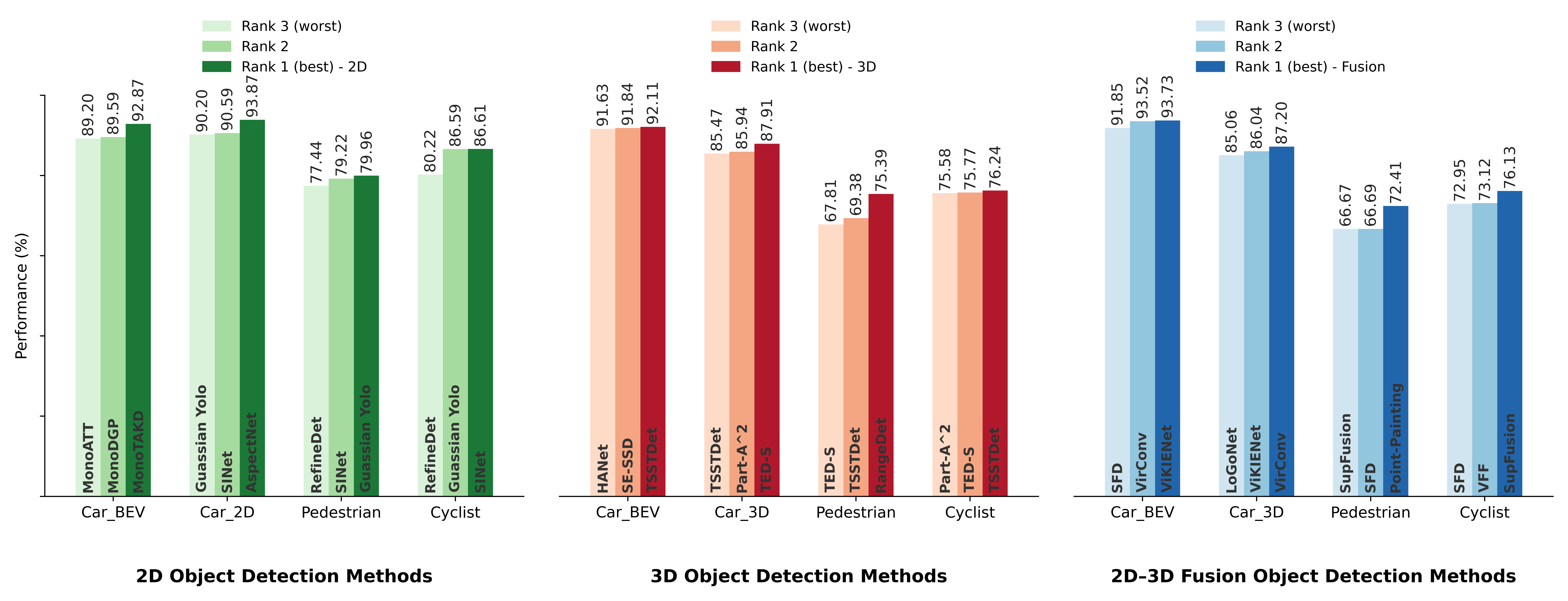}}
    \vspace{-0.3cm}
    \caption{Performance comparison of the top three algorithms from each detection category (2D, 3D, and 2D–3D fusion object detection) across four evaluation metrics: $AP_{BEV}$ $Car$, $AP_{3D}$ $Car$, $AP_{3D}$ $Pedestrian$, and $AP_{3D}$ $Cyclist$ on the KITTI dataset.}
    \label{fig: comparison}
\end{figure}

Figure \ref{fig: comparison} presents a performance comparison of the top three algorithms from each detection category (2D, 3D, and 2D–3D fusion), evaluated across four key metrics: $AP_{BEV}$ $Car$, $AP_{3D}$ $Car$, $AP_{3D}$ $Pedestrian$, and $AP_{3D}$ $Cyclist$. The results highlight the strengths and weaknesses of each category, illustrating how different detection paradigms excel in specific object classes or evaluation perspectives. This comparative view not only identifies the highest-performing methods but also provides insight into the trade-offs between 2D, 3D, and fusion-based approaches.

The comparative analysis across 2D, 3D, and 2D–3D fusion object detection methods highlights modality-driven performance trends. The 2D detectors, evaluated on a dedicated 2D dataset, achieve strong accuracy in the image domain, particularly for the $AP_{2D}$ $Car$ class, where the top-ranked model surpasses 93\% performance. However, their capabilities are inherently constrained in reconstructing depth and precise spatial localization due to the absence of explicit 3D information. On the other hand, LiDAR-based 3D detectors directly operate in 3D space, delivering competitive spatial accuracy, especially in $AP_{BEV}$ $Car$ and $AP_{3D}$ $Car$ metrics, but often exhibit lower performance in detecting small or partially occluded objects, such as pedestrians, where point cloud sparsity and occlusion effects degrade recall. Fusion-based methods leverage complementary strengths: images' dense texture and semantic cues enhance object classification, while LiDAR provides accurate geometry and scale estimation. They achieved the highest $AP_{BEV}$ $Car$ (93.73\%) and $AP_{3D}$ $Car$ (87.20\%) scores, a result attributed to the combined benefits of depth perception and classification from cross-modal cues.

Despite the overall advantage, the fusion improvements are marginal in pedestrian detection. This limited gain is likely attributable to multiple factors: (1) the low point density for pedestrians in LiDAR, especially at range, reduces the geometric detail available for fusion; (2) the relatively small pixel footprint of pedestrians in images limits the discriminative value of texture cues; and (3) misalignment errors from calibration and synchronization between modalities have a greater proportional effect on small-scale targets, diminishing fusion efficacy. Furthermore, since the 2D detectors were trained on a purely 2D dataset, their high pedestrian and cyclist detection scores in the image domain do not fully translate into 3D accuracy when projected or fused, given that depth cues must be inferred indirectly. Consequently, while fusion consistently outperforms single-modality approaches in vehicle-related tasks, the marginal pedestrian improvement underscores the persistent challenge of small-object 3D detection, even under multimodal paradigms.

\begin{figure}[h]
    \centering
    \centerline{\includegraphics[width=0.70\textwidth]{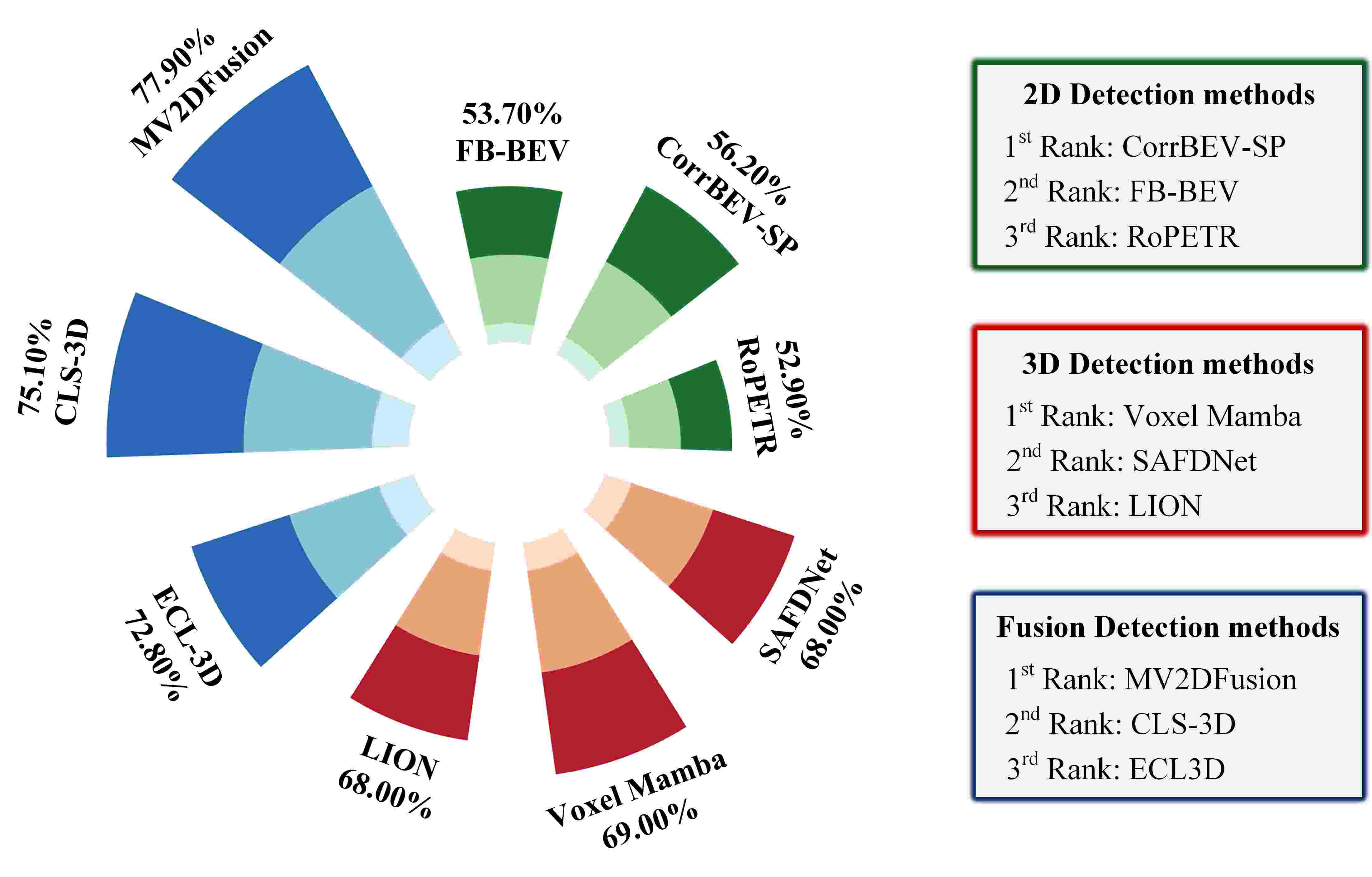}}
    \vspace{-0.2cm}
    \caption{Performance comparison of the top three algorithms from each detection category (2D, 3D, and 2D–3D fusion object detection) across $mAP$ evaluation metric on the NuScenes dataset.}
    \label{fig: nucomparison}
\end{figure}

Additionally, Figure \ref{fig: nucomparison} presents a performance comparison of the top three algorithms from each detection category (2D, 3D, and 2D–3D fusion), evaluated across the $mAP$ metric. The results highlight the strengths and weaknesses of each category, illustrating how different detection paradigms excel in specific object classes or evaluation perspectives. This comparative view not only identifies the highest-performing methods within each category but also provides insight into the trade-offs between 2D, 3D, and fusion-based approaches for AV perception. On the nuScenes benchmark, the mean Average Precision (mAP) values show a clear separation between fusion-based, LiDAR-only, and camera-only detection strategies. The highest score is achieved by MV2DFusion at 77.90\%, ahead of the best LiDAR method, Voxel Mamba (69.00\%), and the leading camera-based approach, CorrBEV-SP (56.20\%). This margin reflects the benefit of combining complementary sensing LiDAR’s precise geometry and range measurements with the rich semantic detail from cameras, allowing fusion models to localize objects better and distinguish classes under varied conditions. LiDAR-only models such as SAFDNet and LION remain strong in spatial reasoning but can miss fine visual cues. At the same time, camera-based methods, even when adapted to a BEV representation, are limited by their indirect depth estimation. 

While the above comparison highlights accuracy-driven differences across 2D, 3D, and fusion-based paradigms, real-world autonomous driving deployment requires an additional and equally important axis: computational and resource efficiency under strict real-time constraints. In practice, perception pipelines must meet bounded end-to-end latency, operate within embedded power and thermal limits, and remain feasible on automotive-grade hardware platforms. In addition, runtime values reported in benchmarks or leaderboards are often measured under heterogeneous settings (e.g., GPU class, inference engine and optimization libraries, batch size, input resolution, and numerical precision), and therefore cannot be treated as a complete or fair indicator of deployability. A deployment-oriented assessment should also consider algorithmic complexity, sensor-driven throughput and synchronization overhead, memory footprint and bandwidth demand, and the calibration costs introduced by multi-sensor perception stacks.

\subsubsection{Computational Cost}
Computational cost is a primary differentiator between 2D, 3D, and fusion detectors, and it is driven by both the network architecture and the data representation of each modality. For 2D detectors, compute typically scales with the input image resolution and the backbone choice (CNN versus Transformer), where higher-resolution feature pyramids improve small-object sensitivity but increase memory traffic. In contrast, LiDAR-based 3D detectors incur additional cost from point encoding and spatial aggregation. Point-based approaches generally scale with the number of points and neighborhood queries, whereas voxel- or pillar-based methods trade geometric fidelity for efficiency by discretizing the scene and enabling (sparse) convolutional processing. Fusion detectors are frequently the most expensive class, as they often execute dual backbones (image and point cloud) and add cross-modal fusion layers (e.g., attention-based alignment, BEV projection, or transformer decoders), which increase both compute and intermediate activation storage. Therefore, a deployment-oriented comparison should report standardized indicators such as parameter count, peak activation memory, and the sensitivity of compute to controllable factors (image resolution, number of points, voxel size, BEV grid size, and temporal window length).

\subsubsection{Inference Latency}
Beyond average FPS, real deployment requires predictable inference-time behavior because the perception--planning--control loop is bounded by strict deadlines. In practice, latency is affected not only by network depth, but also by preprocessing and representation steps (e.g., voxelization, BEV rasterization, NMS strategy, and multi-sensor buffering). Importantly, modality-dependent input rates introduce inherent timing constraints: cameras typically provide higher frame rates (e.g., 30--60~Hz), while LiDAR commonly operates at lower rates (e.g., 10--20~Hz), and radar may fall in a similar range (e.g., 10--30~Hz). Consequently, fusion pipelines may incur additional end-to-end delay due to timestamp alignment, buffering, and interpolation between asynchronous sensor streams. Moreover, 3D pipelines can exhibit higher latency variance across frames because the number of active points/voxels and scene density (traffic, weather artifacts) can fluctuate, impacting sparse-kernel efficiency and memory access patterns. For a fair comparison, inference-time evaluation should be reported with batch size fixed to one, fixed precision (FP16/INT8 where applicable), and measured end-to-end latency that includes preprocessing, fusion synchronization, and post-processing, rather than reporting network-only runtime.

\subsubsection{Energy Efficiency}
Energy efficiency must be evaluated at the system level since perception cost is coupled with sensing. Even if a model is computationally optimized, the overall power budget can be dominated by high-power sensors and their downstream processing demands. As summarized in Table~2, LiDAR is typically associated with high power consumption and high computational load, while cameras exhibit low power consumption but still impose high computational load due to dense visual processing; radar tends to require moderate power with low computational load. Therefore, fusion-based stacks should be assessed by the combined sensing and inference budget, not accuracy alone. In deployment, this motivates energy-aware strategies such as conditional sensor activation (e.g., radar-dominant operation in poor visibility), adaptive compute (dynamic resolution, sparse queries, early-exit policies), and lightweight fusion mechanisms (late fusion or gated fusion) that reduce unnecessary cross-modal computation when conditions are benign. When possible, reporting energy per frame (J/frame) or power under steady-state operation (W) on representative automotive hardware provides a more meaningful measure than FLOPs alone.

\subsubsection{Hardware Requirements}
Hardware feasibility depends on computing throughput, memory capacity, and kernel support for the dominant operators in each modality. 2D-only detectors are generally the most portable, benefiting from mature acceleration paths for convolutions and attention variants, and often achieving substantial speedups under quantization and operator fusion. LiDAR-based 3D detectors, especially those using sparse convolutions or complex point neighborhood aggregation, can be bottlenecked by memory bandwidth and the availability of optimized sparse-kernel implementations on the target runtime. Fusion detectors typically impose the highest requirements because they maintain dual feature hierarchies and additional fusion modules, leading to increased memory footprint and bandwidth pressure. Moreover, multimodal pipelines also increase system integration complexity due to multi-sensor data ingestion, synchronization, and fault handling. As a result, deployment-oriented comparisons should explicitly state the assumed compute platform class (embedded GPU/NPU/ASIC), memory constraints, and whether the method relies on operations that are not uniformly supported across inference toolchains (e.g., sparse 3D kernels, custom voxelization operators, or heavy cross-attention in BEV space).

\subsubsection{Calibration and Synchronization}
Calibration and synchronization constitute an often overlooked but critical component of deployability, especially for fusion systems. Accurate extrinsic calibration is required to align camera and LiDAR/radar measurements into a consistent coordinate frame, and timing synchronization is essential to mitigate motion-induced misalignment during ego-motion or dynamic traffic. Table 2 indicates that calibration complexity is typically \textit{high} for cameras and LiDAR, while radar and ultrasonic sensors tend to exhibit lower calibration complexity. In long-term fleet deployment, calibration is not a one-time engineering task; it must remain stable under vibration, temperature variation, sensor aging, and maintenance events, and it can introduce recurring operational costs. Fusion methods that assume precise alignment may therefore experience performance degradation when calibration drifts or when sensors operate asynchronously. From a system perspective, this encourages robust fusion designs that tolerate imperfect alignment (e.g., uncertainty-aware fusion, temporal modeling, and motion-compensated alignment) and emphasizes the need to evaluate fusion models under realistic synchronization noise and calibration perturbations rather than idealized settings.

\begin{figure}[h]
    \centering
    \centerline{\includegraphics[width=0.75\textwidth]{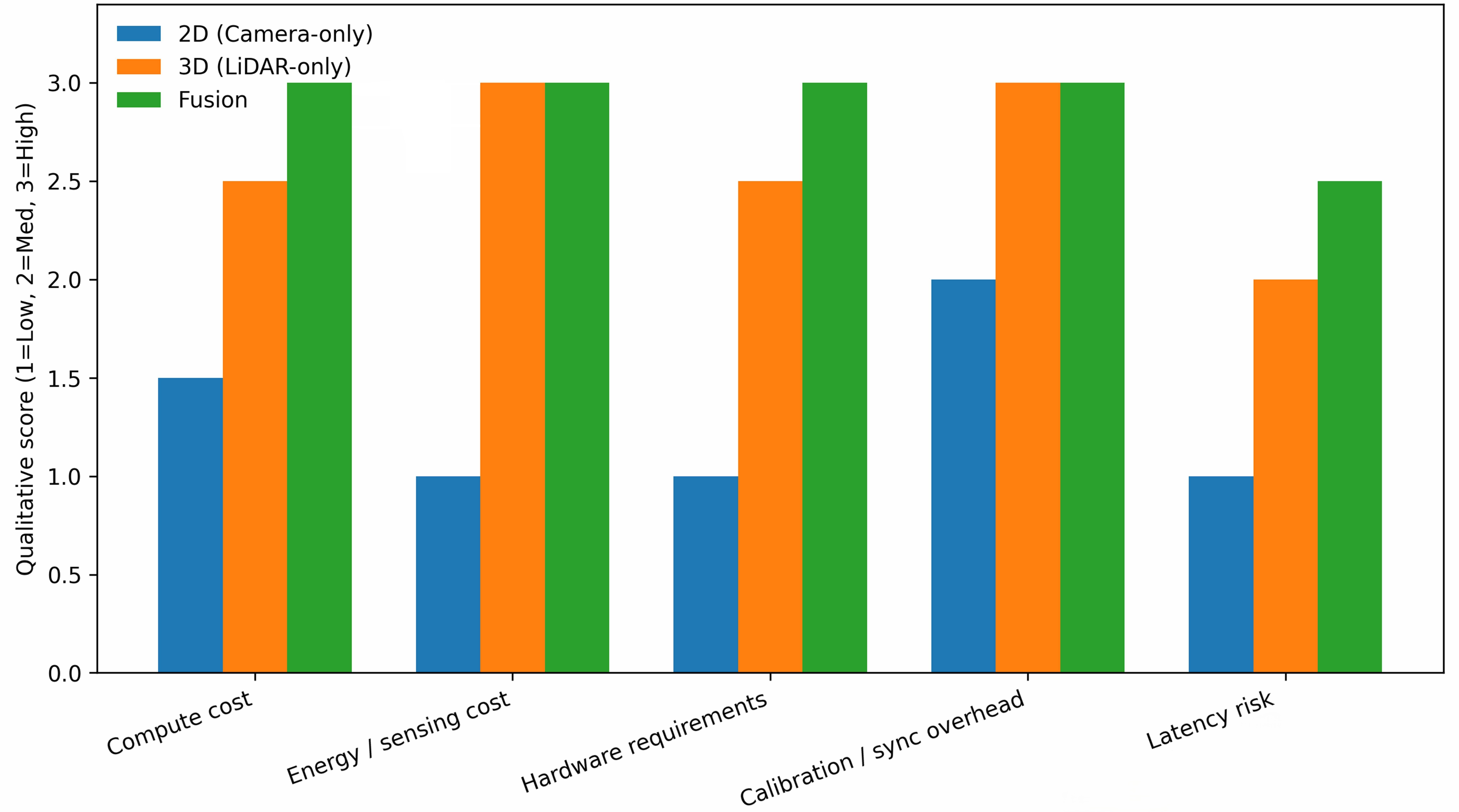}}
    \vspace{-0.2cm}
    \caption{Deployment-oriented trade-off analysis of 2D, 3D, and fusion detection paradigms, summarizing qualitative differences in computational cost, energy/sensing cost, hardware requirements, calibration/synchronization overhead, and latency risk.
}
    \label{fig:nucomparison}
\end{figure}

Figure \ref{fig:nucomparison} provides a qualitative comparison of 2D, 3D, and fusion paradigms across deployment-critical factors such as computational cost, latency risk, energy/sensing cost, hardware requirements, and calibration/synchronization overhead. Overall, 2D methods typically provide the most computationally efficient solutions and the lowest hardware barrier, but they remain constrained by depth ambiguity and limited 3D localization fidelity. 3D LiDAR methods offer stronger geometric consistency and more reliable spatial reasoning, yet they often entail higher sensing cost, power consumption, and computational load due to point processing and BEV/voxel feature construction. Fusion methods can achieve the strongest performance in vehicle-centric metrics by leveraging complementary sensor cues, but they must justify the added system complexity---including higher latency, larger memory footprint, and calibration/synchronization overhead---to be practical in real deployment. Consequently, a deployment-oriented evaluation should report not only accuracy, but also standardized compute indicators (e.g., Params and FLOPs/MACs), end-to-end latency measured under controlled settings (batch size $=1$, fixed precision, and consistent input resolution), peak memory usage, and energy/power estimates on representative automotive-grade hardware.

\subsection{LLMs/VLMs-based Methods}
\label{sec: LLMs/VLMs}

LLMs are being recognized as powerful tools for autonomous driving systems, since they facilitate semantic understanding, contextual reasoning and support the extraction of useful information from language data. Their capabilities have been demonstrated across various aspects of autonomous driving, including interpreting sensor data and generating scene summaries, as well as assisting with trajectory planning and decision-making under dynamic traffic conditions. LLMs take a fundamentally different approach from traditional computer vision systems. While conventional CV models depend entirely on what they can "see" in images, LLMs work with human language. Meanwhile, VLMs are designed to understand and process both images and text together, allowing for more complete and meaningful interpretation of complex driving environments and scenarios. Unlike traditional CV models that only analyze image pixels or LLMs that focus on text, VLMs handle both at the same time. They can describe images, answer questions based on what they "see," and retrieve relevant visuals given a text query.

\begin{figure}[H]
    \centering
    \centerline{\includegraphics[width=1\textwidth]{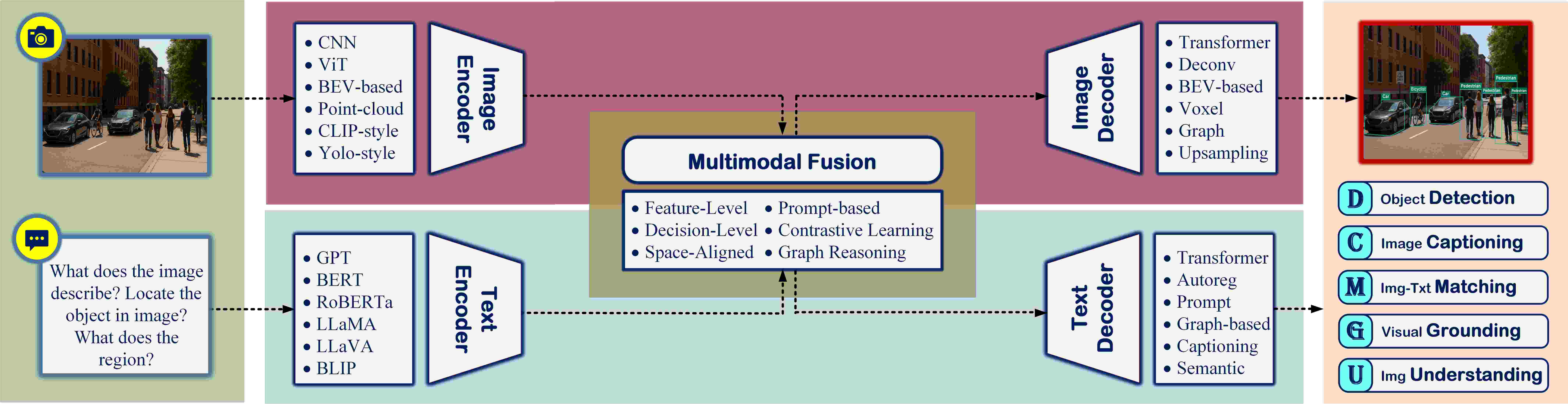}}
    \caption{Overall framework of the VLMs in the context of autonomous driving systems.}
    \label{fig: VLM architecture}
\end{figure}

Figure \ref{fig: VLM architecture} demonstrates an overview of VLM systems tailored for autonomous driving applications. The framework begins with both image and text inputs, which are first processed by image and text encoders to extract high-level semantic features. These features are then integrated through a multimodal fusion module, enabling the model to align and reason across visual and linguistic modalities. The fused representation is subsequently passed through image and text decoders, depending on the task, to generate meaningful task-specific outputs such as object detection, scene captioning, trajectory prediction, or semantic reasoning. At the heart of this architecture, VLMs enable key capabilities, including perception, interaction, prediction, reasoning, alignment, and generation, bridging the gap between low-level sensor data and high-level decision-making.

\vspace{-0.5cm}
Nowadays, there has been significant progress in LLMs/VLMs, and a wide variety of models have been developed for different domains and applications. Each model has different sizes, supported modalities, inference costs, and design architectures. Table \ref{Table: LLM-VLM-General} provides an in-depth overview of the most prominent LLM/VLM models, highlighting their developers, architectures, modalities, number of parameters, and key innovations.


\begin{table}[htbp]
\caption{Summary and analysis of the significant LLM and VLM models, highlighting their advancements and key innovations.}
\vspace{-0.4cm}
\label{Table: LLM-VLM-General}
\begin{center}
\resizebox{\textwidth}{!}{
\setlength{\tabcolsep}{7pt}
\begin{tabular}{lllllllll}

\toprule
\textbf{Ref}  & \textbf{Year}  &  \textbf{Name} & \textbf{Developer}  & \textbf{Architecture}   &\textbf{Modalities} & \textbf{Parameters} & \textbf{Key Innovations}\\[1mm]

\midrule
\myrowcolour
\multicolumn{8}{c}{\textbf{Large-Language Models (LLMs)}} \\
\midrule

\citep{comanici2025gemini}    &2025     &Gemini 2.5   &Google DeepMind   &Multimodal-  &Text, Vision      &$\sim$100\,B  
    &\makebox[0pt][l]{$\square$}\raisebox{.15ex}{\hspace{0.1em}$\checkmark$}Multilingual Processing\\[1mm]
    &&&&Transformer
    &&&\makebox[0pt][l]{$\square$}\raisebox{.15ex}{\hspace{0.1em}$\checkmark$}Multimodal Input\\[2mm]
    
\citep{bae2024enhancing}     &2025      &Claude 4          &Anthropic        &Constitutional-        &Text         &$>$1\,T    
    &\makebox[0pt][l]{$\square$}\raisebox{.15ex}{\hspace{0.1em}$\checkmark$}Safety-focused LLM\\[1mm]
    &&&&Transformer
    &&&\makebox[0pt][l]{$\square$}\raisebox{.15ex}{\hspace{0.1em}$\checkmark$}Improved Reasoning\\[2mm]

\citep{xAI}       &2025      &Grok 4          &xAI            &GPT-style       &Text         &$\sim$1\,B-100\,B    
    &\makebox[0pt][l]{$\square$}\raisebox{.15ex}{\hspace{0.1em}$\checkmark$}Twitter-integrated\\[1mm]
    &&&&Transformer
    &&&\makebox[0pt][l]{$\square$}\raisebox{.15ex}{\hspace{0.1em}$\checkmark$}Improved Factuality\\[2mm]

\citep{hui2024qwen2}     &2025      &Qwen-2.5          &Alibaba            &Multimodal-        &Text, Vision          &+110\,B
    &\makebox[0pt][l]{$\square$}\raisebox{.15ex}{\hspace{0.1em}$\checkmark$}Multilingual Processing\\[1mm]
    &&&&Transformer
    &&&\makebox[0pt][l]{$\square$}\raisebox{.15ex}{\hspace{0.1em}$\checkmark$}Multimodal Input\\[2mm]

    \citep{OpenAI}     &2024      &GPT-o3         &OpenAI            &GPT4-style        &Text, Vision         &$>$1\,T     
    &\makebox[0pt][l]{$\square$}\raisebox{.15ex}{\hspace{0.1em}$\checkmark$}Multimodal Input\\[1mm]
    &&&&Transformer  &
    &&\makebox[0pt][l]{$\square$}\raisebox{.15ex}{\hspace{0.1em}$\checkmark$}High-speed \\[2mm]

    \citep{dubey2024llama}     &2024      &LLaMA 3          &Meta AI            &Decoder-only         &Text         &$\sim$8\,B-70\,B   
    &\makebox[0pt][l]{$\square$}\raisebox{.15ex}{\hspace{0.1em}$\checkmark$}Improved Scaling\\[1mm]
    &&&&Transformer
    &&&\makebox[0pt][l]{$\square$}\raisebox{.15ex}{\hspace{0.1em}$\checkmark$}Training Stability\\[2mm]
    
    \citep{team2024gemini}     &2024      &Gemini 1.5          &Google DeepMind      &Multimodal-     &Text, Vision     &$\sim$100\,B-1\,T   
    &\makebox[0pt][l]{$\square$}\raisebox{.15ex}{\hspace{0.1em}$\checkmark$}Context Understanding\\[1mm]
    &&&&Transformer
    &&&\makebox[0pt][l]{$\square$}\raisebox{.15ex}{\hspace{0.1em}$\checkmark$}Multilingual Processing\\[2mm]

    \citep{Anthropic}     &2024      &Claude 3          &Anthropic            &Constitutional-        &Text         &$\sim$100\,B-1\,T    
    &\makebox[0pt][l]{$\square$}\raisebox{.15ex}{\hspace{0.1em}$\checkmark$}Vision-language Integration\\[1mm]
    &&&&Transformer
    &&&\makebox[0pt][l]{$\square$}\raisebox{.15ex}{\hspace{0.1em}$\checkmark$}Enhanced Reasoning\\[2mm]
    
    \citep{lu2024deepseek}     &2024      &DeepSeek-VL          &DeepSeek AI            &Vision-Language       &Text, Vision         &$\sim$1\,B-100\,B
    &\makebox[0pt][l]{$\square$}\raisebox{.15ex}{\hspace{0.1em}$\checkmark$}Multimodal Tuning\\[1mm]
    &&&&Transformer
    &&&\makebox[0pt][l]{$\square$}\raisebox{.15ex}{\hspace{0.1em}$\checkmark$}Vision-language Integration\\[2mm]

    \citep{hurst2024gpt}     &2024      &GPT-4o         &OpenAI            &Multimodal-         &Text, Audio         &$\sim$1.8\,T  
    &\makebox[0pt][l]{$\square$}\raisebox{.15ex}{\hspace{0.1em}$\checkmark$}Real-time Conversation\\[1mm]
    &&&&Transformer  &
    &&\makebox[0pt][l]{$\square$}\raisebox{.15ex}{\hspace{0.1em}$\checkmark$}End-to-end Audio\\[2mm]

    \citep{abdin2024phi}     &2024      &Phi-3          &Microsoft            &Compact-         &Text         &$\sim$3.8\,B   
    &\makebox[0pt][l]{$\square$}\raisebox{.15ex}{\hspace{0.1em}$\checkmark$}Small-scale\\[1mm]
    &&&&Transformer
    &&&\makebox[0pt][l]{$\square$}\raisebox{.15ex}{\hspace{0.1em}$\checkmark$}Code Understanding\\[2mm]

        \citep{touvron2023llama}     &2023      &LLaMA 2          &Meta AI            &Decoder-only      &Text     &$\sim$7\,B-70\,M     
    &\makebox[0pt][l]{$\square$}\raisebox{.15ex}{\hspace{0.1em}$\checkmark$}Chat Fine-tuning\\[1mm]
    &&&&Transformer
    &&&\makebox[0pt][l]{$\square$}\raisebox{.15ex}{\hspace{0.1em}$\checkmark$}Logic Capabilities\\[2mm]

        \citep{achiam2023gpt}     &2023      &GPT-4          &OpenAI            &Multimodal-         &Text, Vision         &$>$1\,T     
    &\makebox[0pt][l]{$\square$}\raisebox{.15ex}{\hspace{0.1em}$\checkmark$}Multimodal Reasoning\\[1mm]
    &&&&Transformer
    &&&\makebox[0pt][l]{$\square$}\raisebox{.15ex}{\hspace{0.1em}$\checkmark$}Vision Support\\[2mm]

        \citep{anil2023palm}     &2023      &PaLM 2          &Google            &Parallel Layers        &Text     &$\sim$1\,B-100\,B    
    &\makebox[0pt][l]{$\square$}\raisebox{.15ex}{\hspace{0.1em}$\checkmark$}Multilingual Reasoning\\[1mm]
    &&&&Transformer
    &&&\makebox[0pt][l]{$\square$}\raisebox{.15ex}{\hspace{0.1em}$\checkmark$}Logic Capabilities\\[2mm]

        \citep{harabagiu2000falcon}     &2023      &Falcon          &TII            &Auto-regressive         &Text         &$\sim$180\,B   
    &\makebox[0pt][l]{$\square$}\raisebox{.15ex}{\hspace{0.1em}$\checkmark$}Massive scale MoE\\[1mm]
    &&&&Transformer
    &&&\makebox[0pt][l]{$\square$}\raisebox{.15ex}{\hspace{0.1em}$\checkmark$}Improved Conversation\\[2mm]

        \citep{radford2018improving}     &2019      &GPT-2          &OpenAI      &Decoder-only   &Text    &$\sim$1.5\,B      
    &\makebox[0pt][l]{$\square$}\raisebox{.15ex}{\hspace{0.1em}$\checkmark$}Generative Pretraining\\[1mm]
    &&&&Transformer
    &&&\makebox[0pt][l]{$\square$}\raisebox{.15ex}{\hspace{0.1em}$\checkmark$}Zero-shot Generalization\\[2mm]
       
        \citep{lewis2019bart}     &2019      &BART       &Meta AI        &Encoder–Decoder        &Text         &$\sim$406\,M  
    &\makebox[0pt][l]{$\square$}\raisebox{.15ex}{\hspace{0.1em}$\checkmark$}BERT-style Encoder\\[1mm]
    &&&&Transformer
    &&&\makebox[0pt][l]{$\square$}\raisebox{.15ex}{\hspace{0.1em}$\checkmark$}GPT-style Decoder\\[2mm]

        \citep{devlin2019bert}     &2018      &BERT          &Google            &Bidirectional-         &Text         &$\sim$340\,M    
    &\makebox[0pt][l]{$\square$}\raisebox{.15ex}{\hspace{0.1em}$\checkmark$}Bidirectional Encoder\\[1mm]
    &&&&Transformer
    &&&\makebox[0pt][l]{$\square$}\raisebox{.15ex}{\hspace{0.1em}$\checkmark$}Masked Modeling\\[2mm]

\midrule 
\myrowcolour
\multicolumn{8}{c}{\textbf{Vision-Language Models (VLMs)}} \\
\midrule

    \citep{ma2025cake}    &2025          &CAKE      &Xiamen       &CLIP +         &Text, Vision         &$\sim$200\,M    
    &\makebox[0pt][l]{$\square$}\raisebox{.15ex}{\hspace{0.1em}$\checkmark$}Category-aware Extraction\\[1mm]
    &&&&Faster R-CNN
    &&&\makebox[0pt][l]{$\square$}\raisebox{.15ex}{\hspace{0.1em}$\checkmark$}Novel-class Generalization
    \\[2mm]

    \citep{zhou2025led}    &2025          &LED      &Rutgers      &LLM-guided             &Text, Vision         &$\sim$230\,M  
    &\makebox[0pt][l]{$\square$}\raisebox{.15ex}{\hspace{0.1em}$\checkmark$}Fuses LLM Hidden States\\[1mm]
    &&&&Transformer
    &&&\makebox[0pt][l]{$\square$}\raisebox{.15ex}{\hspace{0.1em}$\checkmark$}Avoids Synthetic Data
    \\[2mm]

    \citep{cheng2024yolo}    &2024       &YOLO-World    &Tencent Lab  &YOLO +         &Text, Vision         &$\sim$100\,M    
    &\makebox[0pt][l]{$\square$}\raisebox{.15ex}{\hspace{0.1em}$\checkmark$}Text Encoder Fusion\\[1mm]
    &&&&Text Encoder
    &&&\makebox[0pt][l]{$\square$}\raisebox{.15ex}{\hspace{0.1em}$\checkmark$}Training-free Deployment
    \\[2mm]

    \citep{liu2024grounding}    &2024          &Grounding-DINO	  &IDEA  &DETR-style            &Text, Vision         &$\sim$172\,M  
    &\makebox[0pt][l]{$\square$}\raisebox{.15ex}{\hspace{0.1em}$\checkmark$}LLM-guided Region\\[1mm]
    &&&&Transformer
    &&&\makebox[0pt][l]{$\square$}\raisebox{.15ex}{\hspace{0.1em}$\checkmark$}Text-guided Transformer
    \\[2mm]

    \citep{wu2023cora}    &2023          &CORA     &CPII       &CLIP Backbone +         &Text, Vision         &$\sim$200\,M  
    &\makebox[0pt][l]{$\square$}\raisebox{.15ex}{\hspace{0.1em}$\checkmark$}CLIP-based Anchor\\[1mm]
    &&&&DETR Transformer
    &&&\makebox[0pt][l]{$\square$}\raisebox{.15ex}{\hspace{0.1em}$\checkmark$}Text-aware Proposal
    \\[2mm]

    \citep{li2022grounded}    &2022          &GLIP    &Microsoft  &Vision-Language          &Text, Vision         &$\sim$180\,M    
    &\makebox[0pt][l]{$\square$}\raisebox{.15ex}{\hspace{0.1em}$\checkmark$}Grounded pretraining\\[1mm]
    &&&&Transformer 
    &&&\makebox[0pt][l]{$\square$}\raisebox{.15ex}{\hspace{0.1em}$\checkmark$}Unified VLM Backbone
    \\[2mm]

    \citep{kuo2022f}    &2022          &F-VLM    &Google     &CLIP-based 	T        &Text, Vision         &$\sim$400\,M   
    &\makebox[0pt][l]{$\square$}\raisebox{.15ex}{\hspace{0.1em}$\checkmark$}Frozen CLIP Backbone\\[1mm]
    &&&&Transformer
    &&&\makebox[0pt][l]{$\square$}\raisebox{.15ex}{\hspace{0.1em}$\checkmark$}Lightweight Fine-tuning
    \\[2mm]

    \citep{kamath2021mdetr}    &2021          &MDETR     &Facebook      &DETR-style         &Text, Vision         &$\sim$132\,M            
    &\makebox[0pt][l]{$\square$}\raisebox{.15ex}{\hspace{0.1em}$\checkmark$}First Multimodal Detection\\[1mm]
    &&&&Transformer
    &&&\makebox[0pt][l]{$\square$}\raisebox{.15ex}{\hspace{0.1em}$\checkmark$}Language-modulated Queries
    \\[2mm]

\bottomrule
\end{tabular}}
\end{center}
\end{table}

From Table \ref{Table: LLM-VLM-General}, it is clear that LLM development has progressed significantly, from early models like GPT-1 and BERT to more advanced architectures such as GPT-4o, Gemini 2.5, and Claude 4. These models differ in architecture type, number of parameters, modality support, and their ability to perform reasoning, generation, or multimodal tasks. Some models, like GPT-3.5, GPT-4, and Claude, support tool-use and real-time applications, while others focus on domain-specific tasks. In the context of autonomous driving, LLMs can be integrated to enhance perception, facilitate natural language interaction, interpret sensor data, and support decision-making processes. Their ability to understand and generate human-like language enables more explainable and context-aware autonomous systems. These abilities make them powerful tools for complementing traditional CV methods, especially when domain knowledge and semantic reasoning are required. Meanwhile, the VLMs developed from early models like CLIP and MDETR to the latest advanced architectures such as Grounding-DINO and CAKE. While earlier models focused on contrastive learning and dual encoders, recent VLMs use advanced technologies such as LLM-guided prompts, grounding mechanisms, and hierarchical reasoning. These enhancements allow VLMs to better align visual and textual modalities, operate effectively in zero-shot settings, and support a wide range of vision-language tasks with higher accuracy and flexibility.

Table \ref{Table:vlm_driving_stacked} provides a comparative analysis of recent vision–language driving methods evaluated on two complementary benchmarks: nuScenes closed-loop planning (reporting L2 trajectory error and collision rate over 1–3s horizons) and Bench2Drive (CARLA) closed-loop driving (driving score, success rate, efficiency, and comfort). The table is intended as a practical resource for readers who want to (i) understand which architectural choices are being made in the driving space, and (ii) interpret performance through the lens of reasoning capacity, 3D grounding, and inference efficiency. In particular, pairing planning accuracy (L2) with safety (collision) on nuScenes helps distinguish methods that reduce geometric error from those that meaningfully improve safety under interaction, while Bench2Drive complements this view with closed-loop behavioral metrics that stress long-horizon decision consistency and controllability in a simulator designed for multi-ability evaluation.

\begin{table}[H]
\centering
\caption{Comparative analysis of VLM-based methods on NuScenes and Bench2Drive.}
\label{Table:vlm_driving_stacked}

\resizebox{\textwidth}{!}{%
  \setlength{\tabcolsep}{4pt}%
  \begin{tabular}{lllllll cccc cccc}
    \toprule
    \textbf{Ref} & \textbf{Method} & \textbf{Year} & \textbf{Venue} & \textbf{Visual Encoder} & \textbf{Language Model} & \textbf{Modality}
    & \multicolumn{8}{c}{\textbf{NuScenes}} \\
    \cmidrule(lr){8-15}
    & & & & & &
    & \multicolumn{4}{c}{\textbf{L2 (m) $\downarrow$}}
    & \multicolumn{4}{c}{\textbf{Collision (\%) $\downarrow$}} \\
    \cmidrule(lr){8-11}\cmidrule(lr){12-15}
    & & & & & &
    & \textbf{1s} & \textbf{2s} & \textbf{3s} & \textbf{Avg.}
    & \textbf{1s} & \textbf{2s} & \textbf{3s} & \textbf{Avg.} \\
    \midrule

    \multirow{1}{*}{\cite{chen2025solve}}
      & SOLVE & 2025 & CVPR & SQ-Former & LLaVA-1.5 & Camera
      & 0.13 & 0.25 & 0.47 & 0.28
      & 0.00 & 0.16 & 0.43 & 0.20 \\[1mm]
    \myrowcolour
      & \multicolumn{1}{l}{\textbf{Innovation:}}
      & \multicolumn{13}{l}{Vision--language-guided trajectory refinement via chain-of-thought planning over trajectory banks.} \\[2mm]

    \multirow{1}{*}{\cite{wang2025omnidrive}}
      & OmniDrive & 2025 & CVPR & EVA-02-L & LLaVA & Camera
      & 0.14 & 0.29 & 0.55 & 0.33
      & 0.00 & 0.13 & 0.78 & 0.30 \\[1mm]
    \myrowcolour
      & \multicolumn{1}{l}{\textbf{Innovation:}}
      & \multicolumn{13}{l}{Counterfactual 3D driving question--answer generation enabling scalable data creation.} \\[2mm]

    \multirow{1}{*}{\cite{han2025dme}}
      & DME-Driver & 2025 & AAAI & CLIP & LLaVA & Camera
      & 0.43 & 0.91 & 1.58 & 0.97
      & 0.04 & 0.14 & 0.64 & 0.27 \\[1mm]
    \myrowcolour
      & \multicolumn{1}{l}{\textbf{Innovation:}}
      & \multicolumn{13}{l}{Human-aligned dual-system driving combining language decisions with executor-level control.} \\[2mm]

    \multirow{1}{*}{\cite{zhou2503opendrivevla}}
      & OpenDriveVLA-7B & 2025 & arXiv & ResNet-101 & Qwen2.5-Instruct & Camera
      & 0.15 & 0.31 & 0.55 & 0.33
      & 0.01 & 0.08 & 0.21 & 0.10 \\[1mm]
    \myrowcolour
      & \multicolumn{1}{l}{\textbf{Innovation:}}
      & \multicolumn{13}{l}{Hierarchical alignment of 2D--3D instance-aware tokens with language for grounded actions.} \\[2mm]

    \multirow{1}{*}{\cite{cao2025fastdrivevla}}
      & FastDriveVLA & 2025 & arXiv & Qwen2.5-VL & Impromptu-VLA & Camera
      & 0.13 & 0.28 & 0.53 & 0.32
      & 0.00 & 0.15 & 0.63 & 0.26 \\[1mm]
    \myrowcolour
      & \multicolumn{1}{l}{\textbf{Innovation:}}
      & \multicolumn{13}{l}{Reconstruction-based foreground-aware token pruning enabling efficient acceleration of driving VLA.} \\[2mm]

    \multirow{1}{*}{\cite{xie2025s4}}
      & S4-Drive & 2025 & CVPR & ViT-G (2B) & PaLI-3-5B & Camera
      & 0.13 & 0.28 & 0.51 & 0.31
      & -- & -- & -- & -- \\[1mm]
    \myrowcolour
      & \multicolumn{1}{l}{\textbf{Innovation:}}
      & \multicolumn{13}{l}{Self-supervised end-to-end motion planning with MLLMs using sparse spatiotemporal volume representations.} \\[2mm]

    \multirow{1}{*}{\cite{hegde2025distilling}}
      & DiMA-Dual+ & 2024 & CVPR & -- & LLaVA-v1.5-7B & Camera
      & 0.14 & 0.27 & 0.46 & 0.29
      & 0.05 & 0.07 & 0.15 & 0.09 \\[1mm]
    \myrowcolour
      & \multicolumn{1}{l}{\textbf{Innovation:}}
      & \multicolumn{13}{l}{Distills multimodal reasoning into structured planners for efficient long-tail-robust driving.} \\[2mm]

    \multirow{1}{*}{\cite{pan2024vlp}}
      & VLP & 2024 & CVPR & BEV & CLIP & Camera
      & 0.30 & 0.53 & 0.84 & 0.55
      & 0.01 & 0.07 & 0.38 & 0.15 \\[1mm]
    \myrowcolour
      & \multicolumn{1}{l}{\textbf{Innovation:}}
      & \multicolumn{13}{l}{Integrates language-model reasoning into BEV driving via adaptive planning and strategic decision layers.} \\[2mm]

    \multirow{1}{*}{\cite{tian2024drivevlm}}
      & DriveVLM & 2024 & arXiv & SigLIP-L-384 & Qwen-VL & Camera
      & 0.18 & 0.34 & 0.68 & 0.40
      & 0.10 & 0.22 & 0.45 & 0.27 \\[1mm]
    \myrowcolour
      & \multicolumn{1}{l}{\textbf{Innovation:}}
      & \multicolumn{13}{l}{Chain-of-thought vision--language reasoning for scene understanding and hierarchical trajectory planning.} \\[2mm]

    \multirow{1}{*}{\cite{tian2024drivevlm}}
      & DriveVLM-Dual & 2024 & arXiv & SigLIP-L-384 & Qwen-VL & Camera
      & 0.15 & 0.29 & 0.48 & 0.31
      & 0.05 & 0.08 & 0.17 & 0.10 \\[1mm]
    \myrowcolour
      & \multicolumn{1}{l}{\textbf{Innovation:}}
      & \multicolumn{13}{l}{Slow--fast dual-system integrating VLM reasoning with classical perception and high-rate refinement.} \\[2mm]

    \multirow{1}{*}{\cite{hwang2024emma}}
      & EMMA & 2024 & arXiv & Gemini & Gemini & Camera
      & 0.14 & 0.29 & 0.54 & 0.33
      & -- & -- & -- & -- \\[1mm]
    \myrowcolour
      & \multicolumn{1}{l}{\textbf{Innovation:}}
      & \multicolumn{13}{l}{End-to-end multimodal driving via unified language representation for planning and road understanding.} \\[2mm]

    \bottomrule
  \end{tabular}%
} 


\resizebox{\textwidth}{!}{%
  \setlength{\tabcolsep}{4pt}%
  \begin{tabular}{lllllll cccc}
    \textbf{Ref} & \textbf{Method} & \textbf{Year} & \textbf{Venue} & \textbf{Visual Encoder} & \textbf{Language Model} & \textbf{Modality}
    & \multicolumn{4}{c}{\textbf{Bench2Drive (CARLA)}} \\
    \cmidrule(lr){8-11}
    & & & & & &
    & \textbf{Driving Score} & \textbf{Success Rate} & \textbf{Efficiency} & \textbf{Comfortness} \\
    \midrule

    \multirow{1}{*}{\cite{zhou2025autovla}}
      & AutoVLA & 2025 & arXiv & Qwen2.5-VL & Qwen2.5-VL-3B & Camera
      & 78.84 & 57.73 & 146.93 & 39.33 \\[1mm]
    \myrowcolour
      & \multicolumn{1}{l}{\textbf{Innovation:}}
      & \multicolumn{9}{l}{Human-aligned dual-system driving coupling language decision reasoning with executor-level control supervision.} \\[2mm]

    \multirow{1}{*}{\cite{fu2025orion}}
      & Orion & 2025 & arXiv & EVA-02-L & Vicuna v1.5 & Camera
      & 77.74 & 54.62 & 151.48 & 17.38 \\[1mm]
    \myrowcolour
      & \multicolumn{1}{l}{\textbf{Innovation:}}
      & \multicolumn{9}{l}{Generative planning framework differentiably aligning VLM reasoning with continuous trajectory action space.} \\[2mm]

    \multirow{1}{*}{\cite{renz2025simlingo}}
      & SimLingo & 2025 & CVPR & InternViT-300M & Qwen2.0-5B-Instruct & Camera
      & 85.07 & 67.27 & 259.23 & 33.67 \\[1mm]
    \myrowcolour
      & \multicolumn{1}{l}{\textbf{Innovation:}}
      & \multicolumn{9}{l}{Closed-loop vision-only driving with explicit language--action alignment via Action Dreaming and CoT commentary.} \\[2mm]

    \bottomrule
  \end{tabular}%
} 

\end{table}

Across the nuScenes entries, Table \ref{Table:vlm_driving_stacked} highlights a clear shift from “language as an auxiliary explanation tool” toward language-conditioned planning pipelines that explicitly use reasoning to refine or structure trajectories. For example, SOLVE \cite{chen2025solve} introduces a trajectory chain-of-thought paradigm that progressively refines candidate trajectories (via a trajectory bank) and improves real-time usability via temporal decoupling between heavy VLM reasoning and faster end-to-end planning. In a related direction, DriveVLM \cite{tian2024drivevlm} frames driving as a multi-stage reasoning process (description → analysis → hierarchical planning) using chain-of-thought style intermediate reasoning, while DriveVLM-Dual \cite{tian2024drivevlm} further adopts a slow–fast dual system to retain the interpretability of VLM reasoning without sacrificing high-rate control refinement. This “reason-then-act” structure is also echoed by approaches that focus on robustness and deployment practicality: DiMA \cite{hegde2025distilling} distills knowledge from a multimodal LLM into an efficient vision-based planner using distillation tasks so that LLM inference is not strictly required at test time, improving long-tail robustness while controlling inference cost.

A second trend captured by the Table \ref{Table:vlm_driving_stacked} is the increasing emphasis on 3D grounding and scalable data generation as bottlenecks for driving-capable VLM/VLA models. OmniDrive \cite{wang2025omnidrive} tackles this by constructing a holistic vision–language dataset and agents using counterfactual 3D driving Q\&A—i.e., systematically generating “what if the ego vehicle did X?” questions that force models to learn decision-sensitive 3D understanding, and showing improved transfer to nuScenes planning. Complementarily, S4-Driver \cite{xie2025s4} argues that many MLLMs are pretrained in 2D perspective space and proposes a sparse spatiotemporal volume representation to lift multi-view/multi-frame observations into a representation better aligned with 3D planning, enabling scalable self-supervised motion planning without necessarily fine-tuning the vision encoder. In parallel, OpenDriveVLA \cite{zhou2503opendrivevla} targets grounded action generation by aligning instance-aware 2D and 3D visual tokens with language and ego states through hierarchical vision–language feature alignment, aiming to reduce the gap between perception representations and language-conditioned trajectory generation. Finally, EMMA \cite{hwang2024emma} pushes the “everything as language” philosophy by representing non-visual inputs (e.g., ego status, navigation) and multiple outputs (planning, objects, road graph) in a unified text space, enabling multi-task co-training under an MLLM foundation and yielding strong planning performance while noting computational cost and limited frame capacity as practical constraints.

Bench2Drive results in Table \ref{Table:vlm_driving_stacked} further emphasize that closing the loop is not only about trajectory regression quality, but also about semantic–action alignment and efficient inference under interaction. ORION \cite{fu2025orion} explicitly frames a key challenge as the mismatch between semantic reasoning space (VLM) and numerical action space (trajectories), and addresses it with a design that combines a history-aware “QT-Former”, an LLM reasoning module, and a generative planner with alignment between reasoning and action for unified optimization across planning/VQA. SimLingo \cite{renz2025simlingo} similarly makes alignment central: it unifies closed-loop driving, vision–language understanding, and language–action alignment, and introduces Action Dreaming to evaluate instruction–action consistency without risky on-policy execution, while remaining camera-only. Finally, Table \ref{Table:vlm_driving_stacked} also surfaces a growing line of work on efficiency as a first-class objective for VLA driving, exemplified by FastDriveVLA, which proposes reconstruction-based, foreground-aware token pruning (ReconPruner) to reduce the high cost of long visual token sequences while preserving driving-relevant information. Collectively, these methods suggest a convergence toward designs that (i) leverage language for structured reasoning and long-tail awareness, (ii) improve 3D grounding and data scalability, and (iii) enforce tight language–action alignment and efficiency—precisely the trade-offs that Table \ref{Table:vlm_driving_stacked} is designed to make visible for readers.

\subsection{Challenges and Future Directions}

LLMs and VLMs introduce significant opportunities for advancing autonomous driving by enabling multimodal perception, contextual reasoning, and decision support beyond the capabilities of traditional unimodal pipelines. Their ability to jointly process language and visual signals, including camera streams, LiDAR/radar-derived representations, HD maps, and V2X messages, can provide richer situational awareness and more flexible interaction with the driving stack. Recent studies suggest these models can support tasks such as open-vocabulary object understanding, scene description and risk explanation, long-tail event reasoning, and natural-language interfaces for querying driving context (e.g., ``Why did the vehicle slow down?'' or ``What hazards are present at this intersection?''). When effectively integrated, LLM/VLM-based perception and reasoning modules have the potential to improve robustness in complex environments, enhance interpretability, and facilitate safer decision-making in both research prototypes and real-world autonomous vehicle systems \cite{boroujeni2026vla4codrive}.

However, in safety-critical autonomy, the adoption of LLMs/VLMs must be evaluated in terms of robustness, verifiability, and controlled failure modes. A primary concern is \textbf{hallucination} and spurious reasoning: models may generate plausible but incorrect objects, traffic rules, causal explanations, or future event hypotheses. In autonomous driving, such failures are particularly dangerous when language outputs are implicitly trusted by downstream planning modules or by human operators. Beyond hallucination, \textbf{miscalibration} is equally critical: models may express high-confidence explanations under uncertainty, masking epistemic gaps and making it difficult to design reliable handoff or fallback behaviors. These risks motivate the need for uncertainty-aware inference, confidence estimation, and explicit verification mechanisms before using model outputs for control-critical decisions.

A second essential challenge is \textbf{robustness under distribution shift} across geography, weather, time-of-day, traffic composition, and sensor artifacts. Most LLMs/VLMs are pre-trained on internet-scale data that underrepresents driving-specific edge cases, such as rare road users, unusual signage, construction zones, emergency interactions, adverse illumination, glare, precipitation, and LiDAR/radar dropouts. As a result, failures often emerge precisely in the scenarios where safety margins are already tight. Addressing this requires systematic domain adaptation with curated driving datasets, scenario-balanced evaluation, and stress testing across Operational Design Domains (ODDs), rather than reporting performance on a narrow benchmark split.

Importantly, LLM/VLM-based driving stacks also expand the \textbf{security exposure} of autonomous vehicles. In addition to naturally occurring failures, these models are vulnerable to adversarial attacks and malicious inputs that can induce unsafe behavior: (i) physical-world adversarial perturbations such as adversarial signs, stickers, or billboards; (ii) sensor- and pipeline-level spoofing including LiDAR/radar interference, camera glare injection, or synchronization attacks; and (iii) language- and instruction-based attacks such as prompt injection through V2X messages, navigation instructions, passenger commands, or roadside text content. Such attacks can manipulate model perception or reasoning (e.g., incorrect hazard attribution or rule interpretation) without obvious pixel-level anomalies. Future research must therefore treat security as a first-class requirement, incorporating adversarial training, input sanitization, robust fusion with redundancy checks, and intrusion-aware monitoring.

A further challenge lies in \textbf{real-time deployment} under strict resource and latency constraints. Autonomous vehicles require low-latency perception and prediction on embedded hardware with limited compute, memory, and power budgets and high reliability requirements. Large multimodal models are expensive to run continuously for detection, tracking, forecasting, and scene-level reasoning, especially under multi-sensor fusion and cooperative perception settings with bandwidth-limited V2X links. Practical deployment will likely require efficiency-oriented designs (e.g., token pruning, distilled student models, sparse or event-triggered reasoning, and modular “slow--fast” architectures), together with systems-level scheduling that prioritizes safety-critical tasks under compute contention.

Finally, the field lacks mature \textbf{assurance and evaluation protocols} for LLM/VLM-driven autonomy. Standard metrics (e.g., mAP, planning L2) do not fully capture whether a model’s reasoning is causally grounded, safety-consistent, and robust under worst-case conditions. Future directions include: (1) uncertainty-aware evaluation that measures calibration and abstention; (2) counterfactual and adversarial stress tests that target safety-relevant failure modes; (3) rule- and map-consistency checking to verify language-based claims against structured world models; (4) runtime monitors and shields that constrain actions to a verified safe set; and (5) auditable reasoning traces that support accountability and post-incident analysis. Overall, while LLMs and VLMs open new frontiers for autonomous driving, their effective adoption will require advances in robust domain adaptation, adversarial resilience, uncertainty-aware reasoning, and tightly controlled integration into existing AV pipelines with clear fallback behaviors.

\vspace{-0.3cm}
\section{Conclusion }
\label{sec: Conclusion}
\vspace{-0.3cm}

\subsection{Discussion and Summary of Key Findings}
This survey provides a comprehensive, structured review of object detection in autonomous vehicles, bridging the domains of sensor technologies, benchmark datasets, and detection methodologies. By systematically analyzing 2D camera-based, 3D LiDAR-based, 2D–3D fusion, and emerging LLM/VLM-based approaches, we have synthesized their strengths, limitations, and optimal use cases. Our unique dataset categorization, covering ego-vehicle, roadside, and cooperative perception datasets, offers a clear framework for evaluating and selecting suitable datasets based on application needs and purposes. Moreover, this work outlines the evolving landscape of AV perception by integrating insights from traditional CV pipelines and recent advances such as Vision Transformers and multimodal reasoning. The results underscore that robust object detection is inherently multimodal, requiring balanced trade-offs between computational efficiency, environmental adaptability, and semantic richness. Ultimately, this survey serves as a reference point for both academic and industry researchers seeking to design, benchmark, and deploy next-generation AV perception systems.

\subsection{Future Research Directions}
Future research in AV object detection should explore dynamic, context-aware sensor fusion pipelines that adaptively reweight multimodal inputs (camera, LiDAR, Radar) in real time based on driving context, environmental conditions, and system uncertainty, which is still largely unexplored in practice. Integrating foundation models trained on massive, cross-domain multimodal datasets could enable AVs to reason about rare and unseen scenarios beyond current benchmark coverage. Another promising yet underdeveloped direction is cross-vehicle collaborative perception enhanced by LLM/VLM reasoning, where vehicles exchange compressed semantic representations instead of raw sensor data to reduce bandwidth while preserving scene understanding. In addition, simulation-to-reality domain adaptation leveraging generative models and neural rendering could close the performance gap between synthetic and real-world data. Finally, a critical but underexplored frontier lies in uncertainty-aware perception systems that can self-assess detection reliability and dynamically adjust planning and control strategies, moving toward safer, more interpretable, and trustworthy AV decision-making.

\vspace{-0.3cm}
\section*{Declaration of Competing Interest}
\vspace{-0.3cm}
The authors declare that they have no known competing financial interests or personal relationships that could have appeared to influence the work reported in this paper.

\vspace{-0.3cm}
\section*{Acknowledgement}
\label{sec:Acknowledgement}
\vspace{-0.3cm}
This material is based upon the work supported by the National Science Foundation (NSF) under Grant Numbers 2008784 and
2204721.

\vspace{-0.3cm}
\setlength{\bibsep}{1pt} 
\begingroup
\small 
\bibliographystyle{elsarticle-num} 
\bibliography{references}

@article{hu2025security,
  title={Security analysis and adaptive false data injection against multi-sensor fusion localization for autonomous driving},
  author={Hu, Linqing and Zhang, Junqi and Zhang, Jie and Cheng, Shaoyin and Wang, Yuyi and Zhang, Weiming and Yu, Nenghai},
  journal={Information Fusion},
  volume={117},
  pages={102822},
  year={2025},
  publisher={Elsevier}
}

@inproceedings{yang2016exploit,
  title={Exploit all the layers: Fast and accurate cnn object detector with scale dependent pooling and cascaded rejection classifiers},
  author={Yang, Fan and Choi, Wongun and Lin, Yuanqing},
  booktitle={Proceedings of the IEEE conference on computer vision and pattern recognition},
  pages={2129--2137},
  year={2016}
}

@article{tian2025uavs,
  title={UAVs meet LLMs: Overviews and perspectives towards agentic low-altitude mobility},
  author={Tian, Yonglin and Lin, Fei and Li, Yiduo and Zhang, Tengchao and Zhang, Qiyao and Fu, Xuan and Huang, Jun and Dai, Xingyuan and Wang, Yutong and Tian, Chunwei and others},
  journal={Information Fusion},
  volume={122},
  pages={103158},
  year={2025},
  publisher={Elsevier}
}

@article{du2025advancements,
  title={Advancements in perception system with multi-sensor fusion for embodied agents},
  author={Du, Hao and Ren, Lu and Wang, Yuanda and Cao, Xiang and Sun, Changyin},
  journal={Information Fusion},
  volume={117},
  pages={102859},
  year={2025},
  publisher={Elsevier}
}

@article{wang2025uncertainty,
  title={Uncertainty quantification for safe and reliable autonomous vehicles: A review of methods and applications},
  author={Wang, Ke and Shen, Chongqiang and Li, Xingcan and Lu, Jianbo},
  journal={IEEE Transactions on Intelligent Transportation Systems},
  year={2025},
  publisher={IEEE}
}

@article{chen2025hierarchical,
  title={Hierarchical deep reinforcement learning based multi-agent game control for energy consumption and traffic efficiency improving of autonomous vehicles},
  author={Chen, Xiang and Wang, Xu and Zhao, Wanzhong and Wang, Chunyan and Cheng, Shuo and Luan, Zhongkai},
  journal={Energy},
  volume={323},
  pages={135669},
  year={2025},
  publisher={Elsevier}
}

@article{zha2025real,
  title={Real-time localization and navigation method for autonomous vehicles based on multi-modal data fusion by integrating memory transformer and DDQN},
  author={Zha, Li and Gong, Chen and Lv, Kunfeng},
  journal={Image and Vision Computing},
  volume={156},
  pages={105484},
  year={2025},
  publisher={Elsevier}
}

@article{wang2025depth,
  title={An in-depth examination of SLAM methods: Challenges, advancements, and applications in complex scenes for autonomous driving},
  author={Wang, Ke and Guo, Juwei and Chen, Kai and Lu, Jianbo},
  journal={IEEE Transactions on Intelligent Transportation Systems},
  year={2025},
  publisher={IEEE}
}

@article{zha2025heterogeneous,
  title={Heterogeneous Multiscale Cooperative Perception for Connected Autonomous Vehicles via V2X Interaction},
  author={Zha, Yuanyuan and Shangguan, Wei and Chen, Junjie and Chai, Linguo and Qiu, Weizhi and L{\'o}pez, Antonio M},
  journal={IEEE Internet of Things Journal},
  year={2025},
  publisher={IEEE}
}

@inproceedings{rs2025embedded,
  title={Embedded Large Language Models for Enhanced Human-Machine Interface in Autonomous Vehicles},
  author={RS, Sandhya Devi and Varshni, S Deva},
  booktitle={2025 International Conference on Multi-Agent Systems for Collaborative Intelligence (ICMSCI)},
  pages={1143--1150},
  year={2025},
  organization={IEEE}
}

@inproceedings{kumar2025improving,
  title={Improving Faster R-CNN for Vehicle Detection under Varying Conditions with Domain Adaptation Technique},
  author={Kumar, Hemant and Mamoria, Pushpa and Dewangan, Deepak Kumar},
  booktitle={2025 Fourth International Conference on Power, Control and Computing Technologies (ICPC2T)},
  pages={1--6},
  year={2025},
  organization={IEEE}
}

@inproceedings{shrivastava2025ai,
  title={AI-Powered Object Detection for Autonomous Vehicles: A Comparative Study of Machine Learning Models},
  author={Shrivastava, Anurag and Kansal, Vipashi and Nagpal, Amandeep and Dixit, Krishna Kant and Rajkumar, K Varada and others},
  booktitle={2025 International Conference on Computational, Communication and Information Technology (ICCCIT)},
  pages={612--617},
  year={2025},
  organization={IEEE}
}

@article{subhedar2025insights,
  title={Insights of semantic segmentation using the DeepLab architecture for autonomous driving},
  author={Subhedar, Javed and Bachute, Mrinal R},
  journal={MethodsX},
  pages={103387},
  year={2025},
  publisher={Elsevier}
}

@inproceedings{chen2025ranging,
  title={Ranging research on Telematics based on Mask R-CNN dual eye stereo vision ranging algorithm},
  author={Chen, Shuangshuang and Li, Xiaojie and Wang, Kai and Sun, Jibin and Yang, Bo},
  booktitle={The International Conference Optoelectronic Information and Optical Engineering (OIOE2024)},
  volume={13513},
  pages={884--889},
  year={2025},
  organization={SPIE}
}

@inproceedings{praveen2025autonomous,
  title={Autonomous Vehicle Navigation Systems: Machine Learning for Real-Time Traffic Prediction},
  author={Praveen, RVS and Hundekari, Sheela and Parida, Prasanta and Mittal, Tanusha and Sehgal, Archana and Bhavana, Munugapati},
  booktitle={2025 International Conference on Computational, Communication and Information Technology (ICCCIT)},
  pages={809--813},
  year={2025},
  organization={IEEE}
}

@article{mohammadi2025detection,
  title={Detection of Multiple Small Biased GPS Spoofing Attacks on Autonomous Vehicles Using Time Series Analysis},
  author={Mohammadi, Ahmad and Ahmari, Reza and Hemmati, Vahid and Owusu-Ambrose, Frederick and Mahmoud, Mahmoud Nabil and Kebria, Parham and Homaifar, Abdollah},
  journal={IEEE Open Journal of Vehicular Technology},
  year={2025},
  publisher={IEEE}
}

@inproceedings{guo2024vlm,
  title={Vlm-auto: Vlm-based autonomous driving assistant with human-like behavior and understanding for complex road scenes},
  author={Guo, Ziang and Yagudin, Zakhar and Lykov, Artem and Konenkov, Mikhail and Tsetserukou, Dzmitry},
  booktitle={2024 2nd International Conference on Foundation and Large Language Models (FLLM)},
  pages={501--507},
  year={2024},
  organization={IEEE}
}

@inproceedings{geiger2012we,
  title={Are we ready for autonomous driving? the kitti vision benchmark suite},
  author={Geiger, Andreas and Lenz, Philip and Urtasun, Raquel},
  booktitle={2012 IEEE conference on computer vision and pattern recognition},
  pages={3354--3361},
  year={2012},
  organization={IEEE}
}

@inproceedings{cordts2016cityscapes,
  title={The cityscapes dataset for semantic urban scene understanding},
  author={Cordts, Marius and Omran, Mohamed and Ramos, Sebastian and Rehfeld, Timo and Enzweiler, Markus and Benenson, Rodrigo and Franke, Uwe and Roth, Stefan and Schiele, Bernt},
  booktitle={Proceedings of the IEEE conference on computer vision and pattern recognition},
  pages={3213--3223},
  year={2016}
}

@inproceedings{huang2018apolloscape,
  title={The apolloscape dataset for autonomous driving},
  author={Huang, Xinyu and Cheng, Xinjing and Geng, Qichuan and Cao, Binbin and Zhou, Dingfu and Wang, Peng and Lin, Yuanqing and Yang, Ruigang},
  booktitle={Proceedings of the IEEE conference on computer vision and pattern recognition workshops},
  pages={954--960},
  year={2018}
}

@inproceedings{caesar2020nuscenes,
  title={nuscenes: A multimodal dataset for autonomous driving},
  author={Caesar, Holger and Bankiti, Varun and Lang, Alex H and Vora, Sourabh and Liong, Venice Erin and Xu, Qiang and Krishnan, Anush and Pan, Yu and Baldan, Giancarlo and Beijbom, Oscar},
  booktitle={Proceedings of the IEEE/CVF conference on computer vision and pattern recognition},
  pages={11621--11631},
  year={2020}
}

@inproceedings{chang2019argoverse,
  title={Argoverse: 3d tracking and forecasting with rich maps},
  author={Chang, Ming-Fang and Lambert, John and Sangkloy, Patsorn and Singh, Jagjeet and Bak, Slawomir and Hartnett, Andrew and Wang, De and Carr, Peter and Lucey, Simon and Ramanan, Deva and others},
  booktitle={Proceedings of the IEEE/CVF conference on computer vision and pattern recognition},
  pages={8748--8757},
  year={2019}
}

@inproceedings{xue2019blvd,
  title={BLVD: Building a large-scale 5D semantics benchmark for autonomous driving},
  author={Xue, Jianru and Fang, Jianwu and Li, Tao and Zhang, Bohua and Zhang, Pu and Ye, Zhen and Dou, Jian},
  booktitle={2019 International Conference on Robotics and Automation (ICRA)},
  pages={6685--6691},
  year={2019},
  organization={IEEE}
}

@article{schafer2018commute,
  title={A commute in data: The comma2k19 dataset},
  author={Schafer, Harald and Santana, Eder and Haden, Andrew and Biasini, Riccardo},
  journal={arXiv preprint arXiv:1812.05752},
  year={2018}
}

@inproceedings{sun2020scalability,
  title={Scalability in perception for autonomous driving: Waymo open dataset},
  author={Sun, Pei and Kretzschmar, Henrik and Dotiwalla, Xerxes and Chouard, Aurelien and Patnaik, Vijaysai and Tsui, Paul and Guo, James and Zhou, Yin and Chai, Yuning and Caine, Benjamin and others},
  booktitle={Proceedings of the IEEE/CVF conference on computer vision and pattern recognition},
  pages={2446--2454},
  year={2020}
}

@inproceedings{pham20203d,
  title={A* 3d dataset: Towards autonomous driving in challenging environments},
  author={Pham, Quang-Hieu and Sevestre, Pierre and Pahwa, Ramanpreet Singh and Zhan, Huijing and Pang, Chun Ho and Chen, Yuda and Mustafa, Armin and Chandrasekhar, Vijay and Lin, Jie},
  booktitle={2020 IEEE International Conference on Robotics and Automation (ICRA)},
  pages={2267--2273},
  year={2020},
  organization={IEEE}
}

@inproceedings{inDdataset,
    title={The inD Dataset: A Drone Dataset of Naturalistic Road User Trajectories at German Intersections},
    author={Bock, Julian and Krajewski, Robert and Moers, Tobias and Runde, Steffen and Vater, Lennart and Eckstein, Lutz},
    booktitle={2020 IEEE Intelligent Vehicles Symposium (IV)},
    pages={1929-1934},
    year={2020},
    doi={10.1109/IV47402.2020.9304839}
}

@inproceedings{exiDdataset,
    title={The exiD Dataset: A Real-World Trajectory Dataset of Highly Interactive Highway Scenarios in Germany},
    author={Moers, Tobias and Vater, Lennart and Krajewski, Robert and Bock, Julian and Zlocki, Adrian and Eckstein, Lutz},
    booktitle={2022 IEEE Intelligent Vehicles Symposium (IV)},
    pages={958-964},
    year={2022},
    doi={10.1109/IV51971.2022.9827305}
}

@article{cao2025kptr,
  title={KPTr: Key point transformer for LiDAR-based 3D object detection},
  author={Cao, Jie and Peng, Yiqiang and Wei, Hongqian and Mo, Lingfan and Fan, Likang and Wang, Longfei},
  journal={Measurement},
  volume={242},
  pages={115820},
  year={2025},
  publisher={Elsevier}
}

@misc{roboflow_self_driving,
  author       = {Roboflow},
  title        = {Self-Driving Car Dataset},
  year         = {2025},
  url          = {https://public.roboflow.com/object-detection/self-driving-car},
  note         = {Accessed: 2025-02-28}
}

@article{zheng2024omnihd,
  title={OmniHD-Scenes: A next-generation multimodal dataset for autonomous driving},
  author={Zheng, Lianqing and Yang, Long and Lin, Qunshu and Ai, Wenjin and Liu, Minghao and Lu, Shouyi and Liu, Jianan and Ren, Hongze and Mo, Jingyue and Bai, Xiaokai and others},
  journal={arXiv preprint arXiv:2412.10734},
  year={2024}
}

@article{yang2024v2x,
  title={V2X-Radar: A Multi-modal Dataset with 4D Radar for Cooperative Perception},
  author={Yang, Lei and Zhang, Xinyu and Li, Jun and Wang, Chen and Song, Zhiying and Zhao, Tong and Song, Ziying and Wang, Li and Zhou, Mo and Shen, Yang and others},
  journal={arXiv preprint arXiv:2411.10962},
  year={2024}
}

@inproceedings{xu2023v2v4real,
  title={V2v4real: A real-world large-scale dataset for vehicle-to-vehicle cooperative perception},
  author={Xu, Runsheng and Xia, Xin and Li, Jinlong and Li, Hanzhao and Zhang, Shuo and Tu, Zhengzhong and Meng, Zonglin and Xiang, Hao and Dong, Xiaoyu and Song, Rui and others},
  booktitle={Proceedings of the IEEE/CVF Conference on Computer Vision and Pattern Recognition},
  pages={13712--13722},
  year={2023}
}

@inproceedings{kent2024msu,
  title={MSU-4S-The Michigan State University Four Seasons Dataset},
  author={Kent, Daniel and Alyaqoub, Mohammed and Lu, Xiaohu and Khatounabadi, Hamed and Sung, Kookjin and Scheller, Cole and Dalat, Alexander and bin Thabit, Asma and Whitley, Roberto and Radha, Hayder},
  booktitle={Proceedings of the IEEE/CVF Conference on Computer Vision and Pattern Recognition},
  pages={22658--22667},
  year={2024}
}

@inproceedings{li2024multiagent,
  title={Multiagent multitraversal multimodal self-driving: Open mars dataset},
  author={Li, Yiming and Li, Zhiheng and Chen, Nuo and Gong, Moonjun and Lyu, Zonglin and Wang, Zehong and Jiang, Peili and Feng, Chen},
  booktitle={Proceedings of the IEEE/CVF Conference on Computer Vision and Pattern Recognition},
  pages={22041--22051},
  year={2024}
}

@inproceedings{alibeigi2023zenseact,
  title={Zenseact open dataset: A large-scale and diverse multimodal dataset for autonomous driving},
  author={Alibeigi, Mina and Ljungbergh, William and Tonderski, Adam and Hess, Georg and Lilja, Adam and Lindstr{\"o}m, Carl and Motorniuk, Daria and Fu, Junsheng and Widahl, Jenny and Petersson, Christoffer},
  booktitle={Proceedings of the IEEE/CVF International Conference on Computer Vision},
  pages={20178--20188},
  year={2023}
}

@inproceedings{diaz2022ithaca365,
  title={Ithaca365: Dataset and driving perception under repeated and challenging weather conditions},
  author={Diaz-Ruiz, Carlos A and Xia, Youya and You, Yurong and Nino, Jose and Chen, Junan and Monica, Josephine and Chen, Xiangyu and Luo, Katie and Wang, Yan and Emond, Marc and others},
  booktitle={Proceedings of the IEEE/CVF Conference on Computer Vision and Pattern Recognition},
  pages={21383--21392},
  year={2022}
}

@article{mao2021one,
  title={One million scenes for autonomous driving: Once dataset},
  author={Mao, Jiageng and Niu, Minzhe and Jiang, Chenhan and Liang, Hanxue and Chen, Jingheng and Liang, Xiaodan and Li, Yamin and Ye, Chaoqiang and Zhang, Wei and Li, Zhenguo and others},
  journal={arXiv preprint arXiv:2106.11037},
  year={2021}
}

@inproceedings{deziel2021pixset,
  title={Pixset: An opportunity for 3d computer vision to go beyond point clouds with a full-waveform lidar dataset},
  author={D{\'e}ziel, Jean--Luc and Merriaux, Pierre and Tremblay, Francis and Lessard, Dave and Plourde, Dominique and Stanguennec, Julien and Goulet, Pierre and Olivier, Pierre},
  booktitle={2021 ieee international intelligent transportation systems conference (itsc)},
  pages={2987--2993},
  year={2021},
  organization={IEEE}
}

@inproceedings{xiao2021pandaset,
  title={Pandaset: Advanced sensor suite dataset for autonomous driving},
  author={Xiao, Pengchuan and Shao, Zhenlei and Hao, Steven and Zhang, Zishuo and Chai, Xiaolin and Jiao, Judy and Li, Zesong and Wu, Jian and Sun, Kai and Jiang, Kun and others},
  booktitle={2021 IEEE international intelligent transportation systems conference (ITSC)},
  pages={3095--3101},
  year={2021},
  organization={IEEE}
}

@article{geyer2020a2d2,
  title={A2d2: Audi autonomous driving dataset},
  author={Geyer, Jakob and Kassahun, Yohannes and Mahmudi, Mentar and Ricou, Xavier and Durgesh, Rupesh and Chung, Andrew S and Hauswald, Lorenz and Pham, Viet Hoang and M{\"u}hlegg, Maximilian and Dorn, Sebastian and others},
  journal={arXiv preprint arXiv:2004.06320},
  year={2020}
}

@article{yu2018bdd100k,
  title={Bdd100k: A diverse driving video database with scalable annotation tooling},
  author={Yu, Fisher and Xian, Wenqi and Chen, Yingying and Liu, Fangchen and Liao, Mike and Madhavan, Vashisht and Darrell, Trevor and others},
  journal={arXiv preprint arXiv:1805.04687},
  volume={2},
  number={5},
  pages={6},
  year={2018}
}

@inproceedings{barnes2020oxford,
  title={The oxford radar robotcar dataset: A radar extension to the oxford robotcar dataset},
  author={Barnes, Dan and Gadd, Matthew and Murcutt, Paul and Newman, Paul and Posner, Ingmar},
  booktitle={2020 IEEE international conference on robotics and automation (ICRA)},
  pages={6433--6438},
  year={2020},
  organization={IEEE}
}

@inproceedings{tang2019cityflow,
  title={Cityflow: A city-scale benchmark for multi-target multi-camera vehicle tracking and re-identification},
  author={Tang, Zheng and Naphade, Milind and Liu, Ming-Yu and Yang, Xiaodong and Birchfield, Stan and Wang, Shuo and Kumar, Ratnesh and Anastasiu, David and Hwang, Jenq-Neng},
  booktitle={Proceedings of the IEEE/CVF conference on computer vision and pattern recognition},
  pages={8797--8806},
  year={2019}
}

@article{howe2021weakly,
  title={Weakly supervised training of monocular 3d object detectors using wide baseline multi-view traffic camera data},
  author={Howe, Matthew and Reid, Ian and Mackenzie, Jamie},
  journal={arXiv preprint arXiv:2110.10966},
  year={2021}
}

@article{zhan2019interaction,
  title={Interaction dataset: An international, adversarial and cooperative motion dataset in interactive driving scenarios with semantic maps},
  author={Zhan, Wei and Sun, Liting and Wang, Di and Shi, Haojie and Clausse, Aubrey and Naumann, Maximilian and Kummerle, Julius and Konigshof, Hendrik and Stiller, Christoph and de La Fortelle, Arnaud and others},
  journal={arXiv preprint arXiv:1910.03088},
  year={2019}
}

@inproceedings{cress2022a9,
  title={A9-dataset: Multi-sensor infrastructure-based dataset for mobility research},
  author={Cre{\ss}, Christian and Zimmer, Walter and Strand, Leah and Fortkord, Maximilian and Dai, Siyi and Lakshminarasimhan, Venkatnarayanan and Knoll, Alois},
  booktitle={2022 IEEE Intelligent Vehicles Symposium (IV)},
  pages={965--970},
  year={2022},
  organization={IEEE}
}

@inproceedings{wang2022ips300+,
  title={Ips300+: a challenging multi-modal data sets for intersection perception system},
  author={Wang, Huanan and Zhang, Xinyu and Li, Zhiwei and Li, Jun and Wang, Kun and Lei, Zhu and Haibing, Ren},
  booktitle={2022 International Conference on Robotics and Automation (ICRA)},
  pages={2539--2545},
  year={2022},
  organization={IEEE}
}

@inproceedings{ye2022rope3d,
  title={Rope3d: The roadside perception dataset for autonomous driving and monocular 3d object detection task},
  author={Ye, Xiaoqing and Shu, Mao and Li, Hanyu and Shi, Yifeng and Li, Yingying and Wang, Guangjie and Tan, Xiao and Ding, Errui},
  booktitle={Proceedings of the IEEE/CVF Conference on Computer Vision and Pattern Recognition},
  pages={21341--21350},
  year={2022}
}

@inproceedings{busch2022lumpi,
  title={Lumpi: The leibniz university multi-perspective intersection dataset},
  author={Busch, Steffen and Koetsier, Christian and Axmann, Jeldrik and Brenner, Claus},
  booktitle={2022 IEEE Intelligent Vehicles Symposium (IV)},
  pages={1127--1134},
  year={2022},
  organization={IEEE}
}

@inproceedings{zimmer2023tumtraf,
  title={Tumtraf intersection dataset: All you need for urban 3d camera-lidar roadside perception},
  author={Zimmer, Walter and Cre{\ss}, Christian and Nguyen, Huu Tung and Knoll, Alois C},
  booktitle={2023 IEEE 26th International Conference on Intelligent Transportation Systems (ITSC)},
  pages={1030--1037},
  year={2023},
  organization={IEEE}
}

@inproceedings{zhu2024roscenes,
  title={Roscenes: A large-scale multi-view 3d dataset for roadside perception},
  author={Zhu, Xiaosu and Sheng, Hualian and Cai, Sijia and Deng, Bing and Yang, Shaopeng and Liang, Qiao and Chen, Ken and Gao, Lianli and Song, Jingkuan and Ye, Jieping},
  booktitle={European Conference on Computer Vision},
  pages={331--347},
  year={2024},
  organization={Springer}
}

@article{ishaq2025drivelmm,
  title={DriveLMM-o1: A Step-by-Step Reasoning Dataset and Large Multimodal Model for Driving Scenario Understanding},
  author={Ishaq, Ayesha and Lahoud, Jean and More, Ketan and Thawakar, Omkar and Thawkar, Ritesh and Dissanayake, Dinura and Ahsan, Noor and Li, Yuhao and Khan, Fahad Shahbaz and Cholakkal, Hisham and others},
  journal={arXiv preprint arXiv:2503.10621},
  year={2025}
}

@inproceedings{sima2024drivelm,
  title={Drivelm: Driving with graph visual question answering},
  author={Sima, Chonghao and Renz, Katrin and Chitta, Kashyap and Chen, Li and Zhang, Hanxue and Xie, Chengen and Bei{\ss}wenger, Jens and Luo, Ping and Geiger, Andreas and Li, Hongyang},
  booktitle={European Conference on Computer Vision},
  pages={256--274},
  year={2024},
  organization={Springer}
}

@article{chiu2025v2v,
  title={V2v-llm: Vehicle-to-vehicle cooperative autonomous driving with multi-modal large language models},
  author={Chiu, Hsu-kuang and Hachiuma, Ryo and Wang, Chien-Yi and Smith, Stephen F and Wang, Yu-Chiang Frank and Chen, Min-Hung},
  journal={arXiv preprint arXiv:2502.09980},
  year={2025}
}

@article{wang2025omnidrive,
  title={OmniDrive: A Holistic Vision-Language Dataset for Autonomous Driving with Counterfactual Reasoning},
  author={Wang, Shihao and Yu, Zhiding and Jiang, Xiaohui and Lan, Shiyi and Shi, Min and Chang, Nadine and Kautz, Jan and Li, Ying and Alvarez, Jose M},
  journal={arXiv preprint arXiv:2504.04348},
  year={2025}
}

@article{tian2024tokenize,
  title={Tokenize the world into object-level knowledge to address long-tail events in autonomous driving},
  author={Tian, Ran and Li, Boyi and Weng, Xinshuo and Chen, Yuxiao and Schmerling, Edward and Wang, Yue and Ivanovic, Boris and Pavone, Marco},
  journal={arXiv preprint arXiv:2407.00959},
  year={2024}
}

@inproceedings{cao2024maplm,
  title={Maplm: A real-world large-scale vision-language benchmark for map and traffic scene understanding},
  author={Cao, Xu and Zhou, Tong and Ma, Yunsheng and Ye, Wenqian and Cui, Can and Tang, Kun and Cao, Zhipeng and Liang, Kaizhao and Wang, Ziran and Rehg, James M and others},
  booktitle={Proceedings of the IEEE/CVF Conference on Computer Vision and Pattern Recognition},
  pages={21819--21830},
  year={2024}
}

@inproceedings{marcu2024lingoqa,
  title={LingoQA: Visual question answering for autonomous driving},
  author={Marcu, Ana-Maria and Chen, Long and H{\"u}nermann, Jan and Karnsund, Alice and Hanotte, Benoit and Chidananda, Prajwal and Nair, Saurabh and Badrinarayanan, Vijay and Kendall, Alex and Shotton, Jamie and others},
  booktitle={European Conference on Computer Vision},
  pages={252--269},
  year={2024},
  organization={Springer}
}

@inproceedings{qian2024nuscenes,
  title={Nuscenes-qa: A multi-modal visual question answering benchmark for autonomous driving scenario},
  author={Qian, Tianwen and Chen, Jingjing and Zhuo, Linhai and Jiao, Yang and Jiang, Yu-Gang},
  booktitle={Proceedings of the AAAI Conference on Artificial Intelligence},
  volume={38},
  number={5},
  pages={4542--4550},
  year={2024}
}

@inproceedings{inoue2024nuscenes,
  title={Nuscenes-mqa: Integrated evaluation of captions and qa for autonomous driving datasets using markup annotations},
  author={Inoue, Yuichi and Yada, Yuki and Tanahashi, Kotaro and Yamaguchi, Yu},
  booktitle={Proceedings of the IEEE/CVF Winter Conference on Applications of Computer Vision},
  pages={930--938},
  year={2024}
}

@inproceedings{chen2025automated,
  title={Automated evaluation of large vision-language models on self-driving corner cases},
  author={Chen, Kai and Li, Yanze and Zhang, Wenhua and Liu, Yanxin and Li, Pengxiang and Gao, Ruiyuan and Hong, Lanqing and Tian, Meng and Zhao, Xinhai and Li, Zhenguo and others},
  booktitle={2025 IEEE/CVF Winter Conference on Applications of Computer Vision (WACV)},
  pages={7817--7826},
  year={2025},
  organization={IEEE}
}

@inproceedings{ding2024holistic,
  title={Holistic autonomous driving understanding by bird's-eye-view injected multi-modal large models},
  author={Ding, Xinpeng and Han, Jianhua and Xu, Hang and Liang, Xiaodan and Zhang, Wei and Li, Xiaomeng},
  booktitle={Proceedings of the IEEE/CVF Conference on Computer Vision and Pattern Recognition},
  pages={13668--13677},
  year={2024}
}

@inproceedings{malla2023drama,
  title={Drama: Joint risk localization and captioning in driving},
  author={Malla, Srikanth and Choi, Chiho and Dwivedi, Isht and Choi, Joon Hee and Li, Jiachen},
  booktitle={Proceedings of the IEEE/CVF winter conference on applications of computer vision},
  pages={1043--1052},
  year={2023}
}

@inproceedings{chen2019cooper,
  title={Cooper: Cooperative perception for connected autonomous vehicles based on 3d point clouds},
  author={Chen, Qi and Tang, Sihai and Yang, Qing and Fu, Song},
  booktitle={2019 IEEE 39th International Conference on Distributed Computing Systems (ICDCS)},
  pages={514--524},
  year={2019},
  organization={IEEE}
}

@article{yuan2021comap,
  title={Comap: A synthetic dataset for collective multi-agent perception of autonomous driving},
  author={Yuan, Yunshuang and Sester, Monika},
  journal={The International Archives of the Photogrammetry, Remote Sensing and Spatial Information Sciences},
  volume={43},
  pages={255--263},
  year={2021},
  publisher={Copernicus Publications G{\"o}ttingen, Germany}
}

@article{arnold2021fast,
  title={Fast and robust registration of partially overlapping point clouds},
  author={Arnold, Eduardo and Mozaffari, Sajjad and Dianati, Mehrdad},
  journal={IEEE Robotics and Automation Letters},
  volume={7},
  number={2},
  pages={1502--1509},
  year={2021},
  publisher={IEEE}
}

@inproceedings{xu2022opv2v,
  title={Opv2v: An open benchmark dataset and fusion pipeline for perception with vehicle-to-vehicle communication},
  author={Xu, Runsheng and Xiang, Hao and Xia, Xin and Han, Xu and Li, Jinlong and Ma, Jiaqi},
  booktitle={2022 International Conference on Robotics and Automation (ICRA)},
  pages={2583--2589},
  year={2022},
  organization={IEEE}
}

@inproceedings{hu2023collaboration,
  title={Collaboration helps camera overtake lidar in 3d detection},
  author={Hu, Yue and Lu, Yifan and Xu, Runsheng and Xie, Weidi and Chen, Siheng and Wang, Yanfeng},
  booktitle={Proceedings of the IEEE/CVF Conference on Computer Vision and Pattern Recognition},
  pages={9243--9252},
  year={2023}
}

@article{wei2023asynchrony,
  title={Asynchrony-robust collaborative perception via bird's eye view flow},
  author={Wei, Sizhe and Wei, Yuxi and Hu, Yue and Lu, Yifan and Zhong, Yiqi and Chen, Siheng and Zhang, Ya},
  journal={Advances in Neural Information Processing Systems},
  volume={36},
  pages={28462--28477},
  year={2023}
}

@inproceedings{axmann2023lucoop,
  title={LUCOOP: Leibniz university cooperative perception and urban navigation dataset},
  author={Axmann, Jeldrik and Moftizadeh, Rozhin and Su, Jingyao and Tennstedt, Benjamin and Zou, Qianqian and Yuan, Yunshuang and Ernst, Dominik and Alkhatib, Hamza and Brenner, Claus and Sch{\"o}n, Steffen},
  booktitle={2023 IEEE Intelligent Vehicles Symposium (IV)},
  pages={1--8},
  year={2023},
  organization={IEEE}
}

@article{lu2024extensible,
  title={An extensible framework for open heterogeneous collaborative perception},
  author={Lu, Yifan and Hu, Yue and Zhong, Yiqi and Wang, Dequan and Wang, Yanfeng and Chen, Siheng},
  journal={arXiv preprint arXiv:2401.13964},
  year={2024}
}

@article{hu2022where2comm,
  title={Where2comm: Communication-efficient collaborative perception via spatial confidence maps},
  author={Hu, Yue and Fang, Shaoheng and Lei, Zixing and Zhong, Yiqi and Chen, Siheng},
  journal={Advances in neural information processing systems},
  volume={35},
  pages={4874--4886},
  year={2022}
}

@article{arnold2020cooperative,
  title={Cooperative perception for 3D object detection in driving scenarios using infrastructure sensors},
  author={Arnold, Eduardo and Dianati, Mehrdad and de Temple, Robert and Fallah, Saber},
  journal={IEEE Transactions on Intelligent Transportation Systems},
  volume={23},
  number={3},
  pages={1852--1864},
  year={2020},
  publisher={IEEE}
}

@inproceedings{bai2022pillargrid,
  title={Pillargrid: Deep learning-based cooperative perception for 3d object detection from onboard-roadside lidar},
  author={Bai, Zhengwei and Wu, Guoyuan and Barth, Matthew J and Liu, Yongkang and Sisbot, Emrah Akin and Oguchi, Kentaro},
  booktitle={2022 IEEE 25th International Conference on Intelligent Transportation Systems (ITSC)},
  pages={1743--1749},
  year={2022},
  organization={IEEE}
}

@inproceedings{yu2022dair,
  title={Dair-v2x: A large-scale dataset for vehicle-infrastructure cooperative 3d object detection},
  author={Yu, Haibao and Luo, Yizhen and Shu, Mao and Huo, Yiyi and Yang, Zebang and Shi, Yifeng and Guo, Zhenglong and Li, Hanyu and Hu, Xing and Yuan, Jirui and others},
  booktitle={Proceedings of the IEEE/CVF Conference on Computer Vision and Pattern Recognition},
  pages={21361--21370},
  year={2022}
}

@inproceedings{yu2023v2x,
  title={V2x-seq: A large-scale sequential dataset for vehicle-infrastructure cooperative perception and forecasting},
  author={Yu, Haibao and Yang, Wenxian and Ruan, Hongzhi and Yang, Zhenwei and Tang, Yingjuan and Gao, Xu and Hao, Xin and Shi, Yifeng and Pan, Yifeng and Sun, Ning and others},
  booktitle={Proceedings of the IEEE/CVF Conference on Computer Vision and Pattern Recognition},
  pages={5486--5495},
  year={2023}
}

@inproceedings{ma2024holovic,
  title={HoloVIC: Large-scale dataset and benchmark for multi-sensor holographic intersection and vehicle-infrastructure cooperative},
  author={Ma, Cong and Qiao, Lei and Zhu, Chengkai and Liu, Kai and Kong, Zelong and Li, Qing and Zhou, Xueqi and Kan, Yuheng and Wu, Wei},
  booktitle={Proceedings of the IEEE/CVF Conference on Computer Vision and Pattern Recognition},
  pages={22129--22138},
  year={2024}
}

@inproceedings{zhu2024otvic,
  title={OTVIC: A Dataset with Online Transmission for Vehicle-to-Infrastructure Cooperative 3D Object Detection},
  author={Zhu, He and Wang, Yunkai and Kong, Quyu and Wei, Yufei and Xia, Xunlong and Deng, Bing and Xiong, Rong and Wang, Yue},
  booktitle={2024 IEEE/RSJ International Conference on Intelligent Robots and Systems (IROS)},
  pages={10732--10739},
  year={2024},
  organization={IEEE}
}

@article{wang2024dair,
  title={Dair-v2xreid: A new real-world vehicle-infrastructure cooperative re-id dataset and cross-shot feature aggregation network perception method},
  author={Wang, Hai and Niu, Yaqing and Chen, Long and Li, Yicheng and Sotelo, Miguel Angel and Li, Zhixiong and Cai, Yingfeng},
  journal={IEEE Transactions on Intelligent Transportation Systems},
  year={2024},
  publisher={IEEE}
}

@inproceedings{zimmer2024tumtraf,
  title={Tumtraf v2x cooperative perception dataset},
  author={Zimmer, Walter and Wardana, Gerhard Arya and Sritharan, Suren and Zhou, Xingcheng and Song, Rui and Knoll, Alois C},
  booktitle={Proceedings of the IEEE/CVF conference on computer vision and pattern recognition},
  pages={22668--22677},
  year={2024}
}

@inproceedings{liu2024v2x,
  title={V2X-DSI: A Density-Sensitive Infrastructure LiDAR Benchmark for Economic Vehicle-to-Everything Cooperative Perception},
  author={Liu, Xinyu and Li, Baolu and Xu, Runsheng and Ma, Jiaqi and Li, Xiaopeng and Li, Jinlong and Yu, Hongkai},
  booktitle={2024 IEEE Intelligent Vehicles Symposium (IV)},
  pages={490--495},
  year={2024},
  organization={IEEE}
}

@article{fan2025benchmark,
  title={A Benchmark for Vision-Centric HD Mapping by V2I Systems},
  author={Fan, Miao and Yu, Shanshan and Xu, Shengtong and Jiang, Kun and Xiong, Haoyi and Liu, Xiangzeng},
  journal={arXiv preprint arXiv:2503.23963},
  year={2025}
}

@article{li2022v2x,
  title={V2X-Sim: Multi-agent collaborative perception dataset and benchmark for autonomous driving},
  author={Li, Yiming and Ma, Dekun and An, Ziyan and Wang, Zixun and Zhong, Yiqi and Chen, Siheng and Feng, Chen},
  journal={IEEE Robotics and Automation Letters},
  volume={7},
  number={4},
  pages={10914--10921},
  year={2022},
  publisher={IEEE}
}

@inproceedings{xu2022v2x,
  title={V2x-vit: Vehicle-to-everything cooperative perception with vision transformer},
  author={Xu, Runsheng and Xiang, Hao and Tu, Zhengzhong and Xia, Xin and Yang, Ming-Hsuan and Ma, Jiaqi},
  booktitle={European conference on computer vision},
  pages={107--124},
  year={2022},
  organization={Springer}
}

@inproceedings{mao2022dolphins,
  title={Dolphins: Dataset for collaborative perception enabled harmonious and interconnected self-driving},
  author={Mao, Ruiqing and Guo, Jingyu and Jia, Yukuan and Sun, Yuxuan and Zhou, Sheng and Niu, Zhisheng},
  booktitle={Proceedings of the Asian Conference on Computer Vision},
  pages={4361--4377},
  year={2022}
}

@inproceedings{xiang2024v2x,
  title={V2x-real: a largs-scale dataset for vehicle-to-everything cooperative perception},
  author={Xiang, Hao and Zheng, Zhaoliang and Xia, Xin and Xu, Runsheng and Gao, Letian and Zhou, Zewei and Han, Xu and Ji, Xinkai and Li, Mingxi and Meng, Zonglin and others},
  booktitle={European Conference on Computer Vision},
  pages={455--470},
  year={2024},
  organization={Springer}
}

@inproceedings{wang2024deepaccident,
  title={Deepaccident: A motion and accident prediction benchmark for v2x autonomous driving},
  author={Wang, Tianqi and Kim, Sukmin and Wenxuan, Ji and Xie, Enze and Ge, Chongjian and Chen, Junsong and Li, Zhenguo and Luo, Ping},
  booktitle={Proceedings of the AAAI Conference on Artificial Intelligence},
  volume={38},
  number={6},
  pages={5599--5606},
  year={2024}
}

@article{karvat2024adver,
  title={Adver-City: Open-Source Multi-Modal Dataset for Collaborative Perception Under Adverse Weather Conditions},
  author={Karvat, Mateus and Givigi, Sidney},
  journal={arXiv preprint arXiv:2410.06380},
  year={2024}
}

@article{li2024multi,
  title={Multi-V2X: A Large Scale Multi-modal Multi-penetration-rate Dataset for Cooperative Perception},
  author={Li, Rongsong and Pei, Xin},
  journal={arXiv preprint arXiv:2409.04980},
  year={2024}
}

@article{chen2024whales,
  title={WHALES: A Multi-agent Scheduling Dataset for Enhanced Cooperation in Autonomous Driving},
  author={Chen, Siwei and Song, Ziyi and Zhou, Sheng and others},
  journal={arXiv preprint arXiv:2411.13340},
  year={2024}
}

@article{huang2024v2x,
  title={V2x-r: Cooperative lidar-4d radar fusion for 3d object detection with denoising diffusion},
  author={Huang, Xun and Wang, Jinlong and Xia, Qiming and Chen, Siheng and Yang, Bisheng and Li, Xin and Wang, Cheng and Wen, Chenglu},
  journal={arXiv preprint arXiv:2411.08402},
  year={2024}
}

@article{zhou2024v2xpnp,
  title={V2xpnp: Vehicle-to-everything spatio-temporal fusion for multi-agent perception and prediction},
  author={Zhou, Zewei and Xiang, Hao and Zheng, Zhaoliang and Zhao, Seth Z and Lei, Mingyue and Zhang, Yun and Cai, Tianhui and Liu, Xinyi and Liu, Johnson and Bajji, Maheswari and others},
  journal={arXiv preprint arXiv:2412.01812},
  year={2024}
}

@article{luo2025mixed,
  title={Mixed Signals: A Diverse Point Cloud Dataset for Heterogeneous LiDAR V2X Collaboration},
  author={Luo, Katie Z and Dao, Minh-Quan and Liu, Zhenzhen and Campbell, Mark and Chao, Wei-Lun and Weinberger, Kilian Q and Malis, Ezio and Fremont, Vincent and Hariharan, Bharath and Shan, Mao and others},
  journal={arXiv preprint arXiv:2502.14156},
  year={2025}
}

@article{gamerdinger2024scope,
  title={SCOPE: A Synthetic Multi-Modal Dataset for Collective Perception Including Physical-Correct Weather Conditions},
  author={Gamerdinger, J{\"o}rg and Teufel, Sven and Schulz, Patrick and Amann, Stephan and Kirchner, Jan-Patrick and Bringmann, Oliver},
  journal={arXiv preprint arXiv:2408.03065},
  year={2024}
}

@article{wang2024rcdn,
  title={RCDN: Towards Robust Camera-Insensitivity Collaborative Perception via Dynamic Feature-based 3D Neural Modeling},
  author={Wang, Tianhang and Lu, Fan and Zheng, Zehan and Li, Zhijun and Chen, Guang and others},
  journal={Advances in Neural Information Processing Systems},
  volume={37},
  pages={22350--22369},
  year={2024}
}

@article{xiang2025v2x,
  title={V2X-ReaLO: An Open Online Framework and Dataset for Cooperative Perception in Reality},
  author={Xiang, Hao and Zheng, Zhaoliang and Xia, Xin and Zhao, Seth Z and Gao, Letian and Zhou, Zewei and Cai, Tianhui and Zhang, Yun and Ma, Jiaqi},
  journal={arXiv preprint arXiv:2503.10034},
  year={2025}
}

@inproceedings{hao2024rcooper,
  title={Rcooper: A real-world large-scale dataset for roadside cooperative perception},
  author={Hao, Ruiyang and Fan, Siqi and Dai, Yingru and Zhang, Zhenlin and Li, Chenxi and Wang, Yuntian and Yu, Haibao and Yang, Wenxian and Yuan, Jirui and Nie, Zaiqing},
  booktitle={Proceedings of the IEEE/CVF Conference on Computer Vision and Pattern Recognition},
  pages={22347--22357},
  year={2024}
}

@article{zhang2024inscope,
  title={InScope: A New Real-world 3D Infrastructure-side Collaborative Perception Dataset for Open Traffic Scenarios},
  author={Zhang, Xiaofei and Li, Yining and Wang, Jinping and Qin, Xiangyi and Shen, Ying and Fan, Zhengping and Tan, Xiaojun},
  journal={arXiv preprint arXiv:2407.21581},
  year={2024}
}

@article{guo2019safe,
  title={Is it safe to drive? An overview of factors, metrics, and datasets for driveability assessment in autonomous driving},
  author={Guo, Junyao and Kurup, Unmesh and Shah, Mohak},
  journal={IEEE Transactions on Intelligent Transportation Systems},
  volume={21},
  number={8},
  pages={3135--3151},
  year={2019},
  publisher={IEEE}
}

@article{wang2025developments,
  title={Developments in 3D Object Detection for Autonomous Driving: A Review},
  author={Wang, Yu and Wang, Shaohua and Li, Yicheng and Liu, Mingchun},
  journal={IEEE Sensors Journal},
  year={2025},
  publisher={IEEE}
}

@article{wang2025survey,
  title={A Survey of the Multi-Sensor Fusion Object Detection Task in Autonomous Driving},
  author={Wang, Hai and Liu, Junhao and Dong, Haoran and Shao, Zheng},
  journal={Sensors},
  volume={25},
  number={9},
  pages={2794},
  year={2025},
  publisher={MDPI}
}

@article{wang2025review,
  title={A review of 3D object detection based on autonomous driving},
  author={Wang, Huijuan and Chen, Xinyue and Yuan, Quanbo and Liu, Peng},
  journal={The Visual Computer},
  volume={41},
  number={3},
  pages={1757--1775},
  year={2025},
  publisher={Springer}
}

@article{song2024robustness,
  title={Robustness-aware 3d object detection in autonomous driving: A review and outlook},
  author={Song, Ziying and Liu, Lin and Jia, Feiyang and Luo, Yadan and Jia, Caiyan and Zhang, Guoxin and Yang, Lei and Wang, Li},
  journal={IEEE Transactions on Intelligent Transportation Systems},
  year={2024},
  publisher={IEEE}
}

@article{alaba2024emerging,
  title={Emerging trends in autonomous vehicle perception: Multimodal fusion for 3D object detection},
  author={Alaba, Simegnew Yihunie and Gurbuz, Ali C and Ball, John E},
  journal={World Electric Vehicle Journal},
  volume={15},
  number={1},
  pages={20},
  year={2024},
  publisher={MDPI}
}

@article{zou2023object,
  title={Object detection in 20 years: A survey},
  author={Zou, Zhengxia and Chen, Keyan and Shi, Zhenwei and Guo, Yuhong and Ye, Jieping},
  journal={Proceedings of the IEEE},
  volume={111},
  number={3},
  pages={257--276},
  year={2023},
  publisher={IEEE}
}

@article{ma20233d,
  title={3d object detection from images for autonomous driving: a survey},
  author={Ma, Xinzhu and Ouyang, Wanli and Simonelli, Andrea and Ricci, Elisa},
  journal={IEEE Transactions on Pattern Analysis and Machine Intelligence},
  volume={46},
  number={5},
  pages={3537--3556},
  year={2023},
  publisher={IEEE}
}

@article{qian20223d,
  title={3D object detection for autonomous driving: A survey},
  author={Qian, Rui and Lai, Xin and Li, Xirong},
  journal={Pattern Recognition},
  volume={130},
  pages={108796},
  year={2022},
  publisher={Elsevier}
}

@article{cui2021deep,
  title={Deep learning for image and point cloud fusion in autonomous driving: A review},
  author={Cui, Yaodong and Chen, Ren and Chu, Wenbo and Chen, Long and Tian, Daxin and Li, Ying and Cao, Dongpu},
  journal={IEEE Transactions on Intelligent Transportation Systems},
  volume={23},
  number={2},
  pages={722--739},
  year={2021},
  publisher={IEEE}
}

@article{feng2020deep,
  title={Deep multi-modal object detection and semantic segmentation for autonomous driving: Datasets, methods, and challenges},
  author={Feng, Di and Haase-Sch{\"u}tz, Christian and Rosenbaum, Lars and Hertlein, Heinz and Glaeser, Claudius and Timm, Fabian and Wiesbeck, Werner and Dietmayer, Klaus},
  journal={IEEE Transactions on Intelligent Transportation Systems},
  volume={22},
  number={3},
  pages={1341--1360},
  year={2020},
  publisher={IEEE}
}

@article{comanici2025gemini,
  title={Gemini 2.5: Pushing the Frontier with Advanced Reasoning, Multimodality, Long Context, and Next Generation Agentic Capabilities},
  author={Comanici, Gheorghe and Bieber, Eric and Schaekermann, Mike and Pasupat, Ice and Sachdeva, Noveen and Dhillon, Inderjit and Blistein, Marcel and Ram, Ori and Zhang, Dan and Rosen, Evan and others},
  journal={arXiv preprint arXiv:2507.06261},
  year={2025}
}

@article{bae2024enhancing,
  title={Enhancing software code vulnerability detection using gpt-4o and claude-3.5 sonnet: A study on prompt engineering techniques},
  author={Bae, Jaehyeon and Kwon, Seoryeong and Myeong, Seunghwan},
  journal={Electronics},
  volume={13},
  number={13},
  pages={2657},
  year={2024},
  publisher={MDPI}
}

@misc{xAI,
  author={{xAI,}},
  title={Grok 4},
  url={https://x.ai/news/grok-4},
  urldate={2025-7-15},
  year={2025}
}

@article{hui2024qwen2,
  title={Qwen2. 5-coder technical report},
  author={Hui, Binyuan and Yang, Jian and Cui, Zeyu and Yang, Jiaxi and Liu, Dayiheng and Zhang, Lei and Liu, Tianyu and Zhang, Jiajun and Yu, Bowen and Lu, Keming and others},
  journal={arXiv preprint arXiv:2409.12186},
  year={2024}
}

@misc{OpenAI,
  author={{OpenAI,}},
  title={GPT-o3},
  url={https://openai.com/index/introducing-o3-and-o4-mini},
  urldate={2025-7-15},
  year={2025}
}

@article{dubey2024llama,
  title={The llama 3 herd of models},
  author={Dubey, Abhimanyu and Jauhri, Abhinav and Pandey, Abhinav and Kadian, Abhishek and Al-Dahle, Ahmad and Letman, Aiesha and Mathur, Akhil and Schelten, Alan and Yang, Amy and Fan, Angela and others},
  journal={arXiv e-prints},
  pages={arXiv--2407},
  year={2024}
}

@article{team2024gemini,
  title={Gemini 1.5: Unlocking multimodal understanding across millions of tokens of context},
  author={Team, Gemini and Georgiev, Petko and Lei, Ving Ian and Burnell, Ryan and Bai, Libin and Gulati, Anmol and Tanzer, Garrett and Vincent, Damien and Pan, Zhufeng and Wang, Shibo and others},
  journal={arXiv preprint arXiv:2403.05530},
  year={2024}
}

@misc{Anthropic ,
  author={{Anthropic,}},
  title={Claude 3 },
  url={https://www.anthropic.com/news/claude-3-family},
  urldate={2025-5-4},
  year={2025}
}

@article{lu2024deepseek,
  title={Deepseek-vl: towards real-world vision-language understanding},
  author={Lu, Haoyu and Liu, Wen and Zhang, Bo and Wang, Bingxuan and Dong, Kai and Liu, Bo and Sun, Jingxiang and Ren, Tongzheng and Li, Zhuoshu and Yang, Hao and others},
  journal={arXiv preprint arXiv:2403.05525},
  year={2024}
}

@article{hurst2024gpt,
  title={Gpt-4o system card},
  author={Hurst, Aaron and Lerer, Adam and Goucher, Adam P and Perelman, Adam and Ramesh, Aditya and Clark, Aidan and Ostrow, AJ and Welihinda, Akila and Hayes, Alan and Radford, Alec and others},
  journal={arXiv preprint arXiv:2410.21276},
  year={2024}
}

@article{abdin2024phi,
  title={Phi-4 technical report},
  author={Abdin, Marah and Aneja, Jyoti and Behl, Harkirat and Bubeck, S{\'e}bastien and Eldan, Ronen and Gunasekar, Suriya and Harrison, Michael and Hewett, Russell J and Javaheripi, Mojan and Kauffmann, Piero and others},
  journal={arXiv preprint arXiv:2412.08905},
  year={2024}
}

@article{touvron2023llama,
  title={Llama 2: Open foundation and fine-tuned chat models},
  author={Touvron, Hugo and Martin, Louis and Stone, Kevin and Albert, Peter and Almahairi, Amjad and Babaei, Yasmine and Bashlykov, Nikolay and Batra, Soumya and Bhargava, Prajjwal and Bhosale, Shruti and others},
  journal={arXiv preprint arXiv:2307.09288},
  year={2023}
}

@article{achiam2023gpt,
  title={Gpt-4 technical report},
  author={Achiam, Josh and Adler, Steven and Agarwal, Sandhini and Ahmad, Lama and Akkaya, Ilge and Aleman, Florencia Leoni and Almeida, Diogo and Altenschmidt, Janko and Altman, Sam and Anadkat, Shyamal and others},
  journal={arXiv preprint arXiv:2303.08774},
  year={2023}
}

@article{anil2023palm,
  title={Palm 2 technical report},
  author={Anil, Rohan and Dai, Andrew M and Firat, Orhan and Johnson, Melvin and Lepikhin, Dmitry and Passos, Alexandre and Shakeri, Siamak and Taropa, Emanuel and Bailey, Paige and Chen, Zhifeng and others},
  journal={arXiv preprint arXiv:2305.10403},
  year={2023}
}

@inproceedings{harabagiu2000falcon,
  title={FALCON: Boosting Knowledge for Answer Engines.},
  author={Harabagiu, Sanda M and Moldovan, Dan I and Pasca, Marius and Mihalcea, Rada and Surdeanu, Mihai and Bunescu, Razvan C and Girju, Roxana and Rus, Vasile and Morarescu, Paul},
  booktitle={TREC},
  volume={9},
  pages={479--488},
  year={2000}
}

@article{radford2018improving,
  title={Improving language understanding by generative pre-training},
  author={Radford, Alec and Narasimhan, Karthik and Salimans, Tim and Sutskever, Ilya and others},
  year={2018},
  publisher={San Francisco, CA, USA}
}

@article{lewis2019bart,
  title={BART: Denoising sequence-to-sequence pre-training for natural language generation, translation, and comprehension},
  author={Lewis, Mike and Liu, Yinhan and Goyal, Naman and Ghazvininejad, Marjan and Mohamed, Abdelrahman and Levy, Omer and Stoyanov, Ves and Zettlemoyer, Luke},
  journal={arXiv preprint arXiv:1910.13461},
  year={2019}
}

@inproceedings{devlin2019bert,
  title={Bert: Pre-training of deep bidirectional transformers for language understanding},
  author={Devlin, Jacob and Chang, Ming-Wei and Lee, Kenton and Toutanova, Kristina},
  booktitle={Proceedings of the 2019 conference of the North American chapter of the association for computational linguistics: human language technologies, volume 1 (long and short papers)},
  pages={4171--4186},
  year={2019}
}

@inproceedings{cheng2024yolo,
  title={Yolo-world: Real-time open-vocabulary object detection},
  author={Cheng, Tianheng and Song, Lin and Ge, Yixiao and Liu, Wenyu and Wang, Xinggang and Shan, Ying},
  booktitle={Proceedings of the IEEE/CVF conference on computer vision and pattern recognition},
  pages={16901--16911},
  year={2024}
}

@inproceedings{ma2025cake,
  title={Cake: Category aware knowledge extraction for open-vocabulary object detection},
  author={Ma, Shiyuan and Qian, Donglin and Ye, Kai and Zhang, Shengchuan},
  booktitle={Proceedings of the AAAI Conference on Artificial Intelligence},
  volume={39},
  number={6},
  pages={5982--5990},
  year={2025}
}

@article{zhou2025led,
  title={Led: Llm enhanced open-vocabulary object detection without human curated data generation},
  author={Zhou, Yang and Zhao, Shiyu and Chen, Yuxiao and Wang, Zhenting and Jin, Can and Metaxas, Dimitris N},
  journal={arXiv preprint arXiv:2503.13794},
  year={2025}
}

@inproceedings{liu2024grounding,
  title={Grounding dino: Marrying dino with grounded pre-training for open-set object detection},
  author={Liu, Shilong and Zeng, Zhaoyang and Ren, Tianhe and Li, Feng and Zhang, Hao and Yang, Jie and Jiang, Qing and Li, Chunyuan and Yang, Jianwei and Su, Hang and others},
  booktitle={European conference on computer vision},
  pages={38--55},
  year={2024},
  organization={Springer}
}

@inproceedings{kamath2021mdetr,
  title={Mdetr-modulated detection for end-to-end multi-modal understanding},
  author={Kamath, Aishwarya and Singh, Mannat and LeCun, Yann and Synnaeve, Gabriel and Misra, Ishan and Carion, Nicolas},
  booktitle={Proceedings of the IEEE/CVF international conference on computer vision},
  pages={1780--1790},
  year={2021}
}

@article{kuo2022f,
  title={F-vlm: Open-vocabulary object detection upon frozen vision and language models},
  author={Kuo, Weicheng and Cui, Yin and Gu, Xiuye and Piergiovanni, AJ and Angelova, Anelia},
  journal={arXiv preprint arXiv:2209.15639},
  year={2022}
}

@inproceedings{li2022grounded,
  title={Grounded language-image pre-training},
  author={Li, Liunian Harold and Zhang, Pengchuan and Zhang, Haotian and Yang, Jianwei and Li, Chunyuan and Zhong, Yiwu and Wang, Lijuan and Yuan, Lu and Zhang, Lei and Hwang, Jenq-Neng and others},
  booktitle={Proceedings of the IEEE/CVF conference on computer vision and pattern recognition},
  pages={10965--10975},
  year={2022}
}

@inproceedings{wu2023cora,
  title={Cora: Adapting clip for open-vocabulary detection with region prompting and anchor pre-matching},
  author={Wu, Xiaoshi and Zhu, Feng and Zhao, Rui and Li, Hongsheng},
  booktitle={Proceedings of the IEEE/CVF conference on computer vision and pattern recognition},
  pages={7031--7040},
  year={2023}
}

@InProceedings{Pu_2025_CVPR,
    author    = {Pu, Fanqi and Wang, Yifan and Deng, Jiru and Yang, Wenming},
    title     = {MonoDGP: Monocular 3D Object Detection with Decoupled-Query and Geometry-Error Priors},
    booktitle = {Proceedings of the IEEE/CVF Conference on Computer Vision and Pattern Recognition (CVPR)},
    month     = {June},
    year      = {2025},
    pages     = {6520-6530}
}

@InProceedings{Liu_2025_CVPR,
    author    = {Liu, Hou-I and Wu, Christine and Cheng, Jen-Hao and Chai, Wenhao and Wang, Shian-Yun and Liu, Gaowen and Latapie, Hugo and Wu, Jhih-Ciang and Hwang, Jenq-Neng and Shuai, Hong-Han and Cheng, Wen-Huang},
    title     = {MonoTAKD: Teaching Assistant Knowledge Distillation for Monocular 3D Object Detection},
    booktitle = {Proceedings of the IEEE/CVF Conference on Computer Vision and Pattern Recognition (CVPR)},
    month     = {June},
    year      = {2025},
    pages     = {22266-22275}
}

@inproceedings{wu2024fd3d,
  title={Fd3d: Exploiting foreground depth map for feature-supervised monocular 3d object detection},
  author={Wu, Zizhang and Gan, Yuanzhu and Wu, Yunzhe and Wang, Ruihao and Wang, Xiaoquan and Pu, Jian},
  booktitle={Proceedings of the AAAI Conference on Artificial Intelligence},
  volume={38},
  number={6},
  pages={6189--6197},
  year={2024}
}

@InProceedings{Yan_2024_CVPR,
    author    = {Yan, Longfei and Yan, Pei and Xiong, Shengzhou and Xiang, Xuanyu and Tan, Yihua},
    title     = {MonoCD: Monocular 3D Object Detection with Complementary Depths},
    booktitle = {Proceedings of the IEEE/CVF Conference on Computer Vision and Pattern Recognition (CVPR)},
    month     = {June},
    year      = {2024},
    pages     = {10248-10257}
}

@InProceedings{Ranasinghe_2024_CVPR,
    author    = {Ranasinghe, Yasiru and Hegde, Deepti and Patel, Vishal M.},
    title     = {MonoDiff: Monocular 3D Object Detection and Pose Estimation with Diffusion Models},
    booktitle = {Proceedings of the IEEE/CVF Conference on Computer Vision and Pattern Recognition (CVPR)},
    month     = {June},
    year      = {2024},
    pages     = {10659-10670}
}

@InProceedings{Zhang_2023_ICCV,
    author    = {Zhang, Renrui and Qiu, Han and Wang, Tai and Guo, Ziyu and Cui, Ziteng and Qiao, Yu and Li, Hongsheng and Gao, Peng},
    title     = {MonoDETR: Depth-guided Transformer for Monocular 3D Object Detection},
    booktitle = {Proceedings of the IEEE/CVF International Conference on Computer Vision (ICCV)},
    month     = {October},
    year      = {2023},
    pages     = {9155-9166}
}

@article{jinrang2023monouni,
  title={Monouni: A unified vehicle and infrastructure-side monocular 3d object detection network with sufficient depth clues},
  author={Jinrang, Jia and Li, Zhenjia and Shi, Yifeng},
  journal={Advances in Neural Information Processing Systems},
  volume={36},
  pages={11703--11715},
  year={2023}
}

@inproceedings{kumar2022deviant,
  title={Deviant: Depth equivariant network for monocular 3d object detection},
  author={Kumar, Abhinav and Brazil, Garrick and Corona, Enrique and Parchami, Armin and Liu, Xiaoming},
  booktitle={European Conference on Computer Vision},
  pages={664--683},
  year={2022},
  organization={Springer}
}

@article{wu2023monopgc,
  title={Monopgc: Monocular 3d object detection with pixel geometry contexts},
  author={Wu, Zizhang and Gan, Yuanzhu and Wang, Lei and Chen, Guilian and Pu, Jian},
  journal={arXiv preprint arXiv:2302.10549},
  year={2023}
}

@InProceedings{Zhou_2023_CVPR,
    author    = {Zhou, Yunsong and Zhu, Hongzi and Liu, Quan and Chang, Shan and Guo, Minyi},
    title     = {MonoATT: Online Monocular 3D Object Detection With Adaptive Token Transformer},
    booktitle = {Proceedings of the IEEE/CVF Conference on Computer Vision and Pattern Recognition (CVPR)},
    month     = {June},
    year      = {2023},
    pages     = {17493-17503}
}

@article{lu2024gupnet++,
  title={GUPNet++: Geometry uncertainty propagation network for monocular 3D object detection},
  author={Lu, Yan and Ma, Xinzhu and Yang, Lei and Zhang, Tianzhu and Liu, Yating and Chu, Qi and He, Tong and Li, Yonghui and Ouyang, Wanli},
  journal={IEEE Transactions on Pattern Analysis and Machine Intelligence},
  year={2024},
  publisher={IEEE}
}

@inproceedings{li2022densely,
  title={Densely constrained depth estimator for monocular 3d object detection},
  author={Li, Yingyan and Chen, Yuntao and He, Jiawei and Zhang, Zhaoxiang},
  booktitle={European Conference on Computer Vision},
  pages={718--734},
  year={2022},
  organization={Springer}
}

@InProceedings{Li_2022_CVPR,
    author    = {Li, Zhuoling and Qu, Zhan and Zhou, Yang and Liu, Jianzhuang and Wang, Haoqian and Jiang, Lihui},
    title     = {Diversity Matters: Fully Exploiting Depth Clues for Reliable Monocular 3D Object Detection},
    booktitle = {Proceedings of the IEEE/CVF Conference on Computer Vision and Pattern Recognition (CVPR)},
    month     = {June},
    year      = {2022},
    pages     = {2791-2800}
}

@InProceedings{Qin_2022_CVPR,
    author    = {Qin, Zequn and Li, Xi},
    title     = {MonoGround: Detecting Monocular 3D Objects From the Ground},
    booktitle = {Proceedings of the IEEE/CVF Conference on Computer Vision and Pattern Recognition (CVPR)},
    month     = {June},
    year      = {2022},
    pages     = {3793-3802}
}

@article{gerhard2025scoping,
  title={A scoping review on the methods used to assess health-related quality of life and disability burden in evaluations of road safety interventions},
  author={Gerhard, Robyn and Gabbe, Belinda J and Cameron, Peter and Newstead, Stuart and Morrison, Christopher N and Clarke, Nyssa and Beck, Ben},
  journal={Journal of Safety Research},
  volume={92},
  pages={459--472},
  year={2025},
  publisher={Elsevier}
}

@inproceedings{huang2025robotron,
  title={RoboTron-Drive: All-in-One Large Multimodal Model for Autonomous Driving},
  author={Huang, Zhijian and Feng, Chengjian and Yan, Feng and Xiao, Baihui and Jie, Zequn and Zhong, Yujie and Liang, Xiaodan and Ma, Lin},
  booktitle={Proceedings of the IEEE/CVF International Conference on Computer Vision},
  pages={8011--8021},
  year={2025}
}

@article{tekkesinoglu2025advancing,
  title={Advancing explainable autonomous vehicle systems: A comprehensive review and research roadmap},
  author={Tekkesinoglu, Sule and Habibovic, Azra and Kunze, Lars},
  journal={ACM Transactions on Human-Robot Interaction},
  volume={14},
  number={3},
  pages={1--46},
  year={2025},
  publisher={ACM New York, NY}
}

@article{thottempudi2025resilient,
  title={Resilient object detection for autonomous vehicles: Integrating deep learning and sensor fusion in adverse conditions},
  author={Thottempudi, Pardhu and Jambek, Asral Bin Bahari and Kumar, Vijay and Acharya, Biswaranjan and Moreira, Fernando},
  journal={Engineering Applications of Artificial Intelligence},
  volume={151},
  pages={110563},
  year={2025},
  publisher={Elsevier}
}

@InProceedings{Lian_2022_CVPR,
    author    = {Lian, Qing and Li, Peiliang and Chen, Xiaozhi},
    title     = {MonoJSG: Joint Semantic and Geometric Cost Volume for Monocular 3D Object Detection},
    booktitle = {Proceedings of the IEEE/CVF Conference on Computer Vision and Pattern Recognition (CVPR)},
    month     = {June},
    year      = {2022},
    pages     = {1070-1079}
}

@inproceedings{brazil2020kinematic,
  title={Kinematic 3d object detection in monocular video},
  author={Brazil, Garrick and Pons-Moll, Gerard and Liu, Xiaoming and Schiele, Bernt},
  booktitle={European Conference on Computer Vision},
  pages={135--152},
  year={2020},
  organization={Springer}
}

@InProceedings{Rukhovich_2022_WACV,
    author    = {Rukhovich, Danila and Vorontsova, Anna and Konushin, Anton},
    title     = {ImVoxelNet: Image to Voxels Projection for Monocular and Multi-View General-Purpose 3D Object Detection},
    booktitle = {Proceedings of the IEEE/CVF Winter Conference on Applications of Computer Vision (WACV)},
    month     = {January},
    year      = {2022},
    pages     = {2397-2406}
}

@InProceedings{Gu_2022_CVPR,
    author    = {Gu, Jiaqi and Wu, Bojian and Fan, Lubin and Huang, Jianqiang and Cao, Shen and Xiang, Zhiyu and Hua, Xian-Sheng},
    title     = {Homography Loss for Monocular 3D Object Detection},
    booktitle = {Proceedings of the IEEE/CVF Conference on Computer Vision and Pattern Recognition (CVPR)},
    month     = {June},
    year      = {2022},
    pages     = {1080-1089}
}

@inproceedings{hong2022cross,
  title={Cross-modality knowledge distillation network for monocular 3d object detection},
  author={Hong, Yu and Dai, Hang and Ding, Yong},
  booktitle={European Conference on Computer Vision},
  pages={87--104},
  year={2022},
  organization={Springer}
}

@inproceedings{peng2022did,
  title={Did-m3d: Decoupling instance depth for monocular 3d object detection},
  author={Peng, Liang and Wu, Xiaopei and Yang, Zheng and Liu, Haifeng and Cai, Deng},
  booktitle={European Conference on Computer Vision},
  pages={71--88},
  year={2022},
  organization={Springer}
}

@InProceedings{Lu_2021_ICCV,
    author    = {Lu, Yan and Ma, Xinzhu and Yang, Lei and Zhang, Tianzhu and Liu, Yating and Chu, Qi and Yan, Junjie and Ouyang, Wanli},
    title     = {Geometry Uncertainty Projection Network for Monocular 3D Object Detection},
    booktitle = {Proceedings of the IEEE/CVF International Conference on Computer Vision (ICCV)},
    month     = {October},
    year      = {2021},
    pages     = {3111-3121}
}

@InProceedings{Ma_2021_CVPR,
    author    = {Ma, Xinzhu and Zhang, Yinmin and Xu, Dan and Zhou, Dongzhan and Yi, Shuai and Li, Haojie and Ouyang, Wanli},
    title     = {Delving Into Localization Errors for Monocular 3D Object Detection},
    booktitle = {Proceedings of the IEEE/CVF Conference on Computer Vision and Pattern Recognition (CVPR)},
    month     = {June},
    year      = {2021},
    pages     = {4721-4730}
}

@InProceedings{Shi_2021_ICCV,
    author    = {Shi, Xuepeng and Ye, Qi and Chen, Xiaozhi and Chen, Chuangrong and Chen, Zhixiang and Kim, Tae-Kyun},
    title     = {Geometry-Based Distance Decomposition for Monocular 3D Object Detection},
    booktitle = {Proceedings of the IEEE/CVF International Conference on Computer Vision (ICCV)},
    month     = {October},
    year      = {2021},
    pages     = {15172-15181}
}

@InProceedings{Zhang_2021_CVPR,
    author    = {Zhang, Yunpeng and Lu, Jiwen and Zhou, Jie},
    title     = {Objects Are Different: Flexible Monocular 3D Object Detection},
    booktitle = {Proceedings of the IEEE/CVF Conference on Computer Vision and Pattern Recognition (CVPR)},
    month     = {June},
    year      = {2021},
    pages     = {3289-3298}
}

@InProceedings{Chen_2021_CVPR,
    author    = {Chen, Hansheng and Huang, Yuyao and Tian, Wei and Gao, Zhong and Xiong, Lu},
    title     = {MonoRUn: Monocular 3D Object Detection by Reconstruction and Uncertainty Propagation},
    booktitle = {Proceedings of the IEEE/CVF Conference on Computer Vision and Pattern Recognition (CVPR)},
    month     = {June},
    year      = {2021},
    pages     = {10379-10388}
}

@InProceedings{Wang_2021_CVPR,
    author    = {Wang, Li and Du, Liang and Ye, Xiaoqing and Fu, Yanwei and Guo, Guodong and Xue, Xiangyang and Feng, Jianfeng and Zhang, Li},
    title     = {Depth-Conditioned Dynamic Message Propagation for Monocular 3D Object Detection},
    booktitle = {Proceedings of the IEEE/CVF Conference on Computer Vision and Pattern Recognition (CVPR)},
    month     = {June},
    year      = {2021},
    pages     = {454-463}
}

@inproceedings{chen2020monopair,
  title={Monopair: Monocular 3d object detection using pairwise spatial relationships},
  author={Chen, Yongjian and Tai, Lei and Sun, Kai and Li, Mingyang},
  booktitle={Proceedings of the IEEE/CVF conference on computer vision and pattern recognition},
  pages={12093--12102},
  year={2020}
}

@inproceedings{li2020rtm3d,
  title={Rtm3d: Real-time monocular 3d detection from object keypoints for autonomous driving},
  author={Li, Peixuan and Zhao, Huaici and Liu, Pengfei and Cao, Feidao},
  booktitle={European Conference on Computer Vision},
  pages={644--660},
  year={2020},
  organization={Springer}
}

@inproceedings{shi2020distance,
  title={Distance-normalized unified representation for monocular 3d object detection},
  author={Shi, Xuepeng and Chen, Zhixiang and Kim, Tae-Kyun},
  booktitle={European Conference on Computer Vision},
  pages={91--107},
  year={2020},
  organization={Springer}
}

@inproceedings{simonelli2020towards,
  title={Towards generalization across depth for monocular 3d object detection},
  author={Simonelli, Andrea and Bulo, Samuel Rota and Porzi, Lorenzo and Ricci, Elisa and Kontschieder, Peter},
  booktitle={European Conference on Computer Vision},
  pages={767--782},
  year={2020},
  organization={Springer}
}

@inproceedings{ye2020monocular,
  title={Monocular 3d object detection via feature domain adaptation},
  author={Ye, Xiaoqing and Du, Liang and Shi, Yifeng and Li, Yingying and Tan, Xiao and Feng, Jianfeng and Ding, Errui and Wen, Shilei},
  booktitle={European Conference on Computer Vision},
  pages={17--34},
  year={2020},
  organization={Springer}
}

@inproceedings{simonelli2019disentangling,
  title={Disentangling monocular 3d object detection},
  author={Simonelli, Andrea and Bulo, Samuel Rota and Porzi, Lorenzo and L{\'o}pez-Antequera, Manuel and Kontschieder, Peter},
  booktitle={Proceedings of the IEEE/CVF international conference on computer vision},
  pages={1991--1999},
  year={2019}
}

@inproceedings{brazil2019m3d,
  title={M3d-rpn: Monocular 3d region proposal network for object detection},
  author={Brazil, Garrick and Liu, Xiaoming},
  booktitle={Proceedings of the IEEE/CVF international conference on computer vision},
  pages={9287--9296},
  year={2019}
}

@InProceedings{Xue_2025_CVPR,
    author    = {Xue, Ziteng and Guo, Mingzhe and Fan, Heng and Zhang, Shihui and Zhang, Zhipeng},
    title     = {CorrBEV: Multi-View 3D Object Detection by Correlation Learning with Multi-modal Prototypes},
    booktitle = {Proceedings of the IEEE/CVF Conference on Computer Vision and Pattern Recognition (CVPR)},
    month     = {June},
    year      = {2025},
    pages     = {27413-27423}
}

@article{ji2025ropetr,
  title={RoPETR: Improving Temporal Camera-Only 3D Detection by Integrating Enhanced Rotary Position Embedding},
  author={Ji, Hang and Ni, Tao and Huang, Xufeng and Shi, Zhan and Luo, Tao and Zhan, Xin and Chen, Junbo},
  journal={arXiv preprint arXiv:2504.12643},
  year={2025}
}

@inproceedings{liu2024ray,
  title={Ray denoising: Depth-aware hard negative sampling for multi-view 3d object detection},
  author={Liu, Feng and Huang, Tengteng and Zhang, Qianjing and Yao, Haotian and Zhang, Chi and Wan, Fang and Ye, Qixiang and Zhou, Yanzhao},
  booktitle={European Conference on Computer Vision},
  pages={200--217},
  year={2024},
  organization={Springer}
}

@InProceedings{Wang_2023_ICCV,
    author    = {Wang, Shihao and Liu, Yingfei and Wang, Tiancai and Li, Ying and Zhang, Xiangyu},
    title     = {Exploring Object-Centric Temporal Modeling for Efficient Multi-View 3D Object Detection},
    booktitle = {Proceedings of the IEEE/CVF International Conference on Computer Vision (ICCV)},
    month     = {October},
    year      = {2023},
    pages     = {3621-3631}
}

@InProceedings{Liu_2023_ICCV,
    author    = {Liu, Yingfei and Yan, Junjie and Jia, Fan and Li, Shuailin and Gao, Aqi and Wang, Tiancai and Zhang, Xiangyu},
    title     = {PETRv2: A Unified Framework for 3D Perception from Multi-Camera Images},
    booktitle = {Proceedings of the IEEE/CVF International Conference on Computer Vision (ICCV)},
    month     = {October},
    year      = {2023},
    pages     = {3262-3272}
}

@inproceedings{li2023bevdepth,
  title={Bevdepth: Acquisition of reliable depth for multi-view 3d object detection},
  author={Li, Yinhao and Ge, Zheng and Yu, Guanyi and Yang, Jinrong and Wang, Zengran and Shi, Yukang and Sun, Jianjian and Li, Zeming},
  booktitle={Proceedings of the AAAI conference on artificial intelligence},
  volume={37},
  number={2},
  pages={1477--1485},
  year={2023}
}

@InProceedings{Liu_2023_ICCV2,
    author    = {Liu, Haisong and Teng, Yao and Lu, Tao and Wang, Haiguang and Wang, Limin},
    title     = {SparseBEV: High-Performance Sparse 3D Object Detection from Multi-Camera Videos},
    booktitle = {Proceedings of the IEEE/CVF International Conference on Computer Vision (ICCV)},
    month     = {October},
    year      = {2023},
    pages     = {18580-18590}
}

@InProceedings{Zong_2023_ICCV,
    author    = {Zong, Zhuofan and Jiang, Dongzhi and Song, Guanglu and Xue, Zeyue and Su, Jingyong and Li, Hongsheng and Liu, Yu},
    title     = {Temporal Enhanced Training of Multi-view 3D Object Detector via Historical Object Prediction},
    booktitle = {Proceedings of the IEEE/CVF International Conference on Computer Vision (ICCV)},
    month     = {October},
    year      = {2023},
    pages     = {3781-3790}
}

@inproceedings{jiang2023polarformer,
  title={Polarformer: Multi-camera 3d object detection with polar transformer},
  author={Jiang, Yanqin and Zhang, Li and Miao, Zhenwei and Zhu, Xiatian and Gao, Jin and Hu, Weiming and Jiang, Yu-Gang},
  booktitle={Proceedings of the AAAI conference on Artificial Intelligence},
  volume={37},
  number={1},
  pages={1042--1050},
  year={2023}
}

@InProceedings{Wang_2023_CVPR,
    author    = {Wang, Yuqi and Chen, Yuntao and Zhang, Zhaoxiang},
    title     = {FrustumFormer: Adaptive Instance-Aware Resampling for Multi-View 3D Detection},
    booktitle = {Proceedings of the IEEE/CVF Conference on Computer Vision and Pattern Recognition (CVPR)},
    month     = {June},
    year      = {2023},
    pages     = {5096-5105}
}

@InProceedings{Wang_2023_ICCV2,
    author    = {Wang, Zitian and Huang, Zehao and Fu, Jiahui and Wang, Naiyan and Liu, Si},
    title     = {Object as Query: Lifting Any 2D Object Detector to 3D Detection},
    booktitle = {Proceedings of the IEEE/CVF International Conference on Computer Vision (ICCV)},
    month     = {October},
    year      = {2023},
    pages     = {3791-3800}
}

@InProceedings{Xiong_2023_CVPR,
    author    = {Xiong, Kaixin and Gong, Shi and Ye, Xiaoqing and Tan, Xiao and Wan, Ji and Ding, Errui and Wang, Jingdong and Bai, Xiang},
    title     = {CAPE: Camera View Position Embedding for Multi-View 3D Object Detection},
    booktitle = {Proceedings of the IEEE/CVF Conference on Computer Vision and Pattern Recognition (CVPR)},
    month     = {June},
    year      = {2023},
    pages     = {21570-21579}
}

@InProceedings{Shu_2023_ICCV,
    author    = {Shu, Changyong and Deng, Jiajun and Yu, Fisher and Liu, Yifan},
    title     = {3DPPE: 3D Point Positional Encoding for Transformer-based Multi-Camera 3D Object Detection},
    booktitle = {Proceedings of the IEEE/CVF International Conference on Computer Vision (ICCV)},
    month     = {October},
    year      = {2023},
    pages     = {3580-3589}
}

@inproceedings{liu2022petr,
  title={Petr: Position embedding transformation for multi-view 3d object detection},
  author={Liu, Yingfei and Wang, Tiancai and Zhang, Xiangyu and Sun, Jian},
  booktitle={European conference on computer vision},
  pages={531--548},
  year={2022},
  organization={Springer}
}

@article{li2022unifying,
  title={Unifying voxel-based representation with transformer for 3d object detection},
  author={Li, Yanwei and Chen, Yilun and Qi, Xiaojuan and Li, Zeming and Sun, Jian and Jia, Jiaya},
  journal={Advances in Neural Information Processing Systems},
  volume={35},
  pages={18442--18455},
  year={2022}
}

@article{li2024bevformer,
  title={Bevformer: learning bird's-eye-view representation from lidar-camera via spatiotemporal transformers},
  author={Li, Zhiqi and Wang, Wenhai and Li, Hongyang and Xie, Enze and Sima, Chonghao and Lu, Tong and Yu, Qiao and Dai, Jifeng},
  journal={IEEE Transactions on Pattern Analysis and Machine Intelligence},
  year={2024},
  publisher={IEEE}
}

@inproceedings{li2023bevstereo,
  title={Bevstereo: Enhancing depth estimation in multi-view 3d object detection with temporal stereo},
  author={Li, Yinhao and Bao, Han and Ge, Zheng and Yang, Jinrong and Sun, Jianjian and Li, Zeming},
  booktitle={Proceedings of the AAAI Conference on Artificial Intelligence},
  volume={37},
  number={2},
  pages={1486--1494},
  year={2023}
}

@inproceedings{wang2022probabilistic,
  title={Probabilistic and geometric depth: Detecting objects in perspective},
  author={Wang, Tai and Xinge, ZHU and Pang, Jiangmiao and Lin, Dahua},
  booktitle={Conference on Robot Learning},
  pages={1475--1485},
  year={2022},
  organization={PMLR}
}

@inproceedings{wang2022detr3d,
  title={Detr3d: 3d object detection from multi-view images via 3d-to-2d queries},
  author={Wang, Yue and Guizilini, Vitor Campagnolo and Zhang, Tianyuan and Wang, Yilun and Zhao, Hang and Solomon, Justin},
  booktitle={Conference on robot learning},
  pages={180--191},
  year={2022},
  organization={PMLR}
}

@inproceedings{chen2022graph,
  title={Graph-DETR3D: rethinking overlapping regions for multi-view 3D object detection},
  author={Chen, Zehui and Li, Zhenyu and Zhang, Shiquan and Fang, Liangji and Jiang, Qinhong and Zhao, Feng},
  booktitle={Proceedings of the 30th ACM International Conference on Multimedia},
  pages={5999--6008},
  year={2022}
}

@InProceedings{Li_2023_ICCV,
    author    = {Li, Zhiqi and Yu, Zhiding and Wang, Wenhai and Anandkumar, Anima and Lu, Tong and Alvarez, Jose M.},
    title     = {FB-BEV: BEV Representation from Forward-Backward View Transformations},
    booktitle = {Proceedings of the IEEE/CVF International Conference on Computer Vision (ICCV)},
    month     = {October},
    year      = {2023},
    pages     = {6919-6928}
}

@InProceedings{Wang_2021_ICCV2,
    author    = {Wang, Tai and Zhu, Xinge and Pang, Jiangmiao and Lin, Dahua},
    title     = {FCOS3D: Fully Convolutional One-Stage Monocular 3D Object Detection},
    booktitle = {Proceedings of the IEEE/CVF International Conference on Computer Vision (ICCV) Workshops},
    month     = {October},
    year      = {2021},
    pages     = {913-922}
}

@article{liang2022aspectnet,
  title={AspectNet: Aspect-Aware Anchor-Free Detector for Autonomous Driving},
  author={Liang, Tianjiao and Bao, Hong and Pan, Weiguo and Fan, Xinyue and Li, Han},
  journal={Applied Sciences},
  volume={12},
  number={12},
  pages={5972},
  year={2022},
  publisher={MDPI}
}

@inproceedings{choi2019gaussian,
  title={Gaussian yolov3: An accurate and fast object detector using localization uncertainty for autonomous driving},
  author={Choi, Jiwoong and Chun, Dayoung and Kim, Hyun and Lee, Hyuk-Jae},
  booktitle={Proceedings of the IEEE/CVF International conference on computer vision},
  pages={502--511},
  year={2019}
}

@article{hu2018sinet,
  title={SINet: A scale-insensitive convolutional neural network for fast vehicle detection},
  author={Hu, Xiaowei and Xu, Xuemiao and Xiao, Yongjie and Chen, Hao and He, Shengfeng and Qin, Jing and Heng, Pheng-Ann},
  journal={IEEE transactions on intelligent transportation systems},
  volume={20},
  number={3},
  pages={1010--1019},
  year={2018},
  publisher={IEEE}
}

@article{yi2019assd,
  title={ASSD: Attentive single shot multibox detector},
  author={Yi, Jingru and Wu, Pengxiang and Metaxas, Dimitris N},
  journal={Computer Vision and Image Understanding},
  volume={189},
  pages={102827},
  year={2019},
  publisher={Elsevier}
}

@inproceedings{zhang2018single,
  title={Single-shot refinement neural network for object detection},
  author={Zhang, Shifeng and Wen, Longyin and Bian, Xiao and Lei, Zhen and Li, Stan Z},
  booktitle={Proceedings of the IEEE conference on computer vision and pattern recognition},
  pages={4203--4212},
  year={2018}
}

@inproceedings{liu2018receptive,
  title={Receptive field block net for accurate and fast object detection},
  author={Liu, Songtao and Huang, Di and others},
  booktitle={Proceedings of the European conference on computer vision (ECCV)},
  pages={385--400},
  year={2018}
}

@inproceedings{wu2017squeezedet,
  title={Squeezedet: Unified, small, low power fully convolutional neural networks for real-time object detection for autonomous driving},
  author={Wu, Bichen and Iandola, Forrest and Jin, Peter H and Keutzer, Kurt},
  booktitle={Proceedings of the IEEE conference on computer vision and pattern recognition workshops},
  pages={129--137},
  year={2017}
}

@inproceedings{cai2016unified,
  title={A unified multi-scale deep convolutional neural network for fast object detection},
  author={Cai, Zhaowei and Fan, Quanfu and Feris, Rogerio S and Vasconcelos, Nuno},
  booktitle={European conference on computer vision},
  pages={354--370},
  year={2016},
  organization={Springer}
}

@article{gao2025algorithm,
  title={An algorithm for road target detection of autonomous vehicles based on improved YOLOv8},
  author={Gao, Jianping and Li, Haotian and Li, Zhe and Xie, Chengwei and Ji, Xiaolei and Zhang, Yuzhuo},
  journal={Scientific Reports},
  volume={15},
  number={1},
  pages={21061},
  year={2025},
  publisher={Nature Publishing Group UK London}
}

@article{yang2024yolov8,
  title={YOLOv8-Lite: A lightweight object detection model for real-time autonomous driving systems},
  author={Yang, Ming and Fan, Xiangyu},
  journal={ICCK Transactions on Emerging Topics in Artificial Intelligence},
  volume={1},
  number={1},
  pages={1--16},
  year={2024},
  publisher={Institute of Emerging and Computer Engineers}
}

@article{jia2023fast,
  title={Fast and accurate object detector for autonomous driving based on improved YOLOv5},
  author={Jia, Xiang and Tong, Ying and Qiao, Hongming and Li, Man and Tong, Jiangang and Liang, Baoling},
  journal={Scientific reports},
  volume={13},
  number={1},
  pages={9711},
  year={2023},
  publisher={Nature Publishing Group UK London}
}

@article{he2023object,
  title={Object detection based on lightweight YOLOX for autonomous driving},
  author={He, Qiyi and Xu, Ao and Ye, Zhiwei and Zhou, Wen and Cai, Ting},
  journal={Sensors},
  volume={23},
  number={17},
  pages={7596},
  year={2023},
  publisher={MDPI}
}

@article{an2025improved,
  title={Improved Vehicle Object Detection Algorithm Based on Swin-YOLOv5s},
  author={An, Haichao and Tang, Jianhua and Fan, Ying and Liu, Meiqin},
  journal={Processes},
  volume={13},
  number={3},
  pages={925},
  year={2025},
  publisher={MDPI}
}

@article{li2024yolo,
  title={YOLO-Vehicle-Pro: A Cloud-Edge Collaborative Framework for Object Detection in Autonomous Driving under Adverse Weather Conditions},
  author={Li, Xiguang and Chen, Jiafu and Sun, Yunhe and Lin, Na and Hawbani, Ammar and Zhao, Liang},
  journal={arXiv preprint arXiv:2410.17734},
  year={2024}
}

@article{wang2024yolov10,
  title={Yolov10: Real-time end-to-end object detection},
  author={Wang, Ao and Chen, Hui and Liu, Lihao and Chen, Kai and Lin, Zijia and Han, Jungong and others},
  journal={Advances in Neural Information Processing Systems},
  volume={37},
  pages={107984--108011},
  year={2024}
}

@inproceedings{wang2024yolov9,
  title={Yolov9: Learning what you want to learn using programmable gradient information},
  author={Wang, Chien-Yao and Yeh, I-Hau and Mark Liao, Hong-Yuan},
  booktitle={European conference on computer vision},
  pages={1--21},
  year={2024},
  organization={Springer}
}

@inproceedings{wang2023yolov7,
  title={YOLOv7: Trainable bag-of-freebies sets new state-of-the-art for real-time object detectors},
  author={Wang, Chien-Yao and Bochkovskiy, Alexey and Liao, Hong-Yuan Mark},
  booktitle={Proceedings of the IEEE/CVF conference on computer vision and pattern recognition},
  pages={7464--7475},
  year={2023}
}

@article{li2022yolov6,
  title={YOLOv6: A single-stage object detection framework for industrial applications},
  author={Li, Chuyi and Li, Lulu and Jiang, Hongliang and Weng, Kaiheng and Geng, Yifei and Li, Liang and Ke, Zaidan and Li, Qingyuan and Cheng, Meng and Nie, Weiqiang and others},
  journal={arXiv preprint arXiv:2209.02976},
  year={2022}
}

@inproceedings{adarsh2020yolo,
  title={YOLO v3-Tiny: Object Detection and Recognition using one stage improved model},
  author={Adarsh, Pranav and Rathi, Pratibha and Kumar, Manoj},
  booktitle={2020 6th international conference on advanced computing and communication systems (ICACCS)},
  pages={687--694},
  year={2020},
  organization={IEEE}
}

@article{ye2025fade3d,
  title={Fade3D: Fast and Deployable 3D Object Detection for Autonomous Driving},
  author={Ye, Wei and Xia, Qiming and Wu, Hai and Dong, Zhen and Zhong, Ruofei and Wang, Cheng and Wen, Chenglu},
  journal={IEEE Transactions on Intelligent Transportation Systems},
  year={2025},
  publisher={IEEE}
}

@inproceedings{wang2023dsvt,
  title={Dsvt: Dynamic sparse voxel transformer with rotated sets},
  author={Wang, Haiyang and Shi, Chen and Shi, Shaoshuai and Lei, Meng and Wang, Sen and He, Di and Schiele, Bernt and Wang, Liwei},
  booktitle={Proceedings of the IEEE/CVF Conference on Computer Vision and Pattern Recognition},
  pages={13520--13529},
  year={2023}
}

@inproceedings{zhao2023ada3d,
  title={Ada3d: Exploiting the spatial redundancy with adaptive inference for efficient 3d object detection},
  author={Zhao, Tianchen and Ning, Xuefei and Hong, Ke and Qiu, Zhongyuan and Lu, Pu and Zhao, Yali and Zhang, Linfeng and Zhou, Lipu and Dai, Guohao and Yang, Huazhong and others},
  booktitle={Proceedings of the IEEE/CVF International Conference on Computer Vision},
  pages={17728--17738},
  year={2023}
}

@inproceedings{deng2021voxel,
  title={Voxel r-cnn: Towards high performance voxel-based 3d object detection},
  author={Deng, Jiajun and Shi, Shaoshuai and Li, Peiwei and Zhou, Wengang and Zhang, Yanyong and Li, Houqiang},
  booktitle={Proceedings of the AAAI conference on artificial intelligence},
  volume={35},
  number={2},
  pages={1201--1209},
  year={2021}
}

@inproceedings{chen2023voxelnext,
  title={Voxelnext: Fully sparse voxelnet for 3d object detection and tracking},
  author={Chen, Yukang and Liu, Jianhui and Zhang, Xiangyu and Qi, Xiaojuan and Jia, Jiaya},
  booktitle={Proceedings of the IEEE/CVF conference on computer vision and pattern recognition},
  pages={21674--21683},
  year={2023}
}

@inproceedings{shi2022pillarnet,
  title={Pillarnet: Real-time and high-performance pillar-based 3d object detection},
  author={Shi, Guangsheng and Li, Ruifeng and Ma, Chao},
  booktitle={European conference on computer vision},
  pages={35--52},
  year={2022},
  organization={Springer}
}

@inproceedings{liu2020tanet,
  title={Tanet: Robust 3d object detection from point clouds with triple attention},
  author={Liu, Zhe and Zhao, Xin and Huang, Tengteng and Hu, Ruolan and Zhou, Yu and Bai, Xiang},
  booktitle={Proceedings of the AAAI conference on artificial intelligence},
  volume={34},
  number={07},
  pages={11677--11684},
  year={2020}
}

@article{shi2020points,
  title={From points to parts: 3d object detection from point cloud with part-aware and part-aggregation network},
  author={Shi, Shaoshuai and Wang, Zhe and Shi, Jianping and Wang, Xiaogang and Li, Hongsheng},
  journal={IEEE transactions on pattern analysis and machine intelligence},
  volume={43},
  number={8},
  pages={2647--2664},
  year={2020},
  publisher={IEEE}
}

@inproceedings{tang2020searching,
  title={Searching efficient 3d architectures with sparse point-voxel convolution},
  author={Tang, Haotian and Liu, Zhijian and Zhao, Shengyu and Lin, Yujun and Lin, Ji and Wang, Hanrui and Han, Song},
  booktitle={European conference on computer vision},
  pages={685--702},
  year={2020},
  organization={Springer}
}

@inproceedings{zheng2021cia,
  title={Cia-ssd: Confident iou-aware single-stage object detector from point cloud},
  author={Zheng, Wu and Tang, Weiliang and Chen, Sijin and Jiang, Li and Fu, Chi-Wing},
  booktitle={Proceedings of the AAAI conference on artificial intelligence},
  volume={35},
  number={4},
  pages={3555--3562},
  year={2021}
}

@inproceedings{shi2020pv,
  title={Pv-rcnn: Point-voxel feature set abstraction for 3d object detection},
  author={Shi, Shaoshuai and Guo, Chaoxu and Jiang, Li and Wang, Zhe and Shi, Jianping and Wang, Xiaogang and Li, Hongsheng},
  booktitle={Proceedings of the IEEE/CVF conference on computer vision and pattern recognition},
  pages={10529--10538},
  year={2020}
}

@inproceedings{sheng2021improving,
  title={Improving 3d object detection with channel-wise transformer},
  author={Sheng, Hualian and Cai, Sijia and Liu, Yuan and Deng, Bing and Huang, Jianqiang and Hua, Xian-Sheng and Zhao, Min-Jian},
  booktitle={Proceedings of the IEEE/CVF international conference on computer vision},
  pages={2743--2752},
  year={2021}
}

@inproceedings{xu2021spg,
  title={Spg: Unsupervised domain adaptation for 3d object detection via semantic point generation},
  author={Xu, Qiangeng and Zhou, Yin and Wang, Weiyue and Qi, Charles R and Anguelov, Dragomir},
  booktitle={Proceedings of the IEEE/CVF International Conference on Computer Vision},
  pages={15446--15456},
  year={2021}
}

@inproceedings{zheng2021se,
  title={SE-SSD: Self-ensembling single-stage object detector from point cloud},
  author={Zheng, Wu and Tang, Weiliang and Jiang, Li and Fu, Chi-Wing},
  booktitle={Proceedings of the IEEE/CVF conference on computer vision and pattern recognition},
  pages={14494--14503},
  year={2021}
}

@inproceedings{he2022voxel,
  title={Voxel set transformer: A set-to-set approach to 3d object detection from point clouds},
  author={He, Chenhang and Li, Ruihuang and Li, Shuai and Zhang, Lei},
  booktitle={Proceedings of the IEEE/CVF conference on computer vision and pattern recognition},
  pages={8417--8427},
  year={2022}
}

@inproceedings{xu2022behind,
  title={Behind the curtain: Learning occluded shapes for 3d object detection},
  author={Xu, Qiangeng and Zhong, Yiqi and Neumann, Ulrich},
  booktitle={Proceedings of the AAAI Conference on Artificial Intelligence},
  volume={36},
  number={3},
  pages={2893--2901},
  year={2022}
}

@inproceedings{yang2023pvt,
  title={Pvt-ssd: Single-stage 3d object detector with point-voxel transformer},
  author={Yang, Honghui and Wang, Wenxiao and Chen, Minghao and Lin, Binbin and He, Tong and Chen, Hua and He, Xiaofei and Ouyang, Wanli},
  booktitle={Proceedings of the IEEE/CVF conference on computer vision and pattern recognition},
  pages={13476--13487},
  year={2023}
}

@article{xia20233,
  title={3-D HANet: A flexible 3-D heatmap auxiliary network for object detection},
  author={Xia, Qiming and Chen, Yidong and Cai, Guorong and Chen, Guikun and Xie, Daoshun and Su, Jinhe and Wang, Zongyue},
  journal={IEEE Transactions on Geoscience and Remote Sensing},
  volume={61},
  pages={1--13},
  year={2023},
  publisher={IEEE}
}

@article{hoang2024tsstdet,
  title={TSSTDet: Transformation-based 3-D object detection via a spatial shape transformer},
  author={Hoang, Hiep Anh and Bui, Duy Cuong and Yoo, Myungsik},
  journal={IEEE Sensors Journal},
  volume={24},
  number={5},
  pages={7126--7139},
  year={2024},
  publisher={IEEE}
}

@article{yan2018second,
  title={Second: Sparsely embedded convolutional detection},
  author={Yan, Yan and Mao, Yuxing and Li, Bo},
  journal={Sensors},
  volume={18},
  number={10},
  pages={3337},
  year={2018},
  publisher={Multidisciplinary Digital Publishing Institute}
}

@inproceedings{he2020structure,
  title={Structure aware single-stage 3d object detection from point cloud},
  author={He, Chenhang and Zeng, Hui and Huang, Jianqiang and Hua, Xian-Sheng and Zhang, Lei},
  booktitle={Proceedings of the IEEE/CVF conference on computer vision and pattern recognition},
  pages={11873--11882},
  year={2020}
}

@inproceedings{sheng2022rethinking,
  title={Rethinking IoU-based optimization for single-stage 3D object detection},
  author={Sheng, Hualian and Cai, Sijia and Zhao, Na and Deng, Bing and Huang, Jianqiang and Hua, Xian-Sheng and Zhao, Min-Jian and Lee, Gim Hee},
  booktitle={European Conference on Computer Vision},
  pages={544--561},
  year={2022},
  organization={Springer}
}

@inproceedings{shi2019pointrcnn,
  title={Pointrcnn: 3d object proposal generation and detection from point cloud},
  author={Shi, Shaoshuai and Wang, Xiaogang and Li, Hongsheng},
  booktitle={Proceedings of the IEEE/CVF conference on computer vision and pattern recognition},
  pages={770--779},
  year={2019}
}

@inproceedings{shi2020point,
  title={Point-gnn: Graph neural network for 3d object detection in a point cloud},
  author={Shi, Weijing and Rajkumar, Raj},
  booktitle={Proceedings of the IEEE/CVF conference on computer vision and pattern recognition},
  pages={1711--1719},
  year={2020}
}

@inproceedings{yang20203dssd,
  title={3dssd: Point-based 3d single stage object detector},
  author={Yang, Zetong and Sun, Yanan and Liu, Shu and Jia, Jiaya},
  booktitle={Proceedings of the IEEE/CVF conference on computer vision and pattern recognition},
  pages={11040--11048},
  year={2020}
}

@inproceedings{zhang2022not,
  title={Not all points are equal: Learning highly efficient point-based detectors for 3d lidar point clouds},
  author={Zhang, Yifan and Hu, Qingyong and Xu, Guoquan and Ma, Yanxin and Wan, Jianwei and Guo, Yulan},
  booktitle={Proceedings of the IEEE/CVF conference on computer vision and pattern recognition},
  pages={18953--18962},
  year={2022}
}

@inproceedings{ding2025det,
  title={AS-Det: Active Sampling for Adaptive 3D Object Detection in Point Clouds},
  author={Ding, Ziheng and Zhang, Xiaze and Jing, Qi and Cheng, Ying and Feng, Rui},
  booktitle={Proceedings of the AAAI Conference on Artificial Intelligence},
  volume={39},
  number={3},
  pages={2762--2770},
  year={2025}
}

@article{tian2022fully,
  title={Fully convolutional one-stage 3d object detection on lidar range images},
  author={Tian, Zhi and Chu, Xiangxiang and Wang, Xiaoming and Wei, Xiaolin and Shen, Chunhua},
  journal={Advances in neural information processing systems},
  volume={35},
  pages={34899--34911},
  year={2022}
}

@inproceedings{hu2022afdetv2,
  title={Afdetv2: Rethinking the necessity of the second stage for object detection from point clouds},
  author={Hu, Yihan and Ding, Zhuangzhuang and Ge, Runzhou and Shao, Wenxin and Huang, Li and Li, Kun and Liu, Qiang},
  booktitle={Proceedings of the AAAI Conference on Artificial Intelligence},
  volume={36},
  number={1},
  pages={969--979},
  year={2022}
}

@inproceedings{chen2022focal,
  title={Focal sparse convolutional networks for 3d object detection},
  author={Chen, Yukang and Li, Yanwei and Zhang, Xiangyu and Sun, Jian and Jia, Jiaya},
  booktitle={Proceedings of the IEEE/CVF conference on computer vision and pattern recognition},
  pages={5428--5437},
  year={2022}
}

@inproceedings{lu2023link,
  title={Link: Linear kernel for lidar-based 3d perception},
  author={Lu, Tao and Ding, Xiang and Liu, Haisong and Wu, Gangshan and Wang, Limin},
  booktitle={Proceedings of the IEEE/CVF Conference on Computer Vision and Pattern Recognition},
  pages={1105--1115},
  year={2023}
}

@article{zhang2023hednet,
  title={Hednet: A hierarchical encoder-decoder network for 3d object detection in point clouds},
  author={Zhang, Gang and Junnan, Chen and Gao, Guohuan and Li, Jianmin and Hu, Xiaolin},
  journal={Advances in Neural Information Processing Systems},
  volume={36},
  pages={53076--53089},
  year={2023}
}

@inproceedings{chen2023largekernel3d,
  title={Largekernel3d: Scaling up kernels in 3d sparse cnns},
  author={Chen, Yukang and Liu, Jianhui and Zhang, Xiangyu and Qi, Xiaojuan and Jia, Jiaya},
  booktitle={Proceedings of the IEEE/CVF conference on computer vision and pattern recognition},
  pages={13488--13498},
  year={2023}
}

@article{fan2024fsd,
  title={Fsd v2: Improving fully sparse 3d object detection with virtual voxels},
  author={Fan, Lue and Wang, Feng and Wang, Naiyan and Zhang, Zhaoxiang},
  journal={IEEE Transactions on Pattern Analysis and Machine Intelligence},
  year={2024},
  publisher={IEEE}
}

@article{zhang2024voxel,
  title={Voxel mamba: Group-free state space models for point cloud based 3d object detection},
  author={Zhang, Guowen and Fan, Lue and He, Chenhang and Lei, Zhen and ZHANG, ZHAO-XIANG and Zhang, Lei},
  journal={Advances in Neural Information Processing Systems},
  volume={37},
  pages={81489--81509},
  year={2024}
}

@article{zhu2019class,
  title={Class-balanced grouping and sampling for point cloud 3d object detection},
  author={Zhu, Benjin and Jiang, Zhengkai and Zhou, Xiangxin and Li, Zeming and Yu, Gang},
  journal={arXiv preprint arXiv:1908.09492},
  year={2019}
}

@article{chen2020every,
  title={Every view counts: Cross-view consistency in 3d object detection with hybrid-cylindrical-spherical voxelization},
  author={Chen, Qi and Sun, Lin and Cheung, Ernest and Yuille, Alan L},
  journal={Advances in Neural Information Processing Systems},
  volume={33},
  pages={21224--21235},
  year={2020}
}

@inproceedings{yin2020lidar,
  title={Lidar-based online 3d video object detection with graph-based message passing and spatiotemporal transformer attention},
  author={Yin, Junbo and Shen, Jianbing and Guan, Chenye and Zhou, Dingfu and Yang, Ruigang},
  booktitle={Proceedings of the IEEE/CVF Conference on Computer Vision and Pattern Recognition},
  pages={11495--11504},
  year={2020}
}

@article{zhang2023fully,
  title={Fully sparse transformer 3-D detector for LiDAR point cloud},
  author={Zhang, Diankun and Zheng, Zhijie and Niu, Haoyu and Wang, Xueqing and Liu, Xiaojun},
  journal={IEEE Transactions on Geoscience and Remote Sensing},
  volume={61},
  pages={1--12},
  year={2023},
  publisher={IEEE}
}

@article{wang2023uni3detr,
  title={Uni3detr: Unified 3d detection transformer},
  author={Wang, Zhenyu and Li, Ya-Li and Chen, Xi and Zhao, Hengshuang and Wang, Shengjin},
  journal={Advances in Neural Information Processing Systems},
  volume={36},
  pages={39876--39896},
  year={2023}
}

@inproceedings{bai2022transfusion,
  title={Transfusion: Robust lidar-camera fusion for 3d object detection with transformers},
  author={Bai, Xuyang and Hu, Zeyu and Zhu, Xinge and Huang, Qingqiu and Chen, Yilun and Fu, Hongbo and Tai, Chiew-Lan},
  booktitle={Proceedings of the IEEE/CVF conference on computer vision and pattern recognition},
  pages={1090--1099},
  year={2022}
}

@article{liu2024lion,
  title={Lion: Linear group rnn for 3d object detection in point clouds},
  author={Liu, Zhe and Hou, Jinghua and Wang, Xinyu and Ye, Xiaoqing and Wang, Jingdong and Zhao, Hengshuang and Bai, Xiang},
  journal={Advances in Neural Information Processing Systems},
  volume={37},
  pages={13601--13626},
  year={2024}
}

@inproceedings{pan20213d,
  title={3d object detection with pointformer},
  author={Pan, Xuran and Xia, Zhuofan and Song, Shiji and Li, Li Erran and Huang, Gao},
  booktitle={Proceedings of the IEEE/CVF conference on computer vision and pattern recognition},
  pages={7463--7472},
  year={2021}
}

@inproceedings{fazlali2022versatile,
  title={A versatile multi-view framework for lidar-based 3d object detection with guidance from panoptic segmentation},
  author={Fazlali, Hamidreza and Xu, Yixuan and Ren, Yuan and Liu, Bingbing},
  booktitle={Proceedings of the IEEE/CVF Conference on Computer Vision and Pattern Recognition},
  pages={17192--17201},
  year={2022}
}

@article{mao2024pillarnest,
  title={Pillarnest: Embracing backbone scaling and pretraining for pillar-based 3d object detection},
  author={Mao, Weixin and Wang, Tiancai and Zhang, Diankun and Yan, Junjie and Yoshie, Osamu},
  journal={IEEE Transactions on Intelligent Vehicles},
  year={2024},
  publisher={IEEE}
}

@inproceedings{zhang2024safdnet,
  title={Safdnet: A simple and effective network for fully sparse 3d object detection},
  author={Zhang, Gang and Chen, Junnan and Gao, Guohuan and Li, Jianmin and Liu, Si and Hu, Xiaolin},
  booktitle={Proceedings of the IEEE/CVF Conference on Computer Vision and Pattern Recognition},
  pages={14477--14486},
  year={2024}
}

@inproceedings{yang2019std,
  title={Std: Sparse-to-dense 3d object detector for point cloud},
  author={Yang, Zetong and Sun, Yanan and Liu, Shu and Shen, Xiaoyong and Jia, Jiaya},
  booktitle={Proceedings of the IEEE/CVF international conference on computer vision},
  pages={1951--1960},
  year={2019}
}

@inproceedings{fan2021rangedet,
  title={Rangedet: In defense of range view for lidar-based 3d object detection},
  author={Fan, Lue and Xiong, Xuan and Wang, Feng and Wang, Naiyan and Zhang, Zhaoxiang},
  booktitle={Proceedings of the IEEE/CVF international conference on computer vision},
  pages={2918--2927},
  year={2021}
}

@inproceedings{huang2025l4dr,
  title={L4dr: Lidar-4dradar fusion for weather-robust 3d object detection},
  author={Huang, Xun and Xu, Ziyu and Wu, Hai and Wang, Jinlong and Xia, Qiming and Xia, Yan and Li, Jonathan and Gao, Kyle and Wen, Chenglu and Wang, Cheng},
  booktitle={Proceedings of the AAAI Conference on Artificial Intelligence},
  volume={39},
  number={4},
  pages={3806--3814},
  year={2025}
}

@inproceedings{zhou2018voxelnet,
  title={Voxelnet: End-to-end learning for point cloud based 3d object detection},
  author={Zhou, Yin and Tuzel, Oncel},
  booktitle={Proceedings of the IEEE conference on computer vision and pattern recognition},
  pages={4490--4499},
  year={2018}
}

@inproceedings{jin2024swiftpillars,
  title={Swiftpillars: High-efficiency pillar encoder for lidar-based 3d detection},
  author={Jin, Xin and Liu, Kai and Ma, Cong and Yang, Ruining and Hui, Fei and Wu, Wei},
  booktitle={Proceedings of the AAAI Conference on Artificial Intelligence},
  volume={38},
  number={3},
  pages={2625--2633},
  year={2024}
}

@inproceedings{yin2021center,
  title={Center-based 3d object detection and tracking},
  author={Yin, Tianwei and Zhou, Xingyi and Krahenbuhl, Philipp},
  booktitle={Proceedings of the IEEE/CVF conference on computer vision and pattern recognition},
  pages={11784--11793},
  year={2021}
}

@inproceedings{lang2019pointpillars,
  title={Pointpillars: Fast encoders for object detection from point clouds},
  author={Lang, Alex H and Vora, Sourabh and Caesar, Holger and Zhou, Lubing and Yang, Jiong and Beijbom, Oscar},
  booktitle={Proceedings of the IEEE/CVF conference on computer vision and pattern recognition},
  pages={12697--12705},
  year={2019}
}

@inproceedings{wu2023transformation,
  title={Transformation-equivariant 3d object detection for autonomous driving},
  author={Wu, Hai and Wen, Chenglu and Li, Wei and Li, Xin and Yang, Ruigang and Wang, Cheng},
  booktitle={Proceedings of the AAAI Conference on Artificial Intelligence},
  volume={37},
  number={3},
  pages={2795--2802},
  year={2023}
}

@article{peng2025elmar,
  title={ELMAR: Enhancing LiDAR Detection with 4D Radar Motion Awareness and Cross-modal Uncertainty},
  author={Peng, Xiangyuan and Tang, Miao and Sun, Huawei and Kay, Bierzynski and Servadei, Lorenzo and Wille, Robert},
  journal={arXiv preprint arXiv:2506.17958},
  year={2025}
}

@article{peng2025moral,
  title={MoRAL: Motion-aware Multi-Frame 4D Radar and LiDAR Fusion for Robust 3D Object Detection},
  author={Peng, Xiangyuan and Wang, Yu and Tang, Miao and Kay, Bierzynski and Servadei, Lorenzo and Wille, Robert},
  journal={arXiv preprint arXiv:2505.09422},
  year={2025}
}

@inproceedings{peng2025mutualforce,
  title={Mutualforce: Mutual-aware enhancement for 4d radar-lidar 3d object detection},
  author={Peng, Xiangyuan and Sun, Huawei and Bierzynski, Kay and Fischbacher, Anton and Servadei, Lorenzo and Wille, Robert},
  booktitle={ICASSP 2025-2025 IEEE International Conference on Acoustics, Speech and Signal Processing (ICASSP)},
  pages={1--5},
  year={2025},
  organization={IEEE}
}

@inproceedings{deng2024robust,
  title={Robust 3d object detection from lidar-radar point clouds via cross-modal feature augmentation},
  author={Deng, Jianning and Chan, Gabriel and Zhong, Hantao and Lu, Chris Xiaoxuan},
  booktitle={2024 IEEE International Conference on Robotics and Automation (ICRA)},
  pages={6585--6591},
  year={2024},
  organization={IEEE}
}

@inproceedings{xu2024rlnet,
  title={Rlnet: Adaptive fusion of 4d radar and lidar for 3d object detection},
  author={Xu, Ruoyu and Xiang, Zhiyu},
  booktitle={European Conference on Computer Vision},
  pages={181--194},
  year={2024},
  organization={Springer}
}

@inproceedings{wang2022interfusion,
  title={InterFusion: Interaction-based 4D radar and LiDAR fusion for 3D object detection},
  author={Wang, Li and Zhang, Xinyu and Xv, Baowei and Zhang, Jinzhao and Fu, Rong and Wang, Xiaoyu and Zhu, Lei and Ren, Haibing and Lu, Pingping and Li, Jun and others},
  booktitle={2022 IEEE/RSJ International Conference on Intelligent Robots and Systems (IROS)},
  pages={12247--12253},
  year={2022},
  organization={IEEE}
}

@inproceedings{yu2025vikienet,
  title={ViKIENet: Towards Efficient 3D Object Detection with Virtual Key Instance Enhanced Network},
  author={Yu, Zhuochen and Qiu, Bijie and Khong, Andy WH},
  booktitle={Proceedings of the Computer Vision and Pattern Recognition Conference},
  pages={11844--11853},
  year={2025}
}

@inproceedings{li2023logonet,
  title={Logonet: Towards accurate 3d object detection with local-to-global cross-modal fusion},
  author={Li, Xin and Ma, Tao and Hou, Yuenan and Shi, Botian and Yang, Yuchen and Liu, Youquan and Wu, Xingjiao and Chen, Qin and Li, Yikang and Qiao, Yu and others},
  booktitle={Proceedings of the IEEE/CVF conference on computer vision and pattern recognition},
  pages={17524--17534},
  year={2023}
}

@article{zhu2022vpfnet,
  title={VPFNet: Improving 3D object detection with virtual point based LiDAR and stereo data fusion},
  author={Zhu, Hanqi and Deng, Jiajun and Zhang, Yu and Ji, Jianmin and Mao, Qiuyu and Li, Houqiang and Zhang, Yanyong},
  journal={IEEE Transactions on Multimedia},
  volume={25},
  pages={5291--5304},
  year={2022},
  publisher={IEEE}
}

@inproceedings{wu2022sparse,
  title={Sparse fuse dense: Towards high quality 3d detection with depth completion},
  author={Wu, Xiaopei and Peng, Liang and Yang, Honghui and Xie, Liang and Huang, Chenxi and Deng, Chengqi and Liu, Haifeng and Cai, Deng},
  booktitle={Proceedings of the IEEE/CVF conference on computer vision and pattern recognition},
  pages={5418--5427},
  year={2022}
}

@inproceedings{wu2023virtual,
  title={Virtual sparse convolution for multimodal 3d object detection},
  author={Wu, Hai and Wen, Chenglu and Shi, Shaoshuai and Li, Xin and Wang, Cheng},
  booktitle={Proceedings of the IEEE/CVF conference on computer vision and pattern recognition},
  pages={21653--21662},
  year={2023}
}

@inproceedings{chen2017multi,
  title={Multi-view 3d object detection network for autonomous driving},
  author={Chen, Xiaozhi and Ma, Huimin and Wan, Ji and Li, Bo and Xia, Tian},
  booktitle={Proceedings of the IEEE conference on Computer Vision and Pattern Recognition},
  pages={1907--1915},
  year={2017}
}

@inproceedings{ku2018joint,
  title={Joint 3d proposal generation and object detection from view aggregation},
  author={Ku, Jason and Mozifian, Melissa and Lee, Jungwook and Harakeh, Ali and Waslander, Steven L},
  booktitle={2018 IEEE/RSJ international conference on intelligent robots and systems (IROS)},
  pages={1--8},
  year={2018},
  organization={IEEE}
}

@inproceedings{xu2018pointfusion,
  title={Pointfusion: Deep sensor fusion for 3d bounding box estimation},
  author={Xu, Danfei and Anguelov, Dragomir and Jain, Ashesh},
  booktitle={Proceedings of the IEEE conference on computer vision and pattern recognition},
  pages={244--253},
  year={2018}
}

@inproceedings{qi2018frustum,
  title={Frustum pointnets for 3d object detection from rgb-d data},
  author={Qi, Charles R and Liu, Wei and Wu, Chenxia and Su, Hao and Guibas, Leonidas J},
  booktitle={Proceedings of the IEEE conference on computer vision and pattern recognition},
  pages={918--927},
  year={2018}
}

@inproceedings{pang2020clocs,
  title={CLOCs: Camera-LiDAR object candidates fusion for 3D object detection},
  author={Pang, Su and Morris, Daniel and Radha, Hayder},
  booktitle={2020 IEEE/RSJ International Conference on Intelligent Robots and Systems (IROS)},
  pages={10386--10393},
  year={2020},
  organization={IEEE}
}

@inproceedings{huang2020epnet,
  title={Epnet: Enhancing point features with image semantics for 3d object detection},
  author={Huang, Tengteng and Liu, Zhe and Chen, Xiwu and Bai, Xiang},
  booktitle={European conference on computer vision},
  pages={35--52},
  year={2020},
  organization={Springer}
}

@inproceedings{li2022voxel,
  title={Voxel field fusion for 3d object detection},
  author={Li, Yanwei and Qi, Xiaojuan and Chen, Yukang and Wang, Liwei and Li, Zeming and Sun, Jian and Jia, Jiaya},
  booktitle={Proceedings of the IEEE/CVF conference on computer vision and pattern recognition},
  pages={1120--1129},
  year={2022}
}

@inproceedings{vora2020pointpainting,
  title={Pointpainting: Sequential fusion for 3d object detection},
  author={Vora, Sourabh and Lang, Alex H and Helou, Bassam and Beijbom, Oscar},
  booktitle={Proceedings of the IEEE/CVF conference on computer vision and pattern recognition},
  pages={4604--4612},
  year={2020}
}

@inproceedings{yang2022graph,
  title={Graph r-cnn: Towards accurate 3d object detection with semantic-decorated local graph},
  author={Yang, Honghui and Liu, Zili and Wu, Xiaopei and Wang, Wenxiao and Qian, Wei and He, Xiaofei and Cai, Deng},
  booktitle={European conference on computer vision},
  pages={662--679},
  year={2022},
  organization={Springer}
}

@inproceedings{zhang2022cat,
  title={Cat-det: Contrastively augmented transformer for multi-modal 3d object detection},
  author={Zhang, Yanan and Chen, Jiaxin and Huang, Di},
  booktitle={Proceedings of the IEEE/CVF Conference on Computer Vision and Pattern Recognition},
  pages={908--917},
  year={2022}
}

@article{mushtaq2025cls,
  title={CLS-3D: Content-Wise LiDAR-Camera Fusion and Slot Reweighting Transformer for 3D Object Detection in Autonomous Vehicles},
  author={Mushtaq, Husnain and Latif, Sohaib and Ilyas, Muhammad Saad Bin and Mohsin, Syed Muhammad and Ali, Mohammed},
  journal={IEEE Access},
  year={2025},
  publisher={IEEE}
}

@article{chen2022msl3d,
  title={MSL3D: 3D object detection from monocular, stereo and point cloud for autonomous driving},
  author={Chen, Wenyu and Li, Peixuan and Zhao, Huaici},
  journal={Neurocomputing},
  volume={494},
  pages={23--32},
  year={2022},
  publisher={Elsevier}
}

@article{chen2023lidar,
  title={LiDAR-camera fusion: Dual transformer enhancement for 3D object detection},
  author={Chen, Mu and Liu, Pengfei and Zhao, Huaici},
  journal={Engineering Applications of Artificial Intelligence},
  volume={120},
  pages={105815},
  year={2023},
  publisher={Elsevier}
}

@article{an2022deep,
  title={Deep structural information fusion for 3D object detection on LiDAR--camera system},
  author={An, Pei and Liang, Junxiong and Yu, Kun and Fang, Bin and Ma, Jie},
  journal={Computer Vision and Image Understanding},
  volume={214},
  pages={103295},
  year={2022},
  publisher={Elsevier}
}

@article{li2023mvmm,
  title={MVMM: Multiview multimodal 3-D object detection for autonomous driving},
  author={Li, Shangjie and Geng, Keke and Yin, Guodong and Wang, Ziwei and Qian, Min},
  journal={IEEE Transactions on Industrial Informatics},
  volume={20},
  number={1},
  pages={845--853},
  year={2023},
  publisher={IEEE}
}

@article{uzair2024channel,
  title={Channel-wise and spatially-guided Multimodal feature fusion network for 3D Object Detection in Autonomous Vehicles},
  author={Uzair, Muhammad and Dong, Jian and Shi, Ronghua and Mushtaq, Husnain and Ullah, Irshad},
  journal={IEEE Transactions on Geoscience and Remote Sensing},
  year={2024},
  publisher={IEEE}
}

@article{shen2025point,
  title={Point-Level Fusion and Channel Attention for 3D Object Detection in Autonomous Driving},
  author={Shen, Juntao and Fang, Zheng and Huang, Jin},
  journal={Sensors},
  volume={25},
  number={4},
  pages={1097},
  year={2025},
  publisher={MDPI}
}

@article{li2025tinyfusiondet,
  title={TinyFusionDet: Hardware-Efficient LiDAR-Camera Fusion Framework for 3D Object Detection at Edge},
  author={Li, Yishi and Zeng, Fanhong and Lai, Rui and Wu, Tong and Guan, Juntao and Zhu, Anfu and Zhu, Zhangming},
  journal={IEEE Transactions on Circuits and Systems for Video Technology},
  year={2025},
  publisher={IEEE}
}

@article{liu2022epnet++,
  title={EPNet++: Cascade bi-directional fusion for multi-modal 3D object detection},
  author={Liu, Zhe and Huang, Tengteng and Li, Bingling and Chen, Xiwu and Wang, Xi and Bai, Xiang},
  journal={IEEE transactions on pattern analysis and machine intelligence},
  volume={45},
  number={7},
  pages={8324--8341},
  year={2022},
  publisher={IEEE}
}

@inproceedings{qin2023supfusion,
  title={SupFusion: Supervised LiDAR-camera fusion for 3D object detection},
  author={Qin, Yiran and Wang, Chaoqun and Kang, Zijian and Ma, Ningning and Li, Zhen and Zhang, Ruimao},
  booktitle={Proceedings of the IEEE/CVF international conference on computer vision},
  pages={22014--22024},
  year={2023}
}

@article{chen2025multi,
  title={Multi-Modal BEV Enhancement Fusion for 3D Object Detection in Autonomous Driving},
  author={Chen, Mu and Yang, Mingchuan and Zhang, Yuan and Han, Tao and Li, Xinchi and Zhao, Huaici and Liu, Pengfei},
  journal={IEEE Transactions on Intelligent Transportation Systems},
  year={2025},
  publisher={IEEE}
}
\endgroup
\end{document}